%% file: main.tex
\documentclass[11pt,twoside,leqno,a4paper]{amsart}

\usepackage{amsthm,amsmath,amsfonts,amssymb}
\usepackage[foot]{amsaddr}
\usepackage{geometry}
\usepackage[mathscr]{euscript}
\usepackage[bb=px]{mathalfa}
\usepackage{graphicx}
\usepackage{booktabs}
\usepackage{tabularx}
\usepackage{longtable}
\usepackage{makecell}
\usepackage{pifont}
\usepackage{algorithm}
\usepackage{xcolor}
\usepackage[hyperfootnotes=false,hypertexnames=false,naturalnames=false,breaklinks]{hyperref}

\usepackage[dvipsnames]{xcolor}
\colorlet{inlinkcolor}{purple}
\colorlet{exlinkcolor}{purple}
\colorlet{citecolor}{teal!75!blue}
\hypersetup{colorlinks=true,
    linkcolor=inlinkcolor,
    citecolor=citecolor,
    urlcolor=exlinkcolor,
    linktoc=page,
    breaklinks=true,
    plainpages=false}

\usepackage[mathscr]{euscript}
\usepackage{enumerate}
\usepackage{mathtools}
\usepackage{xcolor}
\newtheorem{theorem}{Theorem}[section]
\newtheorem{proposition}[theorem]{Proposition}
\newtheorem{lemma}[theorem]{Lemma}
\newtheorem{corollary}[theorem]{Corollary}
\theoremstyle{definition}
\newtheorem{definition}[theorem]{Definition}

\newtheorem{assumption}{Assumption}
\theoremstyle{remark}
\newtheorem{remark}[theorem]{Remark}
\theoremstyle{definition}
\newtheorem{condition}{Condition}
\numberwithin{equation}{section}

\input{command}

\usepackage{algpseudocode}

\title[Characteristic Learning for Provable One-Step Generation]{\large Characteristic Learning for Provable One-Step Generation}

\author[Z. Ding]{Zhao Ding$^{1,2}$}
\email{zhao.ding@polyu.edu.hk}
\author[C. Duan]{Chenguang Duan$^{3,2}$}
\email{duan@igpm.rwth-aachen.de}
\author[Y. Jiao]{Yuling Jiao$^{4,5,6}$}
\email{yulingjiaomath@whu.edu.cn}
\author[R. Li]{Ruoxuan Li$^{7,2}$}
\email{ruoxuanli.math@amss.ac.cn}
\author[J.Z. Yang]{Jerry Zhijian Yang$^{5,8,2,6}$}
\email{zjyang.math@whu.edu.cn}
\author[P. Zhang]{Pingwen Zhang$^{8,9}$}
\email{pzhang@pku.edu.cn}
\address{1. Department of Applied Mathematics, Hong Kong Polytechnic University, Hong Kong SAR, 2. School of Mathematics and Statistics, Wuhan University, Wuhan 430072, China, 3. Institut f\"ur Geometrie und Praktische Mathematik, RWTH Aachen University, Im S\"usterfeld 2, 52072 Aachen, Germany, 4. School of Artificial Intelligence, Wuhan University, Wuhan 430072, China, 5. National Center for Applied Mathematics in Hubei, Wuhan University, Wuhan 430072, China, 6. Hubei Key Laboratory of Computational Science, Wuhan University, Wuhan 430072, China, 7. Institute of Applied Mathematics, Academy of Mathematics and Systems Science, Chinese Academy of Sciences, Beijing, China, 8. Wuhan Institute for Math \& AI, Wuhan University, Wuhan 430072, China, 9. School of Mathematical Sciences, Peking University, Beijing 100871, China.}

\date{\today}

\begin{document}
\maketitle

\begin{abstract}
We propose the characteristic generator, an one-step generative model that combines the sampling efficiency of generative adversarial networks (GANs) with the training stability of flow-based models. The proposed model is based on characteristics along which probability-density transport is governed by ordinary differential equations (ODEs). Specifically, we first estimate the underlying velocity field and numerically solve the probability-flow ODE using an exponential integrator, thereby obtaining discrete approximations of the characteristic trajectories. We then train a deep neural network to approximate these trajectories, yielding a one-step transport map from a Gaussian prior to the target distribution. Theoretically, we analyze the errors introduced by velocity-field estimation, numerical discretization, and characteristic approximation, and establish a nonasymptotic convergence rate in the Wasserstein-2 distance under mild assumptions on the data distribution. Moreover, under a low-dimensional linear-subspace assumption, we show that the convergence rate depends on the intrinsic data dimension rather than the potentially much larger ambient dimension, demonstrating the model's ability to alleviate the curse of dimensionality. Experiments on synthetic and real-world datasets show that the characteristic generator produces high-quality, high-resolution samples using only one or a few neural-network evaluations.
\end{abstract}

\input{introduction}

\input{analysis}

\input{numerical}

\input{conclusion}

\bibliographystyle{abbrv}
\bibliography{reference}

\input{appendix}

\end{document}

%% file: command.tex
\DeclareMathOperator{\pr}{Pr}
\DeclareMathOperator{\vcdim}{VCdim}
\DeclareMathOperator{\supp}{supp}
\DeclareMathOperator{\law}{Law}
\DeclareMathOperator{\ac}{ac}
\DeclareMathOperator{\lip}{Lip}
\DeclareMathOperator{\relu}{ReLU}
\DeclareMathOperator{\requ}{ReQU}
\DeclareMathOperator{\op}{op}

\DeclareMathOperator{\cov}{Cov}
\DeclareMathOperator{\var}{Var}

\DeclareMathOperator{\unif}{Unif}
\DeclareMathOperator{\vel}{vel}
\DeclareMathOperator{\flow}{flow}

\DeclareMathOperator{\ei}{EI}
\DeclareMathOperator{\off}{off}
\DeclareMathOperator{\Int}{int}
\DeclareMathOperator{\loc}{loc}
\DeclareMathOperator{\glo}{glo}
\def\d{\,\mathrm{d}}
\def\dt{\,\mathrm{d}t}
\def\ds{\,\mathrm{d}s}
\def\dx{\,\mathrm{d}x}

\def\dsdt{\,\mathrm{d}s\mathrm{d}t}
\DeclareMathOperator{\ess}{ess}
\def\esssup{\mathop{\ess\,\sup}}
\def\what{\widehat}
\def\bbone{\mathbbm{1}}
\usepackage[bb=px]{mathalfa}

\usepackage{bm,bbm}

\def\calD{{\mathcal{D}}}
\def\calE{{\mathcal{E}}}

\def\calL{{\mathcal{L}}}

\def\calO{{\mathcal{O}}}

\def\calR{{\mathcal{R}}}
\def\calS{{\mathcal{S}}}
\def\calT{{\mathcal{T}}}

\def\calX{{\mathcal{X}}}

\def\bbB{{\mathbb{B}}}

\def\bbE{{\mathbb{E}}}

\def\bbN{{\mathbb{N}}}

\def\bbR{{\mathbb{R}}}

\usepackage{mathrsfs}

\def\scrD{{\mathscr{D}}}

\def\scrF{{\mathscr{F}}}
\def\scrG{{\mathscr{G}}}
\def\scrH{{\mathscr{H}}}

\def\euD{{\EuScript{D}}}

\def\euS{{\EuScript{S}}}

\def\euZ{{\EuScript{Z}}}

\DeclareMathOperator*{\argmin}{arg\,min}
\DeclareMathOperator{\sign}{sign}

\def\cskip{c_{\text{skip}}}
\def\cout{c_{\text{out}}}
\def\cin{c_{\text{in}}}
\def\cnoise{c_{\text{noise}}}

\newcommand{\td}[1]{\tilde{#1}}

%% file: introduction.tex
\section{Introduction}

\par Generative models aim to learn an underlying target distribution from observed data and generate new samples from the learned distribution. They have found broad applications in image and video generation~\cite{radford2016unsupervised,meng2022sdedit,Ho2022Video}, text-to-image synthesis~\cite{Ramesh2021Zero,ramesh2022hierarchical,Kang2023Scaling}, and speech synthesis~\cite{kong2021diffwave,chen2021wavegrad}. Among the most influential and widely adopted approaches are generative adversarial networks (GANs)~\cite{Goodfellow2014GAN} and their variants~\cite{Arjovsky2017Wasserstein}. A key advantage of GANs is their high sampling efficiency, since generating a new sample typically requires only a single evaluation of the trained generator. Despite their remarkable empirical success~\cite{Reed2016Generative} and growing theoretical understanding~\cite{Liang2021How,Liu2021Nonasymptotic,Huang2022Error,Zhou2023deep}, GANs remain subject to inherent training instabilities~\cite{Salimans2016Improved}.

\par In recent years, diffusion models~\cite{Ho2020Denoising,Song2021Maximum,song2021score,karras2022elucidating} and flow-based models~\cite{liu2022flow,lipman2023flow} have emerged as powerful generative models. These models have also laid the foundation for the development of generative AI models, such as DALL-E~\cite{Ramesh2021Zero,ramesh2022hierarchical}, Midjourney, Stable Diffusion~\cite{esser2024scaling}, and Sora~\cite{videoworldsimulators2024}. Theoretical analysis for these methods has been studied by~\cite{Oko2023Diffusion,Lee2022Convergence,Lee2023Convergence,chen2023sampling,chen2023probability,benton2024linear,benton2024error,gao2024convergence,wu2024theoretical}. Although diffusion or flow-based models outperform GANs in generation quality across various tasks~\cite{Dhariwal2021Diffusion}, they require hundreds or even thousands of sequential steps involving large neural network evaluations for sampling. As a consequence, their sampling speed is much slower compared to one-step GANs.

\par The instability of GANs and the inefficiency of sampling in diffusion or flow-based models have emerged as significant bottlenecks in practical applications of generative models. This raises two crucial questions:
\begin{quote}
\emph{How can we develop an one-step generative model that combines the efficient sampling of GANs with the stable performance of diffusion or flow-based models? How can we establish a rigorous error analysis for this generative model?}
\end{quote}

\par Several recent papers have made progress in addressing the first question using various techniques such as distillation~\cite{luhman2021knowledge,salimans2022progressive,Song2023Consistency,zhou2024score}, operator learning~\cite{Zheng2023Fast}, or trajectory models~\cite{kim2024consistency,ren2024hypersd}. For a more detailed discussion, please refer to Section~\ref{section:related:work:fast}. Despite these recent advancements, a unifying mathematical framework for designing and analyzing the one-step generative models remains largely limited~\cite{li2024mathematical}. This paper aims to fill this gap and provide a positive answer to the aforementioned questions. Specifically, we introduce a comprehensive framework, known as the characteristic generator, aiming to streamline the sampling process using a single evaluation of the neural network. Our model merges the sampling efficiency of GANs with the promising performance of flow-based models. Furthermore, we present a rigorous error analysis for the characteristic generator. Through numerical experiments, we validate that our approach generates high-quality samples that are comparable to those generated through flow-based models, all while requiring just a single evaluation of the neural network.

\subsection{Contributions}
\par Our contributions are summarized as follows:
\begin{enumerate}[(i)]
\item We introduce the ``characteristic generator,'' a one-step generative model that forms the basis of a unified mathematical framework for recent flow-based distillation methods. Our approach involves modeling the probability density transport equation with stochastic interpolants, which yields a probability flow ODE. By learning to approximate the characteristic curves of the probability transportation, the generator can directly transform a prior distribution to the target distribution without iterative simulation.
\item We provide a rigorous error analysis for the characteristic generator under minimal assumptions on data distribution. Specifically, we derive a convergence rate $\calO(n^{-\frac{2}{d+3}})$ for denoiser matching (Theorem~\ref{theorem:velocity:rate}). Additionally, we propose an error bound for the Wasserstein-2 distance between the distribution of data generated by the exponential integrator (EI) and the target distribution (Theorem~\ref{theorem:error:euler}), which is of independent interest. Finally, we present a non-asymptotic convergence rate for the characteristic generator in the averaged Wasserstein-2 distance (Theorem~\ref{theorem:error:characteristic}). These findings also provide valuable theoretical understanding for distillation~\cite{salimans2022progressive,Song2023Consistency}, operator learning~\cite{Zheng2023Fast}, or trajectory model~\cite{kim2024consistency,ren2024hypersd}.
\item We prove that the characteristic generator mitigates the curse of dimensionality (Corollaries~\ref{corollary:rate:linear:manifold} and~\ref{corollary:rate:follmer:manifold}) under a low-dimensional linear-subspace assumption. We show the convergence rate depends on the data's low intrinsic dimension rather than the high ambient dimension. To our knowledge, this is the first end-to-end analysis that formally explains the strong performance of flow-based one-step generative models on high-dimensional data.
\item We conduct extensive experiments on synthetic and real-world data (CIFAR-10, CelebA-HQ), demonstrating that the characteristic generator produces high-quality samples in a single network evaluation. Our model significantly outperforms prior non-adversarial one-step models~\cite{salimans2022progressive,Song2023Consistency,kim2024consistency} on CIFAR-10. Furthermore, it achieves performance competitive with state-of-the-art methods~\cite[CTM]{kim2024consistency} in just a few iterations, without requiring additional adversarial training. The successful application to large-scale and high-resolution image generation ($256\times256$ and $512\times512$) further validates the scalability of our approach.
\end{enumerate}

\subsection{Main Results}

\par Let $\mu_{0}\in P_{\ac}(\bbR^{d})$ be a known prior distribution, i.e., a standard Gaussian, and let $\mu_{1}\in P_{\ac}(\bbR^{d})$ be the target distribution with an unknown density function $\rho_{1}$. Suppose one has access to finite i.i.d. data samples $X_{1}^{1},\ldots,X_{1}^{n}$ from $\mu_{1}$. In generative learning, we aim to learn a push-forward map $G^{*}:\bbR^{d}\rightarrow\bbR^{d}$ that transports the prior distribution $\mu_{0}$ onto the target distribution $\mu_{1}$. This relationship is defined by the normalizing equation~\cite{Rozen2021Moser}:
\begin{equation}\label{eq:normalizing}
G_{\sharp}^{*}\mu_{0}=\mu_{1}.
\end{equation}
Formally, the goal of the generative learning is to find an estimator $\what{G}$ of the push-forward operator $G^{*}$ based on finite samples $X_{1}^{1},\ldots,X_{1}^{n}$ drawn from $\mu_{1}$.

\begin{remark}[Regularity of the push-forward map]
The learnability of the push-forward map $G^{*}$ requires its regularity. Without crucial properties like boundedness and Lipschitz continuity, the map can be ill-behaved, making it practically impossible for function approximators like neural networks to learn. In this work, we define $G^{*}$ via a probability flow ODE~\eqref{eq:PF:ODE}, and establish the necessary regularity conditions in Section~\ref{section:analysis:property}.
\end{remark}

\par In this work, we construct the desired push-forward operator via a probability flow ODE:
\begin{equation}\label{eq:PF:ODE}
\begin{aligned}
\frac{\d}{\dt}x(t)&=b^{*}(t,x(t)), \quad t\in(0,1), \\
x(0) &=x_{0}\sim\mu_{0},
\end{aligned}
\end{equation}
where the velocity field $b^{*}$ is given as~\eqref{eq:velocity}, and $x(t)$ is known as the characteristic curve of the continuity equation~\eqref{eq:transport}. Denote by $\mu_{t}$ the law of $x(t)$ for each $t\in(0,1)$, and define the flow map $g_{t,s}^{*}$ as $g_{t,s}^{*}(x(t))=x(s)$ for each $0\leq t\leq s\leq1$, or equivalently~\eqref{eq:characteristic:flow}. The flow map transports particles along the characteristic curves, and satisfies $(g_{t,s}^{*})_{\sharp}\mu_{t}=\mu_{s}$. Our target push-forward map is defined as $G^{*}\coloneqq g_{0,1}^{*}$. Let $\what{E}_{0,T}$ be the exponential integrator approximation~\eqref{eq:characteristic:ei:solution} to the flow map $g_{0,T}^{*}$. Denote by $\what{g}_{t,s}$ the estimator of $g_{t,s}^{*}$ as~\eqref{eq:charateristic:erm}, which is referred to as the characteristic generator.

\par Our theoretical results are established under the following assumptions on the prior and target distributions, which will be discussed in Section~\ref{section:analysis:assumptions}.

\begin{assumption}\label{assumption:init:dist:gaussian}
The prior distribution $\mu_{0}=\mathcal{N}(0,I_{d})$.
\end{assumption}

\begin{assumption}\label{assumption:target:dist}
{There exists an unknown constant $\sigma>0$, such that
\begin{equation*}
\mu_{1}(x)=\mathcal{N}(0,\sigma^{2}I_{d})\ast\nu\coloneqq \int\varphi_{d}(x;x^{\prime},\sigma^{2}I_{d})\d\nu(x^{\prime}),
\end{equation*}
where $\varphi_{d}(\cdot;m,\Sigma)$ represents the probability density of a $d$-dimensional Gaussian distribution with mean $m$ and covariance matrix $\Sigma$, and $\d\nu(x)=p(x)\d x$ with $\supp(\nu)\subset[0,1]^{d}$.}
\end{assumption}

\par Assumption~\ref{assumption:target:dist} requires the target distribution to be a Gaussian convolution. That is, for a random variable $X\sim\mu_{1}$, there exist two independent random variables $Z\in[0,1]^{d}$ and $W\sim\mathcal{N}(0,I_{d})$, such that $X\stackrel{\d}{=}Z+\sigma W$. This assumption is essential as it ensures desirable properties of the probability flow ODE, such as bounded moments and the Lipschitz property of the velocity field. Further details can be found in Sections~\ref{section:analysis:assumptions} and~\ref{section:analysis:property}. It is noteworthy that this assumption can be considered relatively mild, given that the smoothed distribution $\mu_1$ is an approximation of the original distribution $\nu$, particularly when the variance $\sigma^{2}$ of the Gaussian distribution is small. In addition, it encompasses non-log-concave distributions, such as Gaussian mixtures~\cite[Appendix C]{Grenioux2024Stochastic}. Similar assumptions have been considered by~\cite{saremi2024chain,Grenioux2024Stochastic,beyler2025convergence}.

\par The following theorem presents the Wasserstein-2 error bound for the exponential integrator (EI) generator~\eqref{eq:characteristic:ei:solution}.

\begin{theorem}[Informal version of Corollary~\ref{corollary:error:euler}]
Suppose that Assumptions~\ref{assumption:init:dist:gaussian} and~\ref{assumption:target:dist} hold. Let $\euS$ be a set of $n$ samples independently and identically drawn from the target distribution $\mu_{1}$. Suppose the depth and width of the denoiser neural network are set properly. Let $K\geq Cn^{\frac{2}{d+3}}$ be the number of time steps of EI. Then the following inequality holds
\begin{equation*}
\bbE_{\euS}\Big[W_{2}^{2}\Big((\what{E}_{0,T})_{\sharp}\mu_{0},\mu_{1}\Big)\Big] 
\leq C_{T}^{1}n^{-\frac{2}{d+3}}\log^{2}n+C_{T}^{2}W_{2}^{2}(\mu_{0},\mu_{1}),
\end{equation*}
where $C_{T}^{1}$ and $C_{T}^{2}$ are two constants depending only on $d$, $\sigma$ and $T$. Further, as the stopping time $T\rightarrow 1$, the constant $C_{T}^{1}$ tends to infinity polynomially while $C_{T}^{2}$ decreases to zero polynomially.
\end{theorem}

\par The averaged Wasserstein-2 bound for characteristic generator~\eqref{eq:charateristic:erm} is stated as follows.

\begin{theorem}[Informal version of Theorem~\ref{theorem:error:characteristic}]
Suppose that Assumptions~\ref{assumption:init:dist:gaussian} and~\ref{assumption:target:dist} hold. Let $\euS$ be a set of $n$ samples independently and identically drawn from the target distribution $\mu_{1}$, and let $\euZ$ be a set of $m$ trajectories generated by EI with $K$ time steps. Suppose the depth and width of the denoiser neural network and the characteristic neural network are set properly, respectively. Then the following inequality holds
\begin{align*}
&\bbE_{\euS}\bbE_{\euZ}\Big[\frac{2}{T^{2}}\int_{0}^{T}\int_{t}^{T}W_{2}^{2}\Big((\what{g}_{t,s})_{\sharp}\mu_{t},\mu_{s}\Big)\dsdt\Big] \\
&\leq C_{T}^{1}\Big\{n^{-\frac{2}{d+3}}\log^{2}n+\frac{1}{K}\Big\}+C\Big\{m^{-\frac{2}{d+4}}\log^{2}m+\frac{\log m}{K}\Big\},
\end{align*}
where $C_{T}^{1}$ is a constant depending on $d$, $\sigma$ and $T$, and $C$ is a constant depending on $d$ and $\sigma$. Further, as the stopping time $T\rightarrow 1$, the constant $C_{T}^{1}$ tends to infinity polynomially.
\end{theorem}

\par  As a consequence of Theorem~\ref{theorem:error:characteristic}, we propose in Table \ref{table:rates} the convergence rate of characteristic generators induced by two special probability flow ODEs.

\begin{table*}[htbp]
\caption{Convergence rates of characteristic generator.}
\centering
\small
\begin{tabular}{lccccc}
\toprule
\textbf{Probability Flow ODE} & $\alpha_{t}$ & $\beta_{t}$ & \textbf{Assumptions} & \textbf{Convergence Rate} &  \\
\midrule
Linear interpolants~\cite{liu2022flow,lipman2023flow} & $1-t$ & $t$ & Assumptions~\ref{assumption:init:dist:gaussian} and~\ref{assumption:target:dist} & $\calO(n^{-\frac{4}{3(d+3)}})$ & Corollary~\ref{corollary:rate:linear} \\
Linear interpolants~\cite{liu2022flow,lipman2023flow} & $1-t$ & $t$ & Assumptions~\ref{assumption:init:dist:gaussian} and~\ref{assumption:target:dist:manifold} & $\calO(n^{-\frac{4}{3(d^{*}+3)}})$ &  Corollary~\ref{corollary:rate:linear:manifold} \\
\midrule
F{\"o}llmer flow~\cite{chang2024deep,jiao2024convergence} & $\sqrt{1-t^{2}}$ & $t$ & Assumptions~\ref{assumption:init:dist:gaussian} and~\ref{assumption:target:dist} & $\calO(n^{-\frac{1}{d+3}})$ & Corollary~\ref{corollary:rate:follmer} \\
F{\"o}llmer flow~\cite{chang2024deep,jiao2024convergence} & $\sqrt{1-t^{2}}$ & $t$ & Assumptions~\ref{assumption:init:dist:gaussian} and~\ref{assumption:target:dist:manifold} & $\calO(n^{-\frac{1}{d^{*}+3}})$ &  Corollary~\ref{corollary:rate:follmer:manifold} \\
\bottomrule
\end{tabular}
\label{table:rates}
\end{table*}

\par The convergence rates under Assumptions~\ref{assumption:init:dist:gaussian} and~\ref{assumption:target:dist} in Table~\ref{table:rates} are direct conclusions of Theorem~\ref{theorem:error:characteristic}, which suffer from the curse of dimensionality (CoD). However, in many practical applications, high-dimensional data --- such as images or text documents --- often lie close to a low-dimensional manifold embedded in the high-dimensional ambient space~\cite{Goodfellow2016deep,bortoli2022convergence,Jiao2023deep,Oko2023Diffusion}. This observation motivates a refined assumption that captures the intrinsic low-dimensional structure of real-world data. We formalize this idea with the following assumption.

\begin{assumption}[Low-dimensional linear sub-space]
\label{assumption:target:dist:manifold}
There exist constants $\sigma>0$ and $d^{*}\ll d$, such that
\begin{equation*}
\mu_{1}=\mathcal{N}(0,\sigma^{2}I_{d})*(P_{\sharp}\tilde{\nu}),
\end{equation*}
where $P\in\bbR^{d \times d^{*}}$ is a column-orthonormal matrix, and $\tilde{\nu}$ is a distribution with $\supp(\tilde{\nu}) \subset [0,1]^{d^{*}}$.
\end{assumption}

\par Note that the distribution $\tilde{\nu}$ in Assumption~\ref{assumption:target:dist:manifold} is supported on a subset of a $d^*$-dimensional hypercube. Since $P$ is column orthogonal, i.e., $P^{\top}P=I_{d^{*}}$, it acts as an isometric embedding of $\mathbb{R}^{d^{*}}$ into the ambient space $\mathbb{R}^{d}$. Consequently, the push-forward measure $\nu \coloneqq  P_{\sharp}\tilde{\nu}$ lies in a $d^*$-dimensional linear sub-space of the ambient space $\mathbb{R}^d$. This assumption is crucial, as it allows us to show that the convergence rate depends only on the intrinsic dimension $d^{*}$ rather than the ambient dimension $d$, thereby effectively mitigating the curse of dimensionality. The convergence rate of characteristic generators under low-dimensional assumption are summarized in Table~\ref{table:rates}.

\par Experiment results and discussions can be found in Section~\ref{section:numerical}. Our code is available online at \url{https://github.com/burning489/CharacteristicGenerator}.

\subsection{Preliminaries and Notations}\label{section:preliminaries}

\subsubsection{Wasserstein-2 distance}
\par Let $P_{\ac}(\bbR^{d})$ be the space of probability measures  on $\bbR^{d}$, which are absolutely continuous with respect to Lebesgue measure. Suppose $\mu_{0},\mu_{1}\in P_{\ac}(\bbR^{d})$ with $\d\mu_{0}(x)=\rho_{0}(x)\dx$ and $\d\mu_{1}(x)=\rho_{1}(x)\dx$. The Wasserstein-2 distance~\cite[Definition 6.1]{Villani2009Optimal} between $\mu_{0}$ and $\mu_{1}$ is defined by the formula
\begin{equation*}
W_{2}(\mu_{0},\mu_{1})=\inf\Big\{\bbE^{1/2}\big[\|X_{0}-X_{1}\|_{2}^{2}\big]: 
\law(X_{0})=\mu_{0},~\law(X_{1})=\mu_{1}\Big\}.
\end{equation*}

\subsubsection{Deep neural networks}

\par A neural network $f:\bbR^{N_{0}}\rightarrow\bbR^{N_{L+1}}$ is a function defined by
\begin{equation*}
f(x)=T_{L}(\varrho(T_{L-1}(\cdots\varrho(T_{0}(x))\cdots))),
\end{equation*}
where the activation function $\varrho$ is applied component-wisely and $T_{\ell}(x)\coloneqq A_{\ell}x+b_{\ell}$ is an affine transformation with $A_{\ell}\in\bbR^{N_{\ell+1}\times N_{\ell}}$ and $b_{\ell}\in\bbR^{N_{\ell}}$ for $\ell=0,\ldots,L$. In this paper, we consider the case where $N_{L+1}=d$. The number $L$ is called the depth of neural networks. Additionally, $S\coloneqq \sum_{\ell=0}^{L}(\|A_{\ell}\|_{0}+\|b_{\ell}\|_{0})$ represents the total number of non-zero weights within the neural network. We denote by $N(L,S)$ the set of neural networks with depth at most $L$ and the number of non-zero weights at most $S$.

\subsubsection{Notations}
\par The set of positive integers is denoted by $\bbN=\{1,2,\ldots\}$. We also denote $\bbN_{0}=\{0\}\cup\bbN_{+}$ for convenience. For a vector $x=(x_{1},\ldots,x_{d})\in\bbR^{d}$, we define its $\ell_{p}$-norms as $\|x\|_{p}^{p}=\sum_{i=1}^{d}|x_{i}|^{p}$ for $1\leq p<\infty$, with $\|x\|_{\infty}=\max_{1\leq i\leq d}|x_{i}|$. Denote by $\langle\cdot,\cdot\rangle$ the inner product between vectors, that is, $\langle x,y\rangle=\sum_{k=1}^{d}x_{k}y_{k}$, where $y=(y_{1},\ldots,y_{d})$. For a matrix $A\in\bbR^{m\times n}$, the operator norm induced by the $\ell_{2}$-norm is defined as $\|A\|_{\op}=\sup_{\|x\|_{2}=1}\|Ax\|_{2}$. Additionally, denote by $\bbB_{R}^{\infty}$ the $\ell_{\infty}$ ball in $\bbR^{d}$ with radius $R$, that is, $\bbB_{R}^{\infty}=\{x\in\bbR^{d}:\|x\|_{\infty}\leq R\}$. For a matrix $A\in\mathbb{R}^{d\times d}$, we say $A\succeq0$ if and only if it is positive definite. Let $A$ and $B$ be two matrices, denote $A\succeq B$ if and only if $(A-B)\succeq 0$. For a function $u(t)$ of time $t$, the time derivative is denoted by $\dot{u}$ or $\partial_{t}u$. Further, let $\ddot{u}$ denote the second-order time derivative. Additionally, we use $\nabla$ and $\nabla\cdot$ to denote the spatial gradient and divergence operators, respectively.

\subsection{Organization}
\par The remainder of this article is organized as follows. Section \ref{section:method} introduces the characteristic generative learning method. A thorough analysis of this method is provided in Section \ref{section:analysis}. Section \ref{section:numerical} presents numerical studies and discussions. Finally, Section \ref{section:conclusion} presents the conclusion and discusses future work. The related work, theoretical proofs, and experimental details are provided in the supplementary materials.

%% file: analysis.tex
%! TEX root=main.tex

\section{Characteristic Generative Learning}\label{section:method}

\par Dating back to~\cite{Moser1965volume,Dacorogna1990partial}, researchers proposed a continuous dynamic-induced approach for solving the normalizing equation~\eqref{eq:normalizing}. In the field of deep generative learning, flow-based models utilize ODE-dynamics to construct probability flows, effectively pushing the prior distribution towards the target distribution. This family of generative models is represented by continuous normalizing flows (CNF)~\cite{Chen2018Neural,grathwohl2019scalable} and their variants~\cite{Gao2019Deep,Rozen2021Moser,Gao2022Deep,lipman2023flow,Neklyudov2023Action,albergo2023building}. The major challenges faced by flow-based models revolve around two key questions:

\begin{quotation}
\emph{\textbf{Q1.} During the training phase, how can we estimate the velocity field of the probability flow ODE? \textbf{Q2.} During the sampling phase, how can we solve the probability flow ODE efficiently?}
\end{quotation}

\par The goal of this section is to propose the characteristic learning that has potential to address the aforementioned questions. In Section~\ref{section:ODE:flow}, we derive a probability flow ODE based on the concept of stochastic interpolants and the method of characteristics. Subsequently, in Section~\ref{section:velocity}, we propose a denoiser matching approach using least-squares regression, which provides an efficient solution to \textbf{Q1}. To tackle \textbf{Q2}, we first solve the probability flow ODE numerically in Section~\ref{section:Euler}. Then Section~\ref{section:characteristic} introduces a regression problem to fit characteristics using the obtained numerical solutions. This leads to an efficient simulation-free sampling method for flow-based generative models. Finally, we summarize the training and sampling algorithms in Section~\ref{section:characteristic}.

\subsection{Characteristics and Probability Flow ODE}\label{section:ODE:flow}

\par In this work, we follow the framework of stochastic interpolant~\cite{albergo2023building,albergo2023stochastic,albergo2023stochasticinterpolants}. Let $X_{0}$ and $X_{1}$ be two independent random variables drawn from $\mu_{0}$ and $\mu_{1}$, respectively. The spatially linear stochastic interpolant $X_{t}$ is the stochastic process defined as
\begin{equation}\label{eq:interpolant}
X_{t}=\alpha_{t}X_{0}+\beta_{t}X_{1}, \quad t\in(0,1),
\end{equation}
where $\alpha_{t}$ and $\beta_{t}$ are two scalar-valued functions satisfying the following condition.

\begin{condition}\label{condition:interpolant}
The interpolant coefficients $\alpha_{t},\beta_{t}\in C([0,1])$ satisfy
\begin{enumerate}[(i)]
\item $\alpha_{0}=\beta_{1}=1$ and $\alpha_{1}=\beta_{0}=0$,
\item $\alpha_{t}^{2}+\beta_{t}^{2}>0$ for each $t\in[0,1]$,
\item $\alpha_{t}$ and $-\beta_{t}$ are monotonically decreasing, and
\item $\dot{\alpha}_{t},\ddot{\alpha}_{t}\in C([0,1))$, $\dot{\alpha}_{t}\alpha_{t}\in {C^1([0,1])}$, $\dot{\beta}_{t},\ddot{\beta}_{t}\in C([0,1])$.
\end{enumerate}
\end{condition}

\par In this paper, we focus on two examples shown in Table~\ref{table:rates}: linear interpolants and F{\"o}llmer flow. Both of them are widely-used in generative learning, such as~\cite{Nichol2021Improved,liu2022flow,albergo2023building,albergo2023stochasticinterpolants,lipman2023flow,chang2024deep}.

\par Denote by $\mu_{t}$ the law of $X_{t}$ for all time $t\in(0,1)$. The following proposition demonstrates that $\mu_{t}$ has a density $\rho_{t}$ that interpolates between $\rho_{0}$ and $\rho_{1}$. Further, the density $\rho_{t}$ satisfies the continuity equation.

\begin{proposition}[Continuity equation]\label{proposition:transport}
The distribution of the stochastic interpolant $X_{t}$ has a density function $\rho(t,x)$ satisfies $\rho(0,x)=\rho_{0}(x)$, $\rho(1,x)=\rho_{1}(x)$, and
\begin{equation*}
\rho(t,x)=\frac{1}{\beta_{t}}\int_{\bbR^{d}}\rho_{0}(x_{0})\rho_{1}\Big(\frac{x-\alpha_{t}x_{0}}{\beta_{t}}\Big)\dx_{0}=\frac{1}{\alpha_{t}}\int_{\bbR^{d}}\rho_{0}\Big(\frac{x-\beta_{t}x_{1}}{\alpha_{t}}\Big)\rho_{1}(x_{1})\dx_{1},
\end{equation*}
for each time $t\in(0,1)$. Further, the density $\rho(t,x)$ solves the continuity equation
\begin{equation}\label{eq:transport}
\partial_{t}\rho(t,x)+\nabla\cdot(b^{*}(t,x)\rho(t,x))=0,  (t,x)\in(0,1)\times\mathbb{R}^{d},
\end{equation}
where the velocity field is defined as
\begin{equation}\label{eq:velocity}
b^{*}(t,x)=\bbE\big[\dot{\alpha}_{t}X_{0}+\dot{\beta}_{t}X_{1} \mid X_{t}=x\big].
\end{equation}
\end{proposition}

\par To generate samples from the target distribution, we interpret the continuity equation~\eqref{eq:transport} through the method of characteristics~\cite[Section~II.2]{Courant1989Methods}. Along each characteristic curve, the continuity equation reduces to the ODE
\begin{equation}\label{eq:characteristic}
\frac{\mathrm{d}}{\mathrm{d}t}x(t) = b^{*}(t,x(t)),
\end{equation}
where $x(t)$ denotes the position of a particle at time $t \in (0,1)$ transported by the velocity field $b^{*}:(0,1)\times\mathbb{R}^{d}\to\mathbb{R}^{d}$. This characteristic ODE is known as the probability flow ODE~\cite{song2021score}. By a classical result in the theory of ODEs~\cite[Lemma~8.1.4]{Ambrosio2005Gradient}, the probability flow ODE~\eqref{eq:characteristic} admits a unique solution defined on the entire interval $[0,1]$, provided that for every compact set
$B \subset \mathbb{R}^{d}$,
\begin{equation}\label{eq:characteristic:well}
\int_{0}^{1}\Bigl(\sup_{B}\|b_{t}^{*}\|_{2} + \mathrm{Lip}(b^{*}_{t}, B)\Bigr)dt < \infty,
\end{equation}
where $b^{*}_{t} \coloneqq b^{*}(t, \cdot)$. The well-posedness of~\eqref{eq:characteristic} under Assumptions~\ref{assumption:init:dist:gaussian} and~\ref{assumption:target:dist} is established in Proposition~\ref{proposition:reg:velocity}.

Further, we define the associated two-parameter continuous normalizing flow $g_{t,s}^{*}:\mathbb{R}^{d}\to\mathbb{R}^{d}$ through the probability flow ODE~\eqref{eq:characteristic}:
\begin{equation}\label{eq:characteristic:flow}
\begin{aligned}
\frac{\mathrm{d}}{\mathrm{d}s}g_{t,s}^{*}(x_{t}) &= b^{*}(s,g_{t,s}^{*}(x_{t})), \quad 0\leq t\leq s\leq 1, \\
g_{t,t}^{*}(x_{t}) &=x_{t},
\end{aligned}
\end{equation}
which transports particles along the characteristic curves: $g_{t,s}^{*}:\bbR^{d}\rightarrow\bbR^{d},~x_{t}\mapsto x_{s}$ for any $\quad 0\leq t\leq s\leq 1$. Notice that the flow $g_{t,s}^{*}$ pushes the distribution $\mu_{t}$ onto $\mu_{s}$, that is,
\begin{equation*}
(g_{t,s}^{*})_{\sharp}\mu_{t} = \mu_{t}\circ g_{t,s}^{-1} = \mu_{s}, \quad 0\leq t\leq s\leq 1.
\end{equation*}
It is evident that the flow $g_{t,s}^{*}$ satisfies the semi-group property as follows.

\begin{proposition}[Semi-group property]\label{proposition:semi:group}
For each $x\in\bbR^{d}$, it holds that
\begin{enumerate}[(i)]
\item $g_{t,t}^{*}(x)-x=0$ for each $0\leq t\leq 1$, and
\item $g_{t,s}^{*}(x)=g_{r,s}^{*}\circ g_{t,r}^{*}(x)$ for each $0\leq t\leq r\leq s\leq 1$.
\end{enumerate}
\end{proposition}

\par Observe that the flow map $g_{0,1}^{*}$ satisfies the normalizing equation~\eqref{eq:normalizing}, and for each fixed $x_{t}$, $\{g_{t,s}(x_{t})\}_{s\geq t}$ is a part of the characteristic curve. Consequently, the generative learning can be reduced to the problem of fitting the characteristic curves by minimizing the following quadratic risk:
\begin{equation}\label{eq:characteristic:LS}
\calR(g)\coloneqq\frac{2}{T^{2}}\int_{0}^{T}\int_{t}^{T}\bbE_{Z_{0}\sim\mu_{0}}\|Z_{s}-g(t,s,Z_{t})\|_{2}^{2}\dsdt,
\end{equation}
where $Z_{t}=g_{0,t}^{*}(Z_{0})$, $Z_{s}=g_{0,s}^{*}(Z_{0})$ and $T\approx 1$ is a pre-specified early-stopping time. The primary goal of our method, ``characteristic learning'', is dedicated to estimate the flow map $g_{t,s}^{*}$ specified by the characteristic curves.

\par Given that the distributions of $Z_{t}$ and $Z_{s}$ in~\eqref{eq:characteristic:LS} are unknown, it is necessary to estimate them prior to minimizing~\eqref{eq:characteristic:LS}. To achieve this, the denoiser is initially estimated in Section~\ref{section:velocity}, followed by the utilization of EI to numerically solve the probability flow ODE in Section~\ref{section:Euler}. The resulting numerical solutions provide approximations of $(Z_{t},Z_{s})$, which are then utilized to approximate the population risk~\eqref{eq:characteristic:LS} in Section~\ref{section:characteristic}.

\subsection{Denoiser Matching}\label{section:velocity}
Substituting~\eqref{eq:interpolant} into~\eqref{eq:velocity}, we have
\begin{equation}\label{eq:b:denoiser}
b^{*}(t,x)=\frac{\dot{\alpha}_{t}}{\alpha_{t}}x-\beta_{t}\Bigl(\frac{\dot{\alpha}_{t}}{\alpha_{t}}-\frac{\dot{\beta}_{t}}{\beta_{t}}\Bigr)D^{*}(t,x),
\end{equation}
where $D^{*}:[0,1]\times\mathbb{R}^{d}\to\mathbb{R}^{d}$ is the denoiser defined as
\begin{equation}\label{eq:denoiser}
D^{*}(t,x) \coloneqq \mathbb{E}[X_{1} \mid X_{t}=x].
\end{equation}
Motivated by the conditional expectation structure of the denoiser~\eqref{eq:denoiser}, we propose the following population risk
\begin{equation}\label{eq:velocity:popu:risk}
\calL(D)=\frac{1}{T}\int_{0}^{T}\bbE\bigl[\|X_{1}-D(t,\alpha_{t}X_{0}+\beta_{t}X_{1})\|_{2}^{2}\bigr]\dt,
\end{equation}
where $D:[0,1]\times\mathbb{R}^{d}\to\mathbb{R}^{d}$ is a measurable function, and the expectation is taken with respect to $(X_{0},X_{1})\sim\mu_{0}\times\mu_{1}$. It is straightforward that the denoiser $D^{*}$ in~\eqref{eq:denoiser} is the unique minimizer of the population risk~\eqref{eq:velocity:popu:risk}.

\par However, the expectation in~\eqref{eq:velocity:popu:risk} does not admit a closed-form, and one needs to approximate it through Monte Carlo methods. Let $\{(X_{0}^{i},X_{1}^{i})\}_{i=1}^{n}$ be a set of independent random copies of $(X_{0},X_{1})\sim\mu_{0}\times\mu_{1}$. Additionally, let $\{t^{i}\}_{i=1}^{n}$ be a set of $n$ i.i.d. random variables drawn from the uniform distribution on $(0,T)$. Denote by $\euS=\{(t^{i},X_{0}^{i},X_{1}^{i})\}_{i=1}^{n}$. Then the corresponding empirical risk of a measurable function $D:[0,1]\times\mathbb{R}^{d}\to\mathbb{R}^{d}$ is defined as
\begin{equation}\label{eq:velocity:er}
\what{\calL}_{n}(D)=\frac{1}{n}\sum_{i=1}^{n}\|X_{1}^{i}-D(t^{i},\alpha_{t^{i}}X_{0}^{i}+\beta_{t^{i}}X_{1}^{i})\|_{2}^{2}.
\end{equation}
The empirical risk minimizer then reads
\begin{equation}\label{eq:velocity:erm}
\what{D}\in\argmin_{D\in\scrD}\what{\calL}_{n}(D),
\end{equation}
where $\scrD$ is a deep neural network class. This approach is known as denoiser matching, which is mathematically equivalent to the velocity/score matching, but with better numerical stability~\cite{karras2022elucidating,kim2024consistency}. In the context of distillation for diffusion models, the denoiser network $\what{D}$ is referred to as the ``teacher'' network.

\subsection{Exponential Integrator Sampling}\label{section:Euler}
In this section, we turn to sample from the estimated probability flow equation. Substituting the semi-linear representation of the velocity~\eqref{eq:b:denoiser} into the probability flow ODE~\eqref{eq:characteristic:flow} yields a semi-linear ODE
\begin{equation}\label{eq:characteristic:semilinear}
\frac{\mathrm{d}}{\mathrm{d}s}g_{t,s}(x_{t}) = \frac{\dot{\alpha}_{s}}{\alpha_{s}}\,g_{t,s}(x_{t}) - \beta_{s}\Bigl(\frac{\dot{\alpha}_{s}}{\alpha_{s}}-\frac{\dot{\beta}_{s}}{\beta_{s}}\Bigr) D_{s}^{*}(g_{t,s}(x_{t})),
\end{equation}
subject to $g_{t,t}(x_{t}) = x_{t}$ for $0 \le t \le s \le T$.

Let $K\in\mathbb{N}_{+}$ denote the total number of steps, so the step size is $h \coloneqq T/K$, and let $\{t_{k} = kh\}_{k=0}^{K}$ be the corresponding grid of time points. We approximate the semi-linear ODE~\eqref{eq:characteristic:semilinear} on each subinterval $kh \le s \le (k+1)h$ through two approximations to the denoiser term $D_{s}^{*}(g_{t,s}(x_{t}))$. First, we replace the unknown ground-truth denoiser $D_{s}^{*}$ by a learned estimator $\widehat{D}_{s}$ in~\eqref{eq:velocity:erm}. Second, we freeze its time index and argument at the left endpoint of the subinterval, evaluating it as $\widehat{D}_{kh}(x_{kh})$ rather than $\widehat{D}_{s}(x_{s})$. Together, these two approximations yield an EI scheme~\cite{Hochbruck2010Exponential}
\begin{equation}\label{eq:characteristic:ei}
\frac{\mathrm{d}}{\mathrm{d}s}\widehat{E}_{kh,s}(x_{kh}) = \frac{\dot{\alpha}_{s}}{\alpha_{s}}\,\widehat{E}_{kh,s}(x_{kh})-\beta_{s}\Bigl(\frac{\dot{\alpha}_{s}}{\alpha_{s}}-\frac{\dot{\beta}_{s}}{\beta_{s}}\Bigr)\widehat{D}_{kh}(\widehat{E}_{kh,kh}(x_{kh})), \nonumber
\end{equation}
where $\widehat{E}_{kh,kh}(x_{kh})=x_{kh}$ and $s\in(kh,(k+1)h)$. Since the right-hand side is now linear with respect to $\widehat{E}_{kh,s}(x_{kh})$, the ODE admits a closed-form solution
\begin{equation}\label{eq:characteristic:ei:solution}
\widehat{E}_{kh,s}(x_{kh}) = \frac{\alpha_{s}}{\alpha_{kh}}x_{kh}-\beta_{kh}\Bigl(\frac{\alpha_{s}}{\alpha_{kh}}-\frac{\beta_{s}}{\beta_{kh}}\Bigr)\widehat{D}_{kh}(x_{kh}).
\end{equation}
See Appendix~\ref{section:proof:discretization:ei} for detailed derivations of EI. Note that EI~\eqref{eq:characteristic:ei} only freezes the nonlinear term at the left endpoint of the subinterval and integrates the linear term exactly. In contrast, forward Euler method introduces discretization error to both linear and nonlinear terms. The EI~\eqref{eq:characteristic:ei} has been utilized in sampling of diffusion models by~\cite{zhang2023fast,lu2022dpm,lu2023dpmsolver,Zheng2023dpmsolverv3}. The resulting flow $\widehat{E}_{kh,s}$ inherits the semigroup property of the ground-truth flow $g_{t,s}^{*}$ in~\eqref{eq:characteristic:flow}.

The EI flow $\widehat{E}_{0,s}$ approximates the ground-truth flow $g_{0,s}^{*}$, with the error bound established in Theorem~\ref{theorem:error:euler}. EI sampling is nonetheless computationally expensive, as it requires a large number of denoiser-network evaluations, which makes it ill-suited to time-sensitive applications. This motivates the development of an efficient, simulation-free approach for evaluating the flow $g_{0,s}^{*}$.

\subsection{Characteristic Fitting}\label{section:characteristic}

\par In order to estimate the flow $g_{t,s}^{*}$ via a simulation-free approach, we leverage a deep neural network to fit characteristics using data samples obtained by the EI flow~\eqref{eq:characteristic:ei}. Let $\euZ=\{\what{Z}_{0}^{i}\}_{i=1}^{m}$ be a set of $m$ random copies of $\what{Z}_{0}\sim\mu_{0}$. Further, one obtains $m$ discrete characteristics $\{(\what{Z}_{k}^{i})_{k=0}^{K}\}_{i=1}^{m}$ by the EI~\eqref{eq:characteristic:ei} with $\what{Z}_{k}^{i}=\what{E}_{0,kh}(\what{Z}_{0}^{i})$. An empirical counterpart of~\eqref{eq:characteristic:LS} is then defined as
\begin{equation}\label{eq:charateristic:er}
\what{\calR}_{m}^{\ei}(g)=\frac{2}{mK^{2}}\sum_{i=1}^{m}\sum_{k=0}^{K-1}\Bigg\{\frac{1}{2}\|\what{Z}_{k}^{i}-g(kh,kh,\what{Z}_{k}^{i})\|_{2}^{2}+\sum_{\ell=k+1}^{K-1}\|\what{Z}_{\ell}^{i}-g(kh,\ell h,\what{Z}_{k}^{i})\|_{2}^{2}\Bigg\}.
\end{equation}
The characteristic generator can be obtained by minimizing this empirical risk
\begin{equation}\label{eq:charateristic:erm}
\what{g}\in\argmin_{g\in\scrG}\what{\calR}_{m}^{\ei}(g),
\end{equation}
where $\scrG$ is a set of deep neural networks. The training procedure of the characteristic generator is summarized as Algorithm~\ref{alg:characteristic}.

\begin{algorithm}
\caption{Training procedure of characteristic generator.}
\label{alg:characteristic}
\begin{algorithmic}[1]
\Require {Denoiser estimator $\what{D}\approx D^{*}$.}
\State {\texttt{\# Exponential integrator sampling}}
\For{$i=1,\ldots,m$}
\State {Initialize a sample: $\what{Z}_{0}^{i}\sim\mu_{0}=\mathcal{N}(0,I_{d})$.}
\For{$k=0,\ldots,K-1$}
\State {$\what{Z}_{k+1}^{i}\leftarrow\what{E}_{kh,(k+1)h}(\what{Z}_{k}^{i})$ via~\eqref{eq:characteristic:ei:solution}.}
\EndFor
\EndFor
\State {\texttt{\# Characteristic fitting}}
\State {Initialize a network $g_{\theta}:\bbR\times\bbR\times\bbR^{d}\rightarrow\bbR^{d}$.}
\Repeat
\State {$\theta\leftarrow\theta-\alpha\nabla_{\theta}\what{\calR}_{m}^{\ei}(g_{\theta})$, where $\alpha$ is the step size.}
\Until {converged}
\Ensure {Characteristic generator $\what{g}\coloneqq g_{\theta}$.}
\end{algorithmic}
\end{algorithm}

\par Note that the estimator $\what{g}_{0,s}:\mathbb{R}^{d}\to\mathbb{R}^{d}$ serves as a neural network approximation to flow map $g_{0,s}^{*}$~\eqref{eq:characteristic:flow}, eliminating the need to simulate the ODE. The induced one-step generation algorithm is summarized in Algorithm~\ref{alg:characteristic:sampling}.

\begin{algorithm}
\caption{One-step sampling via CG.}
\label{alg:characteristic:sampling}
\begin{algorithmic}[1]
\Require {Characteristic generator $\what{g}\approx g^{*}$.}
\State {Initialize a sample $\what{Z}_{0}^{\mathrm{CG}}\sim\mu_{0}=\mathcal{N}(0,I_{d})$.}
\State {$\what{Z}^{\mathrm{CG}}\leftarrow\what{g}_{0,Kh}(\what{Z}_{0}^{\mathrm{CG}})$.}
\Ensure {Generated sample $\what{Z}^{\mathrm{CG}}$.}
\end{algorithmic}
\end{algorithm}

\par The characteristic generator is not restricted to one-step generation as Algorithm~\ref{alg:characteristic:sampling}. Indeed, it can be employed iteratively to produce refined sampling algorithms, thereby enhancing the quality of generation, albeit with a slight increase in computational cost. Algorithm~\ref{alg:characteristic:sampling:fine} presents the comprehensive procedure for achieving this fine-grained sampling.

\begin{algorithm}
\caption{Fine-grained sampling via CG.}
\label{alg:characteristic:sampling:fine}
\begin{algorithmic}[1]
\Require {Characteristic generator $\what{g}\approx g^{*}$.}
\State {Sample initial value $\what{Z}_{0}^{\mathrm{CG}}\sim\mu_{0}=\mathcal{N}(0,I_{d})$.}
\State {Choose a sequence of time points $0=t_{0}<\cdots<t_{L}=T$.}
\For {$\ell=0,\ldots,L-1$}
\State {$\what{Z}_{\ell+1}^{\mathrm{CG}}\leftarrow\what{g}_{t_{\ell},t_{\ell+1}}(\what{Z}_{\ell}^{\mathrm{CG}})$.}
\EndFor
\Ensure {Generated sample $\what{Z}_{L}^{\mathrm{CG}}$.}
\end{algorithmic}
\end{algorithm}

\par Given a trained denoiser estimator, a comparison between two sampling approaches: EI~\eqref{eq:characteristic:ei} and characteristic generator~\eqref{eq:charateristic:erm} is shown in Table~\ref{table:ei:cg}.

\begin{table}[htbp]
\caption{Comparison between EI and the characteristic generator.}
\centering
\small
\begin{tabular}{lccc}
\toprule
& & \textbf{Training} & \textbf{Sampling} \\
\midrule
Exponential integrator  & \eqref{eq:characteristic:ei}  & Not required & High NFE \\
Characteristic generator & \eqref{eq:charateristic:erm} & Required     & One or few NFE \\
\bottomrule
\end{tabular}
\label{table:ei:cg}
\end{table}

\par During the sampling phase, one only needs to evaluate the characteristic generator once or a few times, in contrast to the EI sampler, which requires a large number of neural network (denoiser) evaluations. Consequently, our approach significantly reduces computational cost and accelerates sampling compared to EI-based methods. The trade-off, however, lies in the training phase: as outlined in Algorithm~\ref{alg:characteristic}, fitting the characteristic generator requires a substantial number of ODE simulations. Nevertheless, the benefits outweigh this cost. In practice, the ODE simulations and the fitting of the characteristic generator can be carried out offline on large-scale, high-performance computing platforms, after which the resulting estimator $\widehat{g}$ can be deployed in computationally constrained and latency-sensitive application scenarios.

% =================================================
\section{Convergence Rates Analysis}\label{section:analysis}

\par In this section, we present a comprehensive convergence rate analysis for the characteristic generator. We begin by illustrating Assumptions~\ref{assumption:init:dist:gaussian} and~\ref{assumption:target:dist} in Section~\ref{section:analysis:assumptions}, and then establish regularity properties of the probability flow ODE~\eqref{eq:characteristic} in Section~\ref{section:analysis:property}. In Section~\ref{section:analysis:velocity} and Section~\ref{section:analysis:euler}, we propose a non-asymptotic error analysis for denoiser matching and EI sampling, respectively. In Section~\ref{section:analysis:characteristic}, a convergence rate analysis for the characteristic generator is established. Finally, in Section~\ref{section:analysis:applications}, we apply the aforementioned analysis to two widely-used types of probability flow ODE: linear interpolant and F{\"o}llmer flow.

\subsection{Discussions of Assumptions}\label{section:analysis:assumptions}

\subsubsection{Assumption~\ref{assumption:init:dist:gaussian}}

\par Assumption~\ref{assumption:init:dist:gaussian} is standard and commonly used in flow-based generative models~\cite{liu2022flow,lipman2023flow,albergo2023building}. The following proposition demonstrates that the velocity field $b^{*}(t,x)$ and the denoiser $D^{*}(t,x)$ are both spatial linear combinations of $x$ and the score function $s^{*}(t,x)=\nabla\log\rho_{t}(x)$. This connection suggests that stochastic interpolants that satisfy Assumption~\ref{assumption:init:dist:gaussian} are closely related to score-based diffusion models~\cite{Ho2020Denoising,song2021score}.

\begin{proposition}[Velocity, denoiser, and score]\label{proposition:velocity:score}
Suppose Assumption~\ref{assumption:init:dist:gaussian} holds. Then for any $(t,x)\in(0,1)\times\bbR^{d}$,
\begin{align*}
b^{*}(t,x)&=\frac{\dot{\beta}_{t}}{\beta_{t}}x+\alpha_{t}^{2}\Big(\frac{\dot{\beta}_{t}}{\beta_{t}}-\frac{\dot{\alpha}_{t}}{\alpha_{t}}\Big)s^{*}(t,x), \\
D^{*}(t,x)&=\frac{1}{\beta_{t}}x+\frac{\alpha_{t}^{2}}{\beta_{t}}s^{*}(t,x).
\end{align*}
\end{proposition}

\subsubsection{Assumption~\ref{assumption:target:dist}}
\par Assumption~\ref{assumption:target:dist} requires the target distribution to be a Gaussian convolution of a compactly supported distribution. Although raw image distributions have compact support, a deterministic flow map transporting a Gaussian prior toward such a target may lack regularity, as discussed in the following remark.

\begin{remark}\label{remark:compact}
Recall that there exists a unique flow map $g_{t,s}^{*}:\mathbb{R}^{d}\to\mathbb{R}^{d}$ of the probability flow ODE~\eqref{eq:characteristic}, provided that~\eqref{eq:characteristic:well} is satisfied. Since $g_{t,s}^{*}$ is a diffeomorphism, it is invertible: for any $0\leq t\leq s\leq 1$, there exists $g_{s,t}^{*}:\mathbb{R}^{d}\to\mathbb{R}^{d}$ such that $g_{s,t}^{*}\circ g_{t,s}^{*} = \mathrm{id}$. For any $z\in\mathbb{R}^{d}$, integrating the ODE backward from $1$ to $0$ yields $x \coloneqq g_{1,0}^{*}(z)\in\mathbb{R}^{d}$, and integrating forward from $0$ to $1$ starting at $x$ recovers $g_{0,1}^{*}\circ g_{1,0}^{*}(z)=z$. Consequently, for any $z \in \mathbb{R}^{d}$ there exists $x\coloneqq g_{1,0}(z)\in\mathbb{R}^{d}$ such that $g_{0,1}(x)=z$, which implies that no well-posed ODE can transport a Gaussian distribution (supported on the entire space $\mathbb{R}^{d}$) to a distribution supported on a proper compact subset of $\mathbb{R}^{d}$.
\end{remark}

\par By instead imposing Gaussian smoothing on the target, the resulting probability flow admits strong regularity properties without further assumptions, as established in Section~\ref{section:analysis:property}.

\par Moreover, Assumption~\ref{assumption:target:dist} accommodates a broad class of target distributions, including Gaussian mixtures~\cite[Appendix~C]{Grenioux2024Stochastic}. In particular, the Gaussian-smoothed distribution $\mathcal{N}(0,\sigma^{2}I_{d})\ast\nu$ approximates the original distribution $\nu$ arbitrarily well for sufficiently small $\sigma>0$, making Assumption~\ref{assumption:target:dist} broadly applicable, including image distributions.

\subsection{Regularity Properties of the Probability Flow ODE}\label{section:analysis:property}

\par In this section, we establish the regularity properties that underpin our convergence analysis. We begin with the regularity of the velocity field $b^{*}$ defined in~\eqref{eq:velocity}.

\begin{proposition}[Regularity of the velocity field]\label{proposition:reg:velocity}\label{proposition:bound:velocity}\label{proposition:lip:spatial:velocity}
Suppose Assumptions~\ref{assumption:init:dist:gaussian} and~\ref{assumption:target:dist} hold. Let $R\in(1,+\infty)$. Then the velocity field $b^{*}$ satisfies the following.
\begin{enumerate}[(a)]
\item For any $(t,x)\in[0,1]\times\bbB_{R}^{\infty}$,
\begin{equation*}
\max_{1\leq k\leq d}|b_{k}^{*}(t,x)|\leq B_{\vel}R.
\end{equation*}
\item For any $t\in[0,1]$ and $x,x^{\prime}\in\bbR^{d}$,
\begin{align*}
\|b^{*}(t,x)-b^{*}(t,x^{\prime})\|_{2}&\leq G_{\rm vel}\|x-x^{\prime}\|_{2}.
\end{align*}
\end{enumerate}
Here the constants $B_{\vel}$ and $G_{\rm vel}$ depend only on $d$ and $\sigma$.
\end{proposition}

\par Proposition~\ref{proposition:reg:velocity} implies~\eqref{eq:characteristic:well} directly, ensuring the uniqueness of the solution of probability flow ODE~\eqref{eq:characteristic}. We next turn to the denoiser $D^{*}$ defined in~\eqref{eq:denoiser}. The following proposition records its boundedness, spatial regularity, and time regularity, all uniform in $t$, which underlie the neural network estimation of $D^{*}$.

\begin{proposition}[Regularity of the denoiser]\label{proposition:reg:denoiser}
Suppose Assumptions~\ref{assumption:init:dist:gaussian} and~\ref{assumption:target:dist} hold. Let $R\in(1,+\infty)$. Then the denoiser $D^{*}$ satisfies the following.
\begin{enumerate}[(a)]
\item For any $(t,x)\in[0,1]\times\bbB_{R}^{\infty}$,
\begin{equation*}
\max_{1\leq k\leq d}|D_{k}^{*}(t,x)|\leq B_{\rm deno}R.
\end{equation*}
\item For any $(t,x)\in[0,1]\times\bbR^{d}$, $\|\nabla D^{*}(t,x)\|_{\op}\leq G_{\rm deno}(t)$, where the spatial envelope is
\begin{align}
G_{\rm deno}(t)
&\coloneqq G_{\rm deno}^{0}(t)+G_{\rm deno}^{1}(t), \quad \text{where}\label{eq:G:deno} \\
G_{\rm deno}^{0}(t) &\coloneqq \frac{\sigma^{2}\beta_{t}}{\alpha_{t}^{2}+\sigma^{2}\beta_{t}^{2}}, \quad G_{\rm deno}^{1}(t) \coloneqq \frac{d\,\alpha_{t}^{2}\beta_{t}}{(\alpha_{t}^{2}+\sigma^{2}\beta_{t}^{2})^{2}}. \nonumber
\end{align}
\item For any $(t,x)\in[0,1]\times\bbB_{R}^{\infty}$,
\begin{equation*}
\|\partial_{t}D^{*}(t,x)\|_{2}\leq D_{\rm deno}R.
\end{equation*}
\end{enumerate}
Here the constants $B_{\rm deno}$ and $D_{\rm deno}$ depend only on $d$ and $\sigma$.
\end{proposition}

\par The spatial envelope $G_{\rm deno}(t)$ in~\eqref{eq:G:deno} is uniformly bounded over $[0,1]$ with $G_{\rm deno}(1)=1$, reflecting that the denoiser degenerates to the identity map as $t\rightarrow1$. This property guarantees the stability of EI, as discussed after Theorem~\ref{theorem:error:euler}.

\par Finally, the regularity of the velocity field propagates to the flow map $g^{*}$, yielding its boundedness, the uniform bound on its spatial gradient, and the bounds on its time derivatives, which together are required to establish the approximation of the characteristic generator.

\begin{proposition}[Regularity of the flow map]\label{proposition:reg:flow}\label{corollary:bound:solution}\label{corollary:lip:time:solution}\label{proposition:lip:spatial:flow}
Suppose Assumptions~\ref{assumption:init:dist:gaussian} and~\ref{assumption:target:dist} hold. Let $R\in(1,+\infty)$. Then for each $0\leq t\leq s\leq 1$, the flow map $g^{*}$ satisfies the following.
\begin{enumerate}[(a)]
\item For any $x\in\bbB_{R}^{\infty}$,
$\max_{1\leq k\leq d}|g_{k}^{*}(t,s,x)|\leq B_{\flow}R$.
\item For any $x\in\bbR^{d}$,
$\|\nabla g^{*}(t,s,x)\|_{\op}\leq\exp(G_{\rm vel}(s-t))$.
\item For any $x\in\bbB_{R}^{\infty}$,
\begin{equation*}
\max\Big\{\|\partial_{t}g^{*}(t,s,x)\|_{2},\|\partial_{s}g^{*}(t,s,x)\|_{2}\Big\}\leq B^{\prime}_{\flow}R.
\end{equation*}
\end{enumerate}
Here the constants $B_{\flow}$ and $B^{\prime}_{\flow}$ depend only on $d$ and $\sigma$.
\end{proposition}

\subsection{Analysis for Denoiser Matching}\label{section:analysis:velocity}

\par In this section, we focus on the time-averaged $L^{2}$-error of the denoiser estimator $\what{D}$ in~\eqref{eq:velocity:erm}, that is,
\begin{equation}\label{eq:measure:velocity}
\calE_{T}(\what{D})=\frac{1}{T}\int_{0}^{T}\bbE_{X_{t}}\Big[\|D^{*}(t,X_{t})-\what{D}(t,X_{t})\|_{2}^{2}\Big]\dt.
\end{equation}
The majority of existing literature on theoretical analysis of diffusion and flow-based generative models commonly assumes that the $L^{2}$-risk of score or velocity matching is sufficiently small~\cite{Lee2022Convergence,Lee2023Convergence,chen2023sampling,chen2023probability,benton2024linear,benton2024error,gao2024convergence,beyler2025convergence}. However, this line of research lacks the ability to quantitatively characterize the statistical convergence rate of velocity/score/denoiser matching with respect to the number of samples. Additionally, no prior theoretical guidance for the selection of neural networks is provided in this literature.

\par In this work, we aim to establish a non-asymptotic error bound for the $L^{2}$-risk of the estimated denoiser. The main result is stated as follows.

\begin{theorem}[Convergence rate for denoiser matching]\label{theorem:velocity:rate}
Suppose Assumptions~\ref{assumption:init:dist:gaussian} and~\ref{assumption:target:dist} hold. Let $T\in(1/2,1)$ and $R\in(1,+\infty)$. Set the hypothesis class $\scrD$ as a deep neural network class, which is defined as
\begin{equation*}
\scrD=\left\{D\in N(L,S):
\begin{aligned}
&\max_{1\leq k\leq d}|D_{k}(t,x)|\leq B_{\rm deno}R, \\
&\|\nabla D\|_{\op}\leq \bar{G}_{\rm deno}^{0}(t)+3\bar{G}_{\rm deno}^{1}(t), \\
&\|\partial_{t}D(t,x)\|_{2}\leq3D_{\rm deno}R
\end{aligned}
\right\},
\end{equation*}
where the bounds hold for $(t,x)\in[0,T]\times\bbB_{R}^{\infty}$, the constants $B_{\rm deno}$, $D_{\rm deno}$, $\bar{G}_{\rm deno}^{0}(t)$ and $\bar{G}_{\rm deno}^{1}(t)$ are as in Proposition~\ref{proposition:reg:denoiser}. The depth and the number of non-zero weights of the neural network are given, respectively, as $L=C$ and $S=Cn^{\frac{d+1}{d+3}}$. Then the corresponding estimator $\what{D}$ satisfies
\begin{align*}
\bbE_{\euS}\big[\calE_{T}(\what{D})\big]\leq Cn^{-\frac{2}{d+3}}\log^{2}n,
\end{align*}
where the constant $C$ only depends on $d$ and $\sigma$.
\end{theorem}

\par The rate of denoiser matching in Theorem~\ref{theorem:velocity:rate} is consistent with the minimax optimal rate $\calO(n^{-\frac{2}{d+3}})$ in nonparametric regression~\cite{Stone1982optimal,Yang1999Information,gyorfi2002distribution, Tsybakov2009Introduction} given that the $(d+1)$-dimensional target function is Lipschitz continuous. Moreover, our theoretical findings align with convergence rates of nonparametric regression using deep neural networks~\cite{bauer2019deep, nakada2020Adaptive, schmidt2020nonparametric, kohler2021rate, Farrell2021Deep, Kohler2022Estimation, Jiao2023deep}.

\begin{remark}\label{remark:Lipschitz}
Note that $\|\nabla D(t,x)\|_{\op}\leq \bar{G}_{\rm deno}(t) \coloneqq \bar{G}_{\rm deno}^{0}(t)+3\bar{G}_{\rm deno}^{1}(t)$ for any $x\in\mathbb{R}^{d}$ and $D\in\scrD$, which mirrors the envelope~\eqref{eq:G:deno} and is exactly the stability condition required by EI: the scheme multiplies per-step amplification factors governed by the spatial Lipschitz constant of the denoiser, over a time grid whose accumulated effective step size diverges as $T\rightarrow1$. The resulting amplification remains bounded under the envelope, but not under a uniform bound alone. In practical applications, various techniques for restricting the Lipschitz constant of deep neural networks have been proposed, such as weight clipping~\cite{Arjovsky2017Wasserstein}, gradient penalty~\cite{Gulrajani2017Improved},  spectral normalization~\cite{miyato2018spectral}, and Lipschitz network~\cite{Zhang2022Rethinking}. In the theoretical perspective, the approximation properties of deep neural network with Lipschitz constraint has been studied by~\cite{chen2022distribution,Huang2022Error,Jiao2023Approximation,ding2024semisupervised}. In this work, an approximation error bound for deep neural networks with Lipschitz constraint is established in supplementary materials.
\end{remark}

\subsection{Analysis for Exponential Integrator Sampling}\label{section:analysis:euler}

\par The main objective of this section is to estimate the Wasserstein-2 error of the EI flow~\eqref{eq:characteristic:ei:solution}.

\begin{theorem}[Error analysis for EI sampling]\label{theorem:error:euler}
Under the same conditions as Theorem~\ref{theorem:velocity:rate}. Let $K\in\bbN_{+}$ be the number of time steps of EI~\eqref{eq:characteristic:ei:solution} with step size $h=T/K$, and define
\begin{equation}\label{eq:Phi:T}
\Phi_{T}\coloneqq\int_{0}^{T}\Big(\dot{\beta}_{t}-\frac{\dot{\alpha}_{t}}{\alpha_{t}}\beta_{t}\Big)^{2}\dt.
\end{equation}
Then the following inequality holds
\begin{equation*}
\bbE_{\euS}\Big[W_{2}^{2}\Big((\what{E}_{0,T})_{\sharp}\mu_{0},\mu_{T}\Big)\Big]
\leq C\Phi_{T}\Big\{n^{-\frac{2}{d+3}}\log^{2}n+\frac{1}{K}\Big\},
\end{equation*}
where the constant $C$ depends only on $d$ and $\sigma$. Further, if $K\geq Cn^{\frac{2}{d+3}}$, then it follows that
\begin{equation*}
\bbE_{\euS}\Big[W_{2}^{2}\Big((\what{E}_{0,T})_{\sharp}\mu_{0},\mu_{T}\Big)\Big]\leq C\Phi_{T}n^{-\frac{2}{d+3}}\log^{2}n.
\end{equation*}
\end{theorem}

\par The first term in the error bound of Theorem~\ref{theorem:error:euler} arises from denoiser matching, while the second term stems from time-discretization. The stopping time $T$ enters only through the coefficient functional $\Phi_{T}$, which quantifies the accumulated weight of the matching error along EI recursion. In particular, the per-step amplification of the recursion remains bounded uniformly in $T\in(0,1)$, owing to the time-dependent Lipschitz envelope of the hypothesis class $\scrD$ discussed in Remark~\ref{remark:Lipschitz}. A uniform Lipschitz bound alone would instead incur an amplification factor diverging as $T\rightarrow1$.

\par Based on Theorem~\ref{theorem:error:euler}, we can derive a Wasserstein-2 error bound between the target distribution and the push-forward distribution of $\mu_{0}$ by the EI flow $\what{E}_{0,T}$ as follows.

\begin{corollary}\label{corollary:error:euler}
Under the same conditions as Theorem~\ref{theorem:error:euler}. Suppose the number of time steps satisfies $K\geq Cn^{\frac{2}{d+3}}$. Then the following inequality holds
\begin{equation*}
\bbE_{\euS}\Big[W_{2}^{2}\Big((\what{E}_{0,T})_{\sharp}\mu_{0},\mu_{1}\Big)\Big]
\leq C\Phi_{T}n^{-\frac{2}{d+3}}\log^{2}(n)
+2\max\{\alpha_{T}^{2},(1-\beta_{T})^{2}\}W_{2}^{2}(\mu_{0},\mu_{1}),
\end{equation*}
where the constant $C$ depends only on $d$ and $\sigma$.
\end{corollary}

\par Corollary~\ref{corollary:error:euler} presents an error bound for flow-based generative models. The first term in the error bound corresponds to Wasserstein-2 error of the EI flow, as derived in Theorem~\ref{theorem:error:euler}. The second term captures the convergence of the interpolant distribution $\mu_{T}$ to the target distribution $\mu_{1}$. It is worth noting that as the stopping time $T$ approaches one, the first term tends to infinity, while the second term simultaneously decreases. This trade-off within the error bound highlights the importance of carefully selecting the stopping time $T$ and provides practical guidance for determining it in real-world applications. Since the optimal early-stopping time depends on the specific choice of the interpolant coefficients, $\alpha_{t}$ and $\beta_{t}$, we derive the optimal early-stopping time $T$ for two important cases: linear interpolant and F{\"o}llmer flow in Corollaries~\ref{corollary:rate:linear} and~\ref{corollary:rate:follmer}, respectively.

\subsection{Analysis for Characteristic Generator}\label{section:analysis:characteristic}

\par Despite the empirical success of simulation-free one-step approaches for the efficient sampling of flow-based generative models~\cite{salimans2022progressive,Song2023Consistency,Zheng2023Fast,kim2024consistency,ren2024hypersd}, the theoretical analysis for these line of methods still remains unclear. In this section, we establish a thorough analysis for the characteristic generator.

\par To measure the error of the characteristic generator, we focus on the time-average squared Wasserstein-2 distance between the target distribution and the push-forward distribution by the characteristic generator $\what{g}$:
\begin{equation}\label{eq:measure:characteristic}
\calD(\what{g})\coloneqq\frac{2}{T^{2}}\int_{0}^{T}\int_{t}^{T}W_{2}^{2}\Big((\what{g}_{t,s})_{\sharp}\mu_{t},\mu_{s}\Big)\dsdt.
\end{equation}
The main result is stated as follows.

\begin{theorem}[Error analysis for characteristic generator]\label{theorem:error:characteristic}
Under the same conditions as Theorem~\ref{theorem:error:euler}. Further, set the hypothesis class $\scrG$ as a deep neural network class, which is defined as
\begin{equation*}
\scrG=\left\{g\in N(L,S):
\begin{aligned}
&\|g(t,s,x)\|_{\infty}\leq B_{\flow}\log^{1/2}m, \\
&\|\partial_{s}g(t,s,x)\|_{2}\leq 6\sqrt{d}{ B^{\prime}_{\flow}}\log^{1/2}m, \\
&\|\partial_{t}g(t,s,x)\|_{2}\leq 6\sqrt{d}{ B^{\prime}_{\flow}}\log^{1/2}m, \\
&\|\nabla g(t,s,x)\|_{\op}\leq 3d\exp(G_{\rm vel}T)
\end{aligned}
\right\},
\end{equation*}
where the constants $B_{\rm flow}$, $B_{\rm flow}^{\prime}$ and $G_{\rm vel}$ are as in Proposition~\ref{proposition:reg:flow}. The depth and the number of non-zero weights of the neural network are given, respectively, as $L=C$ and $S=Cm^{\frac{d+2}{d+4}}$. Then it follows that
\begin{equation*}
\bbE_{\euS}\bbE_{\euZ}\big[\calD(\what{g})\big]\leq C\Phi_{T}\Big\{n^{-\frac{2}{d+3}}\log^{2}n+\frac{1}{K}\Big\}
+C\Big\{m^{-\frac{2}{d+4}}\log^{2}m+\frac{\log m}{K}\Big\},
\end{equation*}
where $\Phi_{T}$ is the coefficient functional~\eqref{eq:Phi:T}, and the constant $C$ depends only on $d$ and $\sigma$. Furthermore, if the number of time steps $K$ for EI and the number of samples $m$ for characteristic fitting satisfy
\begin{equation}\label{eq:theorem:error:characteristic:1}
K\geq Cn^{\frac{2}{d+3}}\log m, \quad
m\geq Cn^{\frac{d+4}{d+3}},
\end{equation}
respectively, then the following inequality holds
\begin{equation*}
\bbE_{\euS}\bbE_{\euZ}\big[\calD(\what{g})\big]\leq C\Phi_{T}n^{-\frac{2}{d+3}}\log^{2}n.
\end{equation*}
\end{theorem}

\par In contrast to the error bound of EI sampling in Theorem~\ref{theorem:error:euler}, the error bound of the characteristic generator in Theorem~\ref{theorem:error:characteristic} incorporates an additional error term $\calO(m^{-\frac{2}{d+4}})$. This error term corresponds to the error of the standard nonparametric regression for characteristic fitting, attaining the minimax optimality~\cite{Stone1982optimal,Yang1999Information,gyorfi2002distribution,Tsybakov2009Introduction} given that $g^{*}$ is $(d+1)$-dimensional Lipschitz continuous function.

\par It is noteworthy that the number of samples $m$ for characteristic fitting can be arbitrarily large because training samples can be generated from copies of $Z_{0}\sim\mu_{0}$ using EI~\eqref{eq:characteristic:ei:solution}. Without loss of generality, we consider the case where $m\gg n$. In this scenario, the error bound in Theorem~\ref{theorem:error:characteristic} aligns with the convergence rate in Theorem~\ref{theorem:error:euler}.

\par In the context of distillation~\cite{salimans2022progressive,Song2023Consistency,kim2024consistency}, the EI flow~\eqref{eq:characteristic:ei:solution} is commonly referred as as the ``teacher'' model. Theorem~\ref{theorem:error:characteristic} guarantees that, when the number $m$ of ``teacher'' samples is sufficiently large, the characteristic generator can generate new samples that are as good as those generated by the ``teacher'' model. However, the theorem also highlights that the ``teacher'' model serves as a bottleneck for the characteristic generator, as it cannot surpass the generative quality of the ``teacher'' model. One potential approach to overcome this bottleneck is by combining these one-step generative models with GANs, as demonstrated by~\cite{lu2023cm,kim2024consistency}.

\begin{remark}
The error bound in Theorem~\ref{theorem:error:characteristic} is an average-in-time guarantee, not a uniform one. This is a direct consequence of our least-squares objective \eqref{eq:characteristic:LS}, which is designed to minimize the error averaged over all time pairs  $(t,s)$. While this formulation is effective for learning the overall map, it naturally leads to a convergence guarantee in an averaged sense. However, it does not imply a stronger, uniform guarantee for a specific time pair. Some potential approaches to obtain a uniform convergence guarantee will be discussed in Section~\ref{section:conclusion}.
\end{remark}

\subsection{Applications}\label{section:analysis:applications}

\par In this section, the theoretical analysis is applied to two types of flow-based method in~Table~\ref{table:rates}: linear interpolant and F{\"o}llmer flow. The convergence rates of them are shown in Corollaries~\ref{corollary:rate:linear} and~\ref{corollary:rate:follmer}, respectively.

\begin{corollary}[Convergence rate of linear interpolant]\label{corollary:rate:linear}
Suppose Assumptions~\ref{assumption:init:dist:gaussian} and~\ref{assumption:target:dist} hold, and let $\alpha_{t}=1-t$ and $\beta_{t}=t$. Set the denoiser network class $\scrD$ as in Theorem~\ref{theorem:velocity:rate}, with depth $L=C$ and number of non-zero weights $S=Cn^{\frac{d+1}{d+3}}$.
Set the stopping time $T$ as $T=1-Cn^{-\frac{2}{3(d+3)}}\log^{\frac{2}{3}}n$. Suppose the number of time steps $K$ satisfies $K\geq Cn^{\frac{2}{d+3}}$.
Then it follows that
\begin{equation*}
\bbE_{\euS}\Big[W_{2}^{2}\Big((\what{E}_{0,T})_{\sharp}\mu_{0},\mu_{1}\Big)\Big]\leq C\,n^{-\frac{4}{3(d+3)}}\,\log^{\frac{4}{3}}n.
\end{equation*}
In addition, set the characteristic network class $\scrG$ as in Theorem~\ref{theorem:error:characteristic}, with depth $L_{g}=C$ and number of non-zero weights $S_{g}=C\,m^{\frac{d+2}{d+4}}$. Further, if $m\geq Cn^{\frac{d+4}{d+3}}$ and $K\geq Cn^{\frac{2}{d+3}}\log m$, then
\begin{equation*}
\bbE_{\euS}\bbE_{\euZ}\big[\calD(\what{g})\big]\leq C\,n^{-\frac{4}{3(d+3)}}\,\log^{\frac{4}{3}}n.
\end{equation*}
Here, the constants $C$ may depend on $d$ and $\sigma$, but are independent of $n$.
\end{corollary}

\begin{corollary}[Convergence rate of F{\"o}llmer flow]\label{corollary:rate:follmer}
Suppose Assumptions~\ref{assumption:init:dist:gaussian} and~\ref{assumption:target:dist} hold, and let $\alpha_{t}=\sqrt{1-t^{2}}$ and $\beta_{t}=t$. Set the denoiser network class $\scrD$ as in Theorem~\ref{theorem:velocity:rate}, with depth $L=C$ and number of non-zero weights $S=Cn^{\frac{d+1}{d+3}}$.
Set the stopping time $T$ as $T=1-Cn^{-\frac{1}{d+3}}\log n$. Suppose the number of time steps $K$ satisfies $K\geq Cn^{\frac{2}{d+3}}$. Then it follows that
\begin{equation*}
\bbE_{\euS}\Big[W_{2}^{2}\Big((\what{E}_{0,T})_{\sharp}\mu_{0},\mu_{1}\Big)\Big]\leq C\,n^{-\frac{1}{d+3}}\,\log n.
\end{equation*}
In addition, set the characteristic network class $\scrG$ as in Theorem~\ref{theorem:error:characteristic}, with depth $L_{g}=C$ and number of non-zero weights $S_{g}=Cm^{\frac{d+2}{d+4}}$. Further, if $m\geq Cn^{\frac{d+4}{d+3}}$ and $K\geq Cn^{\frac{2}{d+3}}\log m$, then
\begin{equation*}
\bbE_{\euS}\bbE_{\euZ}\big[\calD(\what{g})\big]\leq C\,n^{-\frac{1}{d+3}}\,\log n.
\end{equation*}
Here, the constants $C$ may depend on $d$ and $\sigma$, but are independent of $n$.
\end{corollary}

\subsection{Mitigate the Curse of Dimensionality}
\label{section:analysis:manifold}
As shown in Theorems~\ref{theorem:velocity:rate} and~\ref{theorem:error:characteristic}, the convergence of both denoiser matching and characteristic fitting suffers from the curse of dimensionality: the sample complexity increases exponentially with the ambient dimension $d$ of the data. However, we will show that under Assumption~\ref{assumption:target:dist:manifold}, the convergence rate of the characteristic generator overcomes this limitation.

\subsubsection{Discussions on Assumption~\ref{assumption:target:dist:manifold}}
Under Assumption~\ref{assumption:target:dist:manifold}, a data point $X_{1} \sim \mu_{1}$ admits the decomposition $X_{1} \stackrel{\mathrm{d}}{=} PZ+\sigma W$, where the latent variable $Z\sim\tilde{\nu}$ and the isotropic Gaussian noise $W\sim\mathcal{N}(0,I_{d})$ are independent. Conceptually, $X_{1}$ comprises a structured signal $PZ$ --- generated by applying the linear decoder $P\in\mathbb{R}^{d\times d^{*}}$ to the low-dimensional representation $Z\in\mathbb{R}^{d^{*}}$ --- and an ambient noise term $\sigma W$. Consequently, the statistical complexity of estimating $\mu_{1}$ should parallel that of estimating the latent distribution $\tilde{\nu}$. We therefore expect the estimation difficulty to scale strictly with the intrinsic dimension $d^{*}$, remaining independent of the ambient dimension $d$.

\par Although our method does not require knowledge of the linear decoder $P\in\mathbb{R}^{d\times d^{*}}$, the following proposition provides a concrete characterization of $P$ within the framework of Assumption~\ref{assumption:target:dist:manifold}, revealing its connection to principal component analysis (PCA). See more discussions in Appendix~\ref{appendix:pca}.

\begin{proposition}\label{proposition:pca}
Suppose Assumption~\ref{assumption:target:dist:manifold} holds. Assume further that $\mathbb{E}_{Z\sim\tilde{\nu}}[Z]=0$ and $\mathrm{Cov}_{Z\sim\tilde{\nu}}(Z)\succ 0$. Let $v_{1},\ldots,v_{d^{*}}\in\mathbb{R}^{d}$ denote the eigenvectors corresponding to the top $d^{*}$ eigenvalues of $\mathrm{Cov}(X_{1})$. Then
\begin{equation*}
\mathrm{span}(v_{1},\ldots,v_{d^{*}})=\mathrm{span}(P).
\end{equation*}
\end{proposition}

\subsubsection{Low-dimensional decomposition}
The key to this result is the following decomposition lemma.

\begin{lemma}[Low-dimensional representation]
\label{lemma:manifold:decomp:main}
Suppose Assumptions~\ref{assumption:init:dist:gaussian} and~\ref{assumption:target:dist:manifold} hold. For any $(t,s,x)$ with $0\le t\le s\le 1$ and $x\in\bbR^d$,
\begin{align*}
b^*(t,x)&\equiv P\,\widetilde{b}^{*}\big(t,P^{\top}x\big)+\frac{\alpha_t \dot{\alpha}_t+\sigma^{2}\beta_t\dot{\beta}_t}{\alpha_t^{2}+\sigma^{2}\beta_t^{2}}\,(I_d-PP^{\top})x, \\
g^*(t,s,x)&\equiv P\,\widetilde{g}^*\big(t,s,P^{\top}x\big)+\sqrt{\frac{\alpha_s^{2}+\sigma^{2}\beta_s^{2}}{\alpha_t^{2}+\sigma^{2}\beta_t^{2}}}\,(I_d-PP^{\top})x{\color{blue},} \\
D^*(t,x)&\equiv P\,\widetilde{D}^{*}\big(t,P^{\top}x\big)+\frac{\sigma^{2}\beta_{t}}{\alpha_{t}^{2}+\sigma^{2}\beta_{t}^{2}}\,(I_d-PP^{\top})x.
\end{align*}
The vector field $\widetilde{b}^{*}:(0,1)\times\bbR^{d^{*}}\rightarrow\bbR^{d^{*}}$ is defined as:
\begin{equation*}
\widetilde{b}^*(t,\widetilde{x})\coloneqq \bbE[\dot{\alpha}_{t}\widetilde{X}_0+\dot{\beta}_{t}\widetilde{X}_{1}|\widetilde{X}_{t}=\widetilde{x}], \quad \widetilde{x} \in \mathbb{R}^{d^*},
\end{equation*}
where $\widetilde{X}_0 \sim \mathcal{N}(0, I_{d^*})$ and $\widetilde{X}_1 \sim \mathcal{N}(0, \sigma^2I_{d^*})*\tilde{\nu}$ are independent, and $\widetilde{X}_{t}\coloneqq \alpha_{t}\widetilde{X}_0+\beta_{t}\widetilde{X}_{1}$. Further, the vector field $\widetilde{g}^{*}:\bbR\times\bbR\times\bbR^{d^{*}}\rightarrow\bbR^{d^{*}}$ is defined as the flow map of $\d\widetilde{x}(t)=\widetilde{b}^{*}(t,\widetilde{x}(t))\dt$, and $\widetilde{D}^{*}(t,\widetilde{x})\coloneqq\bbE[\widetilde{X}_{1}|\widetilde{X}_{t}=\widetilde{x}]$ is the denoiser associated with the $d^{*}$-dimensional interpolant $\widetilde{X}_{t}$.
\end{lemma}

\par This lemma is the cornerstone of our result. It reveals that both $b^{*}$ and $g^{*}$ have a crucial low-dimensional structure. They decompose into two orthogonal parts:
\begin{enumerate}[(i)]
\item A complex, non-linear component governed by the low-dimensional functions $\widetilde{b}^{*}$ and $\widetilde{g}^{*}$, which only act on the $d^{*}$-dimensional latent vector $z\coloneqq P^{\top}x$.
\item A simple, linear component that acts on the space orthogonal to the linear sub-space.
\end{enumerate}

\par This decomposition implies that the learning problem is greatly simplified. Instead of learning a function in the ambient $d$ dimensions, we only need to learn the low-dimensional components $\widetilde{D}^{*}$ and $\widetilde{g}^{*}$ in the intrinsic $d^{*}$ dimensions. This is precisely why our method's convergence rate depends on $d^{*}$ instead of $d$, effectively mitigating the curse of dimensionality. The necessary regularity conditions for these low-dimensional functions are established in the appendix.

\subsubsection{Applications}
Now, we formally state the convergence results of linear interpolant and F\"ollmer flow under low-dimensional hypothesis, which explicitly alleviate the curse of dimensionality.

\begin{corollary}[Convergence of linear interpolant under low-dimensional hypothesis]\label{corollary:rate:linear:manifold}
Suppose Assumptions~\ref{assumption:init:dist:gaussian} and~\ref{assumption:target:dist:manifold} hold. Set the denoiser network class $\scrD$ properly with depth $L=C$ and number of non-zero weights $S=Cn^{\frac{d^*+1}{d^*+3}}$.
Set the stopping time $T$ as $T=1-Cn^{-\frac{2}{3(d^*+3)}}\log^{\frac{2}{3}}n$. Suppose the number of time steps $K$ satisfies $K\geq Cn^{\frac{2}{d^*+3}}$.
Then it follows that
\begin{equation*}
\bbE_{\euS}\Big[W_{2}^{2}\Big((\what{E}_{0,T})_{\sharp}\mu_{0},\mu_{1}\Big)\Big]\leq C\,n^{-\frac{4}{3(d^*+3)}}\,\log^{\frac{4}{3}}n.
\end{equation*}
In addition, set the characteristic network class $\scrG$ with depth $L_{g}=C$ and number of non-zero weights $S_{g}=C\,m^{\frac{d^*+2}{d^*+4}}$. Further, if $m\geq Cn^{\frac{d^*+4}{d^*+3}}$ and $K\geq Cn^{\frac{2}{d^*+3}}\log m$, then
\begin{equation*}
\bbE_{\euS}\bbE_{\euZ}\big[\calD(\what{g})\big]\leq C\,n^{-\frac{4}{3(d^*+3)}}\,\log^{\frac{4}{3}}n.
\end{equation*}
Here, the constants $C$ may depend on $d$, $d^*$, and $\sigma$, but are independent of $n$.
\end{corollary}

\begin{corollary}[Convergence of F\"ollmer flow under low-dimensional hypothesis]\label{corollary:rate:follmer:manifold}
Suppose Assumptions~\ref{assumption:init:dist:gaussian} and~\ref{assumption:target:dist:manifold} hold. Set the denoiser network class $\scrD$ properly with depth $L=C$ and number of non-zero weights $S=Cn^{\frac{d^*+1}{d^*+3}}$.
Set the stopping time $T$ as $T=1-Cn^{-\frac{1}{d^*+3}}\log n$. Suppose the number of time steps $K$ satisfies $K\geq Cn^{\frac{2}{d^*+3}}$. Then it follows that
\begin{equation*}
\bbE_{\euS}\Big[W_{2}^{2}\Big((\what{E}_{0,T})_{\sharp}\mu_{0},\mu_{1}\Big)\Big]\leq C\,n^{-\frac{1}{d^*+3}}\,\log n.
\end{equation*}
In addition, set the characteristic network class $\scrG$ with depth $L_{g}=C$ and number of non-zero weights $S_{g}=Cm^{\frac{d^*+2}{d^*+4}}$. Further, if $m\geq Cn^{\frac{d^*+4}{d^*+3}}$ and $K\geq Cn^{\frac{2}{d^*+3}}\log m$, then
\begin{equation*}
\bbE_{\euS}\bbE_{\euZ}\big[\calD(\what{g})\big]\leq C\,n^{-\frac{1}{d^*+3}}\,\log n.
\end{equation*}
Here, the constants $C$ may depend on $d$, $d^*$, and $\sigma$, but are independent of $n$.
\end{corollary}

\subsubsection{Discussions on Corollaries~\ref{corollary:rate:linear:manifold} and~\ref{corollary:rate:follmer:manifold}}
\begin{enumerate}[(i)]
\item In contrast to the results in Corollaries~\ref{corollary:rate:linear} and~\ref{corollary:rate:follmer}, the convergence rates established under the low-dimensional sub-space assumption depend on the intrinsic dimension $d^{*}$ rather than the ambient dimension $d$. This demonstrates that our characteristic generator can mitigate the curse of dimensionality whenever the data possess intrinsic low-dimensional structure.
\item The convergence rates in Corollaries~\ref{corollary:rate:linear:manifold} and~\ref{corollary:rate:follmer:manifold} hold without knowledge of the linear decoder $P \in \mathbb{R}^{d \times d^{*}}$. That is, under Assumption~\ref{assumption:target:dist:manifold}, the empirical risk minimizers~\eqref{eq:velocity:erm} and~\eqref{eq:charateristic:erm} automatically adapt to the low-dimensional linear sub-space without any modification to the algorithm.
\item\label{item:practice} Corollaries~\ref{corollary:rate:linear:manifold} and~\ref{corollary:rate:follmer:manifold} characterize the convergence of empirical risk minimizers and therefore abstract away optimization error and computational constraints. In practice, however, models are trained under limited computational budgets and optimization may not converge to the global minimum. A standard strategy to alleviate this difficulty is to first learn an encoder that maps the data to a low-dimensional latent space, and then to train a generative model within that latent space. Given a sufficiently accurate encoder, learning in the low-dimensional latent space is computationally cheaper and converges more reliably than learning directly in the ambient space.
\item From Proposition~\ref{proposition:pca}, one practical approach is to use PCA to estimate the unknown linear decoder $P \in \mathbb{R}^{d \times d^{*}}$ and its corresponding encoder $P^{\top} \in \mathbb{R}^{d^{*} \times d}$. See Appendix~\ref{appendix:pca} for more discussions.
\item In practice, nonlinear encoder-decoder pairs --- such as the latent VAE used in our experiments --- are considerably more expressive than the linear model furnished by PCA.
\item While the Assumption~\ref{assumption:target:dist:manifold} is restricted to the linear low-dimensional structure, extending the analysis of Corollaries~\ref{corollary:rate:linear:manifold} and~\ref{corollary:rate:follmer:manifold} to capture nonlinear low-dimensional structure is an important direction for future work.
\end{enumerate}

%% file: numerical.tex
% !TEX root = main.tex
\section{Numerical Studies}\label{section:numerical}

In this section, we delve into the numerical performance of characteristic learning. We use the Fr{\'e}chet inception distance (FID) to measure sampling quality on image data. Lower FID means better performance. See Appendix~\ref{sec:extra experiment details} for the list of full hyper-parameters employed in our numerical experiments. In addition, we provide some remarks on the training time consumption of our numerical experiments in Appendix~\ref{sec:remark time}.

\subsection{Synthetic 2-dimensional dataset}

\par
\begin{figure}[htbp]
\centering
\includegraphics[width=0.60\textwidth]{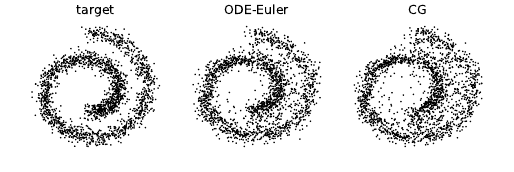}
\caption{Original Swiss roll samples and samples generated by ODE model with Euler method (ODE-Euler) and the characteristic generator (CG).} \label{fig:swissroll}
\end{figure}

 On a 2-dimensional dataset where the target distribution shapes like a Swiss roll, the characteristic generator trained by Algorithm~\ref{alg:characteristic} works well. We display the original dataset and samples generated by Euler method (NFE=100) and the characteristic generator (NFE=1) in Figure~\ref{fig:swissroll}.

\par From Figure~\ref{fig:swissroll}, it can be observed that the generative quality of the one-step characteristic generator closely resembles that of Euler method with 100 function evaluations (NFE). This indicates that the original training procedure (Algorithm~\ref{alg:characteristic}) can yield commendable generation outcomes when dealing with relatively simple target distribution.

\subsection{Real dataset: MNIST and CIFAR-10}

\par To improve the empirical performance of the vanilla characteristic learning in Section~\ref{section:characteristic}, we present some technical tricks to exploit as much of the structure of the problem as possible, without changing its mathematical nature. Our main idea is to reparameterize the probability flow ODE~\eqref{eq:characteristic:semilinear} as
\begin{equation}\label{eq:char:param}
g(t,s,x)= \frac{\alpha_s}{\alpha_t} x + \alpha_s (\gamma_s - \gamma_t) D_{\calS}(t,s,x), \, \gamma_{t}\coloneqq\frac{\beta_{t}}{\alpha_{t}},
\end{equation}
where $D_{\mathcal{S}}$ is a deep neural network known as the ``student'' model. Since the exact denoiser $D^{*}$ is unknown, the student network $D_{\calS}$ has to be trained from a denoiser estimator $\what{D}\approx D^{*}$ in~\eqref{eq:velocity:erm}. Therefore, $\what{D}$ is referred to as the teacher model and denoted by $D_{\calT}\coloneqq\what{D}$ thereafter.

\par We next design the objective functional for the student model $D_{\calS}$ based on~\eqref{eq:characteristic:LS} to utilize as much of the structure of the problem as possible. Note that the exact solution of~\eqref{eq:characteristic:semilinear} admits the integral representation $g^{*}_{t,s}(x)=\frac{\alpha_{s}}{\alpha_{t}}x+\alpha_{s}\int_{t}^{s}D^{*}(\tau,x(\tau))\,\d\gamma_{\tau}$, so that $D_{\calS}(t,s,x)$ in~\eqref{eq:char:param} plays the role of the weighted average of the denoiser over $[t,s]$. First look into the local property of the integral:
\begin{equation*}
\lim_{s\rightarrow t^{+}}  \frac{\int_t^s D^*(\tau, x(\tau)) \d \gamma_\tau}{\gamma_s - \gamma_t} = D^{*}(t, x(t)).
\end{equation*}
This inspires us to reuse the denoiser matching objective functional~\eqref{eq:velocity:popu:risk} as the following local risk to ensure the local consistency of $D_{\calS}$:
\begin{equation}\label{eq:char:pop:risk:1}
\calR_{\loc}(D_{\calS})=\int_{0}^{T}\bbE\Big[\|X_{1}-D_{\calS}(t,t,X_{t})\|_{2}^{2}\Big]\dt.
\end{equation}
On the other hand, we require the outputs of generator~\eqref{eq:char:param} to align with the solutions given by~\eqref{eq:characteristic:ei:solution}, and satisfy the semi-group property in Proposition~\ref{proposition:semi:group}. This implies the following global risk:
\begin{equation}\label{eq:char:pop:risk:2}
\calR_{\glo}(D_{\calS})=\int_{0}^{T} \!\! \int_{t}^{T} \!\! \int_{s}^{T} \!\! \bbE\Big[\|g_{s,T}^{\off}\circ g_{u,s}\circ g_{t,u}^{\Int}(X_{t})
-g_{s,T}^{\off}\circ g_{t,s}^{\off}(X_t)\|_{2}^{2}\Big]{\d u \ds \dt},
\end{equation}
where $g$ is defined by \eqref{eq:char:param}, $g^{\Int}$ denotes the exponential integrator solution given by the teacher model $D_{\calT}$, and $g^{\off}$ denotes an offline copy of $g$, introduced for long-time stability. The population risk~\eqref{eq:char:pop:risk:2} can be considered as a semi-group-consistent variant of the original objective functional~\eqref{eq:characteristic:LS}, ensuring the long-range consistency of the generator. Combining the short-range denoiser matching risk~\eqref{eq:char:pop:risk:1} and the long-range characteristic fitting risk~\eqref{eq:char:pop:risk:2}, we obtain the final training procedure for the characteristic generator. We conclude the practical characteristic learning algorithm in Algorithm~\ref{alg:image:cl}. The one-step sampling procedure is the same as that in Algorithm~\ref{alg:characteristic}. For better sampling quality, one can divide the time interval into pieces as Algorithm~\ref{alg:characteristic:sampling:fine}.

\begin{algorithm}
\label{alg:image:cl}
\begin{algorithmic}[1]
\Require {Observations $X_{1}\sim\mu_{1}$, and pre-trained denoiser $D_{\calT}$.}
\State {Initialize the student network $D_{\calS,\phi}:\bbR\times\bbR\times\bbR^{d}\rightarrow\bbR^{d}$.}
\State {Choose the loss parameter $\lambda$.}
\Repeat
\State {\textcolor{gray!45!black}{\# Short-range denoiser matching}}
\State {Draw $X_{0}\sim\mu_{0}=N(0,I_{d})$ and $t\sim\unif[0,T]$.}
\State {Construct stochastic interpolant $X_{t}=\alpha_{t}X_{0}+\beta_{t}X_{1}$.}
\State {$\what{\calR}_{\loc}(D_{\calS,\phi})\leftarrow\|D_{\calS,\phi}(t,t,X_{t})-X_{0}\|_{2}^2$.}
\State {\textcolor{gray!45!black}{\# Long-range characteristic matching}}
\State {Draw $s\sim\unif[t,T]$ and $u\sim\unif[s,T]$.}
\State {$\what{\calR}_{\glo}(D_{\calS,\phi})\leftarrow\|(g_{s,T}^{\off}\circ g_{u,s}\circ g_{t,u}^{\Int}-g_{s,T}^{\off}\circ g_{t,s}^{\off})(X_t)\|_{2}^{2}$.}
\State {\textcolor{gray!45!black}{\# Gradient descent update}}
\State {$\phi\leftarrow\phi-\alpha\nabla_{\phi}\{\lambda\what{\calR}_{\loc}(D_{\calS,\phi})+\what{\calR}_{\glo}(D_{\calS,\phi})\}$.}
\Until {converged}
\Ensure {$\what{g}(t,s,x) = \frac{\alpha_s}{\alpha_t} x + \alpha_s \left( \gamma_s - \gamma_t \right) D_{\calS,\phi}(t,s,x)$.}
\end{algorithmic}
\caption{Practical training procedure.}
\end{algorithm}

\begin{remark}[Gap between theory and practice]
Although the local and global risks in~\eqref{eq:char:pop:risk:1} and~\eqref{eq:char:pop:risk:2} are designed based on~\eqref{eq:characteristic:LS} and share the same underlying mathematical structure, the error analysis for the characteristic generator developed in this work does not fully align with these local-global risks. A rigorous treatment of this gap is left to future work.
\end{remark}

\begin{remark}[Comparison with CTM~\cite{kim2024consistency}]
\cite{kim2024consistency} proposed a similar method, but our approach differs from the CTM in two significant aspects. First, the integral scheme $g^{\Int}$ employed by CTM is Euler-based. While Euler method coincides with the first-order exponential integrator for VE-ODE~\cite{song2021score}, its numerical stability cannot be guaranteed for general probability flow ODEs. In contrast, the exponential integrator used in our method may ease potential training instability as it fully exploits the semi-linearity of the ODE system. An ablation study on Euler method and exponential integrator is provided in Appendix~\ref{app:ablation}. Secondly, CTM relies on GAN training in their approach, borrowing a pre-trained discriminator and treating $g$ as the generator. This reliance on GAN training may cause potential training instability and require extra training of the discriminator. Our method, however, does not require this additional GAN training burden. Detailed comparisons between characteristic generator and CTM can be found in Appendix~\ref{section:comparison:ctm}.
\end{remark}

\begin{figure}[thb] \centering
\includegraphics[width=0.75\textwidth]{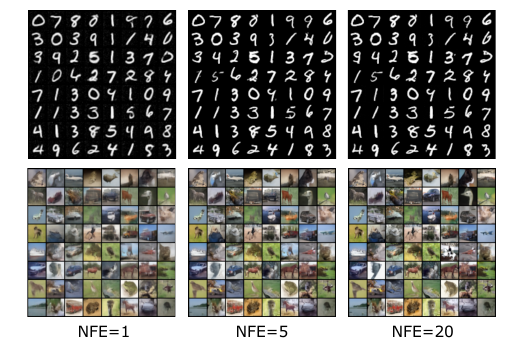}
\caption{Samples generated by the characteristic generator on CIFAR-10.} \label{fig:cg}
\end{figure}

Generated images by the characteristic generator are displayed in Figure~\ref{fig:cg} (MNIST on top and CIFAR-10 on bottom).
The experimental results illustrate that one-step generation has high generation quality, which can be significantly improved by iterating the characteristic generator by a few steps.

\begin{figure}[htb] \centering
\includegraphics[width=0.75\textwidth]{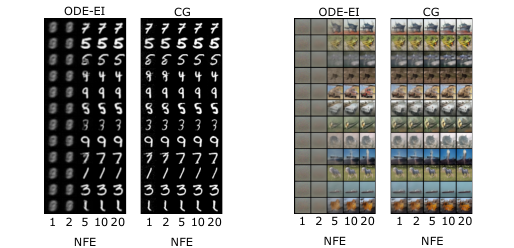}
\caption{Comparison of samples generated by ODE with exponential integrator (ODE-EI) and characteristic generators (CG) under different NFE on MNIST and CIFAR-10.}
\label{fig:comparison}
\end{figure}

We compare the images generated by the numerical sampler and characteristic generator with different numbers of function evaluations (NFE) in Figure~\ref{fig:comparison}. The numerical sampler fails to accurately generate images at small NFE values such as NFE=1 and NFE=2. In fact, with NFE=1, the solution is actually close to the mean of the target distribution. It is necessary to have five or more NFE for the numerical ODE solvers to work properly. In contrast, the characteristic generator is capable of generating high-quality images even with NFE=1. In essence, the characteristic generator distills the knowledge of a precise, multi-step solver into a fast, single step network. Our theory correctly predicts that its error should be much lower than that of a naive one-step exponential integrator, which aligns perfectly with our empirical findings in Figure~\ref{fig:comparison}. See Appendix~\ref{section:comparison} for detailed discussions.

Furthermore, we compare the convergence of FID by NFE for the characteristic generator in Table~\ref{tab:FID:NFE}. It is evident that the characteristic generator converges faster than the numerical sampler and achieves better scores at smaller NFE values.

\begin{table}[htb]
\caption{Comparison of FID by NFE between the exponential integrator (EI) and characteristic generator (CG) on MNIST and CIFAR-10.}
\label{tab:FID:NFE}
\centering
{\small
\begin{tabular}{ccccccc}
\toprule
Dataset & Method & NFE=1 &  NFE=2 &  NFE=5 &  NFE=10 & NFE=20 \\
\toprule
MNIST & ODE-EI & 46.55 & 46.71 & 2.69 & 0.66 & 0.28 \\
MNIST & CG & 1.97 & 1.15 & 0.28 & 0.20 & 0.13 \\
\midrule
CIFAR-10 & ODE-EI & 14.06 & 15.42 & 5.38 & 3.16 & 2.50 \\
CIFAR-10 & CG & 4.59 & 3.50 & 2.90 & 2.76 & 2.63 \\
\bottomrule
\end{tabular}}
\end{table}

Table~\ref{tab:comparison} presents the FID on CIFAR-10 achieved by various generative models. The proposed characteristic generator demonstrates superior generation quality compared to models without GAN, regardless of whether it is one-step or few-step generation. Notably, our proposed method achieves a comparable FID to the state-of-the-art method CTM~\cite{kim2024consistency}, without the requirement of additional GAN training as employed by CTM. Moreover, our proposed characteristic generator with NFE=4 achieves similar or even superior generation performance compared to GAN models.

\begin{table}[htbp]
\caption{Performance comparisons on CIFAR-10.}
\label{tab:comparison}
\centering
{\small
\begin{tabular}{ccc}
\toprule[1pt]
Model & NFE $\downarrow$ & FID $\downarrow$ \\
\toprule[1pt]
\bf{GAN Models} & & \\
BigGAN~\cite{brock2018large} & 1 & 8.51 \\
StyleGAN-Ada~\cite{karras2019style} & 1 & 2.92 \\
\midrule[1pt]
\bf{Diffusion + Sampler} & & \\
DDPM~\cite{Ho2020Denoising} & 1000 & 3.17 \\
DDIM~\cite{song2021denoising} & 100 & 4.16 \\
Score SDE~\cite{song2021score} & 2000 & 2.20 \\
EDM~\cite{karras2022elucidating} & 35 & 2.01 \\
\midrule[1pt]
\bf{Diffusion + Distillation} & & \\
KD~\cite{luhman2021knowledge} & 1 & 9.36 \\
DFNO~\cite{Zheng2023Fast} & 1 & 5.92 \\
Rectified Flow~\cite{liu2022rectified} & 1 & 4.85 \\
PD~\cite{salimans2022progressive} & 1 & 9.12 \\
CD (\cite{Song2023Consistency}, retrained by~\cite{kim2024consistency}) & 1 & 10.53 \\
CTM (without GAN)~\cite[Table 3]{kim2024consistency} & 1 & 5.19 \\
CG (ours) & 1 & \textbf{4.59} \\
\midrule[0.5pt]
PD~\cite{salimans2022progressive} & 2 & 4.51 \\
CTM (without GAN)~\cite[Table 3]{kim2024consistency} & 18 & 3.00 \\
CG (ours) & 2 & \textbf{3.50} \\
CG (ours) & 4 & \textbf{2.83} \\
\midrule[1pt]
\bf{Diffusion + Distillation + GAN} & & \\
CD (with GAN)~\cite{lu2023cm} & 1 & 2.65 \\
CTM (with GAN)~\cite{kim2024consistency} & 1 & 1.98 \\
CTM (with GAN)~\cite{kim2024consistency} & 2 & 1.87 \\
\bottomrule[1pt]
\end{tabular}}
\end{table}

\subsection{High-resolution generation on CelebA HQ}

\par We explore applying the proposed method on higher resolution image datasets such as CelebA HQ (down-sampled to $256 \times 256$ and $512 \times 512$), and eliminate the dependence on the teacher model, resulting in a self-distillation scheme. An ablation study on the comparison of student-teacher distillation and self-distillation is provided in Appendix~\ref{section:ablation:student:teacher}.

\par The main idea is to define a teacher-free reference solution $\overline{g}_{t, s}(X_t)$ from a characteristic estimator $\what{g}$, by
\begin{equation}\label{eq:self-dist}
\overline{g}_{t, s} (X_t) = \what{g}_{u, s} \circ \what{g}_{t, u}(X_t), \ u \in (t, s).
\end{equation}
With the reference solution $\overline{g}$, we can get rid of the extra teacher network when training the characteristic generator, significantly reducing the required memory.
Also, the formation of $\overline{g}$ naturally incorporates the semigroup constraint, as the optimization target is to minimize the distance between $\overline{g}$ and $\what{g}$.
Notice that we do not specify the choice of $u$ in \eqref{eq:self-dist} as it should be arbitrary.
Recent progress~\cite{frans2025one} finds $u = (t+s)/2$ works well enough and shows some symmetry, and we follow this setup. During the early stage of training, the local consistency loss plays the dominant role as the global counterpart relies on the reference solution, which requires the estimator $\what{g}$ fitted on a smaller time scale.
During the late stage of training, the local consistency is already ensured, and the global consistency loss gradually takes effect.

\begin{figure}[htbp]
\centering
\includegraphics[width=0.75\textwidth]{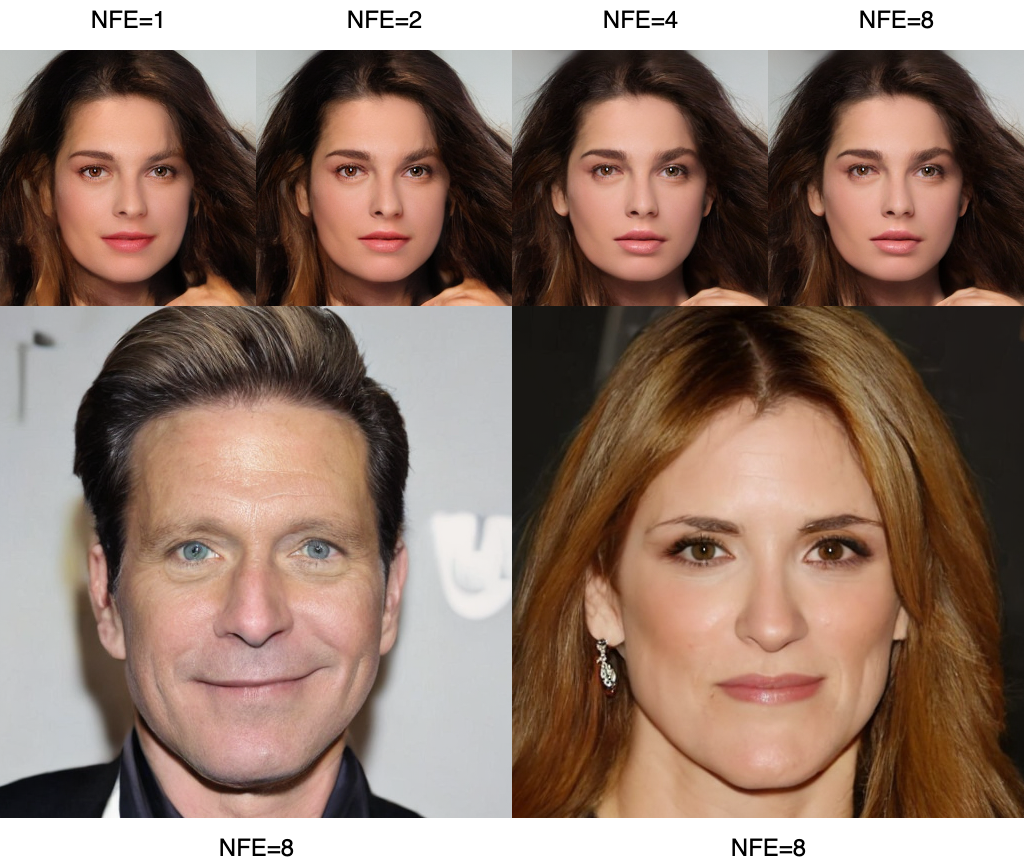}
\caption{Generated images on CelebAHQ-256 (top) and CelebAHQ-512(bottom) (scaled down by a factor of 0.28 to fit the page).}
\label{fig:celeba}
\end{figure}

We perform the self-distillation variant on the CelebA HQ dataset (down-sampled to $256 \times 256$ and $512 \times 512$). We carry out the training and generation on the $8 \times$ down-sampled latent space using the sd-vae-ft-mse autoencoder~\cite{rombach2022high}. We also move from the U-Net architecture to the emerging DiT architectures~\cite{peebles2023scalable} (B/2 on the 256 resolution and XL/2 on the 512 resolution). We report the FID in Table~\ref{tab:celeba} and the generated images in Figure~\ref{fig:celeba}.

\begin{table}[htb]
\caption{Comparison of FID by NFE by the characteristic generator on CelebAHQ-256 and CelebAHQ-512.}
\label{tab:celeba}
\centering
\small
\begin{tabular}{ccccc}
\toprule
Resolution & NFE=1 &  NFE=2 &  NFE=4 &  NFE=8 \\
\toprule
256 &  30.25 & 19.02 & 15.82 & 14.29  \\
512 &  31.18 & 22.45 & 18.61 & 15.95 \\
\bottomrule
\end{tabular}
\end{table}

\subsection{Large-scale dataset: ImageNet}

\par We further explore applying characteristic generator to the ImageNet dataset. ImageNet serves as the primary standardized benchmark for class-conditional image generation. It contains roughly 1.28 million training images spanning 1,000 distinct object categories. It forces a generative model to learn an immense variety of visual structures simultaneously.

\par Similar to the CelebA dataset, we use a pre-trained VAE model to map the original pixel distribution to the latent space, where each latent sample is of shape $4 \times 32 \times 32$.
We adopt the DiT-B/2 architecture for the generator network, with about 130 million parameters.
To illustrate the efficacy of our proposed method, we train the generator from scratch by self-distillation for 80 epochs. The training batch size is 256. Other training hyper-parameters are listed in Appendix~\ref{sec:extra experiment details}.

\begin{figure}[htbp]
\centering
\includegraphics[width=0.75\linewidth]{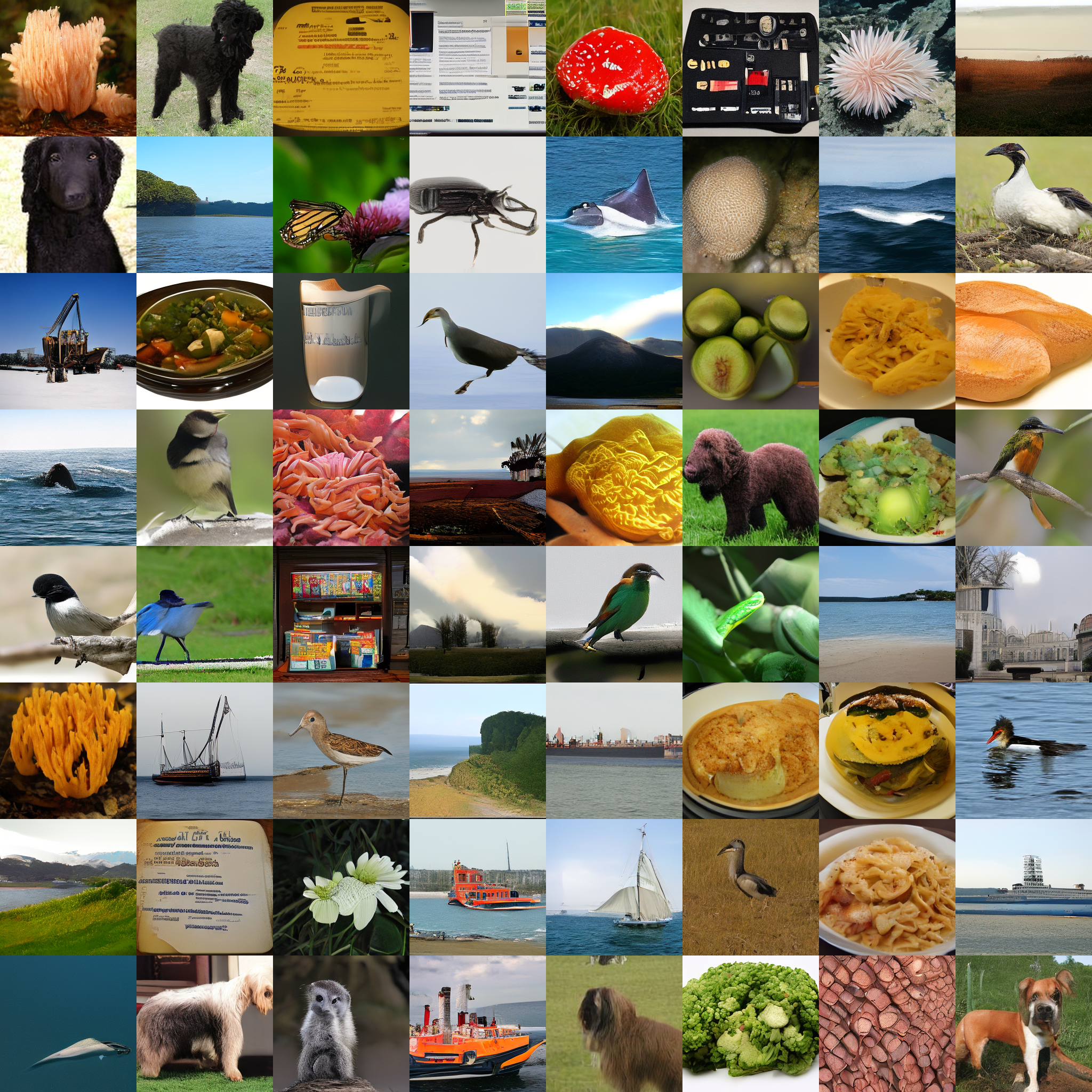}
\caption{Generated samples of 4-NFE characteristic generator on the $256 \times 256$ ImageNet dataset (scaled to fit the page).}
\label{fig:imagenet}
\end{figure}

During inference, we follow baselines like Shortcut Models~\cite{frans2025one} and MeanFlow~\cite{geng2026mean}, and set the classifier-free guidance scale to 1.5. We report the FID of our proposed method with NFE=1, 2, 4, 8 in Table~\ref{tab:fid:imagenet}, showing the improvements in generation quality as NFE increases. The generated images are shown in Figure~\ref{fig:imagenet}.

\begin{table}[htbp]
\centering
\caption{FID of characteristic generators on $256 \times 256$ ImageNet trained for 80 epochs.}
\begin{tabular}{lcccc}
\hline
NFE & 1 & 2 & 4 & 8 \\
\hline
FID & 54.90 & 40.29 & 33.28 & 29.79 \\
\hline
\end{tabular}
\label{tab:fid:imagenet}
\end{table}

\par Under the same network architecture (DiT-B) and training budget, the 1-NFE FID of the characteristic generator is somewhat higher than that of Shortcut Models (FID $= 40.3$)~\cite{frans2025one} and MeanFlow (FID $= 33.33$)~\cite{geng2026mean}. However, since our method learns the full characteristic function rather than being tailored specifically to one-step generation, its full potential is best realized when a small number of additional function evaluations are allowed. Indeed, as shown in Table~\ref{tab:fid:imagenet}, with only a modest increase in inference cost, our method quickly closes this gap and achieves comparable or superior performance.

\begin{remark}
DiT-B is a medium-level network, and the 80 training epochs might not be sufficient for the Transformer to capture the diverse structure of the ImageNet dataset efficiently. Due to limited computational resources, we currently conduct our experiments on the DiT-B scale network. In future work, we plan to scale up to larger architectures such as DiT-XL and train for more epochs to investigate the corresponding scaling laws of our model.
\end{remark}

\subsection{Ablation studies of the early-stopping time}

The theoretical requirement to stop strictly before $1$ is well-founded: taking $T\to1$ can induce numerical instability (due to the rapidly varying denoiser near $t=1$, Proposition~\ref{proposition:reg:denoiser}), and similar behavior has been observed empirically in prior work~\cite{Kim2022Soft}. To mitigate this in practice, we leverage modern architectures and stabilization techniques—most notably a DiT backbone with adaptive layer normalization and sophisticated time-step embeddings—which make the model robust when $T\approx1$.

To directly assess sensitivity, we conducted an ablation on the $256\times256$ CelebA dataset with the characteristic generator (NFE=10, DiT-B/2). Table~\ref{tab:T} reports FIDs as $T$ varies; as $T$ decreases from $1$ to $123/128$, FID increases moderately and remains within an acceptable range.

\begin{table}[ht]
\caption{FIDs under different $T$ (CelebAHQ-256).}
\begin{center}
\begin{tabular}{cccccc}
\hline
$T$ & 127/128 & 126/128 & 125/128 & 124/128 & 123/128 \\
\hline
FID & 14.12 & 14.25 & 14.49 & 14.85 & 15.36 \\
\hline
\end{tabular}
\end{center}
\label{tab:T}
\end{table}

\begin{figure}
\begin{center}
\includegraphics[width=0.20\linewidth]{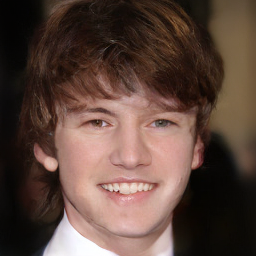}%
\includegraphics[width=0.20\linewidth]{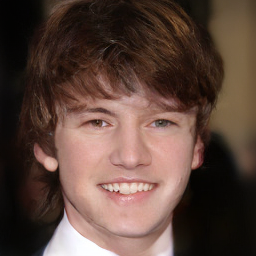}%
\includegraphics[width=0.20\linewidth]{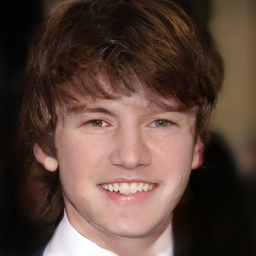}%
\includegraphics[width=0.20\linewidth]{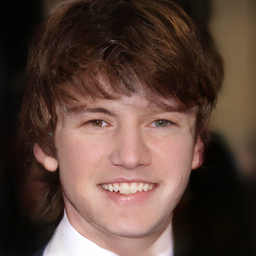}%
\includegraphics[width=0.20\linewidth]{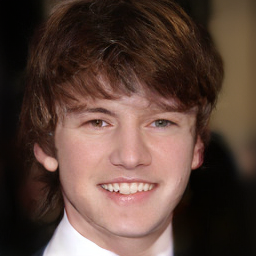}\\
\makebox[0.20\linewidth][c]{$\scriptstyle T=123/128$}%
\makebox[0.20\linewidth][c]{$\scriptstyle T=124/128$}%
\makebox[0.20\linewidth][c]{$\scriptstyle T=125/128$}%
\makebox[0.20\linewidth][c]{$\scriptstyle T=126/128$}%
\makebox[0.20\linewidth][c]{$\scriptstyle T=127/128$}%
\end{center}
\caption{Generated example images under different $T$ (CelebAHQ-256).}
\label{fig:T}
\end{figure}

For illustration, we also provide example images generated under different stopping $T$ in Figure~\ref{fig:T}. These images are visually similar across different choices of $T$.

Overall, these results confirm that our practical implementation is robust: with the above stabilization, generation remains stable and not overly sensitive to the exact choice of $T$ when $T$ is close to $1$.

%% file: conclusion.tex
\section{Conclusions and Future Work}\label{section:conclusion}

\par In this paper, we propose the characteristic generator, a novel one-step generative model that combines sampling efficiency with high generation quality. In terms of theoretical analysis, we have examined the errors in velocity matching, Euler discretization, and characteristic fitting, enabling us to establish a non-asymptotic convergence rate for the characteristic generator in $2$-Wasserstein distance. This analysis represents the first comprehensive investigation into simulation-free one-step generative models, refining the error analysis of flow-based generative models in prior research. We have validated the effectiveness of our method through experiments on synthetic and real datasets. The results demonstrate that the characteristic generator achieves high generation quality with just a single evaluation of the neural network. This highlights the efficiency and stability of our model in generating high-quality samples.

\par Finally, we would like to emphasize that our framework of one-step generation is highly versatile and can be extended to conditional generative learning directly. By incorporating the encoder-decoder technique, our approach can be generalized to the latent space, enabling its application in a wide range of practical scenarios, including video generation. The characteristic generator improved by these techniques lays a technical foundation for deploying large-scale generative models on end devices.

\par On the theoretical front, we intend to exploit the regularity of the velocity field in our error analysis for velocity matching. This will allow us to achieve a faster convergence rate. Additionally, we plan to analyze higher order and more stable numerical schemes, such as the high-order exponential integrator, in order to provide a comprehensive understanding of their effectiveness. Furthermore, we aim to establish a theoretical foundation for the role of semi-group penalties in the characteristic fitting. This will contribute to a deeper understanding of their impact and significance in our framework. A significant extension would be to move from our current average-in-time error guarantee (Theorem~\ref{theorem:error:characteristic}) to a stronger, uniform-in-time bound. Achieving this would likely require designing novel objective functionals for the characteristic fitting, potentially based on $L^{\infty}$-risk minimization or adversarial (minimax) formulations. Investigating the theoretical properties and empirical performance of these approaches is an important open question.

This work provides both theoretical analysis and experimental validation. However, it is important to clarify the role of our theoretical bounds, as their direct numerical verification is challenging due to the well-known gap between deep learning theory and practice. This gap arises from several fundamental factors: (i) Our bounds, like many in the field, serve to characterize scaling laws and qualitative relationships, e.g., how error scales with dimension, sample size, or the number of Euler steps. They are not intended to be numerically predictive of the exact empirical error. The bounds often contain large, abstract constants that are intractable to compute in practice, making a direct numerical comparison infeasible. (ii) Since the theoretical bounds often represent a worst-case scenario, there exists a gap between the empirical error and the theoretical bound. (iii) Our theory analyzes the properties of the empirical risk minimizer, whereas in practice, non-convex optimization methods find local minima. This creates a gap between the object of our theoretical analysis and the model obtained empirically. Bridging this theory-practice gap and developing bounds that more closely reflect the performance of practically trained models remains a crucial challenge for the field.

%% file: appendix.tex
\section*{Outline of the Appendices}

The supplementary material provides detailed derivations, proofs, and auxiliary results, organized into the following appendices:
\begin{enumerate}[(I)]
\item A discussion of related work is provided in Appendix~\ref{section:related:work}.
\item Appendix~\ref{section:natation} summarizes the notation used throughout the paper.
\item Appendix~\ref{section:supp:lemmas} provides auxiliary definitions and lemmas.
\item Appendix~\ref{section:proof:method} presents the proofs for the theoretical results in Section~\ref{section:method}.
\item Appendix~\ref{section:proof:properties} establishes regularity properties of the probability flow ODE and provides the proofs for the propositions in Section~\ref{section:analysis:property}.
\item Appendix~\ref{section:proof:theorem:velocity} provides the proofs for the results in Section~\ref{section:analysis:velocity}.
\item Appendix~\ref{section:proof:discretization} provides the proofs for the results in Section~\ref{section:analysis:euler}.
\item Appendix~\ref{section:proof:analysis:characteristic} provides the proofs for the results in Section~\ref{section:analysis:characteristic}.
\item Appendix~\ref{section:proof:analysis:manifold} provides the proofs for the results in Section~\ref{section:analysis:manifold}.
\item Appendix~\ref{section:oracle} details a generalization error analysis for nonparametric regression, which is used to establish the oracle inequalities for velocity matching (Appendix~\ref{section:proof:theorem:velocity}) and characteristic fitting (Appendix~\ref{section:proof:analysis:characteristic}).
\item Appendix~\ref{section:appendix:app:lip} establishes approximation results for deep neural networks with a Lipschitz constraint, which are also used for the convergence rates of velocity matching (Appendix~\ref{section:proof:theorem:velocity}) and characteristic fitting (Appendix~\ref{section:proof:analysis:characteristic}).
\item Appendix~\ref{app:denoiser_param} specifies the parameterization of the denoiser employed in Section \ref{section:numerical}.
\item Appendix~\ref{app:ablation} provides ablation studies on the CIFAR10 dataset.
\item Appendix~\ref{sec:extra experiment details} reports the full hyperparameters used in our numerical experiments, including model size, training iterations, batch size, learning rate, optimizer, and augmentation choices.
\item Appendix~\ref{sec:remark time} provides some remarks on the training time consumption of our numerical experiments.
\item Appendix~\ref{section:comparison:ctm} conducts a thorough comparison of the implementation details between our method and CTM.
\item Appendix~\ref{section:comparison} conducts a thorough theoretical comparison between the one-step Euler sampler and characteristic generator.
\end{enumerate}

\section{Related Work}\label{section:related:work}

\subsection{Fast sampling method for diffusion and flow-based models}\label{section:related:work:fast}

\par Diffusion and flow-based models have demonstrated impressive generative performance across various applications. However, their iterative sampling process requires a substantial number of evaluations of the score or velocity neural network, which currently limits their real-time application. In recent years, there has been a surge of fast sampling methods aimed at accelerating the sampling process of diffusion or flow-based models.

\par The sampling process of the diffusion or flow-based model can be considered as numerically solving SDE or ODE. Therefore, one approach to address this issue is to develop acceleration algorithms for these equations~\cite{zhang2023fast,lu2022dpm,lu2023dpmsolver,Zheng2023dpmsolverv3,gao2024convergence,li2024accelerating}. For instance,~\cite{lu2022dpm} effectively utilizes the semi-linear structure of the probability flow ODE by employing an exponential integrator. Furthermore, this algorithm incorporates adaptive step size schedules and high-order approximations. While these strategies can achieve high-quality generation requiring 10-15 neural network evaluations, generating samples in a single step still poses a significant challenge.

\par There is another line of recent works that aim to propose a simulation-free one-step sampling method. It is important to note that SDE has probabilistic trajectories, while the trajectory of ODE is deterministic, which is known as ``self-consistency''~\cite{Song2023Consistency}. This line of work is referred to the distillation, which can be divided into two distinct categories.

\par In the first category, researchers aim to train a deep neural network that maps noise to the endpoint of the probability flow ODE, while disregarding the information at intermediate time points. This class of methods includes knowledge distillation (KD)~\cite{luhman2021knowledge}, Euler particle transport (EPT)~\cite{Gao2022Deep}, rectified flow~\cite{liu2022flow}, and diffusion model sampling with neural operator (DSNO)~\cite{Zheng2023Fast}. Unfortunately, these methods are hindered by training instability and low generation quality, as they solely focus on long-range information and are unable to capture the short-range structure at intermediate time points.

\par The second category of distillation, which is highly relevant to our work, aims to fit the characteristics at each time point using deep neural networks, as demonstrated by~\cite{salimans2022progressive,Song2023Consistency,kim2024consistency,zhou2024score,luo2023diff,yin2024onestep,luo2024one,geng2026mean,frans2025one}.
In detail, KL-based DI~\cite{luo2023diff} introduces Integral Kullback-Leibler (IKL) divergence, which is more robust than the original KL divergence. It derives a clean gradient signal by comparing the score predicted by the pre-trained diffusion model against the score of the student generator. This method is generator-agnostic, as long as the generated samples are differentiable with respect to its model parameters.
DMD~\cite{yin2024onestep} minimizes the KL divergence between the output distribution of the one-step student generator and the reference target distribution of the pre-trained multi-step teacher. It implements a ``fake'' teacher score estimator network trained on student outputs, and pairs it with a trajectory regression loss.
Rather than minimizing standard KL divergence, Fisher divergence-based SiD~\cite{zhou2024score} minimizes the Fisher divergence between the teacher's target distribution and the student generator's output distribution across noisy timesteps. It empolys three score-related identities to approximate the loss. The pre-trained teacher model acts as a score estimator that provides gradient direction, eliminating the need for real training images or offline trajectory datasets.
SiM~\cite{luo2024one} introduces a class of score-based divergences between the score function of the student generator and that of the pre-trained teacher model. It allows arbitrary distance functions between the two score functions, offering optimization control over sample quality versus distribution coverage.
MeanFlow~\cite{geng2026mean} directly models the average velocity across time intervals. It introduces a loss function that enforces the intrinsic mathematical relation between average and instantaneous velocities. MeanFlow can be trained entirely from scratch, requiring no pre-trained teacher models.
Shortcut Models~\cite{frans2025one} parameterize the network not only on time, but also on the desired jump distance, explicitly conditioning the model to learn ``shortcuts'' across the generative trajectory. Instead of distillation from a frozen pre-trained teacher model, it also allows self-distillation.
These methods take into account both the long-range and short-range structures of the original probability flow, resulting in a high quality of one-step generation. Furthermore, these models exhibit the potential for further enhancement through a few-step iteration process. However, despite their impressive generation quality and training stability, these methods have not yet undergone rigorous theoretical analysis. In contrast, our paper presents a comprehensive framework for generative models utilizing characteristic matching and establishes a rigorous convergence analysis, providing theoretical guarantees for these methods. Additionally, we incorporate the exponential integrator into the characteristic matching procedure. Notably, our characteristic generator surpasses the generation quality achieved by~\cite{salimans2022progressive,Song2023Consistency,kim2024consistency} without the assistance of GANs.

\subsection{Error analysis for diffusion and flow-based models}

\par Although a large body of literature has been devoted to the theoretical analysis for diffusion and flow-based generative models, a majority of these works rely on intractable assumptions, such as the regularity of the probability flow SDEs or ODEs. In contrast, our theoretical findings are established under fewer and milder assumptions on the prior and target distribution (Assumptions~\ref{assumption:init:dist:gaussian} and~\ref{assumption:target:dist}).

\par The errors of diffusion-based and flow-based one-step generative models primarily focus on three aspects: velocity matching error, discretization error, and characteristic fitting error. Existing literature on theoretical analysis of these generative models commonly assumes that the $L^{2}$-risk of score or velocity matching is sufficiently small~\cite{Lee2022Convergence,Lee2023Convergence,chen2023sampling,chen2023probability,benton2024linear,benton2024error}. However, only a limited number of studies have specifically investigated the convergence rates of score matching~\cite{Oko2023Diffusion,Chen2023Score,han2024neural} and velocity matching~\cite{chang2024deep,gao2024convergence2,jiao2024convergence}. The convergence rate of the velocity, as derived in Theorem~\ref{theorem:velocity:rate}, achieves minimax optimality under the assumption of Lipschitz continuity of the target function, which improves upon the rates proposed by~\cite{Chen2023Score,chang2024deep}. The discretization error of the numerical sampler has been explored for both diffusion models~\cite{Lee2022Convergence,Lee2023Convergence,chen2023sampling,benton2024linear}, and flow-based methods~\cite{chen2023probability,gao2024convergence,li2024towards,li2024accelerating}. To the best of our knowledge, Theorem~\ref{theorem:error:characteristic} is the first to systematically analyze these three errors, providing theoretical guidance for selecting suitable neural networks and determining the number of numerical discretization steps.

\section{Table of Notations}
\label{section:natation}
Table \ref{tab:notations} summarizes the notations used in Sections~\ref{section:method} and ~\ref{section:analysis} for easy reference and cross
checking.
\begin{longtable}{cl}
\caption{The list of notations used in Sections~\ref{section:method}, and ~\ref{section:analysis}.} \label{tab:notations}\\
\toprule
\textbf{Symbols} & \textbf{Description} \\
\midrule
\endfirsthead
\multicolumn{2}{l}{\small Continued from previous page} \\
\toprule
\textbf{Symbols} & \textbf{Description} \\
\midrule
\endhead
\midrule
\multicolumn{2}{r}{\small Continued on next page} \\
\endfoot
\bottomrule
\endlastfoot

$\mu_{0}$ & Prior distribution, $\mu_{0}=N(0,I_{d})$ (Assumption~\ref{assumption:init:dist:gaussian}). \\
$\mu_{1}$ & Target distribution, $\mu_1 = N(0,\sigma^{2}I_{d})*\nu$ (Assumption~\ref{assumption:target:dist}). \\
$\rho_{0},\rho_{1}$ & Densities of $\mu_{0}$ and $\mu_{1}$. \\
$\mu_{t},\,\rho(t,x)$ & Interpolant distribution and its density, see Prop.~\ref{proposition:transport}. \\
$(g_{t,s}^{*})_{\sharp}\mu_{t}$ & Push-forward of $\mu_{t}$ by the flow $g_{t,s}^{*}$, exactly $\mu_s$ (\S\ref{section:ODE:flow}). \\
$\varphi_{d}$ & Standard Gaussian density in $\bbR^{d}$ (Assumption~\ref{assumption:target:dist}). \\
$\nu$ & Base distribution with $\supp(\nu)\subset[0,1]^{d}$ (Assumption~\ref{assumption:target:dist}). \\
\midrule

$X_{0},X_{1}$ & Independent samples from $\mu_{0}$ and $\mu_{1}$. \\
$X_{t}$ & Stochastic interpolant $X_{t}=\alpha(t)X_{0}+\beta(t)X_{1}$ (Eq.~\eqref{eq:interpolant}). \\
$\alpha(t),\beta(t)$ & Interpolant coefficients (Cond.~\ref{condition:interpolant}); write $\alpha_{t},\beta_{t}$. \\
$s^{*}(t,x)$ & Score $\nabla\log\rho_{t}(x)$ (\S\ref{section:analysis:assumptions}). \\
$b^{*}(t,x)$ & Velocity field (Eq.~\eqref{eq:velocity}); transport Eq.~\eqref{eq:transport}. \\
$g_{t,s}^{*}$ & Continuous probability flow map: $x_{t}\mapsto x_{s}$ (\S\ref{section:ODE:flow}). \\
$Z_{t},Z_{s}$ & Flowed variables $Z_{t}=g_{0,t}^{*}(Z_{0})$, $Z_{s}=g_{0,s}^{*}(Z_{0})$ (\S\ref{section:ODE:flow}). \\
$T$ & Stopping time in $(1/2,1)$ used for training/evaluation. \\
\midrule
$\euS,\,n$ & Training dataset for {denoiser} matching $\{(t^{(i)},X_{0}^{(i)},X_{1}^{(i)})\}_{i=1}^{n}$ and its size. \\
$\calL(D)$ & Population risk for denoiser matching (Eq.~\eqref{eq:velocity:popu:risk}). \\
{$\what{\calL}_{n}(D)$} & {Empirical risk for denoiser matching (Eq.~\eqref{eq:velocity:er}).} \\
{$\scrD$} & {Hypothesis class for denoiser networks (Thm.~\ref{theorem:velocity:rate}).} \\
{$f^{*}$} & {Posterior mean of $U$, the core of the denoiser representation (Prop.~\ref{proposition:reg:denoiser}).} \\
{$\what{D}$} & {Denoiser estimator via ERM (Eq.~\eqref{eq:velocity:erm}).} \\
{$\calE_{T}(\what{D})$} & {Time-averaged denoiser error (Eq.~\eqref{eq:measure:velocity}).} \\
{$\what{\mu}_{K}$} & {Push-forward of $\mu_{0}$ under the exponential integrator $\what{E}_{0,T}$ (\S\ref{section:analysis:euler}).} \\
\midrule
$K,\tau,t_{k}$ & Number of steps, step size $\tau=T/K$, and grid $t_{k}=k\tau$ (\S\ref{section:Euler}). \\
{$\what{E}_{t,s}$} & {Exponential integrator flow map (Eq.~\eqref{eq:characteristic:ei:solution}).} \\
\midrule
$\euZ,\,m$ & Teacher trajectories and their number for fitting (\S\ref{section:characteristic}). \\
$\calR(g)$ & Population characteristic fitting risk (Eq.~\eqref{eq:characteristic:LS}). \\
{$\what{\calR}_{T,m,K}^{\ei}(g)$} & {Empirical characteristic risk using exponential integrator data (Eq.~\eqref{eq:charateristic:er}).} \\
$\scrG$ & Hypothesis class for characteristic networks (Thm.~\ref{theorem:error:characteristic}). \\
$\what{g}_{t,s}$ & Learned characteristic generator (ERM; Eq.~\eqref{eq:charateristic:erm}). \\
$\calD(\what{g})$ & Time-averaged squared Wasserstein metric for $\what{g}$ (Eq.~\eqref{eq:measure:characteristic}). \\
\midrule  {$B_{\vel},B_{\rm deno},B_{\flow},B'_{\flow}$} & {Local bounds for velocity/denoiser/flow and time-derivatives (\S\ref{section:analysis:property}, Prop.~\ref{proposition:reg:denoiser}).} \\
{$G_{\rm vel},\,G_{\rm deno}(t)$} & {Spatial Lipschitz bounds for velocity and denoiser (\S\ref{section:analysis:property}).} \\
{$D_{\rm deno}$} & {Time-regularity constant for $\partial_{t}D^{*}$ (Prop.~\ref{proposition:reg:denoiser}).} \\
$C$ & Universal constants depending on $d,\sigma$ (various theorems). \\
\midrule
$d^{*}$ & Intrinsic dimension with $d^{*}\ll d$ (\S\ref{section:analysis:manifold}). \\
$P\in\bbR^{d\times d^{*}}$ & Column-orthonormal embedding matrix (\S\ref{section:analysis:manifold}).
\\
$\tilde{\nu},\,P_{\sharp}\tilde{\nu}$ & Base distribution on $[0,1]^{d^{*}}$ and its push-forward to $\bbR^d$. \\
$\tilde b^{*},\,\tilde g^{*},\,\tilde D^{*}$ & Intrinsic-coordinate velocity/flow/denoiser (Lemma~\ref{lemma:manifold:decomp:main}). \\
\end{longtable}

\section{Supplemental Definitions and Lemmas}
\label{section:supp:lemmas}
\par In this section, we introduce some supplemental definitions and lemmas that are used in the proofs.

\par We first introduce sub-Gaussian random variable~\cite{vershynin2018High,Wainwright2019high}.

\begin{definition}[Sub-Gaussian]
A random variable $X$ with mean $\mu=\bbE[X]$ is sub-Gaussian if there is a positive number $\sigma$ such that
\begin{equation*}
\log\bbE[\exp(\lambda(X-\mu))]\leq\frac{\sigma^{2}\lambda^{2}}{2}, \quad \lambda\in\bbR.
\end{equation*}
Here the constant $\sigma$ is referred to as the variance proxy.
\end{definition}

\par The following results (Lemmas~\ref{lemma:subGaussian:1} to~\ref{lemma:lin:comb:subgaussian}) shows some important properties of sub-Gaussian variables, whose proofs can be found in~\cite[Section 2.1]{Wainwright2019high}.

\begin{lemma}[Chernoff bound]\label{lemma:subGaussian:1}
Let $X$ be a $\sigma^{2}$-sub-Gaussian random variable with zero mean. Then it holds that
\begin{equation*}
\pr\big\{|X|\geq t\big\}\leq2\exp\Big(-\frac{t^{2}}{2\sigma^{2}}\Big).
\end{equation*}
\end{lemma}

\begin{lemma}\label{lemma:subGaussian:2}
Let $X$ be a $\sigma^{2}$-sub-Gaussian random variable with zero mean. Then it holds that
\begin{equation*}
\bbE\Big[\exp\Big(\frac{\lambda X^{2}}{2\sigma^{2}}\Big)\Big]\leq\frac{1}{\sqrt{1-\lambda}}, \quad \lambda\in[0,1).
\end{equation*}
\end{lemma}

\begin{lemma}\label{lemma:lin:comb:subgaussian}
Let $X_{0}$ be a $\sigma_{0}^{2}$-sub-Gaussian, and let $X_{1}$ be a $\sigma_{1}^{2}$-sub-Gaussian independent of $X_{0}$. Then the random variable $\alpha X_{0}+\beta X_{1}$ is sub-Gaussian with variance proxy $\alpha^{2}\sigma_{0}^{2}+\beta^{2}\sigma_{1}^{2}$.
\end{lemma}

\par Lemmas~\ref{lemma:max:subGaussian:proba} and~\ref{lemma:subgaussian:max:exp} show bounds of the tail probability and the expectation of the maximum of $N$ sub-Gaussian variables.

\begin{lemma}\label{lemma:max:subGaussian:proba}
Let $\{X_{n}\}_{n=1}^{N}$ be a set of $\sigma^{2}$-sub-Gaussian random variables with zero mean, then it follows that
\begin{equation*}
\pr\Big\{\max_{1\leq n\leq N}|X_{n}|\geq t\Big\}\leq 2N\exp\Big(-\frac{t^{2}}{2\sigma^{2}}\Big).
\end{equation*}
\end{lemma}

\par The following lemma states that the expectation of the maximum of the squares of $N$ sub-Gaussian variables is bounded by $\log N$.

\begin{lemma}\label{lemma:subgaussian:max:exp}
Let $\{X_{n}\}_{n=1}^{N}$ be a set of $\sigma^{2}$-sub-Gaussian random variables with zero mean, then it follows that
\begin{equation*}
\bbE\Big[\max_{1\leq n\leq N}X_{n}^{2}\Big]\leq 4\sigma^{2}(\log N+1).
\end{equation*}
\end{lemma}

\begin{proof}[Proof of Lemma~\ref{lemma:max:subGaussian:proba}]
It is straightforward that
\begin{equation*}
\pr\Big\{\max_{1\leq n\leq N}|X_{n}|\geq t\Big\}\leq
\sum_{n=1}^{N}\pr\big\{|X_{n}|\geq t\big\}\leq 2N\exp\Big(-\frac{t^{2}}{2\sigma^{2}}\Big),
\end{equation*}
where the last inequality holds from Lemma~\ref{lemma:subGaussian:1}. This completes the proof.
\end{proof}

\begin{proof}[Proof of Lemma~\ref{lemma:subgaussian:max:exp}]
By Jensen's inequality, it is straightforward that
\begin{equation*}
\exp\Big(\frac{\lambda}{2\sigma^{2}}\bbE\Big[\max_{1\leq n\leq N}\xi_{n}^{2}\Big]\Big)
\leq\bbE\Big[\max_{1\leq n\leq N}\exp\Big(\frac{\lambda\xi_{n}^{2}}{2\sigma^{2}}\Big)\Big]
\leq N\bbE\Big[\exp\Big(\frac{\lambda\xi_{1}^{2}}{2\sigma^{2}}\Big)\Big]
\leq\frac{N}{\sqrt{1-\lambda}},
\end{equation*}
where the last inequality holds from Lemma~\ref{lemma:subGaussian:2} for each $\lambda\in[0,1)$. Letting $\lambda=1/2$ yields the desired inequality.
\end{proof}

\begin{lemma}[Fourth moment of standard Gaussian]\label{lemma:4th:moment:Gaussian}
Let $\epsilon\sim N(0,I_{d})$. Then $\bbE[\|\epsilon\|_{2}^{4}]=d^{2}+2d$.
\end{lemma}

\begin{proof}[Proof of Lemma~\ref{lemma:4th:moment:Gaussian}]
It is straightforward that
\begin{align*}
\bbE[\|\epsilon\|_{2}^{4}]=\bbE\Big[\sum_{k=1}^{d}\epsilon_{k}^{4}+\sum_{k\neq\ell}\epsilon_{k}^{2}\epsilon_{\ell}^{2}\Big]=\sum_{k=1}^{d}\bbE[\epsilon_{k}^{4}]+\sum_{k\neq\ell}\bbE[\epsilon_{k}^{2}]\bbE[\epsilon_{\ell}^{2}]=d^{2}+2d,
\end{align*}
where we used the fact that $\bbE[X^{4}]=3$ for $X\sim N(0,1)$.
\end{proof}

\par We next introduce the notation of covering number and Vapnik-Chervonenkis dimension (VC-dimension), both of which measure the complexity of a function class~\cite{mohri2018foundations,Vaart2023Weak}. They are used in the error analysis for velocity matching (Section~\ref{section:proof:theorem:velocity}) and characteristic fitting (Section~\ref{section:proof:analysis:characteristic}).

\begin{definition}[Covering number]
Let $\scrF$ be a class of functions mapping from $\bbR^{d}$ to $\bbR$. Suppose $\euD=\{X^{(i)}\}_{i=1}^{n}$ is a set of samples in $\bbR^{d}$. Define the $L^{\infty}(\euD)$-norm of the function $f\in\scrF$ as $\|f\|_{L^{\infty}(\euD)}=\max_{1\leq i\leq n}|f(X^{(i)})|$. A function set $\scrF_{\delta}$ is called an $L^{\infty}(\euD)$ $\delta$-cover of $\scrF$ if for each $f\in\scrF$, there exits $f_{\delta}\in\scrF_{\delta}$ such that $\|f-f_{\delta}\|_{L^{\infty}(\euD)}\leq\delta$. Furthermore,
\begin{equation*}
N(\delta,\scrF,L^{\infty}(\euD))=\inf\Big\{|\scrF_{\delta}|:\scrF_{\delta}\text{ is a $L^{\infty}(\euD)$ $\delta$-cover of $\scrF$}\Big\}
\end{equation*}
is called the $L^{\infty}(\euD)$ $\delta$-covering number of $\scrF$.
\end{definition}

\begin{definition}[VC-dimension]\label{def:vcdim}
Let $\scrF$ be a class of functions from $\bbR^{d}$ to $\{\pm1\}$. For any non-negative integer $m$, we define the growth function of $\scrF$ as
\begin{equation*}
\Pi_{\scrF}(m)=\max_{\{X^{(i)}\}_{i=1}^{m}\subset\bbR^{d}}\big|\{(f(X^{(1)}),\ldots,f(X^{(m)})):f\in\scrF\}\big|.
\end{equation*}
A set $\{X^{(i)}\}_{i=1}^{m}$ is said to be shattered by $\scrF$ when $|\{(f(X^{(1)}),\ldots,f(X^{(m)})):f\in\scrF\}|=2^{m}$. The Vapnik-Chervonenkis dimension of $\scrF$, denoted $\vcdim(\scrF)$, is the size of the largest set that can be shattered by $\scrF$, that is, $\vcdim(\scrF)=\max\{m:\Pi_{\scrF}(m)=2^{m}\}$. For a class $\scrF$ of real-valued functions, we define $\vcdim(\scrF)=\vcdim(\sign(\scrF))$.
\end{definition}

\begin{lemma}\label{lemma:covering:lip}
Let $\scrF$ be a class of functions mapping from $\bbR^{d}$ to $\bbR$, and let $\scrH$ be a function class defined as $\scrH=\{(x,f)\mapsto h(f,x)\in\bbR:f\in\scrF\}$. Suppose $\euD=\{X^{(i)}\}_{i=1}^{n}$ is a set of samples in $\bbR^{d}$. If there exists a constant $L>0$ such that for each $f,f^{\prime}\in\scrF$,
\begin{equation*}
\max_{1\leq i\leq n}|h(f,X^{(i)})-h(f^{\prime},X^{(i)})|\leq L\max_{1\leq i\leq n}|f(X^{(i)})-f^{\prime}(X^{(i)})|,
\end{equation*}
then the following inequality holds for each $\delta>0$,
\begin{equation*}
N(L\delta,\scrH,L^{\infty}(\euD))\leq N(\delta,\scrF,L^{\infty}(\euD)).
\end{equation*}
\end{lemma}

\begin{proof}[Proof of Lemma~\ref{lemma:covering:lip}]
Let $\scrF_{\delta}$ be an $L^{\infty}(\euD)$ $\delta$-cover of $\scrF$ with $|\scrF_{\delta}|=N(L\delta,\scrH,L^{\infty}(\euD))$. Define $\scrH_{\delta}=\{(x,f)\mapsto h(f,x)\in\bbR:f\in\scrF_{\delta}\}$. Then for each $h(f,\cdot)\in\scrH$, there exists $h(f_{\delta},\cdot)\in\scrH_{\delta}$, such that
\begin{equation*}
\max_{1\leq i\leq n}|h(f,X^{(i)})-h(f_{\delta},X^{(i)})|\leq L\max_{1\leq i\leq n}|f(X^{(i)})-f_{\delta}(X^{(i)})|\leq L\delta.
\end{equation*}
Thus $\scrH_{\delta}$ is an $L^{\infty}(\euD)$ $(L\delta)$-cover of $\scrH$. This completes the proof.
\end{proof}

\par We then bound the covering number by VC-dimension as following lemma.

\begin{lemma}[{\cite[Theorem 12.2]{Anthony1999neural}}]\label{lemma:covering:vc}
Let $\scrF$ be a class of functions mapping from $\bbR^{d}$ to $[0,B]$. Then it follows that for each $n\geq\vcdim(\scrF)$,
\begin{equation*}
\sup_{\euD\in(\bbR^{d})^{n}}\log N(\delta,\scrF,L^{\infty}(\euD))\leq\vcdim(\scrF)\log\Big(\frac{enB}{\delta\vcdim(\scrF)}\Big).
\end{equation*}
\end{lemma}

\par The following lemma provides a VC-dimension bound for neural network classes with piecewise-polynomial activation functions.

\begin{lemma}[{\cite[Theorem 7]{Bartlett2019nearly}}]\label{lemma:vcdim}
The VC-dimension of a neural network class with piecewise-polynomial activation functions is bounded as follows
\begin{equation*}
\vcdim(N(L,S))\leq CLS\log(S),
\end{equation*}
where $C$ is an absolute constant.
\end{lemma}

\section{Proof of Results in Section~\ref{section:method}}\label{section:proof:method}

\par The proof of Proposition~\ref{proposition:transport} follows from the proof of~\cite[Theorem 2]{albergo2023stochastic} and~\cite[Theorem 2.6]{albergo2023stochasticinterpolants}.

\begin{proof}[Proof of Proposition~\ref{proposition:transport}]
The characteristic function of $X_{t}$~\eqref{eq:interpolant} is given as
\begin{align*}
\varphi_{X_{t}}(\xi)
&=\bbE_{X_{t}}[\exp(i\xi\cdot X_{t})] \\
&=\int_{\bbR^{d}\times\bbR^{d}}\exp(i\xi\cdot(\alpha_{t}x_{0}+\beta_{t}x_{1}))\rho_{0}(x_{0})\rho_{1}(x_{1})\dx_{0}\dx_{1} \\
&=\int_{\bbR^{d}}\exp(i\xi\cdot(\alpha_{t}x_{0}))\rho_{0}(x_{0})\dx_{0}\int_{\bbR^{d}}\exp(i\xi\cdot(\beta_{t}x_{1}))\rho_{1}(x_{1})\dx_{1} \\
&=\bbE_{X_{0}}[\exp(i\alpha_{t}\xi\cdot X_{0})]\bbE_{X_{1}}[\exp(i\beta_{t}\xi\cdot X_{1})]=\varphi_{X_{0}}(\alpha_{t}\xi)\varphi_{X_{1}}(\beta_{t}\xi),
\end{align*}
where $\xi\in\bbR^{d}$, and we used the fact that $X_{0}$ is independent of $X_{1}$. On the other hand,
\begin{align*}
&\int_{\bbR^{d}}\exp(i\xi\cdot x)\Big(\frac{1}{\beta_{t}}\int_{\bbR^{d}}\rho_{0}(x_{0})\rho_{1}\Big(\frac{x-\alpha_{t}x_{0}}{\beta_{t}}\Big)\dx_{0}\Big)\dx \\
&=\frac{1}{\beta_{t}}\int_{\bbR^{d}\times\bbR^{d}}\exp(i\alpha_{t}\xi\cdot x_{0})\exp(i\xi\cdot(x-\alpha_{t}x_{0}))\rho_{0}(x_{0})\rho_{1}\Big(\frac{x-\alpha_{t}x_{0}}{\beta_{t}}\Big)\dx_{0}\dx \\
&=\int_{\bbR^{d}}\exp(i\alpha_{t}\xi\cdot x_{0})\rho_{0}(x_{0})\Big\{\int_{\bbR^{d}}\exp\Big(i\xi\cdot(x-\alpha_{t}x_{0})\Big)\rho_{1}\Big(\frac{x-\alpha_{t}x_{0}}{\beta_{t}}\Big)\d\Big(\frac{x-\alpha_{t}x_{0}}{\beta_{t}}\Big)\Big\}\dx_{0} \\
&=\int_{\bbR^{d}}\exp(i\alpha_{t}\xi\cdot x_{0})\rho_{0}(x_{0})\Big\{\int_{\bbR^{d}}\exp(i\beta_{t}\xi\cdot x_{1})\rho_{1}(x_{1})\dx_{1}\Big\}\dx_{0}=\varphi_{X_{0}}(\alpha_{t}\xi)\varphi_{X_{1}}(\beta_{t}\xi).
\end{align*}
Combining the above two equality deduces that for each $\xi\in\bbR^{d}$,
\begin{equation*}
\varphi_{X_{t}}(\xi)=\int_{\bbR^{d}}\exp(i\xi\cdot x)\Big(\frac{1}{\beta_{t}}\int_{\bbR^{d}}\rho_{0}(x_{0})\rho_{1}\Big(\frac{x-\alpha_{t}x_{0}}{\beta_{t}}\Big)\dx_{0}\Big)\dx.
\end{equation*}
By using Fourier inversion theorem, we obtain the density function of $X_{t}$ as
\begin{equation*}
\rho_{t}(x)=\frac{1}{\beta_{t}}\int_{\bbR^{d}}\rho_{0}(x_{0})\rho_{1}\Big(\frac{x-\alpha_{t}x_{0}}{\beta_{t}}\Big)\dx_{0}.
\end{equation*}
By a same argument, we have
\begin{equation*}
\rho_{t}(x)=\frac{1}{\alpha_{t}}\int_{\bbR^{d}}\rho_{0}\Big(\frac{x-\beta_{t}x_{1}}{\alpha_{t}}\Big)\rho_{1}(x_{1})\dx_{1}.
\end{equation*}

\par We next turn to verify that the density function solves the transport equation. Let $\phi$ be an arbitrary smooth testing function. By the definition of the interpolant~\eqref{eq:interpolant}, it holds that
\begin{equation*}
\bbE_{X_{t}}\big[\phi(X_{t})\big]=\bbE_{(X_{0},X_{1})}\big[\phi(\alpha_{t}X_{0}+\beta_{t}X_{1})\big].
\end{equation*}
Taking derivative with respect to $t$ on the left-hand side of the equality yields
\begin{equation}\label{eq:proof:proposition:transport:2}
\frac{\partial}{\partial t}\Big(\int_{\bbR^{d}}\phi(x)\rho_{t}(x)\dx\Big)=\int_{\bbR^{d}}\phi(x)\partial_{t}\rho_{t}(x)\dx.
\end{equation}
Similarly, for the right-hand side of the equality, we have
\begin{align}
&\frac{\partial}{\partial t}\Big(\int_{\bbR^{d}\times \bbR^{d}}\phi(\alpha_{t}x_{0}+\beta_{t}x_{2})\rho_{0}(x_{0})\rho_{1}(x_{1})\dx_{0}\dx_{1}\Big) \nonumber \\
&=\int_{\bbR^{d}\times \bbR^{d}}(\dot{\alpha}_{t}x_{0}+\dot{\beta}_{t}x_{1})\cdot\nabla\phi(\alpha_{t}x_{0}+\beta_{t}x_{1})\rho_{0}(x_{0})\rho_{1}(x_{1})\dx_{0}\dx_{1} \nonumber \\
&=\int_{\bbR^{d}}\bbE[\dot{\alpha}_{t}X_{0}+\dot{\beta}_{t}X_{1}|X_{t}=x]\cdot\nabla\phi(x)\rho_{t}(x)\dx \nonumber \\
&=\int_{\bbR^{d}}b_{t}^{*}(x)\cdot\nabla\phi(x)\rho_{t}(x)\dx=-\int_{\bbR^{d}}\phi(x)\nabla\cdot(b_{t}^{*}(x)\rho_{t}(x))\dx, \label{eq:proof:proposition:transport:3}
\end{align}
where the first equality follows from the chain rule, the third equality holds from the definition of $b_{t}^{*}$~\eqref{eq:velocity}, and the last equality is due to integration by parts and the divergence theorem~\cite[Theorem 1 in Section C.2]{evans2010partial}. Combining~\eqref{eq:proof:proposition:transport:2} and~\eqref{eq:proof:proposition:transport:3} gives the desired transport equation.
\end{proof}

\section{Properties of the Probability Flow ODE}\label{section:proof:properties}

\par In this section, we present some auxiliary properties of the probability flow ODE in Section~\ref{section:proof:properties:1}. Proofs of Proposition~\ref{proposition:velocity:score} and the propositions in Section~\ref{section:analysis:property} are given in Section~\ref{section:proof:properties:2}.

{
\par Throughout this section, the interpolation coefficients $(\alpha_{t},\beta_{t})$ are assumed to satisfy Condition~\ref{condition:interpolant}, and we write
\begin{equation}\label{eq:Delta}
\Delta_{t}\coloneqq\alpha_{t}^{2}+\sigma^{2}\beta_{t}^{2}, \quad t\in[0,1].
\end{equation}
Since $\Delta_{t}\geq\min\{1,\sigma^{2}\}(\alpha_{t}^{2}+\beta_{t}^{2})$ and $t\mapsto\Delta_{t}$ is continuous on the compact interval $[0,1]$, Condition~\ref{condition:interpolant}~(ii) implies
\begin{equation}\label{eq:delta:0}
\delta_{0}\coloneqq\min_{t\in[0,1]}\Delta_{t}>0.
\end{equation}
Further, Condition~\ref{condition:interpolant}~(iv) ensures that the quantities
\begin{equation}\label{eq:interpolant:constant}
A_{1}\coloneqq\sup_{t\in[0,1]}|\dot{\alpha}_{t}\alpha_{t}|, \quad
A_{2}\coloneqq\sup_{t\in[0,1]}\Big|\frac{\d}{\d t}(\dot{\alpha}_{t}\alpha_{t})\Big|, \quad
B_{1}\coloneqq\sup_{t\in[0,1]}|\dot{\beta}_{t}|, \quad
B_{2}\coloneqq\sup_{t\in[0,1]}|\ddot{\beta}_{t}|
\end{equation}
are all finite. Moreover, Condition~\ref{condition:interpolant}~(i) and~(iii) imply $0\leq\alpha_{t},\beta_{t}\leq1$ for each $t\in[0,1]$. In addition, we assume that $\alpha_{t}>0$ for each $t\in[0,1)$, and $\beta_{t}>0$ for each $t\in(0,1]$, as is the case for both the linear interpolant and the F{\"o}llmer flow. Unless otherwise specified, the constants in this section depend only on $d$, $\sigma$ and the coefficients $(\alpha_{t},\beta_{t})$, where the dependence on the coefficients is only through $\delta_{0}$, $A_{1}$, $A_{2}$, $B_{1}$ and $B_{2}$.
}

\subsection{Auxiliary Properties}\label{section:proof:properties:1}

\par Lemmas~\ref{lemma:conditional:score} to~\ref{lemma:velocity:gradient} are established under Assumption~\ref{assumption:init:dist:gaussian}.

\begin{lemma}\label{lemma:conditional:score}
Suppose Assumption~\ref{assumption:init:dist:gaussian} holds. Then the conditional score function is given as
\begin{equation*}
\nabla_{x}\log\rho_{t|1}(x|x_{1})=-\frac{1}{\alpha_{t}^{2}}x+\frac{\beta_{t}}{\alpha_{t}^{2}}x_{1}, \quad t\in(0,1).
\end{equation*}
\end{lemma}

\begin{proof}[Proof of Lemma~\ref{lemma:conditional:score}]
Given $X_{1}=x_{1}$, using the definition of stochastic interpolant~\eqref{eq:interpolant} implies
\begin{equation*}
(X_{t}|X_{1}=x_{1})\sim N(\beta_{t}x_{1},\alpha_{t}^{2}I_{d}), \quad t\in(0,1),
\end{equation*}
which implies the desired result immediately.
\end{proof}

\begin{lemma}\label{lemma:conditional:exp}
Suppose Assumption~\ref{assumption:init:dist:gaussian} holds. Then the following holds
\begin{equation*}
\bbE[X_{1}|X_{t}=x]=\frac{1}{\beta_{t}}x+\frac{\alpha_{t}^{2}}{\beta_{t}}\nabla_{x}\log\rho_{t}(x).
\end{equation*}
\end{lemma}

\begin{proof}[Proof of Lemma~\ref{lemma:conditional:exp}]
It is straightforward that
\begin{align*}
\nabla_{x}\log\rho_{t}(x)
&=\frac{\nabla_{x}\rho_{t}(x)}{\rho_{t}(x)}=\frac{1}{\rho_{t}(x)}\nabla_{x}\Big(\int_{\bbR^{d}}\rho_{t|1}(x|x_{1})\rho_{1}(x_{1})\dx_{1}\Big) \\
&=\int_{\bbR^{d}}\frac{\rho_{t|1}(x|x_{1})\rho_{1}(x_{1})}{\rho_{t}(x)}\nabla_{x}\log\rho_{t|1}(x|x_{1})\dx_{1} \\
&=\int_{\bbR^{d}}\rho_{1|t}(x_{1}|x)\Big(-\frac{1}{\alpha_{t}^{2}}x+\frac{\beta_{t}}{\alpha_{t}^{2}}x_{1}\Big)\dx_{1} \\
&=-\frac{1}{\alpha_{t}^{2}}x+\frac{\beta_{t}}{\alpha_{t}^{2}}\bbE[X_{1}|X_{t}=x],
\end{align*}
where the we used the definition of conditional density and Lemma~\ref{lemma:conditional:score}. This completes the proof.
\end{proof}

{
\begin{lemma}\label{lemma:velocity:gradient}
Suppose Assumption~\ref{assumption:init:dist:gaussian} holds. Then it follows that
\begin{equation*}
\nabla\bbE[X_{1}|X_{t}=x]=\frac{\beta_{t}}{\alpha_{t}^{2}}\cov(X_{1}|X_{t}=x).
\end{equation*}
\end{lemma}

\begin{proof}[Proof of Lemma~\ref{lemma:velocity:gradient}]
It is straightforward that
\begin{align}
&\nabla_{x}\bbE[X_{1}|X_{t}=x]=\nabla_{x}\Big(\frac{1}{{\rho_{t}(x)}}\int_{\bbR^{d}}x_{1}\rho_{t|1}(x|x_{1})\rho_{1}(x_{1})\dx_{1}\Big) \nonumber \\
&=-\frac{\nabla_{x}\rho_{t}(x)}{{\rho_{t}^{2}(x)}}\int_{\bbR^{d}}x_{1}^{T}\rho_{t|1}(x|x_{1})\rho_{1}(x_{1})\dx_{1}+\frac{1}{{\rho_{t}(x)}}\int_{\bbR^{d}}\nabla_{x}\rho_{t|1}(x|x_{1})x_{1}^{T}\rho_{1}(x_{1})\dx_{1} \nonumber \\
&=-\nabla_{x}\log\rho_{t}(x)\int_{\bbR^{d}}x_{1}^{T}\rho_{1|t}(x_{1}|x)\dx_{1}+\int_{\bbR^{d}}\nabla_{x}\log\rho_{t|1}(x|x_{1})x_{1}^{T}\rho_{1|t}(x_{1}|x)\dx_{1}, \label{eq:proof:lemma:velocity:gradient:2}
\end{align}
where we used the definition of the conditional density that
\begin{equation*}
\rho_{1|t}(x_{1}|x)=\frac{\rho_{t|1}(x|x_{1})\rho_{1}(x_{1})}{\rho_{t}(x)}.
\end{equation*}
For the first term in the right-hand side of~\eqref{eq:proof:lemma:velocity:gradient:2}, applying Lemma~\ref{lemma:conditional:exp} implies
\begin{align}
&-\nabla_{x}\log\rho_{t}(x)\int_{\bbR^{d}}x_{1}^{T}\rho_{1|t}(x_{1}|x)\dx_{1} \nonumber \\
&=\Big(\frac{1}{\alpha_{t}^{2}}x-\frac{\beta_{t}}{\alpha_{t}^{2}}\bbE[X_{1}|X_{t}=x]\Big)\bbE[X_{1}|X_{t}=x]^{T}. \label{eq:proof:lemma:velocity:gradient:3}
\end{align}
For the second term, it follows from Lemma~\ref{lemma:conditional:score} that
\begin{align}
&\int_{\bbR^{d}}\nabla_{x}\log\rho_{t|1}(x|x_{1})x_{1}^{T}\rho_{1|t}(x_{1}|x)\dx_{1} \nonumber \\
&=-\frac{1}{\alpha_{t}^{2}}x\bbE[X_{1}|X_{t}=x]^{T}+\frac{\beta_{t}}{\alpha_{t}^{2}}\bbE[X_{1}X_{1}^{T}|X_{t}=x]. \label{eq:proof:lemma:velocity:gradient:4}
\end{align}
Plugging~\eqref{eq:proof:lemma:velocity:gradient:3} and~\eqref{eq:proof:lemma:velocity:gradient:4} into~\eqref{eq:proof:lemma:velocity:gradient:2} yields
\begin{equation*}
\nabla\bbE[X_{1}|X_{t}=x]=\frac{\beta_{t}}{\alpha_{t}^{2}}\Big(\bbE[X_{1}X_{1}^{T}|X_{t}=x]-\bbE[X_{1}|X_{t}=x]\bbE[X_{1}|X_{t}=x]^{T}\Big)=\frac{\beta_{t}}{\alpha_{t}^{2}}\cov(X_{1}|X_{t}=x).
\end{equation*}
This completes the proof.
\end{proof}
}

{\par Lemmas~\ref{lemma:condition:1t} to~\ref{lemma:conditional:exp:time} are established under Assumptions~\ref{assumption:init:dist:gaussian} and~\ref{assumption:target:dist}. By Assumption~\ref{assumption:target:dist}, write $X_{1}=U+\sigma Z$, where $U\sim\nu$ and $Z\sim N(0,I_{d})$ are independent.}

\begin{lemma}\label{lemma:condition:1t}
{
Suppose Assumptions~\ref{assumption:init:dist:gaussian} and~\ref{assumption:target:dist} hold. Then for each $(t,x)\in(0,1)\times\bbR^{d}$, it follows that
\begin{equation*}
(X_{1}|X_{t}=x)\stackrel{\d}{=}\frac{\alpha_{t}^{2}}{\alpha_{t}^{2}+\sigma^{2}\beta_{t}^{2}}U_{t,x}+\sqrt{\frac{\sigma^{2}\alpha_{t}^{2}}{\alpha_{t}^{2}+\sigma^{2}\beta_{t}^{2}}}\epsilon+\frac{\sigma^{2}\beta_{t}}{\alpha_{t}^{2}+\sigma^{2}\beta_{t}^{2}}x,
\end{equation*}
where $U_{t,x}\sim\rho_{u|t}(\cdot|x)$ is distributed according to the conditional distribution of $U$ given $X_{t}=x$, $\epsilon\sim N(0,I_{d})$, and $U_{t,x}$ and $\epsilon$ are independent. Moreover, $\supp(U_{t,x})=\supp(\nu)$.}
\end{lemma}

{
\begin{proof}[Proof of Lemma~\ref{lemma:condition:1t}]
According to the definition of stochastic interpolant~\eqref{eq:interpolant}, together with Assumption~\ref{assumption:target:dist}, we have
\begin{align*}
X_{t}=\alpha_{t}X_{0}+\beta_{t}X_{1}\stackrel{\d}{=}\alpha_{t}X_{0}+\beta_{t}(U+\sigma Z)=\alpha_{t}X_{0}+\sigma\beta_{t}Z+\beta_{t}U,
\end{align*}
where $U\sim\nu$; $X_{0},Z\sim N(0,I_{d})$; and $U$, $X_{0}$ and $Z$ are independent. Hence, we get
\begin{align}
\label{eq:proof:lemma:condition:1t:1}
&\rho_{t|1}(x|x_{1})=\frac{1}{(2\pi)^{d/2}\alpha_{t}^{d}}\exp\Big(-\frac{1}{2\alpha_{t}^{2}}\|x-\beta_{t}x_{1}\|_{2}^{2}\Big), \\
\label{eq:proof:lemma:condition:1t:2}
&\rho_{1|u}(x_{1}|u)=\frac{1}{(2\pi)^{d/2}\sigma^{d}}\exp\Big(-\frac{1}{2\sigma^{2}}\|x_{1}-u\|_{2}^{2}\Big), \\
\label{eq:proof:lemma:condition:1t:3}
&\rho_{t|u}(x|u)=\frac{1}{(2\pi)^{d/2}(\alpha_{t}^{2}+\sigma^{2}\beta_{t}^{2})^{d/2}}\exp\Big(-\frac{1}{2(\alpha_{t}^{2}+\sigma^{2}\beta_{t}^{2})}\|x-\beta_{t}u\|_{2}^{2}\Big),
\end{align}
where $\rho_{1|u}$ and $\rho_{t|u}$ denote the conditional densities of $X_{1}$ and $X_{t}$ given $U=u$, respectively.
\par We first identify the conditional distribution of $U$ given $X_{t}=x$, denoted by $\rho_{u|t}(\d u|x)$. Using Bayes' rule and~\eqref{eq:proof:lemma:condition:1t:3} implies
\begin{align}
\rho_{u|t}(\d u|x)
=\frac{\rho_{t|u}(x|u)\d\nu(u)}{\int_{\bbR^{d}}\rho_{t|u}(x|u^{\prime})\d\nu(u^{\prime})}
=\frac{\exp\Big(\frac{\beta_{t}}{\alpha_{t}^{2}+\sigma^{2}\beta_{t}^{2}}\langle x,u\rangle-\frac{\beta_{t}^{2}}{2(\alpha_{t}^{2}+\sigma^{2}\beta_{t}^{2})}\|u\|_{2}^{2}\Big)\d\nu(u)}{\int_{\bbR^{d}}\exp\Big(\frac{\beta_{t}}{\alpha_{t}^{2}+\sigma^{2}\beta_{t}^{2}}\langle x,u^{\prime}\rangle-\frac{\beta_{t}^{2}}{2(\alpha_{t}^{2}+\sigma^{2}\beta_{t}^{2})}\|u^{\prime}\|_{2}^{2}\Big)\d\nu(u^{\prime})}, \label{eq:proof:lemma:condition:1t:4}
\end{align}
where the second equality holds by expanding the square in~\eqref{eq:proof:lemma:condition:1t:3} and cancelling the factor $\exp(-\|x\|_{2}^{2}/(2(\alpha_{t}^{2}+\sigma^{2}\beta_{t}^{2})))$, which does not depend on $u$. Observe that the density of $\rho_{u|t}(\d u|x)$ with respect to $\nu$ in~\eqref{eq:proof:lemma:condition:1t:4} is positive and finite for each $u\in\bbR^{d}$. Then the measures $\rho_{u|t}(\cdot|x)$ and $\nu$ are mutually absolutely continuous, which implies
\begin{equation}\label{eq:proof:lemma:condition:1t:5}
\supp(\rho_{u|t}(\cdot|x))=\supp(\nu).
\end{equation}
\par We next turn to the conditional distribution of $X_{1}$ given $X_{t}=x$ and $U=u$. Combining~\eqref{eq:proof:lemma:condition:1t:1},~\eqref{eq:proof:lemma:condition:1t:2} and~\eqref{eq:proof:lemma:condition:1t:3} implies
\begin{align}
\rho_{1|t,u}(x_{1}|x,u)
&=\frac{\rho_{t|1,u}(x|x_{1},u)\rho_{1|u}(x_{1}|u)}{\rho_{t|u}(x|u)}
=\frac{\rho_{t|1}(x|x_{1})\rho_{1|u}(x_{1}|u)}{\rho_{t|u}(x|u)} \nonumber \\
&=\frac{1}{(2\pi)^{d/2}}\Big(\frac{\alpha_{t}^{2}+\sigma^{2}\beta_{t}^{2}}{\sigma^{2}\alpha_{t}^{2}}\Big)^{d/2}\exp\Big(-\frac{\alpha_{t}^{2}+\sigma^{2}\beta_{t}^{2}}{2\sigma^{2}\alpha_{t}^{2}}\Big\|x_{1}-\frac{\sigma^{2}\beta_{t}x+\alpha_{t}^{2}u}{\alpha_{t}^{2}+\sigma^{2}\beta_{t}^{2}}\Big\|_{2}^{2}\Big), \label{eq:proof:lemma:condition:1t:6}
\end{align}
where the first equality holds from Bayes' rule, the second equality holds since $X_{t}=\alpha_{t}X_{0}+\beta_{t}X_{1}$ and $X_{0}$ is independent of $(U,X_{1})$, so that $X_{t}$ and $U$ are conditionally independent given $X_{1}$, and the third equality follows from completing the square with respect to $x_{1}$. In other words,
\begin{equation*}
(X_{1}|X_{t}=x,U=u)\sim N\Big(\frac{\sigma^{2}\beta_{t}x+\alpha_{t}^{2}u}{\alpha_{t}^{2}+\sigma^{2}\beta_{t}^{2}},\frac{\sigma^{2}\alpha_{t}^{2}}{\alpha_{t}^{2}+\sigma^{2}\beta_{t}^{2}}I_{d}\Big).
\end{equation*}
\par Finally, integrating~\eqref{eq:proof:lemma:condition:1t:6} with respect to $\rho_{u|t}(\d u|x)$ shows that the conditional distribution of $X_{1}$ given $X_{t}=x$ coincides with the distribution of
\begin{equation*}
\frac{\alpha_{t}^{2}}{\alpha_{t}^{2}+\sigma^{2}\beta_{t}^{2}}U_{t,x}+\sqrt{\frac{\sigma^{2}\alpha_{t}^{2}}{\alpha_{t}^{2}+\sigma^{2}\beta_{t}^{2}}}\epsilon+\frac{\sigma^{2}\beta_{t}}{\alpha_{t}^{2}+\sigma^{2}\beta_{t}^{2}}x,
\end{equation*}
where $U_{t,x}\sim\rho_{u|t}(\d u|x)$ and $\epsilon\sim N(0,I_{d})$ are two independent random variables. Together with~\eqref{eq:proof:lemma:condition:1t:5}, this completes the proof.
\end{proof}
}

{
\begin{lemma}[Conditional expectation]\label{lemma:conditional:exp:bound}
Suppose Assumptions~\ref{assumption:init:dist:gaussian} and~\ref{assumption:target:dist} hold. Then the following inequalities hold for each $t\in(0,1)$ and $x\in\bbR^{d}$,
\begin{equation*}
|\bbE[X_{1,k}|X_{t}=x]|\leq M(1+|x_{k}|), \quad 1\leq k\leq d,
\end{equation*}
where $M$ is a constant only depending on $\sigma$ and the coefficients $(\alpha_{t},\beta_{t})$.
\end{lemma}

\begin{proof}[Proof of Lemma~\ref{lemma:conditional:exp:bound}]
According to Lemma~\ref{lemma:condition:1t}, it holds that
\begin{equation}\label{eq:proof:lemma:conditional:exp:bound:1}
\bbE[X_{1}|X_{t}=x]=\frac{\alpha_{t}^{2}}{\alpha_{t}^{2}+\sigma^{2}\beta_{t}^{2}}\bbE[U_{t,x}]+\frac{\sigma^{2}\beta_{t}}{\alpha_{t}^{2}+\sigma^{2}\beta_{t}^{2}}x.
\end{equation}
By Assumption~\ref{assumption:target:dist} and Lemma~\ref{lemma:condition:1t}, $U_{t,x}\in\supp(\nu)\subseteq[0,1]^{d}$ almost surely, and thus $|\bbE[U_{t,x,k}]|\leq1$ for each $1\leq k\leq d$. Combining this with $\alpha_{t}^{2}/(\alpha_{t}^{2}+\sigma^{2}\beta_{t}^{2})\leq1$, $0\leq\beta_{t}\leq1$ and~\eqref{eq:delta:0} yields the desired inequalities with $M=\max\{1,\sigma^{2}/\delta_{0}\}$. This completes the proof.
\end{proof}
}

\begin{lemma}[Conditional covariance]\label{lemma:conditional:cov:bound}
Suppose Assumptions~\ref{assumption:init:dist:gaussian} and~\ref{assumption:target:dist} hold. Then the following inequalities hold for each $t\in(0,1)$ and $x\in\bbR^{d}$,
\begin{equation*}
\frac{\sigma^{2}\alpha_{t}^{2}}{\alpha_{t}^{2}+\sigma^{2}\beta_{t}^{2}}I_{d}\preceq\cov(X_{1}|X_{t}=x)\preceq \frac{\sigma^{2}\alpha_{t}^{2}}{\alpha_{t}^{2}+\sigma^{2}\beta_{t}^{2}}I_{d}+d\Big(\frac{\alpha_{t}^{2}}{\alpha_{t}^{2}+\sigma^{2}\beta_{t}^{2}}\Big)^{2}I_{d}.
\end{equation*}
\end{lemma}

{
\begin{proof}[Proof of Lemma~\ref{lemma:conditional:cov:bound}]
It is straightforward from Lemma~\ref{lemma:condition:1t} that
\begin{equation*}
\bbE[X_{1}|X_{t}=x]=\frac{\alpha_{t}^{2}}{\alpha_{t}^{2}+\sigma^{2}\beta_{t}^{2}}\bbE[U_{t,x}]+\frac{\sigma^{2}\beta_{t}}{\alpha_{t}^{2}+\sigma^{2}\beta_{t}^{2}}x,
\end{equation*}
which implies
\begin{align*}
&\bbE[X_{1}|X_{t}=x]\bbE[X_{1}|X_{t}=x]^{T} \\
&=\Big(\frac{\alpha_{t}^{2}}{\alpha_{t}^{2}+\sigma^{2}\beta_{t}^{2}}\Big)^{2}\bbE[U_{t,x}]\bbE[U_{t,x}]^{T}+\frac{\alpha_{t}^{2}}{\alpha_{t}^{2}+\sigma^{2}\beta_{t}^{2}}\frac{\sigma^{2}\beta_{t}}{\alpha_{t}^{2}+\sigma^{2}\beta_{t}^{2}}\Big(\bbE[U_{t,x}]x^{T}+x\bbE[U_{t,x}]^{T}\Big)+\Big(\frac{\sigma^{2}\beta_{t}}{\alpha_{t}^{2}+\sigma^{2}\beta_{t}^{2}}\Big)^{2}xx^{T}.
\end{align*}
On the other hand, using Lemma~\ref{lemma:condition:1t} deduces
\begin{align*}
\bbE[X_{1}X_{1}^{T}|X_{t}=x]
&=\frac{\alpha_{t}^{2}}{\alpha_{t}^{2}+\sigma^{2}\beta_{t}^{2}}\frac{\sigma^{2}\beta_{t}}{\alpha_{t}^{2}+\sigma^{2}\beta_{t}^{2}}\Big(\bbE[U_{t,x}]x^{T}+x\bbE[U_{t,x}]^{T}\Big)+\Big(\frac{\sigma^{2}\beta_{t}}{\alpha_{t}^{2}+\sigma^{2}\beta_{t}^{2}}\Big)^{2}xx^{T} \\
&\quad+\Big(\frac{\alpha_{t}^{2}}{\alpha_{t}^{2}+\sigma^{2}\beta_{t}^{2}}\Big)^{2}\bbE[U_{t,x}U_{t,x}^{T}]+\frac{\sigma^{2}\alpha_{t}^{2}}{\alpha_{t}^{2}+\sigma^{2}\beta_{t}^{2}}I_{d}.
\end{align*}
Combining the above two equalities yields
\begin{align*}
\cov(X_{1}|X_{t}=x)
&=\bbE[X_{1}X_{1}^{T}|X_{t}=x]-\bbE[X_{1}|X_{t}=x]\bbE[X_{1}|X_{t}=x]^{T} \\
&=\Big(\frac{\alpha_{t}^{2}}{\alpha_{t}^{2}+\sigma^{2}\beta_{t}^{2}}\Big)^{2}\cov(U_{t,x})+\frac{\sigma^{2}\alpha_{t}^{2}}{\alpha_{t}^{2}+\sigma^{2}\beta_{t}^{2}}I_{d}.
\end{align*}
According to Lemma~\ref{lemma:condition:1t}, the random variable $\|U_{t,x}\|_{\infty}\leq1$ and thus $\cov(U_{t,x})\preceq dI_{d}$. Consequently,
\begin{equation*}
\frac{\sigma^{2}\alpha_{t}^{2}}{\alpha_{t}^{2}+\sigma^{2}\beta_{t}^{2}}I_{d}\preceq\cov(X_{1}|X_{t}=x)\preceq \frac{\sigma^{2}\alpha_{t}^{2}}{\alpha_{t}^{2}+\sigma^{2}\beta_{t}^{2}}I_{d}+d\Big(\frac{\alpha_{t}^{2}}{\alpha_{t}^{2}+\sigma^{2}\beta_{t}^{2}}\Big)^{2}I_{d},
\end{equation*}
for each $(t,x)\in(0,1)\times\bbR^{d}$. This completes the proof.
\end{proof}
}

{
\par The next lemma concerns the posterior mean of $U$ given $X_{t}=x$, namely the mean of the conditional distribution $\rho_{u|t}(\d u|x)$ in~\eqref{eq:proof:lemma:condition:1t:4}. Define
\begin{equation}\label{eq:pq}
p_{t}\coloneqq\frac{\beta_{t}}{\alpha_{t}^{2}+\sigma^{2}\beta_{t}^{2}}, \quad q_{t}\coloneqq\frac{\beta_{t}^{2}}{\alpha_{t}^{2}+\sigma^{2}\beta_{t}^{2}}, \quad t\in[0,1].
\end{equation}
Since the density in~\eqref{eq:proof:lemma:condition:1t:4} depends on $x$ only through $p_{t}x$, the posterior mean factors through the statistic $p_{t}x$; the resulting function and its uniform regularity are the key to the regularity of the denoiser.

\begin{lemma}[Regularity of the posterior mean]\label{lemma:conditional:exp:time}
Suppose Assumptions~\ref{assumption:init:dist:gaussian} and~\ref{assumption:target:dist} hold. For each $(t,\vartheta)\in[0,1]\times\bbR^{d}$, define
\begin{equation}\label{eq:tilted:measure}
f^{*}(t,\vartheta)\coloneqq\frac{\int_{\bbR^{d}}u\exp\big(\langle\vartheta,u\rangle-\frac{q_{t}}{2}\|u\|_{2}^{2}\big)\d\nu(u)}{\int_{\bbR^{d}}\exp\big(\langle\vartheta,u^{\prime}\rangle-\frac{q_{t}}{2}\|u^{\prime}\|_{2}^{2}\big)\d\nu(u^{\prime})}.
\end{equation}
Then the following holds.
\begin{enumerate}[(a)]
\item For each $(t,x)\in(0,1)\times\bbR^{d}$, the posterior mean of $U$ satisfies $\bbE[U_{t,x}]=f^{*}(t,p_{t}x)$, where $U_{t,x}\sim\rho_{u|t}(\cdot|x)$ is given in Lemma~\ref{lemma:condition:1t}. Moreover, $\|f^{*}(t,\vartheta)\|_{\infty}\leq1$ for each $(t,\vartheta)\in[0,1]\times\bbR^{d}$.
\item The map $f^{*}$ is differentiable on $(0,1)\times\bbR^{d}$.
\item For each $(t,\vartheta)\in(0,1)\times\bbR^{d}$ and $1\leq k,\ell\leq d$,
\begin{align}
&|\partial_{\vartheta_{\ell}}f_{k}^{*}(t,\vartheta)|\leq1, \quad
\|\nabla f^{*}(t,\vartheta)\|_{\op}\leq d, \notag \\
&|\partial_{t}f_{k}^{*}(t,\vartheta)|\leq\frac{F_{\rm deno}}{\sqrt{d}}, \quad
\|\partial_{t}f^{*}(t,\vartheta)\|_{2}\leq F_{\rm deno}, \label{eq:lemma:conditional:exp:time:bound}
\end{align}
where $F_{\rm deno}\coloneqq(A_{1}+B_{1})d^{3/2}/\delta_{0}^{2}$.
\end{enumerate}
\end{lemma}

\begin{proof}[Proof of Lemma~\ref{lemma:conditional:exp:time}]
Throughout the proof, write
\begin{equation}\label{eq:proof:lemma:conditional:exp:time:1}
f^{*}(t,\vartheta)=\frac{F(t,\vartheta)}{Z(t,\vartheta)}, \quad
F(t,\vartheta)\coloneqq\int_{\bbR^{d}}u\exp(\eta(t,\vartheta,u))\d\nu(u), \quad
Z(t,\vartheta)\coloneqq\int_{\bbR^{d}}\exp(\eta(t,\vartheta,u))\d\nu(u),
\end{equation}
where $\eta(t,\vartheta,u)\coloneqq\langle\vartheta,u\rangle-\frac{q_{t}}{2}\|u\|_{2}^{2}$, which is exactly~\eqref{eq:tilted:measure}. The normalized integrand defines the probability measure $\nu_{t,\vartheta}(\d u)\coloneqq\exp(\eta(t,\vartheta,u))\d\nu(u)/Z(t,\vartheta)$, and we write $U_{t,\vartheta}\sim\nu_{t,\vartheta}$, so that $f^{*}(t,\vartheta)=\bbE[U_{t,\vartheta}]$.
\par We first prove part~(a). By~\eqref{eq:pq}, the density of the conditional distribution $\rho_{u|t}(\cdot|x)$ with respect to $\nu$ in~\eqref{eq:proof:lemma:condition:1t:4} is proportional to $\exp(\langle p_{t}x,u\rangle-\frac{q_{t}}{2}\|u\|_{2}^{2})$, whence $\rho_{u|t}(\cdot|x)=\nu_{t,p_{t}x}$ and $\bbE[U_{t,x}]=f^{*}(t,p_{t}x)$ for each $(t,x)\in(0,1)\times\bbR^{d}$. Moreover, the density of $\nu_{t,\vartheta}$ with respect to $\nu$ is positive and finite for each $u\in\bbR^{d}$. Then the measures $\nu_{t,\vartheta}$ and $\nu$ are mutually absolutely continuous, which implies $U_{t,\vartheta}\in\supp(\nu)\subseteq[0,1]^{d}$ almost surely by Assumption~\ref{assumption:target:dist}, and thus $\|f^{*}(t,\vartheta)\|_{\infty}\leq1$. This proves part~(a).
\par We next prove part~(b). We first compute the time derivatives of the coefficients~\eqref{eq:pq}. By the quotient rule and $\dot{\Delta}_{t}=2\dot{\alpha}_{t}\alpha_{t}+2\sigma^{2}\dot{\beta}_{t}\beta_{t}$, the coefficients $p_{t}$ and $q_{t}$ are differentiable on $(0,1)$ with
\begin{align}
&\dot{p}_{t}=\frac{\dot{\beta}_{t}\Delta_{t}-\beta_{t}\dot{\Delta}_{t}}{\Delta_{t}^{2}}
=\frac{\dot{\beta}_{t}\alpha_{t}^{2}-2\dot{\alpha}_{t}\alpha_{t}\beta_{t}-\sigma^{2}\dot{\beta}_{t}\beta_{t}^{2}}{(\alpha_{t}^{2}+\sigma^{2}\beta_{t}^{2})^{2}}, \notag \\
&\dot{q}_{t}=\frac{2\dot{\beta}_{t}\beta_{t}\Delta_{t}-\beta_{t}^{2}\dot{\Delta}_{t}}{\Delta_{t}^{2}}
=\frac{2\beta_{t}(\dot{\beta}_{t}\alpha_{t}^{2}-\dot{\alpha}_{t}\alpha_{t}\beta_{t})}{(\alpha_{t}^{2}+\sigma^{2}\beta_{t}^{2})^{2}}. \label{eq:pq:dot}
\end{align}
Observe that $\dot{\alpha}_{t}$ appears in~\eqref{eq:pq:dot} only through the product $\dot{\alpha}_{t}\alpha_{t}$. Hence it follows from~\eqref{eq:delta:0},~\eqref{eq:interpolant:constant} and $0\leq\alpha_{t},\beta_{t}\leq1$ that
\begin{equation}\label{eq:proof:lemma:conditional:exp:time:2}
|\dot{p}_{t}|\leq\frac{(1+\sigma^{2})B_{1}+2A_{1}}{\delta_{0}^{2}}, \quad |\dot{q}_{t}|\leq\frac{2(A_{1}+B_{1})}{\delta_{0}^{2}}, \quad t\in(0,1).
\end{equation}
Similarly, $p_{t}\leq1/\delta_{0}$ and $q_{t}\leq1/\delta_{0}$ for each $t\in[0,1]$.
\par We next differentiate~\eqref{eq:proof:lemma:conditional:exp:time:1}. Since $\supp(\nu)\subseteq[0,1]^{d}$ by Assumption~\ref{assumption:target:dist}, it holds that $\|u\|_{2}\leq\sqrt{d}$ and $\eta(t,\vartheta,u)\leq\sqrt{d}\|\vartheta\|_{2}$ for $\nu$-almost every $u$, and thus
\begin{align*}
&|\partial_{t}\exp(\eta(t,\vartheta,u))|=\frac{|\dot{q}_{t}|}{2}\|u\|_{2}^{2}\exp(\eta(t,\vartheta,u))\leq\frac{d|\dot{q}_{t}|}{2}\exp\big(\sqrt{d}\|\vartheta\|_{2}\big), \\
&\|\nabla_{\vartheta}\exp(\eta(t,\vartheta,u))\|_{2}=\|u\|_{2}\exp(\eta(t,\vartheta,u))\leq\sqrt{d}\exp\big(\sqrt{d}\|\vartheta\|_{2}\big),
\end{align*}
which are bounded uniformly in $t\in(0,1)$ for each fixed $\vartheta$ by~\eqref{eq:proof:lemma:conditional:exp:time:2}. Then the dominated convergence theorem permits differentiation under the integral sign in~\eqref{eq:proof:lemma:conditional:exp:time:1} with respect to both $\vartheta$ and $t$. Since $Z(t,\vartheta)\geq\exp(-\sqrt{d}\|\vartheta\|_{2}-dq_{t}/2)>0$, the map $f^{*}=F/Z$ is differentiable on $(0,1)\times\bbR^{d}$, which proves part~(b).
\par It remains to prove part~(c). Using the quotient rule yields
\begin{align*}
\nabla f^{*}(t,\vartheta)
=\frac{\nabla F(t,\vartheta)}{Z(t,\vartheta)}-\frac{F(t,\vartheta)(\nabla Z(t,\vartheta))^{T}}{Z^{2}(t,\vartheta)}
=\bbE\big[U_{t,\vartheta}U_{t,\vartheta}^{T}\big]-\bbE[U_{t,\vartheta}]\bbE[U_{t,\vartheta}]^{T}=\cov(U_{t,\vartheta}),
\end{align*}
and similarly,
\begin{align*}
\partial_{t}f^{*}(t,\vartheta)
=\frac{\partial_{t}F(t,\vartheta)}{Z(t,\vartheta)}-\frac{F(t,\vartheta)\partial_{t}Z(t,\vartheta)}{Z^{2}(t,\vartheta)}
=-\frac{\dot{q}_{t}}{2}\Big(\bbE\big[U_{t,\vartheta}\|U_{t,\vartheta}\|_{2}^{2}\big]-\bbE[U_{t,\vartheta}]\bbE\big[\|U_{t,\vartheta}\|_{2}^{2}\big]\Big),
\end{align*}
In other words,
\begin{equation}\label{eq:lemma:conditional:exp:time}
\nabla f^{*}(t,\vartheta)=\cov(U_{t,\vartheta}), \quad
\partial_{t}f^{*}(t,\vartheta)=-\frac{\dot{q}_{t}}{2}\Big(\bbE\big[U_{t,\vartheta}\|U_{t,\vartheta}\|_{2}^{2}\big]-\bbE[U_{t,\vartheta}]\bbE\big[\|U_{t,\vartheta}\|_{2}^{2}\big]\Big).
\end{equation}
Since $U_{t,\vartheta}\in[0,1]^{d}$ almost surely, the Cauchy-Schwarz inequality implies that, for each $1\leq k,\ell\leq d$,
\begin{equation*}
|\partial_{\vartheta_{\ell}}f_{k}^{*}(t,\vartheta)|=\big|\cov(U_{t,\vartheta,k},U_{t,\vartheta,\ell})\big|\leq\bbE^{1/2}\big[U_{t,\vartheta,k}^{2}\big]\bbE^{1/2}\big[U_{t,\vartheta,\ell}^{2}\big]\leq1.
\end{equation*}
Moreover, $\|U_{t,\vartheta}\|_{\infty}\leq1$ implies $\cov(U_{t,\vartheta})\preceq dI_{d}$ as in the proof of Lemma~\ref{lemma:conditional:cov:bound}; since the matrix $\cov(U_{t,\vartheta})$ is symmetric and positive semi-definite, it follows from~\eqref{eq:lemma:conditional:exp:time} that $\|\nabla f^{*}(t,\vartheta)\|_{\op}\leq d$. For the time derivative, the Cauchy-Schwarz inequality gives, for each $1\leq k\leq d$,
\begin{align*}
\Big|\bbE\big[U_{t,\vartheta,k}\|U_{t,\vartheta}\|_{2}^{2}\big]-\bbE[U_{t,\vartheta,k}]\bbE\big[\|U_{t,\vartheta}\|_{2}^{2}\big]\Big|
&=\Big|\bbE\Big[\big(U_{t,\vartheta,k}-\bbE[U_{t,\vartheta,k}]\big)\big(\|U_{t,\vartheta}\|_{2}^{2}-\bbE[\|U_{t,\vartheta}\|_{2}^{2}]\big)\Big]\Big| \\
&\leq\Big(\bbE\big[U_{t,\vartheta,k}^{2}\big]\Big)^{1/2}\Big(\bbE\big[\|U_{t,\vartheta}\|_{2}^{4}\big]\Big)^{1/2}\leq d,
\end{align*}
where the last inequality used $|U_{t,\vartheta,k}|\leq1$ and $\|U_{t,\vartheta}\|_{2}^{2}\leq d$ almost surely. Combining this with~\eqref{eq:lemma:conditional:exp:time} and~\eqref{eq:proof:lemma:conditional:exp:time:2} yields
\begin{equation*}
|\partial_{t}f_{k}^{*}(t,\vartheta)|\leq\frac{|\dot{q}_{t}|}{2}\,d\leq\frac{(A_{1}+B_{1})d}{\delta_{0}^{2}}=\frac{F_{\rm deno}}{\sqrt{d}}, \quad
\|\partial_{t}f^{*}(t,\vartheta)\|_{2}\leq\frac{|\dot{q}_{t}|}{2}\,d^{3/2}\leq F_{\rm deno},
\end{equation*}
which proves part~(c). This completes the proof.
\end{proof}
}

\subsection{Proof of Results in Section~\ref{section:analysis:property}}\label{section:proof:properties:2}

{\par Then we show proofs of Proposition~\ref{proposition:velocity:score} and the propositions in Section~\ref{section:analysis:property}.}

{
\begin{proof}[Proof of Proposition~\ref{proposition:velocity:score}]
Using the definition of velocity~\eqref{eq:velocity}, we have
\begin{align}
b^{*}(t,x)
&=\bbE[\dot{\alpha}_{t}X_{0}+\dot{\beta}_{t}X_{1}|X_{t}=x]=\bbE\Big[\dot{\beta}_{t}X_{1}+\frac{\dot{\alpha}_{t}}{\alpha_{t}}(X_{t}-\beta_{t}X_{1})\Big|X_{t}=x\Big] \nonumber \\
&=\beta_{t}\Big(\frac{\dot{\beta}_{t}}{\beta_{t}}-\frac{\dot{\alpha}_{t}}{\alpha_{t}}\Big)\bbE[X_{1}|X_{t}=x]+\frac{\dot{\alpha}_{t}}{\alpha_{t}}x, \label{eq:b*}
\end{align}
where the second equality holds from the definition of stochastic interpolant~\eqref{eq:interpolant}. Then applying Lemma~\ref{lemma:conditional:exp} to~\eqref{eq:b*} implies
\begin{equation*}
b^{*}(t,x)=\alpha_{t}^{2}\Big(\frac{\dot{\beta}_{t}}{\beta_{t}}-\frac{\dot{\alpha}_{t}}{\alpha_{t}}\Big)\nabla_{x}\log\rho_{t}(x)+\frac{\dot{\beta}_{t}}{\beta_{t}}x,
\end{equation*}
which shows the first claim. Moreover, combining the definition of the denoiser~\eqref{eq:denoiser} with Lemma~\ref{lemma:conditional:exp} yields
\begin{equation*}
D^{*}(t,x)=\bbE[X_{1}|X_{t}=x]=\frac{1}{\beta_{t}}\big(x+\alpha_{t}^{2}\nabla_{x}\log\rho_{t}(x)\big),
\end{equation*}
which is the Tweedie formula. This completes the proof.
\end{proof}
}

{
\begin{proof}[Proof of Proposition~\ref{proposition:reg:velocity}]
We first prove part~(a). Using the definition of velocity~\eqref{eq:velocity}, we have
\begin{align}
b^{*}(t,x)
&=\bbE[\dot{\alpha}_{t}X_{0}+\dot{\beta}_{t}X_{1}|X_{t}=x]=\bbE\Big[\dot{\beta}_{t}X_{1}+\frac{\dot{\alpha}_{t}}{\alpha_{t}}(X_{t}-\beta_{t}X_{1})\Big|X_{t}=x\Big] \nonumber \\
&=\beta_{t}\Big(\frac{\dot{\beta}_{t}}{\beta_{t}}-\frac{\dot{\alpha}_{t}}{\alpha_{t}}\Big)\bbE[X_{1}|X_{t}=x]+\frac{\dot{\alpha}_{t}}{\alpha_{t}}x \nonumber \\
&=\beta_{t}\Big(\frac{\dot{\beta}_{t}}{\beta_{t}}-\frac{\dot{\alpha}_{t}}{\alpha_{t}}\Big)\Big(\frac{\alpha_{t}^{2}}{\alpha_{t}^{2}+\sigma^{2}\beta_{t}^{2}}\bbE[U_{t,x}]+\frac{\sigma^{2}\beta_{t}}{\alpha_{t}^{2}+\sigma^{2}\beta_{t}^{2}}x\Big)+\frac{\dot{\alpha}_{t}}{\alpha_{t}}x \nonumber \\
&=\frac{\alpha_{t}^{2}\dot{\beta}_{t}-\dot{\alpha}_{t}\alpha_{t}\beta_{t}}{\alpha_{t}^{2}+\sigma^{2}\beta_{t}^{2}}\bbE[U_{t,x}]+\frac{\dot{\alpha}_{t}\alpha_{t}+\sigma^{2}\dot{\beta}_{t}\beta_{t}}{\alpha_{t}^{2}+\sigma^{2}\beta_{t}^{2}}x, \label{eq:proof:proposition:reg:velocity:1}
\end{align}
where the second equality holds from the definition of stochastic interpolant~\eqref{eq:interpolant}, the fourth equality follows from Lemma~\ref{lemma:condition:1t}, and the last equality holds by a direct calculation. By~\eqref{eq:delta:0},~\eqref{eq:interpolant:constant} and $0\leq\alpha_{t},\beta_{t}\leq1$, the coefficients in~\eqref{eq:proof:proposition:reg:velocity:1} satisfy
\begin{equation}\label{eq:proof:proposition:reg:velocity:2}
\Big|\frac{\alpha_{t}^{2}\dot{\beta}_{t}-\dot{\alpha}_{t}\alpha_{t}\beta_{t}}{\alpha_{t}^{2}+\sigma^{2}\beta_{t}^{2}}\Big|\leq\frac{A_{1}+B_{1}}{\delta_{0}}, \quad
\Big|\frac{\dot{\alpha}_{t}\alpha_{t}+\sigma^{2}\dot{\beta}_{t}\beta_{t}}{\alpha_{t}^{2}+\sigma^{2}\beta_{t}^{2}}\Big|\leq\frac{A_{1}+\sigma^{2}B_{1}}{\delta_{0}}, \quad t\in(0,1),
\end{equation}
where $\dot{\alpha}_{t}$ appears only through the product $\dot{\alpha}_{t}\alpha_{t}$. Since $|\bbE[U_{t,x,k}]|\leq1$ for each $1\leq k\leq d$, it follows that for each $(t,x)\in(0,1)\times\bbB_{R}^{\infty}$,
\begin{equation*}
|b_{k}^{*}(t,x)|\leq\frac{A_{1}+B_{1}}{\delta_{0}}+\frac{A_{1}+\sigma^{2}B_{1}}{\delta_{0}}R\leq B_{\vel}R, \quad B_{\vel}\coloneqq\frac{2A_{1}+(1+\sigma^{2})B_{1}}{\delta_{0}},
\end{equation*}
where the last inequality used $R>1$. The cases $t\in\{0,1\}$ follow since the coefficients in~\eqref{eq:proof:proposition:reg:velocity:1} extend continuously to $[0,1]$ by Condition~\ref{condition:interpolant}~(iv). This proves part~(a).
\par We next prove part~(b). It follows from~\eqref{eq:b*} and Lemma~\ref{lemma:velocity:gradient} that
\begin{equation}\label{eq:proof:proposition:reg:velocity:3}
\nabla b^{*}(t,x)=\frac{\dot{\alpha}_{t}}{\alpha_{t}}I_{d}+\Big(\frac{\dot{\beta}_{t}}{\beta_{t}}-\frac{\dot{\alpha}_{t}}{\alpha_{t}}\Big)\frac{\beta_{t}^{2}}{\alpha_{t}^{2}}\cov(X_{1}|X_{t}=x).
\end{equation}
Notice that $\dot{\beta}_{t}\geq0$ and $\dot{\alpha}_{t}\leq0$ by Condition~\ref{condition:interpolant}~(iii), and thus the factor in front of the conditional covariance in~\eqref{eq:proof:proposition:reg:velocity:3} is non-negative. For the upper bound, combining~\eqref{eq:proof:proposition:reg:velocity:3} with Lemma~\ref{lemma:conditional:cov:bound} implies
\begin{align}
\nabla b^{*}(t,x)
&\preceq\frac{\dot{\alpha}_{t}}{\alpha_{t}}I_{d}+\Big(\frac{\dot{\beta}_{t}}{\beta_{t}}-\frac{\dot{\alpha}_{t}}{\alpha_{t}}\Big)\frac{\beta_{t}^{2}}{\alpha_{t}^{2}}\Big(\frac{\sigma^{2}\alpha_{t}^{2}}{\alpha_{t}^{2}+\sigma^{2}\beta_{t}^{2}}I_{d}+d\Big(\frac{\alpha_{t}^{2}}{\alpha_{t}^{2}+\sigma^{2}\beta_{t}^{2}}\Big)^{2}I_{d}\Big) \nonumber \\
&=\Big(\frac{\dot{\alpha}_{t}\alpha_{t}+\sigma^{2}\dot{\beta}_{t}\beta_{t}}{\alpha_{t}^{2}+\sigma^{2}\beta_{t}^{2}}+d\frac{\alpha_{t}\beta_{t}(\alpha_{t}\dot{\beta}_{t}-\dot{\alpha}_{t}\beta_{t})}{(\alpha_{t}^{2}+\sigma^{2}\beta_{t}^{2})^{2}}\Big)I_{d}. \label{eq:proof:proposition:reg:velocity:4}
\end{align}
By a similar argument, we have
\begin{equation}\label{eq:proof:proposition:reg:velocity:5}
\nabla b^{*}(t,x)\succeq\frac{\dot{\alpha}_{t}}{\alpha_{t}}I_{d}+\Big(\frac{\dot{\beta}_{t}}{\beta_{t}}-\frac{\dot{\alpha}_{t}}{\alpha_{t}}\Big)\frac{\beta_{t}^{2}}{\alpha_{t}^{2}}\frac{\sigma^{2}\alpha_{t}^{2}}{\alpha_{t}^{2}+\sigma^{2}\beta_{t}^{2}}I_{d}=\frac{\dot{\alpha}_{t}\alpha_{t}+\sigma^{2}\dot{\beta}_{t}\beta_{t}}{\alpha_{t}^{2}+\sigma^{2}\beta_{t}^{2}}I_{d}.
\end{equation}
Since the matrix $\nabla b^{*}(t,x)$ is symmetric by~\eqref{eq:proof:proposition:reg:velocity:3}, combining~\eqref{eq:proof:proposition:reg:velocity:4} and~\eqref{eq:proof:proposition:reg:velocity:5} with~\eqref{eq:delta:0},~\eqref{eq:interpolant:constant} and $0\leq\alpha_{t},\beta_{t}\leq1$ yields
\begin{equation}\label{eq:proof:proposition:reg:velocity:6}
\|\nabla b^{*}(t,x)\|_{\op}\leq\Big|\frac{\dot{\alpha}_{t}\alpha_{t}+\sigma^{2}\dot{\beta}_{t}\beta_{t}}{\alpha_{t}^{2}+\sigma^{2}\beta_{t}^{2}}\Big|+d\Big|\frac{\alpha_{t}\beta_{t}(\alpha_{t}\dot{\beta}_{t}-\dot{\alpha}_{t}\beta_{t})}{(\alpha_{t}^{2}+\sigma^{2}\beta_{t}^{2})^{2}}\Big|\leq\frac{A_{1}+\sigma^{2}B_{1}}{\delta_{0}}+\frac{d(A_{1}+B_{1})}{\delta_{0}^{2}},
\end{equation}
for each $(t,x)\in(0,1)\times\bbR^{d}$. The cases $t\in\{0,1\}$ follow by the same continuity argument as in part~(a); hence $\|\nabla b^{*}(t,x)\|_{\op}\leq G_{\rm vel}$ for each $(t,x)\in[0,1]\times\bbR^{d}$, where $G_{\rm vel}\coloneqq(A_{1}+\sigma^{2}B_{1})/\delta_{0}+d(A_{1}+B_{1})/\delta_{0}^{2}$.
\par We now deduce the Lipschitz continuity in part~(b). It is straightforward that for each $t\in[0,1]$ and $x,x^{\prime}\in\bbR^{d}$,
\begin{align}
\|b^{*}(t,x)-b^{*}(t,x^{\prime})\|_{2}
&=\Big\|\int_{0}^{1}\frac{\d}{\d\tau}b^{*}(t,x^{\prime}+\tau(x-x^{\prime}))\d\tau\Big\|_{2} \nonumber \\
&=\Big\|\int_{0}^{1}\nabla b^{*}(t,x^{\prime}+\tau(x-x^{\prime}))\d\tau(x-x^{\prime})\Big\|_{2} \nonumber \\
&\leq\int_{0}^{1}\|\nabla b^{*}(t,x^{\prime}+\tau(x-x^{\prime}))\|_{\op}\d\tau\|x-x^{\prime}\|_{2} \nonumber \\
&\leq\Big(\int_{0}^{1}G_{\rm vel}\d\tau\Big)\|x-x^{\prime}\|_{2}=G_{\rm vel}\|x-x^{\prime}\|_{2}, \label{eq:proof:proposition:reg:velocity:7}
\end{align}
where the first inequality follows from the definition of the operator norm $\|\cdot\|_{\op}$ and Jensen's inequality, and the second inequality is due to~\eqref{eq:proof:proposition:reg:velocity:6}. This proves part~(b), and completes the proof.
\end{proof}
}

{
To prove Proposition \ref{proposition:reg:denoiser}, we first need the following lemma.
\begin{lemma}[Regularity of the denoiser]\label{lemma:reg:denoiser}
Suppose Assumptions~\ref{assumption:init:dist:gaussian} and~\ref{assumption:target:dist} hold. Define the preconditioning coefficients, for $t\in[0,1]$,
\begin{equation}\label{eq:denoiser:precondition}
a_{t}=\frac{\alpha_{t}^{2}}{\Delta_{t}}, \quad
c_{t}=\frac{\sigma^{2}\beta_{t}}{\Delta_{t}}, \quad
p_{t}=\frac{\beta_{t}}{\Delta_{t}}, \quad t\in[0,1].
\end{equation}
In particular, $p_{t}$ coincides with the coefficient defined in~\eqref{eq:pq}. Then for each $(t,x)\in(0,1)\times\bbR^{d}$, the denoiser admits the representation
\begin{equation}\label{eq:denoiser:representation}
D^{*}(t,x)=c_{t}x+a_{t}f^{*}(t,p_{t}x),
\end{equation}
where the function $f^{*}$ satisfies, for each $(t,x)\in(0,1)\times\bbR^{d}$,
\begin{align}
\|f^{*}(t,x)\|_{\infty}\leq1, \quad \|\nabla f^{*}(t,x)\|_{\op}\leq d, \quad \|\partial_{t}f^{*}(t,x)\|_{2}\leq F_{\rm deno}. \label{eq:reg:core}
\end{align}
Here the constant $F_{\rm deno}$ is from Lemma \ref{lemma:conditional:exp:time}(c).
\end{lemma}
}

{
\begin{proof}[Proof of Proposition~\ref{lemma:reg:denoiser}]
We first establish the representation~\eqref{eq:denoiser:representation}. According to the definition of the denoiser~\eqref{eq:denoiser}, Lemma~\ref{lemma:condition:1t} and~\eqref{eq:denoiser:precondition},
\begin{equation}\label{eq:proof:lemma:reg:denoiser:1}
D^{*}(t,x)=\bbE[X_{1}|X_{t}=x]=a_{t}\bbE[U_{t,x}]+c_{t}x, \quad t\in(0,1).
\end{equation}
By part~(a) of Lemma~\ref{lemma:conditional:exp:time}, $\bbE[U_{t,x}]=f^{*}(t,p_{t}x)$ with $f^{*}$ defined therein, which proves~\eqref{eq:denoiser:representation}. The regularity~\eqref{eq:reg:core} of $f^{*}$ is exactly parts~(a) and~(c) of Lemma~\ref{lemma:conditional:exp:time}. This completes the proof.
\end{proof}
}

{
\par Now, we can prove Proposition~\ref{proposition:reg:denoiser}.

\begin{proof}[Proof of Proposition~\ref{proposition:reg:denoiser}]
Part~(a) follows from the representation~\eqref{eq:denoiser:representation}, the bound $\|f^{*}\|_{\infty}\leq1$ in~\eqref{eq:reg:core}, and $a_{t}\leq1$, $c_{t}\leq\sigma^{2}/\delta_{0}$ (by~\eqref{eq:delta:0} and $0\leq\alpha_{t},\beta_{t}\leq1$), which give $|D_{k}^{*}(t,x)|\leq1+(\sigma^{2}/\delta_{0})R\leq B_{\rm deno}R$ with $B_{\rm deno}\coloneqq1+\sigma^{2}/\delta_{0}$, using $R>1$; at the endpoints, $D^{*}(0,\cdot)\equiv\bbE[X_{1}]=\bbE[U]\in[0,1]^{d}$ and $D^{*}(1,\cdot)$ is the identity map, so the same bound holds trivially. Part~(b) follows by differentiating~\eqref{eq:denoiser:representation} with respect to $x$, which gives $\nabla D^{*}(t,x)=c_{t}I_{d}+a_{t}p_{t}\nabla f^{*}(t,p_{t}x)$, together with $\|\nabla f^{*}\|_{\op}\leq d$ in~\eqref{eq:reg:core}, whence $\|\nabla D^{*}(t,x)\|_{\op}\leq c_{t}+d\,a_{t}p_{t}=G_{\rm deno}(t)$.
\par For part~(c), by the quotient rule, in which $\dot{\alpha}_{t}$ appears only through the product $\dot{\alpha}_{t}\alpha_{t}$ as in the proof of Lemma~\ref{lemma:conditional:exp:time}, the coefficients~\eqref{eq:denoiser:precondition} are differentiable on $(0,1)$ and satisfy
\begin{equation}\label{eq:proof:proposition:reg:denoiser:6}
|\dot{a}_{t}|\leq\frac{2\sigma^{2}(A_{1}+B_{1})}{\delta_{0}^{2}}, \quad
|\dot{c}_{t}|\leq\frac{\sigma^{2}((1+\sigma^{2})B_{1}+2A_{1})}{\delta_{0}^{2}}, \quad t\in(0,1).
\end{equation}
Then differentiating~\eqref{eq:denoiser:representation} with respect to $t$ by the chain rule and part~(b) of Lemma~\ref{lemma:conditional:exp:time} yields
\begin{equation}\label{eq:proof:proposition:reg:denoiser:7}
\partial_{t}D^{*}(t,x)=\dot{c}_{t}x+\dot{a}_{t}f^{*}(t,p_{t}x)+a_{t}\big(\partial_{t}f^{*}(t,p_{t}x)+\dot{p}_{t}\nabla f^{*}(t,p_{t}x)\,x\big).
\end{equation}
Plugging~\eqref{eq:proof:proposition:reg:denoiser:6},~\eqref{eq:proof:lemma:conditional:exp:time:2}, parts~(a) and~(c) of Lemma~\ref{lemma:conditional:exp:time} and $a_{t}\leq1$ into~\eqref{eq:proof:proposition:reg:denoiser:7}, together with $\|x\|_{2}\leq\sqrt{d}R$ and $R>1$, yields $\|\partial_{t}D^{*}(t,x)\|_{2}\leq D_{\rm deno}R$ for each $(t,x)\in(0,1)\times\bbB_{R}^{\infty}$, where the constant $D_{\rm deno}$ only depends on $d$, $\sigma$ and the coefficients $(\alpha_{t},\beta_{t})$. This completes the proof.
\end{proof}
}

{
\begin{proof}[Proof of Proposition~\ref{proposition:reg:flow}]
We first prove part~(a). For each $1\leq k\leq d$, it follows from~\eqref{eq:characteristic:flow} that
\begin{equation}\label{eq:proof:proposition:reg:flow:1}
g_{k}^{*}(t,s,x)=x_{k}+\int_{t}^{s}b_{k}^{*}(\tau,g^{*}(t,\tau,x))\d\tau.
\end{equation}
Using the triangular inequality,~\eqref{eq:proof:proposition:reg:velocity:1} and $|\bbE[U_{\tau,y,k}]|\leq1$ for each $y\in\bbR^{d}$, we have
\begin{align*}
|g_{k}^{*}(t,s,x)|
&\leq|x_{k}|+\int_{t}^{s}|b_{k}^{*}(\tau,g^{*}(t,\tau,x))|\d\tau \\
&\leq|x_{k}|+\int_{t}^{s}\Big|\frac{\alpha_{\tau}^{2}\dot{\beta}_{\tau}-\dot{\alpha}_{\tau}\alpha_{\tau}\beta_{\tau}}{\alpha_{\tau}^{2}+\sigma^{2}\beta_{\tau}^{2}}\Big|\d\tau+\int_{t}^{s}\Big|\frac{\dot{\alpha}_{\tau}\alpha_{\tau}+\sigma^{2}\dot{\beta}_{\tau}\beta_{\tau}}{\alpha_{\tau}^{2}+\sigma^{2}\beta_{\tau}^{2}}\Big||g_{k}^{*}(t,\tau,x)|\d\tau.
\end{align*}
Applying Gronwall's inequality~\cite[Section B.2]{evans2010partial} and~\eqref{eq:proof:proposition:reg:velocity:2} yields
\begin{align*}
|g_{k}^{*}(t,s,x)|
&\leq\Big(|x_{k}|+\int_{t}^{s}\Big|\frac{\alpha_{\tau}^{2}\dot{\beta}_{\tau}-\dot{\alpha}_{\tau}\alpha_{\tau}\beta_{\tau}}{\alpha_{\tau}^{2}+\sigma^{2}\beta_{\tau}^{2}}\Big|\d\tau\Big)\exp\Big(\int_{t}^{s}\Big|\frac{\dot{\alpha}_{\tau}\alpha_{\tau}+\sigma^{2}\dot{\beta}_{\tau}\beta_{\tau}}{\alpha_{\tau}^{2}+\sigma^{2}\beta_{\tau}^{2}}\Big|\d\tau\Big) \\
&\leq\Big(R+\frac{A_{1}+B_{1}}{\delta_{0}}\Big)\exp\Big(\frac{A_{1}+\sigma^{2}B_{1}}{\delta_{0}}\Big)\leq B_{\flow}R,
\end{align*}
where the second inequality used $0\leq s-t\leq1$ and $x\in\bbB_{R}^{\infty}$, and the last inequality holds from $R>1$ with $B_{\flow}\coloneqq(1+(A_{1}+B_{1})/\delta_{0})\exp((A_{1}+\sigma^{2}B_{1})/\delta_{0})$. This proves part~(a).
\par We next prove part~(b). Let $x,x^{\prime}\in\bbR^{d}$, and denote by $x(\cdot)\coloneqq g^{*}(t,\cdot,x)$ and $x^{\prime}(\cdot)\coloneqq g^{*}(t,\cdot,x^{\prime})$ the solutions of the ODE~\eqref{eq:characteristic} starting from $x$ and $x^{\prime}$ at time $t$, respectively. Then it follows that
\begin{equation*}
\frac{\d}{\d s}(x_{k}(s)-x_{k}^{\prime}(s))=b_{k}^{*}(s,x(s))-b_{k}^{*}(s,x^{\prime}(s)), \quad 1\leq k\leq d.
\end{equation*}
By the definition of $\ell_{2}$-norm, we have
\begin{align*}
\frac{\d}{\d s}\|x(s)-x^{\prime}(s)\|_{2}
&=\frac{1}{2\|x(s)-x^{\prime}(s)\|_{2}}\sum_{k=1}^{d}\frac{\d}{\d s}(x_{k}(s)-x_{k}^{\prime}(s))^{2} \\
&=\frac{1}{\|x(s)-x^{\prime}(s)\|_{2}}\sum_{k=1}^{d}(x_{k}(s)-x_{k}^{\prime}(s))(b_{k}^{*}(s,x(s))-b_{k}^{*}(s,x^{\prime}(s))) \\
&\leq\|b^{*}(s,x(s))-b^{*}(s,x^{\prime}(s))\|_{2}\leq G_{\rm vel}\|x(s)-x^{\prime}(s)\|_{2},
\end{align*}
where the first inequality follows from Cauchy-Schwarz inequality, and the second inequality is due to the Lipschitz continuity of the velocity field in part~(b) of Proposition~\ref{proposition:reg:velocity}. Then applying Gronwall's inequality~\cite[Section B.2]{evans2010partial} yields
\begin{equation*}
\|g^{*}(t,s,x)-g^{*}(t,s,x^{\prime})\|_{2}\leq\exp(G_{\rm vel}(s-t))\|x-x^{\prime}\|_{2}.
\end{equation*}
Consequently, the map $x\mapsto g^{*}(t,s,x)$ is Lipschitz continuous with Lipschitz constant $\exp(G_{\rm vel}(s-t))$, which implies $\|\nabla g^{*}(t,s,x)\|_{\op}\leq\exp(G_{\rm vel}(s-t))$ for each $x\in\bbR^{d}$. This proves part~(b).
\par Finally, we prove part~(c). Taking the partial derivative with respect to $s$ in~\eqref{eq:proof:proposition:reg:flow:1} yields $\partial_{s}g^{*}(t,s,x)=b^{*}(s,g^{*}(t,s,x))$. Since $x\in\bbB_{R}^{\infty}$, part~(a) implies $g^{*}(t,s,x)\in\bbB_{B_{\flow}R}^{\infty}$, and hence part~(a) of Proposition~\ref{proposition:reg:velocity} gives
\begin{equation*}
\|\partial_{s}g^{*}(t,s,x)\|_{2}\leq\sqrt{d}\max_{1\leq k\leq d}|b_{k}^{*}(s,g^{*}(t,s,x))|\leq\sqrt{d}B_{\vel}B_{\flow}R.
\end{equation*}
\par Now, we consider the partial derivative of $g^{*}(t,s,x)$ with respect to $t$, which leads to
\begin{align}\label{eq:par_t g*}
\partial_{t}g^{*}(t,s,x)=-b^{*}(t,x)+\int_{t}^{s}\nabla b^{*}(\tau,g^{*}(t,\tau,x))\partial_{t}g^{*}(t,\tau,x)\d\tau.
\end{align}
Note that $\partial_{t}g^{*}(t,t,x)\coloneqq\partial_{t}g^{*}(t,s,x)|_{s=t}=-b^{*}(t,x)$. Denote by $\zeta(t,s,x)\coloneqq\partial_{t}g^{*}(t,s,x)$, and further take the partial derivative with respect to $s$ on both sides of~\eqref{eq:par_t g*}:
\begin{align*}
\partial_{s}\zeta(t,s,x)=\nabla b^{*}(s,g^{*}(t,s,x))\zeta(t,s,x).
\end{align*}
Then, it holds that
\begin{align*}
\partial_{s}\|\zeta(t,s,x)\|_{2}^{2}=2\big\langle\zeta(t,s,x),\partial_{s}\zeta(t,s,x)\big\rangle=2\big\langle\zeta(t,s,x),\nabla b^{*}(s,g^{*}(t,s,x))\zeta(t,s,x)\big\rangle.
\end{align*}
Using Cauchy-Schwarz inequality and part~(b) of Proposition~\ref{proposition:reg:velocity}, we have that
\begin{align*}
\partial_{s}\|\zeta(t,s,x)\|_{2}^{2}\leq2\|\nabla b^{*}(s,g^{*}(t,s,x))\|_{\op}\|\zeta(t,s,x)\|_{2}^{2}\leq2G_{\rm vel}\|\zeta(t,s,x)\|_{2}^{2}.
\end{align*}
With Gronwall's inequality~\cite[Section B.2]{evans2010partial}, it holds that
\begin{align*}
\|\zeta(t,s,x)\|_{2}^{2}\leq\|\zeta(t,t,x)\|_{2}^{2}\exp\Big(\int_{t}^{s}2G_{\rm vel}\d\tau\Big)=\|b^{*}(t,x)\|_{2}^{2}\exp(2G_{\rm vel}(s-t)).
\end{align*}
Thus, we conclude that
\begin{equation*}
\|\partial_{t}g^{*}(t,s,x)\|_{2}\leq\|b^{*}(t,x)\|_{2}\exp(G_{\rm vel}(s-t))\leq\sqrt{d}B_{\vel}R\exp(G_{\rm vel}),
\end{equation*}
where the last inequality holds from part~(a) of Proposition~\ref{proposition:reg:velocity},~$0\leq s-t\leq1$ and $x\in\bbB_{R}^{\infty}$. Setting $$B_{\flow}^{\prime}\coloneqq\sqrt{d}B_{\vel}\max\{B_{\flow},\exp(G_{\rm vel})\}$$ completes the proof.
\end{proof}
}

{
\par Finally, we record a uniform bound on the time derivative of the velocity field, which is used in the analysis of subsequent sections.

\begin{proposition}\label{proposition:lip:time:velocity}
Suppose Assumptions~\ref{assumption:init:dist:gaussian} and~\ref{assumption:target:dist} hold. Let $R\in(1,+\infty)$. Then for each $(t,x)\in(0,1)\times\bbB_{R}^{\infty}$,
\begin{equation*}
\|\partial_{t}b^{*}(t,x)\|_{2}\leq DR,
\end{equation*}
where $D$ is a constant only depending on $d$, $\sigma$ and the coefficients $(\alpha_{t},\beta_{t})$.
\end{proposition}

\begin{proof}[Proof of Proposition~\ref{proposition:lip:time:velocity}]
Write the representation~\eqref{eq:proof:proposition:reg:velocity:1} as
\begin{equation}\label{eq:proof:proposition:lip:time:velocity:1}
b^{*}(t,x)=\psi_{t}\bbE[U_{t,x}]+\chi_{t}x, \quad
\psi_{t}\coloneqq\frac{\alpha_{t}^{2}\dot{\beta}_{t}-\dot{\alpha}_{t}\alpha_{t}\beta_{t}}{\alpha_{t}^{2}+\sigma^{2}\beta_{t}^{2}}, \quad
\chi_{t}\coloneqq\frac{\dot{\alpha}_{t}\alpha_{t}+\sigma^{2}\dot{\beta}_{t}\beta_{t}}{\alpha_{t}^{2}+\sigma^{2}\beta_{t}^{2}}.
\end{equation}
In the following, denote by $(\dot{\alpha}_{t}\alpha_{t})^{\prime}$ the derivative of $t\mapsto\dot{\alpha}_{t}\alpha_{t}$. By Condition~\ref{condition:interpolant}~(iv), the numerators in~\eqref{eq:proof:proposition:lip:time:velocity:1} are differentiable on $(0,1)$ with
\begin{align}
\frac{\d}{\d t}\big(\alpha_{t}^{2}\dot{\beta}_{t}-\dot{\alpha}_{t}\alpha_{t}\beta_{t}\big)
&=2\dot{\alpha}_{t}\alpha_{t}\dot{\beta}_{t}+\alpha_{t}^{2}\ddot{\beta}_{t}-(\dot{\alpha}_{t}\alpha_{t})^{\prime}\beta_{t}-\dot{\alpha}_{t}\alpha_{t}\dot{\beta}_{t}
=\dot{\alpha}_{t}\alpha_{t}\dot{\beta}_{t}+\alpha_{t}^{2}\ddot{\beta}_{t}-(\dot{\alpha}_{t}\alpha_{t})^{\prime}\beta_{t}, \label{eq:proof:proposition:lip:time:velocity:2} \\
\frac{\d}{\d t}\big(\dot{\alpha}_{t}\alpha_{t}+\sigma^{2}\dot{\beta}_{t}\beta_{t}\big)
&=(\dot{\alpha}_{t}\alpha_{t})^{\prime}+\sigma^{2}\big(\ddot{\beta}_{t}\beta_{t}+\dot{\beta}_{t}^{2}\big). \label{eq:proof:proposition:lip:time:velocity:3}
\end{align}
Hence, applying the quotient rule to~\eqref{eq:proof:proposition:lip:time:velocity:1}, together with~\eqref{eq:delta:0},~\eqref{eq:interpolant:constant}, $|\dot{\Delta}_{t}|\leq2(A_{1}+\sigma^{2}B_{1})$ and $0\leq\alpha_{t},\beta_{t}\leq1$, implies
\begin{equation}\label{eq:proof:proposition:lip:time:velocity:4}
|\dot{\psi}_{t}|\leq\frac{A_{1}B_{1}+B_{2}+A_{2}}{\delta_{0}}+\frac{2(A_{1}+B_{1})(A_{1}+\sigma^{2}B_{1})}{\delta_{0}^{2}}, \quad
|\dot{\chi}_{t}|\leq\frac{A_{2}+\sigma^{2}(B_{2}+B_{1}^{2})}{\delta_{0}}+\frac{2(A_{1}+\sigma^{2}B_{1})^{2}}{\delta_{0}^{2}},
\end{equation}
for each $t\in(0,1)$. Then differentiating~\eqref{eq:proof:proposition:lip:time:velocity:1} with respect to $t$ and using Lemma~\ref{lemma:conditional:exp:time} yields
\begin{equation*}
\partial_{t}b^{*}(t,x)=\dot{\psi}_{t}\bbE[U_{t,x}]+\psi_{t}\partial_{t}\bbE[U_{t,x}]+\dot{\chi}_{t}x.
\end{equation*}
Combining $\|\bbE[U_{t,x}]\|_{2}\leq\sqrt{d}$, $|\psi_{t}|\leq(A_{1}+B_{1})/\delta_{0}$,~\eqref{eq:lemma:conditional:exp:time:bound},~\eqref{eq:proof:proposition:lip:time:velocity:4} and $\|x\|_{2}\leq\sqrt{d}R$ implies
\begin{equation*}
\|\partial_{t}b^{*}(t,x)\|_{2}\leq|\dot{\psi}_{t}|\sqrt{d}+\frac{(A_{1}+B_{1})M}{\delta_{0}}(1+\sqrt{d}R)+|\dot{\chi}_{t}|\sqrt{d}R\leq DR,
\end{equation*}
for each $(t,x)\in(0,1)\times\bbB_{R}^{\infty}$, where $M$ is the constant in Lemma~\ref{lemma:conditional:exp:time}, and the last inequality holds from $R>1$ with a constant $D$ only depending on $d$, $\sigma$ and the coefficients $(\alpha_{t},\beta_{t})$. This completes the proof.
\end{proof}
}

\section{Proof of Results in Section~\ref{section:analysis:velocity}}\label{section:proof:theorem:velocity}

\par In the section, we present the proofs of Theorem~\ref{theorem:velocity:rate}.

\subsection{Proof of Theorem~\ref{theorem:velocity:rate}}

{
\par In this section, we prove Theorem~\ref{theorem:velocity:rate}. Specifically, we propose the oracle inequality in Lemma~\ref{lemma:oracle:velocity}, which decomposes the $L^{2}$-risk into the approximation error, the generalization error, and the truncation error. Then we provide an approximation error bound in Lemma~\ref{lemma:approx:velocity}. By making a trade-off between three errors, we finally obtain the convergence rate for the denoiser estimator, which completes the proof of Theorem~\ref{theorem:velocity:rate}. Finally, Lemma~\ref{lemma:matching:grid} relates the time-averaged risk to its values on the sampling grid, which is the form of the matching error consumed by the discretization analysis in Appendix~\ref{section:proof:discretization}.

\par Recall the weighted $L^{2}$-risk~\eqref{eq:measure:velocity} of a measurable function $D:\bbR\times\bbR^{d}\rightarrow\bbR^{d}$ as
\begin{equation*}
\calE_{T}(D)=\frac{1}{T}\int_{0}^{T}\bbE_{X_{t}\sim\mu_{t}}\Big[\|D(t,X_{t})-D^{*}(t,X_{t})\|_{2}^{2}\Big]\dt,
\end{equation*}
and for the sake of notation simplicity, define the truncated $L^{2}$-risk with truncation parameter $R>1$ as
\begin{equation*}
\calE_{T,R}(D)=\frac{1}{T}\int_{0}^{T}\bbE_{X_{t}\sim\mu_{t}}\Big[\|D(t,X_{t})-D^{*}(t,X_{t})\|_{2}^{2}\bbone\{\|X_{t}\|_{\infty}\leq R\}\Big]\dt.
\end{equation*}
Throughout this section, we adopt the notation~\eqref{eq:Delta},~\eqref{eq:delta:0} and~\eqref{eq:interpolant:constant}, recall the preconditioning coefficients $(a_{t},c_{t},p_{t})$ in~\eqref{eq:denoiser:precondition} together with $q_{t}$ in~\eqref{eq:pq}, and recall the posterior mean $f^{*}$ and the constant $F_{\rm deno}=(A_{1}+B_{1})d^{3/2}/\delta_{0}^{2}$ in Lemma~\ref{lemma:conditional:exp:time}.

\begin{lemma}[Oracle inequality for denoiser estimation]\label{lemma:oracle:velocity}
Suppose that Assumptions~\ref{assumption:init:dist:gaussian} and~\ref{assumption:target:dist} hold. Let $T\in(1/2,1)$ and $R\in(1,+\infty)$. Further, assume that for each $D\in\scrD$,
\begin{equation}\label{eq:lemma:oracle:velocity:0}
\max_{1\leq k\leq d}|D_{k}(t,x)|\leq B_{\scrD}(1+\|x\|_{\infty}), \quad (t,x)\in[0,T]\times\bbR^{d}.
\end{equation}
Then the following inequality holds for each $n\geq\max_{1\leq k\leq d}\vcdim(\Pi_{k}\scrD)$,
\begin{equation*}
\bbE_{\euS}\big[\calE_{T}(\what{D})\big]\leq 2\inf_{D\in\scrD}\calE_{T,R}(D)+CR^{2}\Big(\max_{1\leq k\leq d}\frac{\vcdim(\Pi_{k}\scrD)}{n\log^{-1}(n)}+\frac{1}{\exp(\theta R^{2})}\Big),
\end{equation*}
where the constant $\theta$ only depends on $\sigma$, and the constant $C$ only depends on $d$, $\sigma$, $B_{\scrD}$ and the coefficients $(\alpha_{t},\beta_{t})$.
\end{lemma}

\begin{proof}[Proof of Lemma~\ref{lemma:oracle:velocity}]

\par Before proceeding, we introduce some notations, aiming to reformulate the original denoiser matching problem to a standard regression model.

\par Given a pair of random variables $(X_{0},X_{1})$ sampled from $\mu_{0}\times\mu_{1}$, recall the stochastic interpolant $X_{t}=\alpha_{t}X_{0}+\beta_{t}X_{1}$. We define the noise term as $\xi_{t}=X_{1}-D^{*}(t,X_{t})$. Since $D^{*}(t,x)=\bbE[X_{1}\big|X_{t}=x]$ by~\eqref{eq:denoiser}, we have $\bbE[\xi_{t}|X_{t}=x]=0$ for each $(t,x)\in[0,T]\times\bbR^{d}$. Therefore, in the rest of this proof, it suffices to consider the following regression model:
\begin{equation}\label{eq:proof:lemma:oracle:velocity:0:0}
X_{1}=D^{*}(t,X_{t})+\xi_{t}, \quad X_{t}\sim\mu_{t},~t\sim\unif[0,T].
\end{equation}

\par Recall the data set $\euS=\{(t^{(i)},X_{0}^{(i)},X_{1}^{(i)})\}_{i=1}^{n}$. Then we define the data set corresponding to the regression model~\eqref{eq:proof:lemma:oracle:velocity:0:0} as $\{(t^{(i)},X_{t}^{(i)},X_{1}^{(i)})\}_{i=1}^{n}$, for which
\begin{equation*}
X_{t}^{(i)}=\alpha(t^{(i)})X_{0}^{(i)}+\beta(t^{(i)})X_{1}^{(i)}.
\end{equation*}
The noise terms $\{\xi_{t}^{(i)}\}_{i=1}^{n}$ can be defined by $\xi_{t}^{(i)}=X_{1}^{(i)}-D^{*}(t^{(i)},X_{t}^{(i)})$ for each $1\leq i\leq n$.

\par We divide the proof into four steps.

\par\noindent\emph{Step 1. Sub-Gaussian noise.}
\par In this step, we show that the noise term $(\xi_{t}|X_{t}=x)$ in~\eqref{eq:proof:lemma:oracle:velocity:0:0} is sub-Gaussian for each $t\in[0,T]$ and $x\in\bbR^{d}$, and aim to estimate its variance proxy. According to Lemma~\ref{lemma:condition:1t}, it holds that
\begin{equation*}
(X_{1}|X_{t}=x)\stackrel{\d}{=}a_{t}U_{t,x}+\sqrt{\frac{\sigma^{2}\alpha_{t}^{2}}{\alpha_{t}^{2}+\sigma^{2}\beta_{t}^{2}}}\epsilon+c_{t}x,
\end{equation*}
where $U_{t,x}\in[0,1]^{d}$ and $\epsilon\sim N(0,I_{d})$ are two independent variables. Then taking expectation on both sides of the equality yields
\begin{equation*}
D^{*}(t,x)=\bbE[X_{1}|X_{t}=x]=a_{t}\bbE[U_{t,x}]+c_{t}x.
\end{equation*}
Therefore, by the definition of the noise term, the following equality holds
\begin{equation*}
(\xi_{t}|X_{t}=x)
=(X_{1}|X_{t}=x)-\bbE[X_{1}|X_{t}=x]
\stackrel{\d}{=}a_{t}\big(U_{t,x}-\bbE[U_{t,x}]\big)+\sqrt{\frac{\sigma^{2}\alpha_{t}^{2}}{\alpha_{t}^{2}+\sigma^{2}\beta_{t}^{2}}}\epsilon.
\end{equation*}
Since that the random variable $U_{t,x}\in[0,1]^{d}$, using Hoeffding's lemma~\cite[Lemma D.1]{mohri2018foundations} implies that $U_{t,x}$ is $1$-sub-Gaussian. Further, applying Lemma~\ref{lemma:lin:comb:subgaussian} deduces that each element of $(\xi_{t}|X_{t}=x)$ is sub-Gaussian for each $t\in[0,T]$ and $x\in\bbR^{d}$ with variance proxy
\begin{equation}\label{eq:proof:lemma:oracle:velocity:1:1}
\sigma_{\scrD}^{2}=\sup_{t\in[0,T]}\Big\{a_{t}^{2}+\frac{\sigma^{2}\alpha_{t}^{2}}{\alpha_{t}^{2}+\sigma^{2}\beta_{t}^{2}}\Big\}\leq1+\sigma^{2},
\end{equation}
where we used $a_{t}\leq1$ and $\sigma^{2}\alpha_{t}^{2}/(\alpha_{t}^{2}+\sigma^{2}\beta_{t}^{2})\leq\sigma^{2}$.

\par\noindent\emph{Step 2. Truncation.}

\par Notice that the denoiser $D^{*}$ is defined on $\bbR^{d}$. It is necessary to restrict the original problem onto a compact subset of $\bbR^{d}$ by the technique of truncation. To begin with, we define the truncated population and empirical excess risks with radius $R>1$, respectively, as
\begin{align*}
\calE_{T,R}(D)&=\frac{1}{T}\int_{0}^{T}\bbE_{X_{t}\sim\mu_{t}}\Big[\|D^{*}(t,X_{t})-D(t,X_{t})\|_{2}^{2}\bbone\{\|X_{t}\|_{\infty}\leq R\}\Big]\dt, \\
\what{\calE}_{T,R,n}(D)&=\frac{1}{n}\sum_{i=1}^{n}\|D^{*}(t^{(i)},X_{t}^{(i)})-D(t^{(i)},X_{t}^{(i)})\|_{2}^{2}\bbone\{\|X_{t}^{(i)}\|_{\infty}\leq R\}.
\end{align*}
The population excess risk of the estimator $\what{D}$ can be decomposed by
\begin{equation}\label{eq:proof:lemma:oracle:velocity:2:1}
\bbE_{\euS}[\calE_{T}(\what{D})]\leq\bbE_{\euS}\Big[\calE_{T}(\what{D})-\calE_{T,R}(\what{D})\Big]+\bbE_{\euS}\Big[\sup_{D\in\scrD}\calE_{T,R}(D)-2\what{\calE}_{T,R,n}(D)\Big]+2\bbE_{\euS}\Big[\what{\calE}_{T,R,n}(\what{D})\Big].
\end{equation}
The first term in the right-hand side of~\eqref{eq:proof:lemma:oracle:velocity:2:1} corresponds to the truncation
error, which is estimated in the rest of this step. The second term of~\eqref{eq:proof:lemma:oracle:velocity:2:1} is studied
in \emph{Step 3}. Finally, we bound the last term of~\eqref{eq:proof:lemma:oracle:velocity:2:1} in \emph{Step 4}.

\par For each hypothesis $D\in\scrD$, it follows that
\begin{align}
&\bbE_{X_{t}\sim\mu_{t}}\Big[\|D^{*}(t,X_{t})-D(t,X_{t})\|_{2}^{2}\bbone\{\|X_{t}\|_{\infty}>R\}\Big] \nonumber \\
&\leq\bbE_{X_{t}\sim\mu_{t}}^{1/2}\Big[\|D^{*}(t,X_{t})-D(t,X_{t})\|_{2}^{4}\Big]\bbE_{X_{t}\sim\mu_{t}}^{1/2}\Big[\bbone\{\|X_{t}\|_{\infty}>R\}\Big] \nonumber \\
&\leq8\Big(\bbE_{X_{t}\sim\mu_{t}}^{1/2}\Big[\|D(t,X_{t})\|_{2}^{4}\Big]+\bbE_{X_{t}\sim\mu_{t}}^{1/2}\Big[\|D^{*}(t,X_{t})\|_{2}^{4}\Big]\Big)\pr^{1/2}\{\|X_{t}\|_{\infty}>R\}, \label{eq:proof:lemma:oracle:velocity:2:2}
\end{align}
where the first inequality follows from Cauchy-Schwarz inequality, and the second inequality is due to the triangular inequality. We first bound the fourth moment of $X_{t}$. According to Assumption~\ref{assumption:target:dist}, we have
\begin{equation*}
X_{t}\stackrel{\d}{=}\alpha_{t}X_{0}+\beta_{t}U+\sigma\beta_{t}\epsilon, \quad U\sim\nu,~\epsilon\sim N(0,I_{d}),
\end{equation*}
which implies
\begin{align*}
\bbE^{1/2}\Big[\|X_{t}\|_{2}^{4}\Big]
&\leq\bbE^{1/2}\Big[(\|\alpha_{t}X_{0}\|_{2}+\|\beta_{t}U\|_{2}+\|\sigma\beta_{t}\epsilon\|_{2})^{4}\Big] \\
&\leq27\Big(\alpha_{t}^{4}\bbE\Big[\|X_{0}\|_{2}^{4}\Big]+\beta_{t}^{4}\bbE\Big[\|U\|_{2}^{4}\Big]+\sigma^{4}\beta_{t}^{4}\bbE\Big[\|\epsilon\|_{2}^{4}\Big]\Big)^{1/2}
\leq Cd(1+\sigma^{2}), \quad t\in[0,T],
\end{align*}
where the second inequality holds from the triangular inequality, and the last inequality follows from Lemma~\ref{lemma:4th:moment:Gaussian} together with $\bbE[\|U\|_{2}^{4}]\leq d^{2}$, $0\leq\alpha_{t},\beta_{t}\leq1$, and an absolute constant $C$. The growth condition~\eqref{eq:lemma:oracle:velocity:0} implies $\|D(t,x)\|_{2}\leq\sqrt{d}B_{\scrD}(1+\|x\|_{\infty})\leq\sqrt{d}B_{\scrD}(1+\|x\|_{2})$ for each $(t,x)\in[0,T]\times\bbR^{d}$, and consequently
\begin{equation}\label{eq:proof:lemma:oracle:velocity:2:3}
\bbE_{X_{t}\sim\mu_{t}}^{1/2}\Big[\|D(t,X_{t})\|_{2}^{4}\Big]
\leq dB_{\scrD}^{2}\bbE^{1/2}\Big[(1+\|X_{t}\|_{2})^{4}\Big]
\leq3dB_{\scrD}^{2}\Big(1+\bbE^{1/2}\big[\|X_{t}\|_{2}^{4}\big]\Big)
\leq Cd^{2}B_{\scrD}^{2}(1+\sigma^{2}),
\end{equation}
where $C$ is an absolute constant. Then we consider the fourth moment of $D^{*}(t,X_{t})$ in~\eqref{eq:proof:lemma:oracle:velocity:2:2}. By using Assumption~\ref{assumption:target:dist}, we have $X_{1}\stackrel{\d}{=}U+\sigma\epsilon$ with $U\sim\nu$ and $\epsilon\sim N(0,I_{d})$, which implies
\begin{equation*}
\bbE^{1/2}\Big[\|X_{1}\|_{2}^{4}\Big]
\leq\bbE^{1/2}\Big[(\|U\|_{2}+\|\sigma\epsilon\|_{2})^{4}\Big]
\leq9\Big(\bbE\Big[\|U\|_{2}^{4}\Big]+\sigma^{4}\bbE\Big[\|\epsilon\|_{2}^{4}\Big]\Big)^{1/2}
\leq Cd(1+\sigma^{2}),
\end{equation*}
where the last inequality follows from Lemma~\ref{lemma:4th:moment:Gaussian} and $\bbE[\|U\|_{2}^{4}]\leq d^{2}$, and $C$ is an absolute constant. Consequently,
\begin{equation}\label{eq:proof:lemma:oracle:velocity:2:4}
\bbE_{X_{t}}^{1/2}\Big[\|D^{*}(t,X_{t})\|_{2}^{4}\Big]
=\bbE_{X_{t}}^{1/2}\Big[\|\bbE[X_{1}|X_{t}]\|_{2}^{4}\Big]
\leq\bbE^{1/2}\Big[\|X_{1}\|_{2}^{4}\Big]\leq Cd(1+\sigma^{2}),
\end{equation}
where the first inequality is due to the definition of the denoiser~\eqref{eq:denoiser}, and the second inequality follows from Jensen's inequality. We next consider the tail probability of $X_{t}$ in~\eqref{eq:proof:lemma:oracle:velocity:2:2}. According to Assumption~\ref{assumption:target:dist}, we find
\begin{equation*}
X_{t}\stackrel{\d}{=}\alpha_{t}X_{0}+\beta_{t}U+\sigma\beta_{t}\epsilon, \quad U\sim\nu,~\epsilon\sim N(0,I_{d}),
\end{equation*}
which implies from Hoeffding's lemma~\cite[Lemma D.1]{mohri2018foundations} and Lemma~\ref{lemma:lin:comb:subgaussian} that $X_{t}$ is a $(\alpha_{t}^{2}+\beta_{t}^{2}+\sigma^{2}\beta_{t}^{2})$-sub-Gaussian random variable. Then it follows from Lemma~\ref{lemma:max:subGaussian:proba} that
\begin{equation}\label{eq:proof:lemma:oracle:velocity:2:5}
\sup_{t\in(0,1)}\pr\big\{\|X_{t}\|_{\infty}>R\big\}\leq2d\sup_{t\in(0,1)}\exp\Big(-\frac{R^{2}}{2(\alpha_{t}^{2}+\beta_{t}^{2}+\sigma^{2}\beta_{t}^{2})}\Big)\leq\frac{2d}{\exp(\theta R^{2})},
\end{equation}
where $\theta$ is a constant only depending on $\sigma$. Substituting~\eqref{eq:proof:lemma:oracle:velocity:2:3},~\eqref{eq:proof:lemma:oracle:velocity:2:4}, and~\eqref{eq:proof:lemma:oracle:velocity:2:5} into~\eqref{eq:proof:lemma:oracle:velocity:2:2} yields
\begin{equation*}
\bbE_{X_{t}\sim\mu_{t}}\Big[\|D^{*}(t,X_{t})-D(t,X_{t})\|_{2}^{2}\bbone\{\|X_{t}\|_{\infty}>R\}\Big]\leq\frac{C}{\exp(\theta R^{2})}, \quad t\in[0,T],
\end{equation*}
where $C$ is a constant only depending on $d$, $\sigma$ and $B_{\scrD}$, and we redefined $\theta$ to absorb the factor $1/2$. As a consequence, for each hypothesis $D\in\scrD$, it follows that
\begin{equation}\label{eq:proof:lemma:oracle:velocity:2:6}
\calE_{T}(D)-\calE_{T,R}(D)\leq\frac{C}{\exp(\theta R^{2})}.
\end{equation}
Hence the first term in the right-hand side of~\eqref{eq:proof:lemma:oracle:velocity:2:1} can be bounded by
\begin{equation}\label{eq:proof:lemma:oracle:velocity:2:7}
\bbE_{\euS}\Big[\calE_{T}(\what{D})-\calE_{T,R}(\what{D})\Big]\leq\frac{C}{\exp(\theta R^{2})}.
\end{equation}

\par We next bound another truncation term by a similar argument, which will be used in \emph{Step 4}. It follows from Cauchy-Schwarz inequality and~\eqref{eq:proof:lemma:oracle:velocity:2:5} that
\begin{equation}\label{eq:proof:lemma:oracle:velocity:2:8}
\bbE_{\euS}\Big[(\xi_{t,k}^{(i)})^{2}\bbone\{\|X_{t}^{(i)}\|_{\infty}>R\}\Big]\leq\bbE_{\euS}^{1/2}\Big[(\xi_{t,k}^{(i)})^{4}\Big]\pr^{1/2}\{\|X_{t}^{(i)}\|_{\infty}>R\}\leq\frac{C}{\exp(\theta R^{2})},
\end{equation}
where $C$ is a constant only depending on $d$ and $\sigma$, and the last inequality follows from the fact that $\xi_{t,k}^{(i)}$ is sub-Gaussian with variance proxy in~\eqref{eq:proof:lemma:oracle:velocity:1:1}.

\par\noindent\emph{Step 3. Relate the truncated population excess risk of the estimator with its empirical counterpart.}

\par In this step, we prove the following inequality:
\begin{equation}\label{eq:proof:lemma:oracle:velocity:3:1}
\bbE_{\euS}\Big[\sup_{D\in\scrD}\calE_{T,R}(D)-2\what{\calE}_{T,R,n}(D)\Big]\leq CR^{2}\sum_{k=1}^{d}\frac{\vcdim(\Pi_{k}\scrD)}{n\log^{-1}(n)},
\end{equation}
where $C$ is a constant only depending on $d$, $\sigma$, $B_{\scrD}$ and the coefficients $(\alpha_{t},\beta_{t})$, and $n\geq\vcdim(\Pi_{k}\scrD)$ for each $1\leq k\leq d$. For simplicity of notation, we define the $k$-th term of excess risks as
\begin{align*}
\calE_{T,R}^{k}(D)&=\frac{1}{T}\int_{0}^{T}\bbE_{X_{t}\sim\mu_{t}}\Big[(D_{k}^{*}(t,X_{t})-D_{k}(t,X_{t}))^{2}\bbone\{\|X_{t}\|_{\infty}\leq R\}\Big]\dt, \\
\what{\calE}_{T,R,n}^{k}(D)&=\frac{1}{n}\sum_{i=1}^{n}(D_{k}^{*}(t^{(i)},X_{t}^{(i)})-D_{k}(t^{(i)},X_{t}^{(i)}))^{2}\bbone\{\|X_{t}^{(i)}\|_{\infty}\leq R\}.
\end{align*}
On the event $\{\|X_{t}\|_{\infty}\leq R\}$, part~(a) of Proposition~\ref{proposition:reg:denoiser} gives $|D_{k}^{*}(t,X_{t})|\leq B_{\rm deno}R$, while the growth condition~\eqref{eq:lemma:oracle:velocity:0} together with $R>1$ gives $|D_{k}(t,X_{t})|\leq2B_{\scrD}R$. Applying Lemma~\ref{lemma:generalization:1} yields the following inequality
\begin{equation*}
\bbE_{\euS}\Big[\sup_{D\in\scrD}\calE_{T,R}^{k}(D)-2\what{\calE}_{T,R,n}^{k}(D)\Big]\leq CR^{2}\frac{\vcdim(\Pi_{k}\scrD)}{n\log^{-1}(n)}, \quad 1\leq k\leq d,
\end{equation*}
which implies~\eqref{eq:proof:lemma:oracle:velocity:3:1} by summing with respect to $1\leq k\leq d$.

\par\noindent\emph{Step 4. Estimate the empirical excess risk.}

\par In this step, we show the following bound for the empirical excess risk of the estimator
\begin{equation}\label{eq:proof:lemma:oracle:velocity:4:0}
\bbE_{\euS}\Big[\what{\calE}_{T,R,n}(\what{D})\Big]\leq2\inf_{D\in\scrD}\calE_{T,R}(D)+CR^{2}\Big(\sum_{k=1}^{d}\frac{\vcdim(\Pi_{k}\scrD)}{n\log^{-1}n}+\frac{1}{\exp(\theta R^{2})}\Big),
\end{equation}
where $C$ is a constant only depending on $d$, $\sigma$, $B_{\scrD}$ and the coefficients $(\alpha_{t},\beta_{t})$, and $n\geq\vcdim(\Pi_{k}\scrD)$ for each $1\leq k\leq d$.

\par Recall the empirical risk $\what{\calL}_{n}$ in~\eqref{eq:velocity:er}, and define its truncated version as
\begin{equation*}
\what{\calL}_{R,n}(D)=\frac{1}{n}\sum_{i=1}^{n}\|X_{1}^{(i)}-D(t^{(i)},X_{t}^{(i)})\|_{2}^{2}\bbone\{\|X_{t}^{(i)}\|_{\infty}\leq R\}.
\end{equation*}
It is straightforward that
\begin{align*}
\what{\calL}_{R,n}(\what{D})
&=\frac{1}{n}\sum_{i=1}^{n}\sum_{k=1}^{d}(\xi_{t,k}^{(i)}+D_{k}^{*}(t^{(i)},X_{t}^{(i)})-\what{D}_{k}(t^{(i)},X_{t}^{(i)}))^{2}\bbone\{\|X_{t}^{(i)}\|_{\infty}\leq R\} \\
&=\what{\calE}_{T,R,n}(\what{D})+\frac{1}{n}\sum_{i=1}^{n}\sum_{k=1}^{d}(\xi_{t,k}^{(i)})^{2}\bbone\{\|X_{t}^{(i)}\|_{\infty}\leq R\} \\
&\quad+\frac{2}{n}\sum_{i=1}^{n}\sum_{k=1}^{d}\xi_{t,k}^{(i)}(D_{k}^{*}(t^{(i)},X_{t}^{(i)})-\what{D}_{k}(t^{(i)},X_{t}^{(i)}))\bbone\{\|X_{t}^{(i)}\|_{\infty}\leq R\}.
\end{align*}
Since $\what{D}$ is a minimizer of $\what{\calL}_{n}(\cdot)$ over the hypothesis class $\scrD$ by~\eqref{eq:velocity:erm}, it holds that $\what{\calL}_{R,n}(\what{D})\leq\what{\calL}_{n}(\what{D})\leq\what{\calL}_{n}(D)$ for each $D\in\scrD$. Consequently,
\begin{equation}\label{eq:proof:lemma:oracle:velocity:4:1}
\begin{aligned}
\bbE_{\euS}\Big[\what{\calE}_{T,R,n}(\what{D})\Big]
&\leq\calL(D)-\bbE_{\euS}\Big[\frac{1}{n}\sum_{i=1}^{n}\sum_{k=1}^{d}(\xi_{t,k}^{(i)})^{2}\bbone\{\|X_{t}^{(i)}\|_{\infty}\leq R\}\Big] \\
&\quad+2\bbE_{\euS}\Big[\frac{1}{n}\sum_{i=1}^{n}\sum_{k=1}^{d}\xi_{t,k}^{(i)}\what{D}_{k}(t^{(i)},X_{t}^{(i)})\bbone\{\|X_{t}^{(i)}\|_{\infty}\leq R\}\Big],
\end{aligned}
\end{equation}
where we used the fact that $\bbE_{\euS}[\what{\calL}_{n}(D)]=\calL(D)$ with $\calL$ as in~\eqref{eq:velocity:popu:risk}, and $\bbE[\xi_{t,k}^{(i)}D_{k}^{*}(t^{(i)},X_{t}^{(i)})]=0$. For the first term in the right-hand side of~\eqref{eq:proof:lemma:oracle:velocity:4:1}, we have
\begin{align}
&\calL(D)-\bbE_{\euS}\Big[\frac{1}{n}\sum_{i=1}^{n}\sum_{k=1}^{d}(\xi_{t,k}^{(i)})^{2}\bbone\{\|X_{t}^{(i)}\|_{\infty}\leq R\}\Big] \nonumber \\
&=\calE_{T}(D)+\bbE_{\euS}\Big[\frac{1}{n}\sum_{i=1}^{n}\sum_{k=1}^{d}(\xi_{t,k}^{(i)})^{2}\bbone\{\|X_{t}^{(i)}\|_{\infty}>R\}\Big]\leq\calE_{T,R}(D)+\frac{2C}{\exp(\theta R^{2})}, \label{eq:proof:lemma:oracle:velocity:4:2}
\end{align}
where the inequality follows from~\eqref{eq:proof:lemma:oracle:velocity:2:6} and~\eqref{eq:proof:lemma:oracle:velocity:2:8}. It remains to bound the second term in the right-hand side of~\eqref{eq:proof:lemma:oracle:velocity:4:1}. Plugging part~(a) of Proposition~\ref{proposition:reg:denoiser}, the growth condition~\eqref{eq:lemma:oracle:velocity:0}, and~\eqref{eq:proof:lemma:oracle:velocity:1:1} into Lemma~\ref{lemma:generalization:2}, we have
\begin{align}
&\bbE_{\euS}\Big[\frac{1}{n}\sum_{i=1}^{n}\sum_{k=1}^{d}\xi_{t,k}^{(i)}\what{D}_{k}(t^{(i)},X_{t}^{(i)})\bbone\{\|X_{t}^{(i)}\|_{\infty}\leq R\}\Big] \nonumber \\
&\leq\frac{1}{4}\bbE_{\euS}\Big[\what{\calE}_{T,R,n}(\what{D})\Big]+CR^{2}\sum_{k=1}^{d}\frac{\vcdim(\Pi_{k}\scrD)}{n\log^{-1}n}, \label{eq:proof:lemma:oracle:velocity:4:3}
\end{align}
where $C$ is a constant only depending on $d$, $\sigma$, $B_{\scrD}$ and the coefficients $(\alpha_{t},\beta_{t})$. Substituting~\eqref{eq:proof:lemma:oracle:velocity:4:2} and~\eqref{eq:proof:lemma:oracle:velocity:4:3} into~\eqref{eq:proof:lemma:oracle:velocity:4:1} yields~\eqref{eq:proof:lemma:oracle:velocity:4:0}.

\par Finally, plugging~\eqref{eq:proof:lemma:oracle:velocity:2:7},~\eqref{eq:proof:lemma:oracle:velocity:3:1}, and~\eqref{eq:proof:lemma:oracle:velocity:4:0} into~\eqref{eq:proof:lemma:oracle:velocity:2:1} completes the proof.
\end{proof}

\begin{lemma}[Approximation error]\label{lemma:approx:velocity}
Suppose that Assumptions~\ref{assumption:init:dist:gaussian} and~\ref{assumption:target:dist} hold. Let $T\in(1/2,1)$ and $R\in(1,+\infty)$. Set the hypothesis class $\scrD$ as a preconditioned deep neural network class, which is defined as
\begin{equation}\label{eq:scrD}
\scrD=\left\{
\begin{aligned}
&D(t,x)=c_{t}x+a_{t}f(t,p_{t}x):~f\in N(L,S), \\
&\|f(t,x)\|_{\infty}\leq1,~\|\nabla f(t,x)\|_{\op}\leq3d, \\
&\|\partial_{t}f(t,x)\|_{2}\leq3F_{\rm deno},~(t,x)\in[0,T]\times\bbR^{d}
\end{aligned}
\right\},
\end{equation}
where $F_{\rm deno}$ is the constant in Lemma~\ref{lemma:conditional:exp:time}, and the depth and the number of parameters of the neural network are given, respectively, as $L=C$ and $S=CN^{d+1}$. Then the following inequality holds for each $N\in\bbN_{+}$,
\begin{equation*}
\inf_{D\in\scrD}\calE_{T,R}(D)\leq CR^{2}N^{-2},
\end{equation*}
where $C$ is a constant only depending on $d$, $\sigma$ and the coefficients $(\alpha_{t},\beta_{t})$.
\end{lemma}

\begin{proof}[Proof of Lemma~\ref{lemma:approx:velocity}]
\par By~\eqref{eq:denoiser:representation}, the ground-truth denoiser shares the preconditioned structure of the hypothesis class~\eqref{eq:scrD}, that is, $D^{*}(t,x)=c_{t}x+a_{t}f^{*}(t,p_{t}x)$ for each $(t,x)\in(0,T]\times\bbR^{d}$, where $f^{*}$ is the posterior mean in Lemma~\ref{lemma:conditional:exp:time}. Hence it suffices to approximate $f^{*}$ by a neural network fulfilling the constraints in~\eqref{eq:scrD}, for which parts~(a) and~(c) of Lemma~\ref{lemma:conditional:exp:time} provide the uniform bounds: for each $(t,\vartheta)\in[0,T]\times\bbR^{d}$ and $1\leq k,\ell\leq d$,
\begin{equation}\label{eq:proof:lemma:approx:velocity:1}
\|f^{*}(t,\vartheta)\|_{\infty}\leq1, \quad
|\partial_{\vartheta_{\ell}}f_{k}^{*}(t,\vartheta)|\leq1, \quad
|\partial_{t}f_{k}^{*}(t,\vartheta)|\leq\frac{F_{\rm deno}}{\sqrt{d}}.
\end{equation}

\par We now construct the neural network. Since $p_{t}=\beta_{t}/\Delta_{t}\leq1/\delta_{0}$ for each $t\in[0,1]$ by~\eqref{eq:pq},~\eqref{eq:delta:0} and $0\leq\beta_{t}\leq1$, write $\bar{R}\coloneqq R/\delta_{0}$ and $\calX\coloneqq[0,T]\times[-\bar{R},\bar{R}]^{d}$, so that $(t,p_{t}x)\in\calX$ whenever $(t,x)\in[0,T]\times\bbB_{R}^{\infty}$. According to Corollary~\ref{corollary:approx}, for each element $1\leq k\leq d$, there exists a real-valued deep neural network $f_{k}$ with depth $L_{k}=\lceil\log_{2}(d+1)\rceil+3$ and number of parameters $S_{k}=(22(d+1)+6)(N+1)^{d+1}$, such that
\begin{equation}\label{eq:proof:lemma:approx:velocity:2}
\|f_{k}-f_{k}^{*}\|_{L^{\infty}(\calX)}\leq2^{d+1}\Big(\frac{T}{N}\|\partial_{t}f_{k}^{*}\|_{L^{\infty}(\calX)}+\sum_{\ell=1}^{d}\frac{2\bar{R}}{N}\|\partial_{\vartheta_{\ell}}f_{k}^{*}\|_{L^{\infty}(\calX)}\Big)\leq\frac{CR}{N},
\end{equation}
where the last inequality follows from~\eqref{eq:proof:lemma:approx:velocity:1}, and $C$ is a constant only depending on $d$, $\sigma$ and the coefficients $(\alpha_{t},\beta_{t})$. Moreover, Corollary~\ref{corollary:approx} guarantees that
\begin{equation}\label{eq:proof:lemma:approx:velocity:3}
\|f_{k}\|_{L^{\infty}(\calX)}=\|f_{k}^{*}\|_{L^{\infty}(\calX)}\leq1, \quad
\|\partial_{t}f_{k}\|_{L^{\infty}(\calX)}\leq3\|\partial_{t}f_{k}^{*}\|_{L^{\infty}(\calX)}, \quad
\|\partial_{\vartheta_{\ell}}f_{k}\|_{L^{\infty}(\calX)}\leq3\|\partial_{\vartheta_{\ell}}f_{k}^{*}\|_{L^{\infty}(\calX)}.
\end{equation}
The bounds in~\eqref{eq:proof:lemma:approx:velocity:3} hold on $\calX$ only. To extend them to $[0,T]\times\bbR^{d}$, define the coordinatewise clipping map $\pi_{\bar{R}}(\vartheta)\coloneqq(\min\{\max\{\vartheta_{\ell},-\bar{R}\},\bar{R}\})_{\ell=1}^{d}$, which is realized exactly by two additional ReLU layers with $\calO(d)$ non-zero weights, and set $f(t,\vartheta)\coloneqq(f_{k}(t,\pi_{\bar{R}}(\vartheta)))_{k=1}^{d}$. Since $\pi_{\bar{R}}$ is the identity on $[-\bar{R},\bar{R}]^{d}$, the approximation bound~\eqref{eq:proof:lemma:approx:velocity:2} is unaffected on $\calX$; since $\pi_{\bar{R}}$ takes values in $[-\bar{R},\bar{R}]^{d}$ and is $1$-Lipschitz in each coordinate, the bounds in~\eqref{eq:proof:lemma:approx:velocity:3} extend to all of $[0,T]\times\bbR^{d}$. Consequently, for each $(t,\vartheta)\in[0,T]\times\bbR^{d}$,
\begin{equation*}
\|f(t,\vartheta)\|_{\infty}\leq1, \quad
\|\nabla f(t,\vartheta)\|_{\op}\leq\Big(\sum_{k=1}^{d}\sum_{\ell=1}^{d}|\partial_{\vartheta_{\ell}}f_{k}(t,\vartheta)|^{2}\Big)^{1/2}\leq3d, \quad
\|\partial_{t}f(t,\vartheta)\|_{2}\leq3F_{\rm deno},
\end{equation*}
where the second inequality bounds the operator norm by the Frobenius norm, and the last two inequalities combine~\eqref{eq:proof:lemma:approx:velocity:3} with~\eqref{eq:proof:lemma:approx:velocity:1}. The vector-valued network $f$ has depth $L=\max_{1\leq k\leq d}L_{k}+C$ and number of parameters $S=\sum_{k=1}^{d}S_{k}+Cd\leq CN^{d+1}$. Hence $D(t,x)\coloneqq c_{t}x+a_{t}f(t,p_{t}x)$ belongs to the hypothesis class $\scrD$.

\par It remains to bound the truncated risk of $D$. For each $(t,x)\in(0,T]\times\bbB_{R}^{\infty}$, we have $(t,p_{t}x)\in\calX$, and thus the representation~\eqref{eq:denoiser:representation} together with $a_{t}\leq1$ implies
\begin{equation*}
\|D(t,x)-D^{*}(t,x)\|_{2}
=a_{t}\|f(t,p_{t}x)-f^{*}(t,p_{t}x)\|_{2}
\leq\Big(\sum_{k=1}^{d}\|f_{k}-f_{k}^{*}\|_{L^{\infty}(\calX)}^{2}\Big)^{1/2}
\leq\frac{C\sqrt{d}R}{N},
\end{equation*}
where the last inequality follows from~\eqref{eq:proof:lemma:approx:velocity:2}. Consequently,
\begin{equation*}
\calE_{T,R}(D)=\frac{1}{T}\int_{0}^{T}\bbE_{X_{t}\sim\mu_{t}}\Big[\|D(t,X_{t})-D^{*}(t,X_{t})\|_{2}^{2}\bbone\{\|X_{t}\|_{\infty}\leq R\}\Big]\dt\leq C^{\prime}R^{2}N^{-2},
\end{equation*}
where $C^{\prime}$ is a constant only depending on $d$, $\sigma$ and the coefficients $(\alpha_{t},\beta_{t})$. This completes the proof.
\end{proof}

{
\par We prove Theorem~\ref{theorem:velocity:rate} with the explicit preconditioned hypothesis class as that in Lemma~\ref{lemma:approx:velocity}
\begin{equation*}
\scrD=\left\{
\begin{aligned}
&D(t,x)=c_{t}x+a_{t}f(t,p_{t}x):~f\in N(L,S), \\
&\|f(t,x)\|_{\infty}\leq1,~\|\nabla f(t,x)\|_{\op}\leq3d, \\
&\|\partial_{t}f(t,x)\|_{2}\leq3F_{\rm deno},~(t,x)\in[0,T]\times\bbR^{d}
\end{aligned}
\right\},
\end{equation*}
This class mirrors the preconditioned representation~\eqref{eq:denoiser:representation}: the network $f$ only needs to fit the regular core $f^{*}$, coinciding with the denoiser parameterization of~\cite{karras2022elucidating} adopted in our implementation. Every $D\in\scrD$ obeys the regularity bounds stated in Theorem~\ref{theorem:velocity:rate}.
}

\begin{proof}[Proof of Theorem~\ref{theorem:velocity:rate}]
Set the hypothesis class $\scrD$ as that in Lemma~\ref{lemma:approx:velocity}. We first verify the growth condition~\eqref{eq:lemma:oracle:velocity:0}. For each $D\in\scrD$, since $\|f(t,\vartheta)\|_{\infty}\leq1$, $a_{t}\leq1$ and $c_{t}\leq\sigma^{2}/\delta_{0}$ by~\eqref{eq:delta:0}, it follows that
\begin{equation*}
\max_{1\leq k\leq d}|D_{k}(t,x)|\leq c_{t}\|x\|_{\infty}+a_{t}\|f(t,p_{t}x)\|_{\infty}\leq\Big(1+\frac{\sigma^{2}}{\delta_{0}}\Big)(1+\|x\|_{\infty}), \quad (t,x)\in[0,T]\times\bbR^{d},
\end{equation*}
that is,~\eqref{eq:lemma:oracle:velocity:0} holds with $B_{\scrD}=1+\sigma^{2}/\delta_{0}$. According to Lemma~\ref{lemma:approx:velocity}, there exists a deep neural network $D\in\scrD$ such that
\begin{equation}\label{eq:proof:theorem:velocity:rate:1}
\inf_{D\in\scrD}\calE_{T,R}(D)\leq CR^{2}N^{-2},
\end{equation}
where $C$ is a constant only depending on $d$, $\sigma$ and the coefficients $(\alpha_{t},\beta_{t})$. On the other hand, we estimate the VC-dimension of $\Pi_{k}\scrD$. Since $\alpha_{t}>0$ for each $t\in[0,T]$, we have $a_{t}>0$ on $[0,T]$. Hence for each $(t,x)\in[0,T]\times\bbR^{d}$ and $y\in\bbR$, the sign of $D_{k}(t,x)-y$ coincides with the sign of $f_{k}(t,p_{t}x)-(y-c_{t}x_{k})/a_{t}$, so that every set of points shattered by the subgraphs of $\Pi_{k}\scrD$ induces a set of the same cardinality shattered by the subgraphs of $\Pi_{k}\scrF$, where $\scrF\subseteq N(L,S)$ denotes the network class in the definition~\eqref{eq:scrD}. Then by applying Lemma~\ref{lemma:vcdim}, the VC-dimension of the class $\scrD$ is given as
\begin{equation}\label{eq:proof:theorem:velocity:rate:2}
\vcdim(\Pi_{k}\scrD)\leq\vcdim(\Pi_{k}\scrF)\leq CN^{d+1}\log N,
\end{equation}
where $C$ is an absolute constant. Plugging~\eqref{eq:proof:theorem:velocity:rate:1} and~\eqref{eq:proof:theorem:velocity:rate:2} into Lemma~\ref{lemma:oracle:velocity} yields
\begin{align*}
\bbE_{\euS}\big[\calE_{T}(\what{D})\big]
\leq CR^{2}\Big(\frac{1}{N^{2}}+\frac{N^{d+1}\log N}{n\log^{-1}n}+\frac{1}{\exp(\theta R^{2})}\Big),
\end{align*}
where $\theta$ is a constant only depending on $\sigma$, and $C$ is a constant only depending on $d$, $\sigma$ and the coefficients $(\alpha_{t},\beta_{t})$. By setting $N=Cn^{\frac{1}{d+3}}$, we obtain that
\begin{equation}\label{eq:proof:theorem:velocity:rate:3}
\bbE_{\euS}\big[\calE_{T}(\what{D})\big]\leq CR^{2}\Big(n^{-\frac{2}{d+3}}\log(n)+\exp(-\theta R^{2})\Big).
\end{equation}
Then by substituting $R^{2}=\log(n)\theta^{-1}$, we obtain the desired result.
\end{proof}

\par The discretization analysis in Appendix~\ref{section:proof:discretization} consumes the matching error through its values on the sampling grid $\{kh\}_{k=0}^{K-1}$, while Theorem~\ref{theorem:velocity:rate} controls the time-averaged risk $\calE_{T}(\what{D})$. The following lemma bridges the two quantities for estimators in the hypothesis class $\scrD$; its right-hand side may serve as the quantity $\varepsilon^{2}$ in Lemma~\ref{lemma:discretization}.

\begin{lemma}[From the time-averaged risk to the grid risk]\label{lemma:matching:grid}
Suppose that Assumptions~\ref{assumption:init:dist:gaussian} and~\ref{assumption:target:dist} hold. Let $T\in(1/2,1)$ and $K\in\bbN_{+}$, and set $h=T/K$. Then for each $\what{D}\in\scrD$ with the hypothesis class $\scrD$ as in Lemma~\ref{lemma:approx:velocity}, the following inequality holds
\begin{equation*}
\frac{1}{K}\sum_{k=0}^{K-1}\bbE_{X_{kh}\sim\mu_{kh}}\Big[\|D^{*}(kh,X_{kh})-\what{D}(kh,X_{kh})\|_{2}^{2}\Big]\leq\calE_{T}(\what{D})+C_{\mathrm{grid}}h,
\end{equation*}
where $C_{\mathrm{grid}}$ is a constant only depending on $d$, $\sigma$ and the coefficients $(\alpha_{t},\beta_{t})$.
\end{lemma}

\begin{proof}[Proof of Lemma~\ref{lemma:matching:grid}]
Write $e(t,x)\coloneqq\what{D}(t,x)-D^{*}(t,x)$ and
\begin{equation*}
\psi(t)\coloneqq\bbE_{X_{t}\sim\mu_{t}}\big[\|e(t,X_{t})\|_{2}^{2}\big]=\bbE_{X_{0}\sim\mu_{0}}\big[\|e(t,g_{0,t}^{*}(X_{0}))\|_{2}^{2}\big], \quad t\in[0,T],
\end{equation*}
where we used $g_{0,t}^{*}(X_{0})\stackrel{\d}{=}X_{t}$. Then it is straightforward that
\begin{equation}\label{eq:proof:lemma:matching:grid:1}
\frac{1}{K}\sum_{k=0}^{K-1}\psi(kh)-\calE_{T}(\what{D})
=\frac{1}{T}\sum_{k=0}^{K-1}\int_{kh}^{(k+1)h}\big(\psi(kh)-\psi(t)\big)\dt
\leq\frac{1}{T}\sum_{k=0}^{K-1}\int_{kh}^{(k+1)h}|\psi(kh)-\psi(t)|\dt.
\end{equation}
It suffices to show that $\psi$ is Lipschitz continuous on $[0,T]$ with a uniform constant. By the chain rule and~\eqref{eq:characteristic:flow}, for almost every $t\in(0,T)$,
\begin{equation}\label{eq:proof:lemma:matching:grid:2}
\frac{\mathrm{d}}{\mathrm{d}t}\big\|e(t,g_{0,t}^{*}(x_{0}))\big\|_{2}^{2}
=2\Big\langle e(t,g_{0,t}^{*}(x_{0})),\,\partial_{t}e(t,g_{0,t}^{*}(x_{0}))+\nabla e(t,g_{0,t}^{*}(x_{0}))\,b^{*}(t,g_{0,t}^{*}(x_{0}))\Big\rangle.
\end{equation}
Write $y\coloneqq g_{0,t}^{*}(x_{0})$ and $R_{y}\coloneqq\max\{1,\|y\|_{\infty}\}$, so that $y\in\bbB_{R_{y}}^{\infty}$ and $R_{y}\leq1+\|y\|_{2}$. We bound the factors in~\eqref{eq:proof:lemma:matching:grid:2} separately. First, part~(a) of Proposition~\ref{proposition:reg:denoiser} gives $\|D^{*}(t,y)\|_{2}\leq\sqrt{d}B_{\rm deno}R_{y}$, while $\|\what{D}(t,y)\|_{2}\leq\sqrt{d}(1+\sigma^{2}/\delta_{0})(1+\|y\|_{\infty})$ as verified in the proof of Theorem~\ref{theorem:velocity:rate}, whence
\begin{equation*}
\|e(t,y)\|_{2}\leq\sqrt{d}\Big(B_{\rm deno}+1+\frac{\sigma^{2}}{\delta_{0}}\Big)(1+\|y\|_{2}).
\end{equation*}
Second, part~(c) of Proposition~\ref{proposition:reg:denoiser} gives $\|\partial_{t}D^{*}(t,y)\|_{2}\leq D_{\rm deno}R_{y}$. For $\what{D}(t,x)=c_{t}x+a_{t}f(t,p_{t}x)$, the chain rule yields
\begin{equation*}
\partial_{t}\what{D}(t,y)=\dot{c}_{t}y+\dot{a}_{t}f(t,p_{t}y)+a_{t}\big(\partial_{t}f(t,p_{t}y)+\dot{p}_{t}\nabla f(t,p_{t}y)\,y\big),
\end{equation*}
and therefore, by the constraints in~\eqref{eq:scrD}, the bounds on $|\dot{a}_{t}|$ and $|\dot{c}_{t}|$ in~\eqref{eq:proof:proposition:reg:denoiser:6}, the bound on $|\dot{p}_{t}|$ in~\eqref{eq:proof:lemma:conditional:exp:time:2}, and $a_{t}\leq1$,
\begin{equation*}
\|\partial_{t}\what{D}(t,y)\|_{2}\leq|\dot{c}_{t}|\|y\|_{2}+\sqrt{d}|\dot{a}_{t}|+3F_{\rm deno}+3d|\dot{p}_{t}|\|y\|_{2}\leq C_{1}(1+\|y\|_{2}),
\end{equation*}
where $C_{1}$ is a constant only depending on $d$, $\sigma$ and the coefficients $(\alpha_{t},\beta_{t})$. Hence
\begin{equation*}
\|\partial_{t}e(t,y)\|_{2}\leq(D_{\rm deno}+C_{1})(1+\|y\|_{2}).
\end{equation*}
Third, by the chain rule, $\nabla\what{D}(t,x)=c_{t}I_{d}+a_{t}p_{t}\nabla f(t,p_{t}x)$, and thus part~(b) of Proposition~\ref{proposition:reg:denoiser},~\eqref{eq:delta:0} and $0\leq\alpha_{t},\beta_{t}\leq1$ imply
\begin{equation*}
\|\nabla e(t,y)\|_{\op}\leq G_{\rm deno}(t)+c_{t}+3d\,a_{t}p_{t}\leq\frac{\sigma^{2}+d}{\delta_{0}}+\frac{\sigma^{2}+3d}{\delta_{0}},
\end{equation*}
where we used $c_{t}\leq\sigma^{2}/\delta_{0}$ and $a_{t}p_{t}=\alpha_{t}^{2}\beta_{t}/\Delta_{t}^{2}\leq1/\delta_{0}$. Fourth, part~(a) of Proposition~\ref{proposition:reg:velocity} gives $\|b^{*}(t,y)\|_{2}\leq\sqrt{d}B_{\vel}R_{y}\leq\sqrt{d}B_{\vel}(1+\|y\|_{2})$. Substituting the above four bounds into~\eqref{eq:proof:lemma:matching:grid:2} yields, for almost every $t\in(0,T)$,
\begin{equation*}
\Big|\frac{\mathrm{d}}{\mathrm{d}t}\big\|e(t,g_{0,t}^{*}(x_{0}))\big\|_{2}^{2}\Big|\leq C_{2}\big(1+\|g_{0,t}^{*}(x_{0})\|_{2}\big)^{2},
\end{equation*}
where $C_{2}$ is a constant only depending on $d$, $\sigma$ and the coefficients $(\alpha_{t},\beta_{t})$. Since $g_{0,t}^{*}(X_{0})\stackrel{\d}{=}X_{t}=\alpha_{t}X_{0}+\beta_{t}X_{1}$ for $(X_{0},X_{1})\sim\mu_{0}\times\mu_{1}$, we have
\begin{equation*}
\bbE\big[\|g_{0,t}^{*}(X_{0})\|_{2}^{2}\big]=\alpha_{t}^{2}\bbE\big[\|X_{0}\|_{2}^{2}\big]+\beta_{t}^{2}\bbE\big[\|X_{1}\|_{2}^{2}\big]\leq d+d(1+\sigma^{2})=d(2+\sigma^{2}),
\end{equation*}
where we used $\bbE[\|X_{0}\|_{2}^{2}]=d$, $\bbE[\|X_{1}\|_{2}^{2}]\leq d(1+\sigma^{2})$ and $0\leq\alpha_{t},\beta_{t}\leq1$. Taking expectation with respect to $X_{0}\sim\mu_{0}$ and integrating in time, we obtain, for each $0\leq s\leq t\leq T$,
\begin{equation*}
|\psi(t)-\psi(s)|\leq\int_{s}^{t}\bbE_{X_{0}\sim\mu_{0}}\Big[\Big|\frac{\mathrm{d}}{\mathrm{d}\tau}\big\|e(\tau,g_{0,\tau}^{*}(X_{0}))\big\|_{2}^{2}\Big|\Big]\,\mathrm{d}\tau\leq C_{\mathrm{grid}}(t-s), \quad C_{\mathrm{grid}}\coloneqq2C_{2}\big(1+d(2+\sigma^{2})\big).
\end{equation*}
Substituting this into~\eqref{eq:proof:lemma:matching:grid:1} yields
\begin{equation*}
\frac{1}{K}\sum_{k=0}^{K-1}\psi(kh)-\calE_{T}(\what{D})\leq\frac{1}{T}\sum_{k=0}^{K-1}\int_{kh}^{(k+1)h}C_{\mathrm{grid}}(t-kh)\dt=\frac{C_{\mathrm{grid}}Kh^{2}}{2T}=\frac{C_{\mathrm{grid}}h}{2}.
\end{equation*}
This completes the proof.
\end{proof}
}

\section{Discretization Error Analysis}\label{section:proof:discretization}

{
\subsection{Derivations of Exponential Integrator}\label{section:proof:discretization:ei}

\par Recall the probaility flow ODE in~\eqref{eq:characteristic:semilinear} on the time subinterval $t\in(kh,(k+1)h)$,
\begin{equation*}
\frac{\mathrm{d}}{\mathrm{d}t}g_{kh,t}^{*}(x_{kh}) = \frac{\dot{\alpha}_{t}}{\alpha_{t}}g_{kh,t}^{*}(x_{kh})-\beta_{t}\Bigl(\frac{\dot{\alpha}_{t}}{\alpha_{t}}-\frac{\dot{\beta}_{t}}{\beta_{t}}\Bigr) D_{t}^{*}(g_{kh,t}^{*}(x_{kh})), \quad x_{kh}\in\mathbb{R}^{d}.
\end{equation*}
Multiplying both sides of the equality by the integrating factor $\alpha_{t}^{-1}$ yields
\begin{equation*}
\frac{\mathrm{d}}{\mathrm{d}t}\Bigl(\frac{1}{\alpha_{t}}g_{kh,t}^{*}(x_{kh})\Bigr) = -\frac{\beta_{t}}{\alpha_{t}}\Bigl(\frac{\dot{\alpha}_{t}}{\alpha_{t}}-\frac{\dot{\beta}_{t}}{\beta_{t}}\Bigr)D_{t}^{*}(g_{kh,t}^{*}(x_{kh})).
\end{equation*}
By integrating both sides of the equality from $t=kh$ to $t=(k+1)h$, one has
\begin{equation*}
\frac{1}{\alpha_{(k+1)h}}g_{kh,(k+1)h}^{*}(x_{kh})-\frac{1}{\alpha_{kh}}g_{kh,kh}^{*}(x_{kh}) = -\int_{kh}^{(k+1)h}\frac{\beta_{t}}{\alpha_{t}}\Bigl(\frac{\dot{\alpha}_{t}}{\alpha_{t}}-\frac{\dot{\beta}_{t}}{\beta_{t}}\Bigr)D_{t}^{*}(g_{kh,t}^{*}(x_{kh}))\,\mathrm{d}t,
\end{equation*}
which implies
\begin{equation}\label{eq:discretization:1}
g_{kh,(k+1)h}^{*}(x_{kh}) = \frac{\alpha_{(k+1)h}}{\alpha_{kh}}x_{kh} - \alpha_{(k+1)h}\int_{kh}^{(k+1)h}\frac{\beta_{t}}{\alpha_{t}}\Bigl(\frac{\dot{\alpha}_{t}}{\alpha_{t}}-\frac{\dot{\beta}_{t}}{\beta_{t}}\Bigr)D_{t}^{*}(g_{kh,t}^{*}(x_{kh}))\,\mathrm{d}t,
\end{equation}
Here we used the fact that $g_{kh,kh}(x_{kh})=x_{kh}$. Equivalently,~\eqref{eq:discretization:1} can be reformulated as
\begin{equation}\label{eq:discretization:2}
g_{0,(k+1)h}^{*}(x_{0}) = \frac{\alpha_{(k+1)h}}{\alpha_{kh}}g_{0,kh}^{*}(x_{0}) - \alpha_{(k+1)h}\int_{kh}^{(k+1)h}\frac{\beta_{t}}{\alpha_{t}}\Bigl(\frac{\dot{\alpha}_{t}}{\alpha_{t}}-\frac{\dot{\beta}_{t}}{\beta_{t}}\Bigr)D_{t}^{*}(g_{0,t}^{*}(x_{0}))\,\mathrm{d}t,
\end{equation}

\par By the same arguments, for the exponential integrator~\eqref{eq:characteristic:ei}, we have
\begin{align}
\widehat{E}_{kh,(k+1)h}(x_{kh})
&= \frac{\alpha_{(k+1)h}}{\alpha_{kh}}x_{kh} - \alpha_{(k+1)h}\int_{kh}^{(k+1)h}\frac{\beta_{t}}{\alpha_{t}}\Bigl(\frac{\dot{\alpha}_{t}}{\alpha_{t}}-\frac{\dot{\beta}_{t}}{\beta_{t}}\Bigr)\widehat{D}_{kh}(\widehat{E}_{kh,kh}(x_{kh}))\,\mathrm{d}t \nonumber \\
&= \frac{\alpha_{(k+1)h}}{\alpha_{kh}}x_{kh} - \alpha_{(k+1)h}\int_{kh}^{(k+1)h}\frac{\beta_{t}}{\alpha_{t}}\Bigl(\frac{\dot{\alpha}_{t}}{\alpha_{t}}-\frac{\dot{\beta}_{t}}{\beta_{t}}\Bigr)\,\mathrm{d}t\widehat{D}_{kh}(x_{kh}) \nonumber \\
&= \frac{\alpha_{(k+1)h}}{\alpha_{kh}}x_{kh} - \beta_{kh}\Bigl(\frac{\alpha_{(k+1)h}}{\alpha_{kh}}-\frac{\beta_{(k+1)h}}{\beta_{kh}}\Bigr)\widehat{D}_{kh}(x_{kh}), \label{eq:discretization:3}
\end{align}
which concludes~\eqref{eq:characteristic:ei:solution}. The equivalent form reads
\begin{equation}\label{eq:discretization:4}
\widehat{E}_{0,(k+1)h}(x_{0}) = \frac{\alpha_{(k+1)h}}{\alpha_{kh}}\widehat{E}_{0,kh}(x_{0}) - \beta_{kh}\Bigl(\frac{\alpha_{(k+1)h}}{\alpha_{kh}}-\frac{\beta_{(k+1)h}}{\beta_{kh}}\Bigr)\widehat{D}_{kh}(\widehat{E}_{0,kh}(x_{0})).
\end{equation}
}

{
\subsection{Error Analysis for Exponential Integrator Sampling}

\par We analyze the discretization error of the exponential integrator for an arbitrary interpolant satisfying Condition~\ref{condition:interpolant}, with the linear interpolant and the F{\"o}llmer flow recovered as special cases. Throughout this subsection, we adopt the notation~\eqref{eq:Delta},~\eqref{eq:delta:0} and~\eqref{eq:interpolant:constant}, and write
\begin{equation}\label{eq:discretization:phi}
\phi_{t}\coloneqq\dot{\beta}_{t}-\frac{\dot{\alpha}_{t}}{\alpha_{t}}\beta_{t}, \quad t\in[0,1),
\end{equation}
which is non-negative by Condition~\ref{condition:interpolant}~(iii). Since the integrand weight in~\eqref{eq:discretization:2} satisfies $-\frac{\beta_{t}}{\alpha_{t}}\bigl(\frac{\dot{\alpha}_{t}}{\alpha_{t}}-\frac{\dot{\beta}_{t}}{\beta_{t}}\bigr)=\phi_{t}/\alpha_{t}$, the evolution equations~\eqref{eq:discretization:2} and~\eqref{eq:discretization:4} can be rewritten, respectively, as
\begin{align}
g_{0,(k+1)h}^{*}(x_{0}) &= \frac{\alpha_{(k+1)h}}{\alpha_{kh}}g_{0,kh}^{*}(x_{0})+\alpha_{(k+1)h}\int_{kh}^{(k+1)h}\frac{\phi_{t}}{\alpha_{t}}D_{t}^{*}(g_{0,t}^{*}(x_{0}))\,\mathrm{d}t, \label{eq:discretization:5} \\
\widehat{E}_{0,(k+1)h}(x_{0}) &= \frac{\alpha_{(k+1)h}}{\alpha_{kh}}\widehat{E}_{0,kh}(x_{0})+w_{k}\widehat{D}_{kh}(\widehat{E}_{0,kh}(x_{0})), \label{eq:discretization:6}
\end{align}
where the step weight is available in closed form for arbitrary coefficients,
\begin{equation}\label{eq:discretization:weight}
w_{k}\coloneqq\alpha_{(k+1)h}\int_{kh}^{(k+1)h}\frac{\phi_{t}}{\alpha_{t}}\,\mathrm{d}t=\alpha_{(k+1)h}\Big(\frac{\beta_{(k+1)h}}{\alpha_{(k+1)h}}-\frac{\beta_{kh}}{\alpha_{kh}}\Big)=\beta_{(k+1)h}-\frac{\alpha_{(k+1)h}}{\alpha_{kh}}\beta_{kh}\geq0,
\end{equation}
with the second equality due to $\phi_{t}/\alpha_{t}=\frac{\mathrm{d}}{\mathrm{d}t}(\beta_{t}/\alpha_{t})$. We further introduce the coefficient functionals
\begin{equation}\label{eq:discretization:PsiPhi}
\Psi_{T}\coloneqq\int_{0}^{T}\phi_{t}\,\mathrm{d}t, \quad \Phi_{T}\coloneqq\int_{0}^{T}\phi_{t}^{2}\,\mathrm{d}t,
\end{equation}
which quantify the accumulation of the local truncation error and of the matching error, respectively.
% For the linear interpolant $\alpha_{t}=1-t$, $\beta_{t}=t$, one has $\phi_{t}=1/(1-t)$, recovering the integrands in the previous, linear-interpolant version of~\eqref{eq:discretization:5} and~\eqref{eq:discretization:6}.
\par We first bound the time regularity of the denoiser along the ground-truth trajectory, which controls the local truncation error. In contrast to the Euler method, the following bound is uniform in $s\in(0,1)$, as a direct consequence of the uniform time regularity of the denoiser in part~(c) of Proposition~\ref{proposition:reg:denoiser}; in particular, no factor $1/(1-s)$ appears.

\begin{lemma}\label{lemma:discretization:denoiser}
Suppose Assumptions~\ref{assumption:init:dist:gaussian} and~\ref{assumption:target:dist} hold. Then for each $s\in(0,1)$,
\begin{equation*}
\Big\|\frac{\mathrm{d}}{\mathrm{d}s}D_{s}^{*}(g_{0,s}^{*})\Big\|_{L^{2}(\mu_{0})} = \mathbb{E}_{X_{0}\sim\mu_{0}}^{1/2}\Big[\Big\|\frac{\mathrm{d}}{\mathrm{d}s}D_{s}^{*}(g_{0,s}^{*}(X_{0}))\Big\|_{2}^{2}\Big] \leq C_{\mathrm{traj}},
\end{equation*}
where $C_{\mathrm{traj}}$ is a constant only depending on $d$, $\sigma$ and the coefficients $(\alpha_{t},\beta_{t})$.
\end{lemma}

\begin{proof}[Proof of Lemma~\ref{lemma:discretization:denoiser}]
According to the chain rule and~\eqref{eq:characteristic:flow}, we have
\begin{align}
\frac{\mathrm{d}}{\mathrm{d}s}D_{s}^{*}(g_{0,s}^{*}(x_{0}))
&= \partial_{s}D_{s}^{*}(g_{0,s}^{*}(x_{0}))+\nabla D_{s}^{*}(g_{0,s}^{*}(x_{0}))\frac{\mathrm{d}}{\mathrm{d}s}g_{0,s}^{*}(x_{0}) \nonumber \\
&= \partial_{s}D_{s}^{*}(g_{0,s}^{*}(x_{0}))+\nabla D_{s}^{*}(g_{0,s}^{*}(x_{0}))b^{*}(s,g_{0,s}^{*}(x_{0})). \label{eq:discretization:denoiser:1}
\end{align}
Write $y\coloneqq g_{0,s}^{*}(x_{0})$ and $R_{y}\coloneqq\max\{1,\|y\|_{\infty}\}$, so that $y\in\bbB_{R_{y}}^{\infty}$ and $R_{y}\leq1+\|y\|_{2}$. We bound the three factors in~\eqref{eq:discretization:denoiser:1}. By part~(c) of Proposition~\ref{proposition:reg:denoiser},
\begin{equation*}
\|\partial_{s}D_{s}^{*}(y)\|_{2}\leq D_{\rm deno}R_{y}\leq D_{\rm deno}(1+\|y\|_{2}).
\end{equation*}
By part~(b) of Proposition~\ref{proposition:reg:denoiser},~\eqref{eq:delta:0} and $0\leq\alpha_{s},\beta_{s}\leq1$, the spatial Jacobian obeys $\|\nabla D_{s}^{*}(y)\|_{\op}\leq G_{\rm deno}(s)\leq(\sigma^{2}+d)/\delta_{0}$. By part~(a) of Proposition~\ref{proposition:reg:velocity},
\begin{equation*}
\|b^{*}(s,y)\|_{2}\leq\sqrt{d}\max_{1\leq k\leq d}|b_{k}^{*}(s,y)|\leq\sqrt{d}B_{\vel}R_{y}\leq\sqrt{d}B_{\vel}(1+\|y\|_{2}).
\end{equation*}
Substituting these three bounds into~\eqref{eq:discretization:denoiser:1} yields
\begin{equation}\label{eq:discretization:denoiser:2}
\Big\|\frac{\mathrm{d}}{\mathrm{d}s}D_{s}^{*}(g_{0,s}^{*}(x_{0}))\Big\|_{2} \leq C(1+\|g_{0,s}^{*}(x_{0})\|_{2}), \quad C\coloneqq D_{\rm deno}+\frac{(\sigma^{2}+d)\sqrt{d}B_{\vel}}{\delta_{0}}.
\end{equation}
Since $g_{0,s}^{*}(X_{0})\stackrel{\mathrm{d}}{=}X_{s}=\alpha_{s}X_{0}+\beta_{s}X_{1}$ for $(X_{0},X_{1})\sim\mu_{0}\times\mu_{1}$, and $X_{0}$, $X_{1}$ are independent with $\bbE[X_{0}]=0$, we have
\begin{equation*}
\bbE[\|g_{0,s}^{*}(X_{0})\|_{2}^{2}]=\alpha_{s}^{2}\bbE[\|X_{0}\|_{2}^{2}]+\beta_{s}^{2}\bbE[\|X_{1}\|_{2}^{2}]\leq d+d(1+\sigma^{2})=d(2+\sigma^{2}),
\end{equation*}
where we used $\bbE[\|X_{0}\|_{2}^{2}]=d$, $\bbE[\|X_{1}\|_{2}^{2}]=\bbE[\|U\|_{2}^{2}]+\sigma^{2}d\leq d(1+\sigma^{2})$ and $0\leq\alpha_{s},\beta_{s}\leq1$. Taking the $L^{2}(\mu_{0})$ norm of~\eqref{eq:discretization:denoiser:2} and applying the Minkowski inequality gives
\begin{equation*}
\Big\|\frac{\mathrm{d}}{\mathrm{d}s}D_{s}^{*}(g_{0,s}^{*})\Big\|_{L^{2}(\mu_{0})} \leq C\big(1+\bbE^{1/2}[\|X_{s}\|_{2}^{2}]\big)\leq C\big(1+\sqrt{d(2+\sigma^{2})}\big)\eqqcolon C_{\mathrm{traj}}.
\end{equation*}
This completes the proof.
\end{proof}

\par We now turn to the discretization error of the exponential integrator. Define
\begin{equation}\label{eq:discretization:cr}
c_{t}\coloneqq\frac{\sigma^{2}\beta_{t}}{\alpha_{t}^{2}+\sigma^{2}\beta_{t}^{2}}, \quad r_{t}\coloneqq\frac{\alpha_{t}^{2}\beta_{t}}{(\alpha_{t}^{2}+\sigma^{2}\beta_{t}^{2})^{2}},
\end{equation}
so that $c_{t}$ coincides with the preconditioning coefficient in~\eqref{eq:denoiser:precondition}, $r_{t}=a_{t}p_{t}$, and $G_{\rm deno}(t)=c_{t}+d\,r_{t}$ in part~(b) of Proposition~\ref{proposition:reg:denoiser}. The structured Lipschitz condition~\eqref{eq:discretization:lip} below requires the spatial Jacobian of the learned denoiser to share the contractive skeleton $c_{t}I_{d}$ of the ground truth; the latter satisfies it with $G=d$ by part~(b) of Proposition~\ref{proposition:reg:denoiser}. This structure is essential: under a plain uniform bound $\|\nabla\widehat{D}\|_{\op}\leq\bar{G}$ with $\bar{G}>1$ the amplification factor in the discrete Gr{\"o}nwall argument below would be of order $\alpha_{T}^{-(\bar{G}-1)}$, which diverges as $T\to1$.

\begin{lemma}[Discretization error]\label{lemma:discretization}
Suppose Assumptions~\ref{assumption:init:dist:gaussian} and~\ref{assumption:target:dist} hold. Assume further that
\begin{equation*}
\frac{1}{K}\sum_{k=0}^{K-1}\mathbb{E}\bigl[\|D_{kh}^{*}(X_{kh})-\widehat{D}_{kh}(X_{kh})\|_{2}^{2}\bigr] \leq \varepsilon^{2},
\end{equation*}
and that, for some constant $G\geq0$,
\begin{equation}\label{eq:discretization:lip}
\|\nabla\widehat{D}(t,x)\|_{\op} \leq c_{t}+r_{t}G, \quad \forall \, (t,x)\in(0,T)\times\mathbb{R}^{d}.
\end{equation}
Then for any $T\in(0,1)$, we have
{
\begin{equation*}
W_{2}^{2}((g_{0,T})_{\sharp}\mu_{0},(\widehat{E}_{0,T})_{\sharp}\mu_{0})\leq\max_{0\leq k\leq K}\big\|g_{0,kh}^{*}-\widehat{E}_{0,kh}\big\|_{L^{2}(\mu_{0})}^{2} \leq C\big(\Psi_{T}^{2}h^{2}+\Phi_{T}\varepsilon^{2}\big),
\end{equation*}
where the constant $C$ only depends on $d$, $\sigma$, $G$ and the coefficients $(\alpha_{t},\beta_{t})$, and is bounded uniformly in $T\in(0,1)$; it is proportional to the square of the amplification factor}
\begin{equation}\label{eq:discretization:Gamma}
\Gamma\coloneqq\exp\Big(\frac{(A_{1}+\sigma^{2}B_{1})T}{\delta_{0}}+\frac{2G}{\delta_{0}^{2}}\Big){,}
\end{equation}
{which collects the per-step amplification of the exponential integrator recursion.}
\end{lemma}

\begin{proof}[Proof of Lemma~\ref{lemma:discretization}]
Write $e_{k}\coloneqq\|g_{0,kh}^{*}-\widehat{E}_{0,kh}\|_{L^{2}(\mu_{0})}$, so that $e_{0}=0$. Subtracting~\eqref{eq:discretization:6} from~\eqref{eq:discretization:5}, inserting the terms $\pm D_{kh}^{*}(g_{0,kh}^{*})$ and $\pm\widehat{D}_{kh}(g_{0,kh}^{*})$, and applying the triangle inequality together with the Minkowski integral inequality, we have
\begin{align}
e_{k+1}
&\leq \frac{\alpha_{(k+1)h}}{\alpha_{kh}}e_{k}+\underbrace{\alpha_{(k+1)h}\int_{kh}^{(k+1)h}\frac{\phi_{t}}{\alpha_{t}}\|D_{t}^{*}(g_{0,t}^{*})-D_{kh}^{*}(g_{0,kh}^{*})\|_{L^{2}(\mu_{0})}\,\mathrm{d}t}_{E_{1}} \nonumber \\
&\quad + \underbrace{w_{k}\|D_{kh}^{*}(g_{0,kh}^{*})-\widehat{D}_{kh}(g_{0,kh}^{*})\|_{L^{2}(\mu_{0})}}_{E_{2}} + \underbrace{w_{k}\|\widehat{D}_{kh}(g_{0,kh}^{*})-\widehat{D}_{kh}(\widehat{E}_{0,kh})\|_{L^{2}(\mu_{0})}}_{E_{3}}, \label{eq:lemma:discretization:1}
\end{align}
where the time index and the spatial argument of the two $\widehat{D}_{kh}$-terms are both frozen at $t=kh$, so that their common weight equals $\alpha_{(k+1)h}\int_{kh}^{(k+1)h}(\phi_{t}/\alpha_{t})\,\mathrm{d}t=w_{k}$ by~\eqref{eq:discretization:weight}.

\par For the summand $E_{1}$ in~\eqref{eq:lemma:discretization:1}, the fundamental theorem of calculus along the trajectory, the Minkowski integral inequality and Lemma~\ref{lemma:discretization:denoiser} imply that, for each $t\in[kh,(k+1)h]$,
\begin{equation*}
\|D_{t}^{*}(g_{0,t}^{*})-D_{kh}^{*}(g_{0,kh}^{*})\|_{L^{2}(\mu_{0})} = \Big\|\int_{kh}^{t}\frac{\mathrm{d}}{\mathrm{d}s}D_{s}^{*}(g_{0,s}^{*})\,\mathrm{d}s\Big\|_{L^{2}(\mu_{0})} \leq \int_{kh}^{t}\Big\|\frac{\mathrm{d}}{\mathrm{d}s}D_{s}^{*}(g_{0,s}^{*})\Big\|_{L^{2}(\mu_{0})}\,\mathrm{d}s \leq C_{\mathrm{traj}}h.
\end{equation*}
Since $\alpha_{t}$ is non-increasing, $\alpha_{(k+1)h}/\alpha_{t}\leq1$ for $t\in[kh,(k+1)h]$, whence
\begin{equation}\label{eq:lemma:discretization:2}
E_{1}\leq C_{\mathrm{traj}}h\,\alpha_{(k+1)h}\int_{kh}^{(k+1)h}\frac{\phi_{t}}{\alpha_{t}}\,\mathrm{d}t\leq C_{\mathrm{traj}}h\int_{kh}^{(k+1)h}\phi_{t}\,\mathrm{d}t.
\end{equation}
The same monotonicity yields $w_{k}\leq\int_{kh}^{(k+1)h}\phi_{t}\,\mathrm{d}t$.

\par For the summand $E_{2}$ in~\eqref{eq:lemma:discretization:1}, the definition of $g^{*}$ gives $g_{0,kh}^{*}(X_{0})\stackrel{\mathrm{d}}{=}X_{kh}$, and thus
\begin{equation}\label{eq:lemma:discretization:3}
E_{2}=w_{k}\,\mathbb{E}^{\frac{1}{2}}\bigl[\|D_{kh}^{*}(X_{kh})-\widehat{D}_{kh}(X_{kh})\|_{2}^{2}\bigr]\eqqcolon w_{k}\eta_{k}.
\end{equation}
For the summand $E_{3}$ in~\eqref{eq:lemma:discretization:1}, the structured Lipschitz condition~\eqref{eq:discretization:lip} implies that $\widehat{D}_{kh}$ is Lipschitz continuous with constant $L_{k}\coloneqq c_{kh}+r_{kh}G$, whence
\begin{equation}\label{eq:lemma:discretization:4}
E_{3}\leq w_{k}L_{k}\|g_{0,kh}^{*}-\widehat{E}_{0,kh}\|_{L^{2}(\mu_{0})}=w_{k}L_{k}e_{k}.
\end{equation}
Substituting~\eqref{eq:lemma:discretization:2},~\eqref{eq:lemma:discretization:3} and~\eqref{eq:lemma:discretization:4} into~\eqref{eq:lemma:discretization:1} yields the recursion
\begin{equation}\label{eq:lemma:discretization:5}
e_{k+1}\leq\rho_{k}e_{k}+C_{\mathrm{traj}}h\int_{kh}^{(k+1)h}\phi_{t}\,\mathrm{d}t+w_{k}\eta_{k}, \quad \rho_{k}\coloneqq\frac{\alpha_{(k+1)h}}{\alpha_{kh}}+w_{k}L_{k}.
\end{equation}

\par We next bound the accumulated amplification $\prod_{k=0}^{K-1}\rho_{k}$. Decompose $\rho_{k}=Q_{k}+w_{k}r_{kh}G$ with $Q_{k}\coloneqq\frac{\alpha_{(k+1)h}}{\alpha_{kh}}+w_{k}c_{kh}$. Substituting~\eqref{eq:discretization:weight}, $c_{kh}=\sigma^{2}\beta_{kh}/\Delta_{kh}$ and $1-\sigma^{2}\beta_{kh}^{2}/\Delta_{kh}=\alpha_{kh}^{2}/\Delta_{kh}$,
\begin{equation*}
Q_{k}=\frac{\alpha_{(k+1)h}}{\alpha_{kh}}\cdot\frac{\alpha_{kh}^{2}}{\Delta_{kh}}+\frac{\sigma^{2}\beta_{kh}\beta_{(k+1)h}}{\Delta_{kh}}=\frac{\alpha_{kh}\alpha_{(k+1)h}+\sigma^{2}\beta_{kh}\beta_{(k+1)h}}{\Delta_{kh}},
\end{equation*}
and therefore, by the elementary inequality $a(b-a)\leq(b^{2}-a^{2})/2$ applied with $(a,b)=(\alpha_{kh},\alpha_{(k+1)h})$ and $(a,b)=(\beta_{kh},\beta_{(k+1)h})$,
\begin{equation*}
Q_{k}-1=\frac{\alpha_{kh}(\alpha_{(k+1)h}-\alpha_{kh})+\sigma^{2}\beta_{kh}(\beta_{(k+1)h}-\beta_{kh})}{\Delta_{kh}}\leq\frac{\Delta_{(k+1)h}-\Delta_{kh}}{2\Delta_{kh}}.
\end{equation*}
Summing over $0\leq k\leq K-1$ and using $\Delta_{kh}\geq\delta_{0}$, $(\Delta_{(k+1)h}-\Delta_{kh})_{+}\leq\int_{kh}^{(k+1)h}|\dot{\Delta}_{t}|\,\mathrm{d}t$ and $|\dot{\Delta}_{t}|\leq2(A_{1}+\sigma^{2}B_{1})$,
\begin{equation}\label{eq:lemma:discretization:6}
\sum_{k=0}^{K-1}(Q_{k}-1)_{+}\leq\frac{1}{2\delta_{0}}\int_{0}^{T}|\dot{\Delta}_{t}|\,\mathrm{d}t\leq\frac{(A_{1}+\sigma^{2}B_{1})T}{\delta_{0}}.
\end{equation}
For the remaining part, writing $w_{k}=(\beta_{(k+1)h}-\beta_{kh})+\frac{\beta_{kh}}{\alpha_{kh}}(\alpha_{kh}-\alpha_{(k+1)h})$ and using $r_{kh}=\alpha_{kh}^{2}\beta_{kh}/\Delta_{kh}^{2}$,
\begin{equation*}
w_{k}r_{kh}=(\beta_{(k+1)h}-\beta_{kh})\frac{\alpha_{kh}^{2}\beta_{kh}}{\Delta_{kh}^{2}}+\frac{\beta_{kh}^{2}\,\alpha_{kh}(\alpha_{kh}-\alpha_{(k+1)h})}{\Delta_{kh}^{2}}\leq\frac{(\beta_{(k+1)h}-\beta_{kh})+(\alpha_{kh}^{2}-\alpha_{(k+1)h}^{2})}{\delta_{0}^{2}},
\end{equation*}
where the inequality used $0\leq\alpha_{t},\beta_{t}\leq1$, $\Delta_{kh}\geq\delta_{0}$ and $\alpha_{kh}(\alpha_{kh}-\alpha_{(k+1)h})\leq\alpha_{kh}^{2}-\alpha_{(k+1)h}^{2}$.
Both differences telescope, and with $\alpha_{0}=1$, $\beta_{0}=0$ and $0\leq\alpha_{T},\beta_{T}\leq1$,
\begin{equation}\label{eq:lemma:discretization:7}
\sum_{k=0}^{K-1}w_{k}r_{kh}G\leq\frac{G(\beta_{T}+1-\alpha_{T}^{2})}{\delta_{0}^{2}}\leq\frac{2G}{\delta_{0}^{2}}.
\end{equation}
Since $1+x\leq e^{x}$ and $w_{k}r_{kh}G\geq0$, combining~\eqref{eq:lemma:discretization:6} and~\eqref{eq:lemma:discretization:7} gives
\begin{equation}\label{eq:lemma:discretization:8}
\prod_{k=0}^{K-1}\rho_{k}\leq\exp\Big(\sum_{k=0}^{K-1}(\rho_{k}-1)_{+}\Big)\leq\exp\Big(\sum_{k=0}^{K-1}\big[(Q_{k}-1)_{+}+w_{k}r_{kh}G\big]\Big)\leq\Gamma.
\end{equation}

\par Iterating the recursion~\eqref{eq:lemma:discretization:5} from $e_{0}=0$ and applying~\eqref{eq:lemma:discretization:8},
\begin{align*}
e_{K}
&\leq\sum_{k=0}^{K-1}\Big(C_{\mathrm{traj}}h\int_{kh}^{(k+1)h}\phi_{t}\,\mathrm{d}t+w_{k}\eta_{k}\Big)\prod_{j=k+1}^{K-1}\rho_{j} \\
&\leq\Gamma\Big(C_{\mathrm{traj}}h\sum_{k=0}^{K-1}\int_{kh}^{(k+1)h}\phi_{t}\,\mathrm{d}t+\sum_{k=0}^{K-1}w_{k}\eta_{k}\Big)=\Gamma\Big(C_{\mathrm{traj}}h\Psi_{T}+\sum_{k=0}^{K-1}w_{k}\eta_{k}\Big),
\end{align*}
where the last equality used~\eqref{eq:discretization:PsiPhi}. For the matching term, the Cauchy--Schwarz inequality together with $w_{k}^{2}\leq h\int_{kh}^{(k+1)h}\phi_{t}^{2}\,\mathrm{d}t$ (again Cauchy--Schwarz) gives
\begin{equation*}
\sum_{k=0}^{K-1}w_{k}\eta_{k}\leq\Big(\sum_{k=0}^{K-1}w_{k}^{2}\Big)^{\frac{1}{2}}\Big(\sum_{k=0}^{K-1}\eta_{k}^{2}\Big)^{\frac{1}{2}}\leq(h\Phi_{T})^{\frac{1}{2}}(K\varepsilon^{2})^{\frac{1}{2}}=\sqrt{T\Phi_{T}}\,\varepsilon,
\end{equation*}
where we used $\sum_{k=0}^{K-1}\eta_{k}^{2}=K\cdot\frac{1}{K}\sum_{k=0}^{K-1}\mathbb{E}\bigl[\|D_{kh}^{*}(X_{kh})-\widehat{D}_{kh}(X_{kh})\|_{2}^{2}\bigr]\leq K\varepsilon^{2}$ and $hK=T$. Consequently,
\begin{equation*}
\|g_{0,T}-\widehat{E}_{0,T}\|_{L^{2}(\mu_{0})}=e_{K}\leq\Gamma\big(C_{\mathrm{traj}}h\Psi_{T}+\sqrt{T\Phi_{T}}\,\varepsilon\big).
\end{equation*}
{Moreover, the same argument applied to the first $k$ steps bounds $e_{k}$ by the same right-hand side for each $0\leq k\leq K$, since all the accumulated sums and products above are non-decreasing in $k$.} Since $(g_{0,T}^{*},\widehat{E}_{0,T})$ is a coupling of $(g_{0,T})_{\sharp}\mu_{0}$ and $(\widehat{E}_{0,T})_{\sharp}\mu_{0}$, and $(u+v)^{2}\leq2u^{2}+2v^{2}$, we conclude
\begin{equation*}
W_{2}^{2}((g_{0,T})_{\sharp}\mu_{0},(\widehat{E}_{0,T})_{\sharp}\mu_{0})\leq e_{K}^{2}\leq{\max_{0\leq k\leq K}e_{k}^{2}\leq}2\Gamma^{2}\big(C_{\mathrm{traj}}^{2}h^{2}\Psi_{T}^{2}+T\Phi_{T}\varepsilon^{2}\big){\leq C\big(\Psi_{T}^{2}h^{2}+\Phi_{T}\varepsilon^{2}\big),}
\end{equation*}
{where the last inequality absorbs $2\Gamma^{2}$, $C_{\mathrm{traj}}^{2}$ and $T$ into the constant $C$.} This completes the proof.
\end{proof}

}

{
\begin{proof}[Proof of Theorem~\ref{theorem:error:euler}]
We apply Lemma~\ref{lemma:discretization} to the estimated denoiser $\what{D}$, which belongs to the hypothesis class $\scrD$ in Theorem~\ref{theorem:velocity:rate}. First, each $D\in\scrD$ is of the preconditioned form $D(t,x)=c_{t}x+a_{t}f(t,p_{t}x)$ with $\|\nabla f(t,x)\|_{\op}\leq3d$, whence
\begin{equation*}
\|\nabla\what{D}(t,x)\|_{\op}\leq c_{t}+3d\,a_{t}p_{t}=c_{t}+3d\,r_{t}, \quad (t,x)\in(0,T)\times\bbR^{d},
\end{equation*}
so that the structured Lipschitz condition~\eqref{eq:discretization:lip} holds with $G=3d$. Second, Lemma~\ref{lemma:matching:grid} bounds the grid risk as
\begin{equation*}
\frac{1}{K}\sum_{k=0}^{K-1}\bbE_{X_{kh}\sim\mu_{kh}}\Big[\|D^{*}(kh,X_{kh})-\what{D}(kh,X_{kh})\|_{2}^{2}\Big]\leq\calE_{T}(\what{D})+C_{\mathrm{grid}}h\eqqcolon\varepsilon^{2}.
\end{equation*}
Since $(g_{0,T})_{\sharp}\mu_{0}=\mu_{T}$, applying Lemma~\ref{lemma:discretization} with $G=3d$ yields
\begin{equation}\label{eq:proof:theorem:error:euler:1}
W_{2}^{2}\Big((\what{E}_{0,T})_{\sharp}\mu_{0},\mu_{T}\Big)\leq\max_{0\leq k\leq K}\big\|g_{0,kh}^{*}-\what{E}_{0,kh}\big\|_{L^{2}(\mu_{0})}^{2}\leq C\big(\Psi_{T}^{2}h^{2}+\Phi_{T}\varepsilon^{2}\big)\leq C\Phi_{T}\Big(\calE_{T}(\what{D})+\frac{1}{K}\Big),
\end{equation}
where the last inequality used $\Psi_{T}^{2}\leq T\Phi_{T}\leq\Phi_{T}$ by the Cauchy--Schwarz inequality, together with $h=T/K\leq1/K$ and $h^{2}\leq h$. Taking expectation with respect to $\euS$ and applying Theorem~\ref{theorem:velocity:rate} give
\begin{equation*}
\bbE_{\euS}\Big[W_{2}^{2}\Big((\what{E}_{0,T})_{\sharp}\mu_{0},\mu_{T}\Big)\Big]\leq C\Phi_{T}\Big\{n^{-\frac{2}{d+3}}\log^{2}n+\frac{1}{K}\Big\},
\end{equation*}
where the constant $C$ only depends on $d$, $\sigma$ and the coefficients $(\alpha_{t},\beta_{t})$. Finally, if $K\geq Cn^{\frac{2}{d+3}}$, the second term in the braces is dominated by the first, which completes the proof.
\end{proof}
}

\begin{proof}[Proof of Corollary~\ref{corollary:error:euler}]
We first show that
\begin{equation}\label{eq:proof:corollary:error:euler:1}
W_{2}(\mu_{T},\mu_{1})\leq\max\{\alpha_{T},1-\beta_{T}\}W_{2}(\mu_{0},\mu_{1}).
\end{equation}
Indeed, let $X_{0}\sim\mu_{0}$ and $X_{1}\sim\mu_{1}$ be two independent random variables. Then $X_{T}=\alpha_{T}X_{0}+\beta_{T}X_{1}$ is a random variable obeying $\mu_{T}$. It follows that
\begin{equation*}
\|X_{T}-X_{1}\|_{2}=\|\alpha_{T}X_{0}-(1-\beta_{T})X_{1}\|_{2}\leq\max\{\alpha_{T},1-\beta_{T}\}\|X_{0}-X_{1}\|_{2}.
\end{equation*}
Taking expectation on both sides of the inequality with respect to $X_{0}$ and $X_{1}$ and recalling the definition of 2-Wasserstein distance implies~\eqref{eq:proof:corollary:error:euler:1}.

\par According to the triangular inequality of the Wasserstein distance~\cite[Chapter 6]{Villani2009Optimal}, we have
\begin{align*}
W_{2}(\what{\mu}_{K},\mu_{1})
&\leq W_{2}(\what{\mu}_{K},\mu_{T})+W_{2}(\mu_{T},\mu_{1}) \\
&\leq W_{2}(\what{\mu}_{K},\mu_{T})+\max\{\alpha_{T},1-\beta_{T}\}W_{2}(\mu_{0},\mu_{1}),
\end{align*}
where $\what{\mu}_{K}\coloneqq(\what{E}_{0,T})_{\sharp}\mu_{0}$ denotes the push-forward distribution of the exponential integrator, and we used inequality~\eqref{eq:proof:corollary:error:euler:1}. Squaring both sides via $(u+v)^{2}\leq2u^{2}+2v^{2}$, taking expectation with respect to $\euS$, and combining with Theorem~\ref{theorem:error:euler} imply the desired result.
\end{proof}

\section{Proof of Results in Section~\ref{section:analysis:characteristic}}\label{section:proof:analysis:characteristic}

\par In this section, we aim to prove Theorem~\ref{theorem:error:characteristic}. Towards this end, we first relate the averaged $2$-Wasserstein distance of the characteristic generator to its $L^{2}$-risk by Lemma~\ref{lemma:W2:ghat}. Then an oracle inequality of $L^{2}$-risk are proposed in Lemma~\ref{lemma:oracle:solution}. Finally, substituting approximation and generalization error bounds into the oracle inequality and using Theorem~\ref{theorem:error:euler} completes the proof.

\begin{lemma}\label{lemma:W2:ghat}
Let $\what{g}_{t,s}$ be the estimator defined as~\eqref{eq:charateristic:erm}. Then it follows that
\begin{align*}
&\frac{2}{T^{2}}\int_{0}^{T}\int_{t}^{T}W_{2}^{2}\Big((\what{g}_{t,s})_{\sharp}\mu_{t},\mu_{s}\Big)\dsdt \\
&\leq\bbE_{Z_{0}\sim\mu_{0}}\Big[\frac{2}{T^{2}}\int_{0}^{T}\int_{t}^{T}\|g^{*}(t,s,Z_{t})-\what{g}(t,s,Z_{t})\|_{2}^{2}\dsdt\Big].
\end{align*}
\end{lemma}

\begin{proof}[Proof of Lemma~\ref{lemma:W2:ghat}]
According to the definition of 2-Wasserstein distance, it follows that
\begin{align*}
W_{2}^{2}\Big((\what{g}_{t,s})_{\sharp}\mu_{t},\mu_{s}\Big)\leq\bbE_{Z_{0}\sim\mu_{0}}\Big[\|\what{g}(t,s,Z_{t})-Z_{s}\|_{2}^{2}\Big].
\end{align*}
Integrating both sides of the inequality with respect to $0\leq t\leq s\leq T$ deduces
\begin{align*}
&\frac{2}{T^{2}}\int_{0}^{T}\int_{t}^{T}W_{2}^{2}\Big((\what{g}_{t,s})_{\sharp}\mu_{t},\mu_{s}\Big)\dsdt \\
&\leq\frac{2}{T^{2}}\int_{0}^{T}\int_{t}^{T}\bbE_{Z_{0}\sim\mu_{0}}\Big[\|Z_{s}-\what{g}(t,s,Z_{t})\|_{2}^{2}\Big]\dsdt \\
&=\frac{2}{T^{2}}\int_{0}^{T}\int_{t}^{T}\bbE_{Z_{t}\sim\mu_{t}}\Big[\|g^{*}(t,s,Z_{t})-\what{g}(t,s,Z_{t})\|_{2}^{2}\Big]\dsdt \\
&=\bbE_{Z_{0}\sim\mu_{0}}\Big[\frac{2}{T^{2}}\int_{0}^{T}\int_{t}^{T}\|g^{*}(t,s,Z_{t})-\what{g}(t,s,Z_{t})\|_{2}^{2}\dsdt\Big],
\end{align*}
which completes the proof.
\end{proof}

\begin{lemma}[Oracle inequality for characteristic fitting]\label{lemma:oracle:solution}
Suppose Assumptions~\ref{assumption:init:dist:gaussian} and~\ref{assumption:target:dist} hold. Let $T\in(1/2,1)$ and $R\in(1,+\infty)$. Further, assume the hypothesis class $\scrG$ satisfies the following conditions for each $0\leq t\leq s\leq T$ and $x\in\bbR^{d}$:
\begin{enumerate}[(i)]
\item $\sup_{1\leq k\leq d}|g_{k}(t,s,x)|\leq B_{\flow}R$,
\item $\|\partial_{t}g(t,s,x)\|_{2},\|\partial_{s}g(t,s,x)\|_{2}\leq 6\sqrt{d}B^{\prime}_{\flow}R$, and
\item $\|\nabla g(t,s,x)\|_{\op}\leq 3d\exp(G_{\rm vel}T)$.
\end{enumerate}
Then the following inequality holds for each $m\geq\max_{1\leq k\leq d}\vcdim(\Pi_{k}\scrG)$,
\begin{equation*}
\bbE_{\euZ}\big[\calR_{T}(\what{g})\big] \leq\inf_{g\in\scrG}\calR_{T,R}(g)+C\bar{e}_{K}^{2}+CR^{2}\max_{1\leq k\leq d}\frac{\vcdim(\Pi_{k}\scrG)}{m\log^{-1}(m)}+\frac{CR^{2}}{K}+\frac{CR^{2}}{\exp(\theta R^{2})},
\end{equation*}
where $\bar{e}_{K}\coloneqq\max_{0\leq k\leq K}\|g_{0,k\tau}^{*}-\what{E}_{0,k\tau}\|_{L^{2}(\mu_{0})}$ denotes the uniform grid error of the exponential integrator~\eqref{eq:characteristic:ei:solution} with step size $\tau=T/K$, the constant $\theta$ only depends on $\sigma$, and the constant $C$ only depends on $d$ and $\sigma$.
\end{lemma}

\begin{proof}[Proof of Lemma~\ref{lemma:oracle:solution}]

\par Recall the set of $m$ random variables $\euZ=\{Z_{0}^{(i)}\}_{i=1}^{m}$ i.i.d. sampled from $\mu_{0}$. Further, let $Z_{t}^{(i)}$ denote the solution of the ODE~\eqref{eq:characteristic} at time $t\in[0,1]$ given initial value $Z_{0}^{(i)}$ for each $1\leq i\leq m$, and let $\what{Z}_{k}^{(i)}$ denote the solution of the exponential integrator~\eqref{eq:characteristic:ei:solution}, namely $\what{Z}_{k}^{(i)}=\what{E}_{0,k\tau}(Z_{0}^{(i)})$, given the same initial value $Z_{0}^{(i)}$ at time $k\tau$ for $0\leq k\leq K$.

\par We first recall the empirical risk of the characteristic fitting:
\begin{equation}\label{eq:proof:lemma:oracle:solution:0:1}
\what{\calR}_{T,m,K}^{\ei}(g)
=\frac{2}{mK^{2}}\sum_{i=1}^{m}\Bigg\{\sum_{k=0}^{K-1}\frac{1}{2}\|\what{Z}_{k}^{(i)}-g(k\tau,k\tau,\what{Z}_{k}^{(i)})\|_{2}^{2}+\sum_{k=0}^{K-1}\sum_{\ell=k+1}^{K-1}\|\what{Z}_{\ell}^{(i)}-g(k\tau,\ell\tau,\what{Z}_{k}^{(i)})\|_{2}^{2}\Bigg\}.
\end{equation}
Then replacing the numerical solutions $\{\what{Z}_{k}^{(i)}:0\leq k\leq K-1\}_{i=1}^{m}$ in the empirical risk by exact solutions $\{Z_{k}^{(i)}:0\leq k\leq K-1\}_{i=1}^{m}$ of the ODE~\eqref{eq:characteristic} yields an auxiliary empirical risk
\begin{equation}\label{eq:proof:lemma:oracle:solution:0:2}
\begin{multlined}
\what{\calR}_{T,m,K}(g)
=\frac{2}{mK^{2}}\sum_{i=1}^{m}\Bigg\{\sum_{k=0}^{K-1}\frac{1}{2}\|g^{*}(k\tau,k\tau,Z_{k\tau}^{(i)})-g(k\tau,k\tau,Z_{k\tau}^{(i)})\|_{2}^{2} \\
+\sum_{k=0}^{K-1}\sum_{\ell=k+1}^{K-1}\|g^{*}(k\tau,\ell\tau,Z_{k\tau}^{(i)})-g(k\tau,\ell\tau,Z_{k\tau}^{(i)})\|_{2}^{2}\Bigg\}.
\end{multlined}
\end{equation}
Next we introduce a spatial truncation to~\eqref{eq:proof:lemma:oracle:solution:0:2}, which implies the following risk:
\begin{equation}\label{eq:proof:lemma:oracle:solution:0:3}
\begin{multlined}
\what{\calR}_{T,R,m,K}(g)=\frac{2}{mK^{2}}\sum_{i=1}^{m}\sum_{k=0}^{K-1}\Bigg\{\frac{1}{2}\|g^{*}(k\tau,k\tau,Z_{k\tau}^{(i)})-g(k\tau,k\tau,Z_{k\tau}^{(i)})\|_{2}^{2} \\
+\sum_{\ell=k+1}^{K-1}\|g^{*}(k\tau,\ell\tau,Z_{k\tau}^{(i)})-g(k\tau,\ell\tau,Z_{k\tau}^{(i)})\|_{2}^{2}\Bigg\}\bbone\{\|Z_{k\tau}^{(i)}\|_{\infty}\leq R\}.
\end{multlined}
\end{equation}
We then define the following semi-discretized risk, which replaces the empirical average with respect to first two variables in~\eqref{eq:proof:lemma:oracle:solution:0:3} by its population :
\begin{equation}\label{eq:proof:lemma:oracle:solution:0:4}
\what{\calR}_{T,R,m}(g)
=\frac{1}{m}\sum_{i=1}^{m}\Big\{\frac{2}{T^{2}}\int_{0}^{T}\int_{t}^{T}\|g^{*}(t,s,Z_{t}^{(i)})-g(t,s,Z_{t}^{(i)})\|_{2}^{2}\bbone\{\|Z_{t}^{(i)}\|_{\infty}\leq R\}\dsdt\Big\}.
\end{equation}
Finally, recall the population risk of the characteristic fitting
\begin{equation}\label{eq:proof:lemma:oracle:solution:0:6}
\calR_{T}(g)=\bbE_{Z_{0}\sim\mu_{0}}\Big[\frac{2}{T^{2}}\int_{0}^{T}\int_{t}^{T}\|g^{*}(t,s,Z_{t})-g(t,s,Z_{t})\|_{2}^{2}\dsdt\Big],
\end{equation}
of which the spatial truncated counterpart is given as
\begin{equation}\label{eq:proof:lemma:oracle:solution:0:5}
\calR_{T,R}(g)
=\bbE_{Z_{0}\sim\mu_{0}}\Big[\frac{2}{T^{2}}\int_{0}^{T}\int_{t}^{T}\|g^{*}(t,s,Z_{t})-g(t,s,Z_{t})\|_{2}^{2}\bbone\{\|Z_{t}\|_{\infty}\leq R\}\dsdt\Big].
\end{equation}

\par According to definitions~\eqref{eq:proof:lemma:oracle:solution:0:1} to~\eqref{eq:proof:lemma:oracle:solution:0:5}, it is straightforward that for each $g\in\scrG$,
\begin{align*}
\calR_{T}(\what{g})
&\leq\Big(\calR_{T}(\what{g})-\calR_{T,R}(\what{g})\Big)+\Big(\calR_{T,R}(\what{g})-2\what{\calR}_{T,R,m}(\what{g})\Big)+2\Big(\what{\calR}_{T,R,m}(\what{g})-\what{\calR}_{T,R,m,K}(\what{g})\Big) \\
&\quad+2\Big(\what{\calR}_{T,R,m,K}(\what{g})-\what{\calR}_{T,m,K}(\what{g})\Big)+\Big(2\what{\calR}_{T,m,K}(\what{g})-\what{\calR}_{T,m,K}^{\ei}(\what{g})\Big)+\what{\calR}_{T,m,K}^{\ei}(g),
\end{align*}
where the inequality follows from the fact that $\what{g}$ is the minimizer of the empirical risk minimizer $\what{\calR}_{T,m,K}^{\ei}$ over the hypothesis class $\scrG$. Taking expectation on both sides of the inequality with respect to $\euZ\sim\mu_{0}^{m}$ yields
\begin{equation}\label{eq:proof:lemma:oracle:solution:0:0}
\begin{aligned}
\bbE_{\euZ}\big[\calR_{T}(\what{g})\big]
&\leq\Big(\calR_{T}(\what{g})-\calR_{T,R}(\what{g})\Big)+\bbE_{\euZ}\Big[\sup_{g\in\scrG}\calR_{T,R}(g)-2\what{\calR}_{T,R,m}(g)\Big] \\
&\quad+2\bbE_{\euZ}\Big[\what{\calR}_{T,R,m}(\what{g})-\what{\calR}_{T,R,m,K}(\what{g})\Big]+2\bbE_{\euZ}\Big[\what{\calR}_{T,R,m,K}(\what{g})-\what{\calR}_{T,m,K}(\what{g})\Big] \\
&\quad+\bbE_{\euZ}\Big[2\what{\calR}_{T,m,K}(\what{g})-\what{\calR}_{T,m,K}^{\ei}(\what{g})\Big]+\inf_{g\in\scrG}\bbE_{\euZ}\Big[\what{\calR}_{T,m,K}^{\ei}(g)\Big],
\end{aligned}
\end{equation}
where we used the inequality $\sup(a+b)\leq\sup(a)+\sup(b)$.

\par Up to now, the $L^{2}$-risk of the estimator $\what{g}$ is divided into six terms by~\eqref{eq:proof:lemma:oracle:solution:0:0}. In the remainder of the proof, we bound six these error terms one by one.

\begin{enumerate}[(i)]
\item The first term in the right-hand side of~\eqref{eq:proof:lemma:oracle:solution:0:0} measures the error caused by the spatial truncation, for which
\begin{equation}\label{eq:proof:lemma:oracle:solution:1:0}
\calR_{T}(\what{g})-\calR_{T,R}(\what{g})
\leq\frac{CR^{2}}{\exp(\theta R^{2})},
\end{equation}
where the constant $\theta$ only depends on $\sigma$, and $C$ only depends on $d$ and $\sigma$.
\item The second term in the right-hand side of~\eqref{eq:proof:lemma:oracle:solution:0:0} is known as the generalization error, for which
\begin{equation}\label{eq:proof:lemma:oracle:solution:2:0}
\bbE_{\euZ}\Big[\sup_{g\in\scrG}\calR_{T,R}(g)-2\what{\calR}_{T,R,m}(g)\Big]\leq CR^{2}\max_{1\leq k\leq d}\frac{\vcdim(\Pi_{k}\scrG)}{m\log^{-1}(m)},
\end{equation}
where the constant $\theta$ only depends on $\sigma$, and $C$ only depends on $d$ and $\sigma$.
\item The third term in the right-hand side of~\eqref{eq:proof:lemma:oracle:solution:0:0} is led by the time discretization. It holds for each $g\in\scrG$ that
\begin{equation}\label{eq:proof:lemma:oracle:solution:3:0}
\what{\calR}_{T,R,m}(g)-\what{\calR}_{T,R,m,K}(g)\leq\frac{CR^{2}}{K},
\end{equation}
where the constant $\theta$ only depends on $\sigma$, and $C$ only depends on $d$ and $\sigma$.
\item The fourth term in the right-hand side of~\eqref{eq:proof:lemma:oracle:solution:0:0} is also a truncation error. By an argument similar to~\eqref{eq:proof:lemma:oracle:solution:1:0}, it holds that
\begin{equation}\label{eq:proof:lemma:oracle:solution:4:0}
\bbE_{\euZ}\Big[\what{\calR}_{T,R,m,K}(\what{g})-\what{\calR}_{T,m,K}(\what{g})\Big]\leq\frac{CR^{2}}{\exp(\theta R^{2})},
\end{equation}
where the constant $\theta$ only depends on $\sigma$, and $C$ only depends on $d$ and $\sigma$.
\item The fifth term in the right-hand side of~\eqref{eq:proof:lemma:oracle:solution:0:0} is caused by the error of the exponential integrator, for which
\begin{equation}\label{eq:proof:lemma:oracle:solution:5:0}
\bbE_{\euZ}\Big[\what{\calR}_{T,m,K}(\what{g})-2\what{\calR}_{T,m,K}^{\ei}(\what{g})\Big]\leq C\bar{e}_{K}^{2},
\end{equation}
where the constant $\theta$ only depends on $\sigma$, and $C$ only depends on $d$ and $\sigma$.
\item The sixth term in the right-hand side of~\eqref{eq:proof:lemma:oracle:solution:0:0} is the empirical risk of the estimator. Using the definition of the empirical risk minimizer, we deduce
\begin{equation}\label{eq:proof:lemma:oracle:solution:6:0}
\begin{aligned}
\bbE_{\euZ}\Big[\what{\calR}_{T,m,K}^{\ei}(g)\Big]
&\leq\inf_{g\in\scrG}\calR_{T,R}(g)+C\bar{e}_{K}^{2}+\frac{CR^{2}}{K}+\frac{CR^{2}}{\exp(\theta R^{2})},
\end{aligned}
\end{equation}
where the constant $\theta$ only depends on $\sigma$, and $C$ only depends on $d$ and $\sigma$.
\end{enumerate}
Plugging~\eqref{eq:proof:lemma:oracle:solution:1:0} to~\eqref{eq:proof:lemma:oracle:solution:6:0} into~\eqref{eq:proof:lemma:oracle:solution:0:0} obtains the desired result.

\par\noindent\emph{Step 1. Estimate the first term in the right-hand side of~\eqref{eq:proof:lemma:oracle:solution:0:0}.}

\par For each hypothesis $g\in\scrG$, by an argument similar to~\eqref{eq:proof:lemma:oracle:velocity:2:2}, we have
\begin{align}
&\bbE_{Z_{t}\sim\mu_{t}}\Big[\|g^{*}(t,s,Z_{t})-g(t,s,Z_{t})\|_{2}^{2}\bbone\{\|Z_{t}\|_{\infty}>R\}\Big] \nonumber \\
&\leq\bbE_{Z_{t}\sim\mu_{t}}^{1/2}\Big[\|g^{*}(t,s,Z_{t})-g(t,s,Z_{t})\|_{2}^{4}\Big]\bbE_{Z_{t}\sim\mu_{t}}^{1/2}\Big[\bbone\{\|Z_{t}\|_{\infty}>R\}\Big] \nonumber \\
&\leq8\Big(\bbE_{Z_{t}\sim\mu_{t}}^{1/2}\Big[\|Z_{s}\|_{2}^{4}\Big]+\bbE_{Z_{t}\sim\mu_{t}}^{1/2}\Big[\|g(t,s,Z_{t})\|_{2}^{4}\Big]\Big)\pr^{1/2}\{\|Z_{t}\|_{\infty}>R\}, \label{eq:proof:lemma:oracle:solution:1:2}
\end{align}
where the first inequality holds from Cauchy-Schwarz inequality, and the second inequality is due to the triangular inequality. By using Assumption~\ref{assumption:target:dist}, we have
\begin{equation*}
Z_{s}\stackrel{\d}{=}X_{s}\stackrel{\d}{=}\alpha_{s}X_{0}+\beta_{s}U+\sigma\beta_{s}\epsilon, \quad U\in\nu,~\epsilon\sim N(0,I_{d}),
\end{equation*}
which implies by an argument similar to~\eqref{eq:proof:lemma:oracle:velocity:2:4} that
\begin{align*}
\bbE_{Z_{s}}^{1/2}\Big[\|Z_{s}\|_{2}^{4}\Big]
&=\bbE_{X_{s}}^{1/2}\Big[\|X_{s}\|_{2}^{4}\Big]\leq\bbE_{(X_{0},U,\epsilon)}^{1/2}\Big[(\|\alpha_{s}X_{0}\|_{2}+\|\beta_{s}U\|_{2}+\|\sigma\beta_{s}\epsilon\|_{2})^{4}\Big] \\
&\leq27\Big(\alpha_{s}^{4}\bbE_{X_{0}}\Big[\|X_{0}\|_{2}^{4}\Big]+\beta_{s}^{4}\bbE_{U}\Big[\|U\|_{2}^{4}\Big]+\sigma^{4}\beta_{s}^{4}\bbE_{\epsilon}\Big[\|\epsilon\|_{2}^{4}\Big]\Big)^{1/2} \\
&\leq81d(\alpha_{s}^{2}+\beta_{s}^{2}+\sigma^{2}\beta_{s}^{2}),
\end{align*}
where the last inequality follows from Lemma~\ref{lemma:4th:moment:Gaussian}. As a consequence,
\begin{equation}\label{eq:proof:lemma:oracle:solution:1:3}
\bbE_{Z_{t}}^{1/2}\Big[\|g^{*}(t,s,Z_{t})\|_{2}^{4}\Big]
=\bbE_{Z_{s}}^{1/2}\Big[\|Z_{s}\|_{2}^{4}\Big]\leq C,
\end{equation}
where the constant $C$ only depends on $d$ and $\sigma$. Additionally, by using the boundedness of $g\in\scrG$, we have
\begin{equation}\label{eq:proof:lemma:oracle:solution:1:4}
\bbE_{Z_{t}}^{1/2}\Big[\|g(t,s,Z_{t})\|_{2}^{4}\Big]\leq dB_{\flow}^{2}R^{2}.
\end{equation}
Further, using~\eqref{eq:proof:lemma:oracle:velocity:2:5} yields
\begin{equation}\label{eq:proof:lemma:oracle:solution:1:5}
\sup_{t\in(0,1)}\pr\big\{\|Z_{t}\|_{\infty}>R\big\}=\sup_{t\in(0,1)}\pr\big\{\|X_{t}\|_{\infty}>R\big\}\leq\frac{2d}{\exp(\theta R^{2})},
\end{equation}
where $\theta$ is a constant only depending on $\sigma$. Substituting inequalities~\eqref{eq:proof:lemma:oracle:solution:1:3},~\eqref{eq:proof:lemma:oracle:solution:1:4} and~\eqref{eq:proof:lemma:oracle:solution:1:5} into~\eqref{eq:proof:lemma:oracle:solution:1:2} deduces
\begin{equation*}
\bbE_{Z_{t}\sim\mu_{t}}\Big[\|g^{*}(t,s,Z_{t})-\what{g}(t,s,Z_{t})\|_{2}^{2}\bbone\{\|Z_{t}\|_{\infty}>R\}\Big]\leq\frac{CR^{2}}{\exp(\theta R^{2})},
\end{equation*}
where $C$ is a constant only depending to $d$ and $\sigma$. Finally, combining the above inequality with definitions~\eqref{eq:proof:lemma:oracle:solution:0:6} and~\eqref{eq:proof:lemma:oracle:solution:0:5} completes the proof of~\eqref{eq:proof:lemma:oracle:solution:1:0}.

\par\noindent\emph{Step 2. Estimate the second term in the right-hand side of~\eqref{eq:proof:lemma:oracle:solution:0:0}.}

For simplicity of notation, we define the $k$-th term of $\calR_{T,R}$~\eqref{eq:proof:lemma:oracle:solution:0:5} and $\what{\calR}_{T,R,m}$~\eqref{eq:proof:lemma:oracle:solution:0:4}, respectively, as
\begin{align*}
\calR_{T,R}^{k}(b)&=\bbE_{Z_{0}\sim\mu_{0}}\Big[\frac{2}{T^{2}}\int_{0}^{T}\int_{t}^{T}(g_{k}^{*}(t,s,Z_{t})-g_{k}(t,s,Z_{t}))^{2}\bbone\{\|Z_{t}\|_{\infty}\leq R\}\dsdt\Big], \\
\what{\calR}_{T,R,m}^{k}(b)&=\frac{1}{m}\sum_{i=1}^{m}\Big\{\frac{2}{T^{2}}\int_{0}^{T}\int_{t}^{T}(g_{k}^{*}(t,s,Z_{t}^{(i)})-g_{k}(t,s,Z_{t}^{(i)}))^{2}\bbone\{\|Z_{t}^{(i)}\|_{\infty}\leq R\}\dsdt\Big\}.
\end{align*}
Applying the boundedness of $g\in\scrG$, Proposition~\ref{corollary:bound:solution}, and Lemma~\ref{lemma:generalization:1} yields
\begin{equation*}
\bbE_{\euZ}\Big[\sup_{g\in\scrG}\calR_{T,R}^{k}(g)-2\what{\calR}_{T,R,m}^{k}(g)\Big]\leq CR^{2}\frac{\vcdim(\Pi_{k}\scrG)}{m\log^{-1}(m)}, \quad 1\leq k\leq d.
\end{equation*}
Summing the above inequalities with respect to $1\leq k\leq d$ completes the proof of~\eqref{eq:proof:lemma:oracle:solution:2:0}.

\par\noindent\emph{Step 3. Estimate the third term in the right-hand side of~\eqref{eq:proof:lemma:oracle:solution:0:0}.}

\par For each fixed $x\in \bbB_{R}^{\infty}$, we define an auxiliary function
\begin{equation*}
u(t,s,x)=\|g^{*}(t,s,x)-\what{g}(t,s,x)\|_{2}^{2}, \quad 0\leq t\leq s\leq T,~x\in\bbB_{R}^{\infty}.
\end{equation*}
It is apparent that the following inequality holds for each $0\leq t\leq s\leq T$ and $x\in \bbB_{R}^{\infty}$,
\begin{equation}\label{eq:proof:lemma:oracle:solution:3:1}
|\partial_{t}u(t,s,x)|\leq2\|g^{*}(t,s,x)-g(t,s,x)\|_{2}\|\partial_{t}g^{*}(t,s,x)-\partial_{t}g(t,s,x)\|_{2}\leq CR^{2},
\end{equation}
where the first inequality follows from Cauchy-Schwarz inequality, and the last inequality used Corollaries~\ref{corollary:bound:solution} and~\ref{corollary:lip:time:solution}, and the definition of hypothesis class $\scrG$. Here the constant $C$ only depends on $d$ and $\sigma$. By the same argument, we have
\begin{equation}\label{eq:proof:lemma:oracle:solution:3:2}
|\partial_{s}u(t,s,x)|\leq CR^{2}, \quad 0\leq t\leq s\leq T,~x\in \bbB_{R}^{\infty}.
\end{equation}
Substituting~\eqref{eq:proof:lemma:oracle:solution:3:1} and~\eqref{eq:proof:lemma:oracle:solution:3:2} into Lemma~\ref{lemma:num:int:err:2D} yields~\eqref{eq:proof:lemma:oracle:solution:3:0}.

\par\noindent\emph{Step 4. Estimate the forth term in the right-hand side of~\eqref{eq:proof:lemma:oracle:solution:0:0}.}

\par We use an argument similar to \emph{Step 1}. For each $0\leq k\leq \ell\leq K-1$, it follows that
\begin{align}
&\bbE_{\euZ}\Big[\|g^{*}(k\tau,\ell\tau,Z_{k\tau}^{(i)})-\what{g}(k\tau,\ell\tau,Z_{k\tau}^{(i)})\|_{2}^{2}\bbone\{\|Z_{k\tau}^{(i)}\|_{\infty}>R\}\Big] \nonumber \\
&\leq8\Big(\bbE_{\euZ}^{1/2}\Big[\|g^{*}(k\tau,\ell\tau,Z_{k\tau}^{(i)})\|_{2}^{4}\Big]+\bbE_{\euZ}^{1/2}\Big[\|\what{g}(k\tau,\ell\tau,Z_{k\tau}^{(i)})\|_{2}^{4}\Big]\Big)\bbE_{\euZ}^{1/2}\Big[\bbone\big\{\|Z_{k\tau}^{(i)}\|_{\infty}>R\big\}\Big] \nonumber \\
&=8\Big(\bbE_{\euZ}^{1/2}\Big[\|Z_{\ell\tau}^{(i)}\|_{2}^{4}\Big]+\bbE_{\euZ}^{1/2}\Big[\|\what{g}(k\tau,\ell\tau,Z_{k\tau}^{(i)})\|_{2}^{4}\Big]\Big)\pr^{1/2}\big\{\|Z_{k\tau}^{(i)}\|_{\infty}>R\big\}, \label{eq:proof:lemma:oracle:solution:4:1}
\end{align}
where we used Cauchy-Schwarz inequality and the triangular inequality. Substituting inequalities~\eqref{eq:proof:lemma:oracle:solution:1:3},~\eqref{eq:proof:lemma:oracle:solution:1:4} and~\eqref{eq:proof:lemma:oracle:solution:1:5} into~\eqref{eq:proof:lemma:oracle:solution:4:1} yields that for $0\leq k\leq \ell\leq K-1$,
\begin{equation}\label{eq:proof:lemma:oracle:solution:4:2}
\bbE_{\euZ}\Big[\|g^{*}(k\tau,\ell\tau,Z_{k\tau}^{(i)})-\what{g}(k\tau,\ell\tau,Z_{k\tau}^{(i)})\|_{2}^{2}\bbone\{\|Z_{k\tau}^{(i)}\|_{\infty}>R\}\Big]\leq\frac{CR^{2}}{\exp(\theta R^{2})},
\end{equation}
where $C$ is a constant only depending on $d$ and $\sigma$. Combining~\eqref{eq:proof:lemma:oracle:solution:4:2} with definitions~\eqref{eq:proof:lemma:oracle:solution:0:2} and~\eqref{eq:proof:lemma:oracle:solution:0:3} deduces~\eqref{eq:proof:lemma:oracle:solution:4:0}.

\par\noindent\emph{Step 5. Estimate the fifth term in the right-hand side of~\eqref{eq:proof:lemma:oracle:solution:0:0}.}

\par For each $1\leq i\leq m$ and $0\leq k\leq\ell\leq K-1$, it follows that
\begin{align*}
&\|\what{Z}_{\ell}^{(i)}-\what{g}(k\tau,\ell\tau,\what{Z}_{k}^{(i)})\|_{2} \\
&\leq\|\what{Z}_{\ell}^{(i)}-Z_{\ell\tau}^{(i)}\|_{2}+\|Z_{\ell\tau}^{(i)}-\what{g}(k\tau,\ell\tau,Z_{k\tau}^{(i)})\|_{2}+\|\what{g}(k\tau,\ell\tau,Z_{k\tau}^{(i)})-\what{g}(k\tau,\ell\tau,\what{Z}_{k}^{(i)})\|_{2} \\
&\leq\|\what{Z}_{\ell}^{(i)}-Z_{\ell\tau}^{(i)}\|_{2}+\|g^{*}(k\tau,\ell\tau,Z_{k\tau}^{(i)})-\what{g}(k\tau,\ell\tau,Z_{k\tau}^{(i)})\|_{2}+\|\what{g}\|_{\lip}\|\what{Z}_{k}^{(i)}-Z_{k\tau}^{(i)}\|_{2},
\end{align*}
where we used the triangular inequality. Squaring both sides of the inequality yields
\begin{align}
&\|\what{Z}_{\ell}^{(i)}-\what{g}(k\tau,\ell\tau,\what{Z}_{k}^{(i)})\|_{2}^{2} \nonumber \\
&\leq4\|\what{Z}_{\ell}^{(i)}-Z_{\ell\tau}^{(i)}\|_{2}^{2}+4\|\what{g}\|_{\lip}^{2}\|\what{Z}_{k}^{(i)}-Z_{k\tau}^{(i)}\|_{2}^{2}+2\|g^{*}(k\tau,\ell\tau,Z_{k\tau}^{(i)})-\what{g}(k\tau,\ell\tau,Z_{k\tau}^{(i)})\|_{2}^{2}. \label{eq:proof:lemma:oracle:solution:5:1}
\end{align}
Substituting~\eqref{eq:proof:lemma:oracle:solution:5:1} into~\eqref{eq:proof:lemma:oracle:solution:0:1} deduces
\begin{align*}
\what{\calR}_{T,m,K}^{\ei}(\what{g})
&\leq\frac{2}{mK^{2}}\sum_{i=1}^{m}\sum_{k=0}^{K-1}\Big\{\frac{1}{2}\Big(4\|\what{Z}_{k}^{(i)}-Z_{k\tau}^{(i)}\|_{2}^{2}+4\|\what{g}\|_{\lip}^{2}\|\what{Z}_{k}^{(i)}-Z_{k\tau}^{(i)}\|_{2}^{2}\Big) \\
&\quad+\sum_{\ell=k+1}^{K-1}\Big(4\|\what{Z}_{\ell}^{(i)}-Z_{\ell\tau}^{(i)}\|_{2}^{2}+4\|\what{g}\|_{\lip}^{2}\|\what{Z}_{k}^{(i)}-Z_{k\tau}^{(i)}\|_{2}^{2}\Big)\Big\}+2\what{\calR}_{T,m,K}(\what{g}){.}
\end{align*}
{Since $\what{Z}_{k}^{(i)}$ and $Z_{k\tau}^{(i)}$ share the same initial value $Z_{0}^{(i)}\sim\mu_{0}$, it follows that
\begin{equation*}
\bbE_{\euZ}\Big[\|\what{Z}_{k}^{(i)}-Z_{k\tau}^{(i)}\|_{2}^{2}\Big]=\big\|g_{0,k\tau}^{*}-\what{E}_{0,k\tau}\big\|_{L^{2}(\mu_{0})}^{2}\leq\bar{e}_{K}^{2}, \quad 0\leq k\leq K.
\end{equation*}
Taking expectation with respect to $\euZ$ on both sides of the previous inequality and using the counting identity $\frac{2}{K^{2}}\sum_{k=0}^{K-1}\big(\frac{1}{2}+K-k-1\big)=1$ then yield
\begin{equation*}
\bbE_{\euZ}\Big[\what{\calR}_{T,m,K}^{\ei}(\what{g})-2\what{\calR}_{T,m,K}(\what{g})\Big]\leq4\big(1+\|\what{g}\|_{\lip}^{2}\big)\bar{e}_{K}^{2}.
\end{equation*}
Plugging $\|\what{g}\|_{\lip}\leq 3d\exp(G_{\rm vel}T)$ implies
\begin{equation}\label{eq:proof:lemma:oracle:solution:5:2}
\bbE_{\euZ}\Big[\what{\calR}_{T,m,K}^{\ei}(\what{g})-2\what{\calR}_{T,m,K}(\what{g})\Big]\leq C\bar{e}_{K}^{2},
\end{equation}
where $C$ is a constant only depending on $d$ and $\sigma$. By the same argument as inequality~\eqref{eq:proof:lemma:oracle:solution:5:2}, we can obtain~\eqref{eq:proof:lemma:oracle:solution:5:0} immediately.}

\par\noindent\emph{Step 6. Estimate the sixth term in the right-hand side of~\eqref{eq:proof:lemma:oracle:solution:0:0}.}

\par For each fixed $g\in\scrG$ independent of $\euZ$, it follows that
\begin{align*}
\bbE_{\euZ}\Big[\what{\calR}_{T,m,K}^{\ei}(g)\Big]
&=\bbE_{\euZ}\Big[\what{\calR}_{T,m,K}^{\ei}(g)-2\what{\calR}_{T,m,K}(g)\Big]+2\bbE_{\euZ}\Big[\what{\calR}_{T,m,K}(g)-\what{\calR}_{T,R,m,K}(g)\Big] \\
&\quad+2\bbE_{\euZ}\Big[\what{\calR}_{T,R,m,K}(g)-\what{\calR}_{T,R,m}(g)\Big]+2\bbE_{\euZ}\Big[\what{\calR}_{T,R,m}(g)\Big],
\end{align*}
where the first term can be estimated by~\eqref{eq:proof:lemma:oracle:solution:5:2}, the second and third terms can be bounded by an argument similar to~\eqref{eq:proof:lemma:oracle:solution:4:0} and~\eqref{eq:proof:lemma:oracle:solution:3:0}, respectively. For the last term, we have $\bbE_{\euZ}[\what{\calR}_{T,R,m}(g)]=\calR_{T,R}(g)$. Combining above results yields~\eqref{eq:proof:lemma:oracle:solution:6:0}.
\end{proof}

\begin{proof}[Proof of Theorem~\ref{theorem:error:characteristic}]
{We first construct a neural network approximating the flow map $g^{*}$. Set $\calX\coloneqq[0,T]\times[0,T]\times[-R,R]^{d}$. Extend $g^{*}$ to $\calX$ by setting $g^{*}(t,s,x)\coloneqq g^{*}(t,\max\{s,t\},x)$ for $s<t$; the extension is Lipschitz on $\calX$, and by Proposition~\ref{proposition:reg:flow}, its first-order derivatives are bounded by $2B^{\prime}_{\flow}R$ in $t$ and $s$, and by $\exp(G_{\rm vel}T)$ in each coordinate of $x$. According to Corollary~\ref{corollary:approx}, for each $1\leq k\leq d$, there exists a real-valued deep neural network $g_{k}$ with depth $L_{k}=\lceil\log_{2}(d+2)\rceil+3$ and number of parameters $S_{k}=(22(d+2)+6)(N+1)^{d+2}$, such that
\begin{equation*}
\|g_{k}-g_{k}^{*}\|_{L^{\infty}(\calX)}\leq2^{d+2}\Big(\frac{2T}{N}\cdot2B^{\prime}_{\flow}R+\sum_{\ell=1}^{d}\frac{2R}{N}\exp(G_{\rm vel}T)\Big)\leq\frac{C_{1}R}{N},
\end{equation*}
and moreover $\|g_{k}\|_{L^{\infty}(\calX)}=\|g_{k}^{*}\|_{L^{\infty}(\calX)}\leq B_{\flow}R$, while the first-order derivatives of $g_{k}$ are bounded by three times those of $g_{k}^{*}$. Set $g(t,s,x)\coloneqq(g_{1},\ldots,g_{d})(t,s,\pi_{R}(x))$, where $\pi_{R}$ is the coordinatewise clipping map at level $R$ in $\bbR^{d}$, realized exactly by two additional ReLU layers with $\calO(d)$ non-zero weights as in the proof of Lemma~\ref{lemma:approx:velocity}, so that $g\in N(L,S)$ with $L=C$ and $S=CN^{d+2}$. Since $\pi_{R}$ is the identity on $\bbB_{R}^{\infty}$, it follows that for each $0\leq t\leq s\leq T$ and $x\in\bbB_{R}^{\infty}$,
\begin{equation*}
\|g(t,s,x)-g^{*}(t,s,x)\|_{2}\leq\Big(\sum_{k=1}^{d}\|g_{k}-g_{k}^{*}\|_{L^{\infty}(\calX)}^{2}\Big)^{1/2}\leq\frac{C_{1}\sqrt{d}R}{N},
\end{equation*}
while $\|g\|_{\infty}\leq B_{\flow}R$, $\max\{\|\partial_{t}g\|_{2},\|\partial_{s}g\|_{2}\}\leq6\sqrt{d}B^{\prime}_{\flow}R$ and $\|\nabla g\|_{\op}\leq3d\exp(G_{\rm vel}T)$ hold on $[0,T]\times[0,T]\times\bbR^{d}$, so that $g$ satisfies conditions~(i) to~(iii) of Lemma~\ref{lemma:oracle:solution} and belongs to the hypothesis class $\scrG$. Consequently, by the definition~\eqref{eq:proof:lemma:oracle:solution:0:5} of the truncated risk,
\begin{equation}\label{eq:proof:theorem:error:characteristic:1}
\inf_{g\in\scrG}\calR_{T,R}(g)\leq\frac{CR^{2}}{N^{2}},
\end{equation}
where $C$ is a constant only depending on $d$ and $\sigma$. On the other hand, by applying Lemma~\ref{lemma:vcdim}, the VC-dimension of this deep neural network class $\scrG$ is given as
\begin{equation}\label{eq:proof:theorem:error:characteristic:2}
\vcdim(\Pi_{k}\scrG)\leq CN^{d+2}\log N,
\end{equation}
where $C$ is an absolute constant. Plugging~\eqref{eq:proof:theorem:error:characteristic:1} and~\eqref{eq:proof:theorem:error:characteristic:2} into Lemma~\ref{lemma:oracle:solution} yields
\begin{align*}
\bbE_{\euZ}\big[\calR_{T}(\what{g})\big]\leq\frac{CR^{2}}{N^{2}}+C\bar{e}_{K}^{2}+CR^{2}\max_{1\leq k\leq d}\frac{N^{d+2}\log N}{m\log^{-1}(m)}+\frac{CR^{2}}{K}+\frac{CR^{2}}{\exp(\theta R^{2})},
\end{align*}
where $C$ is a constant only depending to $d$ and $\sigma$. By setting $N=Cm^{\frac{1}{d+4}}$ and $R^{2}=\log(m)\theta^{-1}$, we have
\begin{align*}
\bbE_{\euZ}\big[\calR_{T}(\what{g})\big]\leq Cm^{-\frac{2}{d+4}}\log^{2}(m)+C\bar{e}_{K}^{2}+\frac{C\log(m)}{K},
\end{align*}
where Lemma~\ref{lemma:W2:ghat} further gives $\calD(\what{g})\leq\calR_{T}(\what{g})$. By inequality~\eqref{eq:proof:theorem:error:euler:1} together with Theorem~\ref{theorem:velocity:rate},
\begin{equation*}
\bbE_{\euS}\big[\bar{e}_{K}^{2}\big]\leq C\Phi_{T}\Big(\bbE_{\euS}\big[\calE_{T}(\what{D})\big]+\frac{1}{K}\Big)\leq C\Phi_{T}\Big(n^{-\frac{2}{d+3}}\log^{2}n+\frac{1}{K}\Big).
\end{equation*}
Taking expectation with respect to $\euS$ and combining the last two displays yields the first claimed bound
\begin{equation*}
\bbE_{\euS}\bbE_{\euZ}\big[\calD(\what{g})\big]\leq C\Phi_{T}\Big\{n^{-\frac{2}{d+3}}\log^{2}n+\frac{1}{K}\Big\}+C\Big\{m^{-\frac{2}{d+4}}\log^{2}m+\frac{\log m}{K}\Big\}.
\end{equation*}
For the second claim, observe that $\phi_{t}\geq\dot{\beta}_{t}$ by Condition~\ref{condition:interpolant}~(iii), so the Cauchy--Schwarz inequality gives $\Phi_{T}\geq T^{-1}\big(\int_{0}^{T}\phi_{t}\dt\big)^{2}\geq\beta_{T}^{2}\geq\beta_{1/2}^{2}>0$ for $T\in(1/2,1)$, a positive constant depending only on the coefficients. Suppose now $K\geq Cn^{\frac{2}{d+3}}\log m$ and $m\geq Cn^{\frac{d+4}{d+3}}$. Then $\Phi_{T}/K\leq\Phi_{T}n^{-\frac{2}{d+3}}$ and $\log(m)/K\leq n^{-\frac{2}{d+3}}\leq\beta_{1/2}^{-2}\Phi_{T}n^{-\frac{2}{d+3}}$. Moreover, since the map $m\mapsto m^{-\frac{2}{d+4}}\log^{2}m$ is non-increasing for $\log m\geq d+4$, it follows that $m^{-\frac{2}{d+4}}\log^{2}m\leq Cn^{-\frac{2}{d+3}}\log^{2}n\leq C\beta_{1/2}^{-2}\Phi_{T}n^{-\frac{2}{d+3}}\log^{2}n$. Combining the above estimates completes the proof.}
\end{proof}

\par We conclude this section by giving an error bound for 2-dimensional numerical integral.

\begin{lemma}\label{lemma:num:int:err:2D}
Let $T>0$ and $K\in\bbN_{+}$. Assume that $u\in W^{1,\infty}([0,T]^{2})$. Define the step size as $\tau=T/K$, and define $\{t_{\ell}=\ell\tau\}_{\ell=0}^{K}$ as the set of time points. Then it follows that
\begin{align*}
&\frac{T^{2}}{K^{2}}\sum_{k=1}^{K}\Big\{\frac{1}{2}u(t_{k-1},t_{k-1})+\sum_{\ell=k+1}^{K}u(t_{k-1},t_{\ell-1})\Big\}-\int_{0}^{T}\int_{t}^{T}u(s,t)\dsdt \\
&\leq\big(\|\partial_{t}u\|_{L^{\infty}([0,T]^{2})}+\|\partial_{s}u\|_{L^{\infty}([0,T]^{2})}\big)\frac{T}{K}.
\end{align*}
\end{lemma}

\begin{proof}[Proof of Lemma~\ref{lemma:num:int:err:2D}]
According to the Taylor expansion of $u(s,t)$ around $(t_{k-1},t_{\ell-1})$ with $1\leq k\leq \ell\leq K$, it follows that
\begin{equation}\label{eq:proof:lemma:num:int:err:2D:1}
u(t,s)=u(t_{k-1},t_{\ell-1})+\partial_{t}u(\theta_{k-1}^{t},t_{\ell-1})(t-t_{k-1})+\partial_{s}u(t_{k-1},\theta_{\ell-1}^{s})(s-t_{\ell-1}),
\end{equation}
where $\theta_{k-1}^{t}\in[t_{k-1},t]$ and $\theta_{\ell-1}^{s}\in[t_{\ell-1},s]$.

\par For $1\leq\ell=k\leq K$, integrating both sides of~\eqref{eq:proof:lemma:num:int:err:2D:1} on $(t,s)\in[t_{k-1},t_{k}]\times[t,t_{k}]$ yields
\begin{align*}
&\int_{t_{k-1}}^{t_{k}}\int_{t}^{t_{k}}u(t,s)\dsdt-u(t_{k-1},t_{k-1})\frac{\tau^{2}}{2} \\
&=\partial_{t}u(\theta_{k-1}^{t},t_{k-1})\int_{t_{k-1}}^{t_{k}}(t_{k}-t)(t-t_{k-1})\dt+\partial_{s}u(t_{k-1},\theta_{k-1}^{s})\int_{t_{k-1}}^{t_{k}}\int_{t}^{t_{k}}(s-t_{k-1})\dsdt \\
&=\partial_{t}u(\theta_{k-1}^{t},t_{k-1})\frac{\tau^{3}}{6}+\partial_{s}u(t_{k-1},\theta_{k-1}^{s})\frac{\tau^{3}}{3},
\end{align*}
where we used the fact that $t_{k}-t_{k-1}=\tau$. By summing both sides of the above equality with respect to $k=1,\ldots,K$, we obtain
\begin{align}
&\frac{T^{2}}{2K^{2}}\sum_{k=1}^{K}u(t_{k-1},t_{k-1})-\sum_{k=1}^{K}\int_{t_{k-1}}^{t_{k}}\int_{t}^{t_{k}}u(t,s)\dsdt \nonumber \\
&\leq\sup_{(t,s)\in[0,T]^{2}}\big\{|\partial_{t}u(t,s)|+|\partial_{s}u(t,s)|\big\}\frac{T^{3}}{3K^{2}}. \label{eq:proof:lemma:num:int:err:2D:2}
\end{align}

\par For $1\leq\ell<k\leq K$, integrating both sides of~\eqref{eq:proof:lemma:num:int:err:2D:1} on $(t,s)\in[t_{k-1},t_{k}]\times[t_{\ell-1},t_{\ell}]$ yields
\begin{align*}
&\int_{t_{k-1}}^{t_{k}}\int_{t_{\ell-1}}^{t_{\ell}}u(t,s)\dsdt-u(t_{k-1},t_{\ell-1})\tau^{2} \\
&=\partial_{t}u(\theta_{k-1}^{t},t_{\ell-1})\tau\int_{t_{k-1}}^{t_{k}}(t-t_{k-1})\dt+\partial_{s}u(t_{k-1},\theta_{\ell-1}^{s})\tau\int_{t_{\ell-1}}^{t_{\ell}}(s-t_{\ell-1})\ds \\
&=\partial_{t}u(\theta_{k-1}^{t},t_{\ell-1})\frac{\tau^{3}}{2}+\partial_{s}u(t_{k-1},\theta_{\ell-1}^{s})\frac{\tau^{3}}{2}.
\end{align*}
By a similar argument to~\eqref{eq:proof:lemma:num:int:err:2D:2}, it follows that
\begin{align}
&\frac{T^{2}}{K^{2}}\sum_{k=1}^{K}\sum_{\ell=k+1}^{K}u(t_{k-1},t_{\ell-1})-\sum_{k=1}^{K}\sum_{\ell=k+1}^{K}\int_{t_{k-1}}^{t_{k}}\int_{t_{\ell-1}}^{t_{\ell}}u(t,s)\dsdt \nonumber \\
&\leq\sup_{(t,s)\in[0,T]^{2}}\big\{|\partial_{t}u(t,s)|+|\partial_{s}u(t,s)|\big\}\frac{T^{3}}{4K}. \label{eq:proof:lemma:num:int:err:2D:3}
\end{align}
Summing~\eqref{eq:proof:lemma:num:int:err:2D:2} and~\eqref{eq:proof:lemma:num:int:err:2D:3} completes the proof.
\end{proof}

{
\section{Proof of Results in Section~\ref{section:analysis:applications}}\label{section:proof:analysis:applications}

\par In this section, we prove Corollaries~\ref{corollary:rate:linear} and~\ref{corollary:rate:follmer}. We first record the closed form of the coefficient functional $\Phi_{T}$ in~\eqref{eq:Phi:T} for the two interpolants, together with a moment bound. For the linear interpolant $\alpha_{t}=1-t$ and $\beta_{t}=t$, we have $\phi_{t}=\dot{\beta}_{t}-\frac{\dot{\alpha}_{t}}{\alpha_{t}}\beta_{t}=\frac{1}{1-t}$, so that
\begin{equation}\label{eq:Phi:linear}
\Phi_{T}=\int_{0}^{T}\frac{\dt}{(1-t)^{2}}=\frac{T}{1-T}\leq\frac{1}{1-T}.
\end{equation}
For the F{\"o}llmer flow $\alpha_{t}=\sqrt{1-t^{2}}$ and $\beta_{t}=t$, we have $\frac{\dot{\alpha}_{t}}{\alpha_{t}}=\frac{-t}{1-t^{2}}$ and thus $\phi_{t}=\frac{1}{1-t^{2}}$, so that
\begin{equation}\label{eq:Phi:follmer}
\Phi_{T}=\int_{0}^{T}\frac{\dt}{(1-t^{2})^{2}}=\frac{T}{2(1-T^{2})}+\frac{1}{4}\log\frac{1+T}{1-T}\leq\frac{1}{1-T}.
\end{equation}
Moreover, under Assumption~\ref{assumption:init:dist:gaussian} together with Assumption~\ref{assumption:target:dist} or Assumption~\ref{assumption:target:dist:manifold}, both $\mu_{0}$ and $\mu_{1}$ have bounded second moments, whence
\begin{equation}\label{eq:W2:moment}
W_{2}^{2}(\mu_{0},\mu_{1})\leq2\bbE\big[\|X_{0}\|_{2}^{2}\big]+2\bbE\big[\|X_{1}\|_{2}^{2}\big]\leq C,
\end{equation}
where the constant $C$ only depends on $d$ and $\sigma$.

\begin{proof}[Proof of Corollary~\ref{corollary:rate:linear}]
For the linear interpolant, $\max\{\alpha_{T}^{2},(1-\beta_{T})^{2}\}=(1-T)^{2}$. Combining Corollary~\ref{corollary:error:euler} with~\eqref{eq:Phi:linear} and~\eqref{eq:W2:moment} yields
\begin{equation*}
\bbE_{\euS}\Big[W_{2}^{2}\Big((\what{E}_{0,T})_{\sharp}\mu_{0},\mu_{1}\Big)\Big]\leq\frac{C}{1-T}n^{-\frac{2}{d+3}}\log^{2}n+C(1-T)^{2}.
\end{equation*}
Substituting the choice $1-T=Cn^{-\frac{2}{3(d+3)}}\log^{\frac{2}{3}}n$, which balances the two terms in the right-hand side, bounds both terms by $Cn^{-\frac{4}{3(d+3)}}\log^{\frac{4}{3}}n$, and proves the first claim. For the second claim, Theorem~\ref{theorem:error:characteristic} with $K\geq Cn^{\frac{2}{d+3}}\log m$ and $m\geq Cn^{\frac{d+4}{d+3}}$ gives
\begin{equation*}
\bbE_{\euS}\bbE_{\euZ}\big[\calD(\what{g})\big]\leq C\Phi_{T}n^{-\frac{2}{d+3}}\log^{2}n\leq\frac{C}{1-T}n^{-\frac{2}{d+3}}\log^{2}n=Cn^{-\frac{4}{3(d+3)}}\log^{\frac{4}{3}}n,
\end{equation*}
which completes the proof.
\end{proof}

\begin{proof}[Proof of Corollary~\ref{corollary:rate:follmer}]
For the F{\"o}llmer flow, $\max\{\alpha_{T}^{2},(1-\beta_{T})^{2}\}=\max\{1-T^{2},(1-T)^{2}\}=1-T^{2}\leq2(1-T)$. Combining Corollary~\ref{corollary:error:euler} with~\eqref{eq:Phi:follmer} and~\eqref{eq:W2:moment} yields
\begin{equation*}
\bbE_{\euS}\Big[W_{2}^{2}\Big((\what{E}_{0,T})_{\sharp}\mu_{0},\mu_{1}\Big)\Big]\leq\frac{C}{1-T}n^{-\frac{2}{d+3}}\log^{2}n+C(1-T).
\end{equation*}
Substituting the choice $1-T=Cn^{-\frac{1}{d+3}}\log n$, which balances the two terms in the right-hand side, bounds both terms by $Cn^{-\frac{1}{d+3}}\log n$, and proves the first claim. For the second claim, Theorem~\ref{theorem:error:characteristic} with $K\geq Cn^{\frac{2}{d+3}}\log m$ and $m\geq Cn^{\frac{d+4}{d+3}}$ gives
\begin{equation*}
\bbE_{\euS}\bbE_{\euZ}\big[\calD(\what{g})\big]\leq C\Phi_{T}n^{-\frac{2}{d+3}}\log^{2}n\leq\frac{C}{1-T}n^{-\frac{2}{d+3}}\log^{2}n=Cn^{-\frac{1}{d+3}}\log n,
\end{equation*}
which completes the proof.
\end{proof}
}

\section{Proof of Results in Section~\ref{section:analysis:manifold}}
\label{section:proof:analysis:manifold}

\subsection{Low-dimensional representation}

{ First, analogous to the construction of $ b^*(t, x) $ and $g^*(t, s, x)$ under Assumptions \ref{assumption:init:dist:gaussian} and \ref{assumption:target:dist}, we construct a $ d^* $-dimensional velocity field $ \widetilde{b}^*(t, \widetilde{x}) $, and its corresponding ODE flow $\widetilde{g}^*(t, s, \td{x})$, together with the associated $d^{*}$-dimensional denoiser $\widetilde{D}^{*}(t,\td{x})$}.

\begin{definition} \label{def:manifold b g}
{
Let $\widetilde{X}_0 \sim \widetilde{\mu}_0 = N(0, I_{d^*})$ be independent of $\widetilde{X}_1 \sim \widetilde{\mu}_1 = N(0, \sigma^2I_{d^*})*\widetilde{\nu}$, where $\widetilde{\nu}$ is from Assumption \ref{assumption:target:dist:manifold}. Let $\widetilde{X}_t = \alpha_t \widetilde{X}_0 + \beta_t \widetilde{X}_1$. Then, we define
\begin{align*}
\widetilde{b}^*(t, \widetilde{x}) := \bbE[\dot{\alpha}_t\widetilde{X}_0 + \dot{\beta}_t\widetilde{X}_1|\widetilde{X}_t=\widetilde{x}] \, , \quad \widetilde{x} \in \mathbb{R}^{d^*} \, .
\end{align*}
Also, we define the ODE flow corresponding to $ \widetilde{b}^*(t, \widetilde{x}) $ as $ \widetilde{g}^*(t, s, \td{x}) $, $0 \le t < s < 1$. Moreover, define the $d^{*}$-dimensional denoiser as $\widetilde{D}^{*}(t,\td{x})\coloneqq\bbE[\widetilde{X}_{1}|\widetilde{X}_{t}=\td{x}]$ for each $(t,\td{x})\in(0,1)\times\bbR^{d^{*}}$.}
\end{definition}

Observe that $\widetilde{X}_{t}$ is a stochastic interpolant satisfying Assumptions~\ref{assumption:init:dist:gaussian} and~\ref{assumption:target:dist} in dimension $d^{*}$, with $\widetilde{\nu}$ in place of $\nu$. By directly transferring the regularity estimates for $ b^* $ and $ g^* $ from Section III.B of the main text, we obtain the following regularity characterization for $ \td{b}^* $ and $ \td{g}^* $. Here, we denote by $\td{\bbB}_{R}^{\infty}$ the $\ell_{\infty}$ ball in ${\bbR}^{d^*}$ with radius $R$.

\begin{lemma}\label{lemma:b_tilde properties}
{
Let $T \in (1/2, 1)$, $R \in (1, +\infty)$ and $(t, \td{x}) \in [0, T] \times \td{\bbB}_{R}^{\infty}$. Then, it holds that
\begin{align*}
\max_{1\leq k\leq d^*}|\td{b}_{k}^{*}(t,\td{x})|\leq \td{B}_{\vel}R \, ; \quad \|\nabla \td{b}^{*}(t,\td{x})\|_{\op}\leq \td{G},
\end{align*}
where the constants $\td{B}_{\vel}$ and $\td{G}$ only depend on $d^*$ and $\sigma$.}
\end{lemma}

\begin{lemma} \label{lemma:g_tilde properties}
Let $T \in (1/2, 1)$, $R \in (1, +\infty)$, $0 \leq t \leq s \leq T$ and $x \in \td{\bbB}_{R}^{\infty}$. Then, it holds that
\begin{align*}
&\max_{1\leq k\leq d^*}|\td{g}_{k}^{*}(t,s,\td{x})|\leq \td{B}_{\flow}R \, ; \qquad \|\nabla \td{g}^{*}(t,s,\td{x})\|_{\op}\leq\exp(\td{G}) \, ; \\
&~~~~~~~~ \max\Big\{\|\partial_{t}\td{g}^{*}(t,s,\td{x})\|_{2},\|\partial_{s}\td{g}^{*}(t,s,\td{x})\|_{2}\Big\}\leq \td{B}^{\prime}_{\flow}R \,,
\end{align*}
where the constants $\td{B}_{\flow}$ and $\td{B}^{\prime}_{\flow}$ only depend on $d^*$ and $\sigma$, and $\td{G}$ is from Lemma \ref{lemma:b_tilde properties}.
\end{lemma}
Under Assumptions \ref{assumption:init:dist:gaussian} and \ref{assumption:target:dist:manifold}, with the aid of $ \td{b}^{*} $ and $ \td{g}^{*} $, we can decompose $ b^* $ as $ b^* = b^*_1 + b^*_2 $ and $ g^* $ as $ g^* = g^*_1 + g^*_2 $, where $ b^*_1 $ and $ g^*_1 $ are directly related to $ \td{b}^* $ and $ \td{g}^* $, respectively, and can be regarded as the components of $ b^* $ and $ g^* $ in the space $ W $, which are essentially functions of dimension $ d^* $. On the other hand, $ b^*_2 $ and $ g^*_2 $ can be considered as the components of $ b^* $ and $ g^* $ in the orthogonal space $ W^\perp $, and can be approximated by small-scale neural networks.
\begin{lemma} \label{lemma:b g manifold decomp}
Under Assumptions 1 and 3, the velocity field $ b^*(t, x) $ and the probabilistic ODE flow $ g^*(t, s, x) $ have the following decomposition:
\begin{align}
& b^*(t, x) = P \td{b}^*\big(t, P^{\top}x \big) + \frac{\alpha_t \dot{\alpha}_t + \sigma^2 \beta_t \dot{\beta}_t}{\alpha_t^2 + \sigma^2 \beta_t^2} \big(I_{d }-PP^{\top}\big)x \, , \label{eq:manifold b} \\
& g^*(t, s, x) = P \td{g}^*\big(t, s, P^{\top}x \big) + \sqrt{\frac{\alpha_s^2 + \sigma^2 \beta_s^2}{\alpha_t^2 + \sigma^2 \beta_t^2}} \big(I_{d }-PP^{\top}\big)x \, , \label{eq:manifold g}
\end{align}
where $P \in \bbR^{d \times d^*}$ is from Assumption \ref{assumption:target:dist:manifold}, $ \td{b}^*(t, \td{x}) $ and $ \td{g}^*(t, s, \td{x}) $ are from Definition \ref{def:manifold b g}.
\end{lemma}
\begin{proof} [Proof of Lemma \ref{lemma:b g manifold decomp}]
By Lemma \ref{lemma:conditional:exp} and \eqref{eq:b*}, it holds that
\begin{align} \label{eq:b* score}
b^*(t, x) = \frac{\dot{\beta}_t}{\beta_t} x + \alpha_t^2 \Big(\frac{\dot{\beta}_t}{\beta_t} - \frac{\dot{\alpha}_t}{\alpha_t}\Big) \nabla_x \log \rho_t(x) \, ,
\end{align}
where $X_t \sim \rho_t(x)$. Letting $\td{X}_t \sim \td{\rho}_t(\td{x})$, we analogously get that
\begin{align} \label{eq:b_tilde* score}
\td{b}^*(t, \td{x}) = \frac{\dot{\beta}_t}{\beta_t} \td{x} + \alpha_t^2 \Big(\frac{\dot{\beta}_t}{\beta_t} - \frac{\dot{\alpha}_t}{\alpha_t}\Big) \nabla_{\td{x}} \log \td{\rho}_t(\td{x}) \, , \quad \td{x} \in \td{\bbR} \, .
\end{align}
By Assumption \ref{assumption:target:dist:manifold},
\begin{align} \label{eq:X_t d}
X_t & =  \alpha_t X_0 + \beta_t X_1 \stackrel{d}{=} \alpha_t X_0 + \beta_t (U + \sigma Z^{\prime}) \\
& \stackrel{d}{=}  \sqrt{\alpha_t^2+\sigma^2 \beta^2_t} Z + \beta_t U = \sqrt{\alpha_t^2+\sigma^2 \beta^2_t} Z +  \beta_t P \td{U} \, . \notag
\end{align}
where $\td{U} \sim \td{\nu}$, $U = P \td{U} \sim \nu$, and $Z, Z^{\prime} \sim N(0, I_d)$ are independent of $U$, $\td{U}$ and $X_0$.  Note that we have applied the sum property of independent Gaussian random variables in the third equality above.
By \eqref{eq:X_t d}, we expand $\rho_t(x)$ as follows:
\begin{align*}
\rho_t(x) & = \frac{1}{\sqrt{2\pi}^d \sqrt{\alpha_t^2+\sigma^2\beta_t^2}^d} \int \exp \bigg\{\frac{-\|x- \beta_t u\|_2^2}{2(\alpha_t^2+\sigma^2\beta_t^2)}\bigg\} p(u) \d u \\
& = \frac{1}{\sqrt{2\pi}^d \sqrt{\alpha_t^2+\sigma^2\beta_t^2}^d} \int \exp \bigg\{\frac{-\|x- \beta_t u\|_2^2}{2(\alpha_t^2+\sigma^2\beta_t^2)}\bigg\} \bigg(\int \rho_{u|\td{u}}(u|\td{u}) q(\td{u}) \d \td{u}\bigg) \d u \\
& = \frac{1}{\sqrt{2\pi}^d \sqrt{\alpha_t^2+\sigma^2\beta_t^2}^d} \int \bigg(\int \exp \bigg\{\frac{-\|x- \beta_t u\|_2^2}{2(\alpha_t^2+\sigma^2\beta_t^2)}\bigg\}  \delta_{P \td{u}}(u) \d u \bigg)  q(\td{u}) \d \td{u} \\
& = \frac{1}{\sqrt{2\pi}^d \sqrt{\alpha_t^2+\sigma^2\beta_t^2}^d} \int \exp \bigg\{\frac{-\|x- \beta_t P \td{u}\|_2^2}{2(\alpha_t^2+\sigma^2\beta_t^2)}\bigg\}   q(\td{u}) \d \td{u} \, ,
\end{align*}
where we have used the conditional density of $U$ given $\td{U} = \td{u}$ is $\delta_{P\td{u}}(u)$. Here, $\delta_a(x)$ denotes the $ d $-dimensional Dirac delta density concentrated at $ a $.
Since we have $x = (I_d - PP^{\top})x+PP^{\top}x$, and $\|P y\|_2 = \|y\|_2$ for any $y \in \bbR^{d^*}$, it further holds that
\begin{align*}
\rho_t(x) & = \frac{1}{\sqrt{2\pi}^d \sqrt{\alpha_t^2+\sigma^2\beta_t^2}^d} \int \exp \bigg\{\frac{-\|(I_d - PP^{\top})x+PP^{\top}x- \beta_t P \td{u}\|_2^2}{2(\alpha_t^2+\sigma^2\beta_t^2)}\bigg\}   q(\td{u}) \d \td{u} \\
& = \frac{1}{\sqrt{2\pi}^d \sqrt{\alpha_t^2+\sigma^2\beta_t^2}^d} \int \exp \bigg\{\frac{-\|(I_d - PP^{\top})x\|_2^2-\|P^{\top}x- \beta_t \td{u}\|_2^2}{2(\alpha_t^2+\sigma^2\beta_t^2)}\bigg\}   q(\td{u}) \d \td{u} \\
& = \td{\rho}_t\big(P^\top x\big) \cdot \frac{1}{\sqrt{2\pi}^{d-d^*} \sqrt{\alpha_t^2+\sigma^2\beta_t^2}^{d-d^*}}  \exp \bigg\{\frac{-\|(I_d - PP^{\top})x\|_2^2}{2(\alpha_t^2+\sigma^2\beta_t^2)}\bigg\} \, .
\end{align*}
Thus, we have
\begin{align} \label{eq:manifold rho_t}
\nabla_x \log \rho_t(x) = P \nabla_{\td{x}} \log \td{\rho}_t(P^\top x) - \frac{(I_d-PP^\top)x}{\alpha_t^2+\sigma^2\beta_t^2} \, .
\end{align}
Combining \eqref{eq:b* score}, \eqref{eq:b_tilde* score} and \eqref{eq:manifold rho_t}, we conclude that
\begin{align} \label{eq:b* manifold decomp}
b^*(t, x) = P \td{b}^*\big(t, P^{\top}x \big) + \frac{\alpha_t \dot{\alpha}_t + \sigma^2 \beta_t \dot{\beta}_t}{\alpha_t^2 + \sigma^2 \beta_t^2} \big(I_{d }-PP^{\top}\big)x \, .
\end{align}

Now, Let $$b_1(t, x) =  P \td{b}^*\big(t, P^{\top}x \big) \, , \quad b_2(t, x) = \frac{\alpha_t \dot{\alpha}_t + \sigma^2 \beta_t \dot{\beta}_t}{\alpha_t^2 + \sigma^2 \beta_t^2} x \, .$$ Denote by $ g_1(t, s, x) $ and $ g_2(t, s, x) $ the ODE flows corresponding to $ b_1(t, x) $ and $ b_2(t, x) $, respectively. Note that $ g_2(t, s, x) $ has an explicit expression, namely:
\begin{align*}
g_2(t, s, x) = x \cdot \exp\bigg\{ \int_t^s \frac{\alpha_{\tau} \dot{\alpha}_{\tau} + \sigma^2 \beta_{\tau} \dot{\beta}_{\tau}}{\alpha_{\tau}^2 + \sigma^2 \beta_{\tau}^2} \d \tau \bigg\} = \sqrt{\frac{\alpha_s^2 + \sigma^2 \beta_s^2}{\alpha_t^2 + \sigma^2 \beta_t^2}}  x \, .
\end{align*}
Let $ x(\tau) $ be any solution trajectory of the ODE $ \d x(\tau) = b^*(\tau, x(\tau)) \d \tau$, where $ 0 \leq \tau \leq 1 $. For any given $ 0 \leq t \leq s \leq 1 $, it holds that
\begin{align} \label{eq:g* manifold decomp 1}
g^*(t, s, x(t)) =  x(s) = PP^{\top} x(s) + \big(I_d - PP^{\top} \big)x(s) \, .
\end{align}
By \eqref{eq:b* manifold decomp} and $P^\top P = I_{d^*}$, we have
\begin{align*}
& PP^{\top} x(s) - PP^{\top} x(t) = PP^{\top} \int_t^s b^*(\tau, x(\tau)) \d \tau \\
&~~~~~~ = PP^{\top} \int_t^s \Big[ P \td{b}^*\big({\tau}, P^{\top}x(\tau) \big) + \frac{\alpha_{\tau} \dot{\alpha}_{\tau} + \sigma^2 \beta_{\tau} \dot{\beta}_{\tau}}{\alpha_{\tau}^2 + \sigma^2 \beta_{\tau}^2} \big(I_{d }-PP^{\top}\big)x(\tau) \Big] \d \tau \\
&~~~~~~ = \int_t^s P \td{b}^*\big({\tau}, P^{\top}(PP^{\top}x(\tau)) \big) \d \tau = \int_t^s b_1\big(\tau, PP^\top x(\tau)\big) \d \tau \, ,
\end{align*}
which means $PP^\top x(\tau)$ is a solution trajectory of the ODE $ \d \bar{x}(\tau) = b_1(\tau, \bar{x}(\tau)) \d \tau$. Thus, it holds that
$$PP^\top x(s) = g_1\big(t, s, PP^\top x(t)\big) \, .$$
Similarly, we can obtain $$\big(I_d - PP^{\top} \big) x(s) = g_2\big(t, s, \big(I_d - PP^{\top} \big) x(t)\big) = \sqrt{\frac{\alpha_s^2 + \sigma^2 \beta_s^2}{\alpha_t^2 + \sigma^2 \beta_t^2}} \big(I_{d }-PP^{\top}\big) x(t) \, .$$
Combining these with \eqref{eq:g* manifold decomp 1}, and due to the arbitrariness of $x(t)$, we can infer that
\begin{align} \label{eq:g* manifold decomp 2}
g^*(t, s, x) = g_1\big(t, s, PP^\top x\big) + \sqrt{\frac{\alpha_s^2 + \sigma^2 \beta_s^2}{\alpha_t^2 + \sigma^2 \beta_t^2}} \big(I_{d }-PP^{\top}\big) x \, .
\end{align}
Further, let $\td{x}(\tau)$ be any solution trajectory of the ODE $\d \td{x}(\tau) = \td{b}^*(\tau, \td{x}(\tau)) \d \tau$, $0 \le \tau \le 1$. For any given $0 \le t \le s \le 1$, it holds that
\begin{align*}
& P\td{x}(s) - P\td{x}(t) = P \int_t^s \td{b}^*(\tau, \td{x}(\tau)) \d \tau =  \int_t^s P\td{b}^*\big(\tau, P^\top(P\td{x}(\tau))\big) \d \tau = \int_t^s b_1(\tau, P\td{x}(\tau)) \d \tau \, ,
\end{align*}
which means $P \td{x}(\tau)$ is a solution trajectory of the ODE $ \d \bar{x}(\tau) = b_1(\tau, \bar{x}(\tau)) \d \tau$. Thus,
\begin{align*}
g_1(t, s, P \td{x}(t)) = P \td{x}(s) = P \td{g}^*(t, s, \td{x}(t)) \,\, \Longrightarrow \,\, g_1(t, s, P \td{x}) = P \td{g}^*(t, s, \td{x}) \, , \quad \forall \td{x} \in \bbR^{d^*} \, .
\end{align*}
For any $x \in \bbR^d$, let $\td{x} = P^\top x$. Then, we have
\begin{align*}
g_1\big(t, s, PP^\top x\big) = P \td{g}^*(t, s, P^\top x) \, .
\end{align*}
Combing this with \eqref{eq:g* manifold decomp 2} concludes the proof.
\end{proof}

\par Lemma~\ref{lemma:b g manifold decomp} reveals a low-dimensional structure in $b^{*}$ and $g^{*}$, enabling neural network approximations with complexity depending on the intrinsic dimension $d^{*}$ rather than the ambient dimension $d$.

{
\par We next show that the denoiser $D^{*}$ inherits the same low-dimensional structure, in a form compatible with the preconditioned representation~\eqref{eq:denoiser:representation}.

\begin{lemma}[Low-dimensional structure of the denoiser]\label{lemma:manifold:core}
Suppose Assumptions~\ref{assumption:init:dist:gaussian} and~\ref{assumption:target:dist:manifold} hold. For each $(t,\td{\vartheta})\in[0,1]\times\bbR^{d^{*}}$, define
\begin{equation}\label{eq:tilted:measure:manifold}
\td{f}^{*}(t,\td{\vartheta})\coloneqq\frac{\int_{\bbR^{d^{*}}}\td{u}\exp\big(\langle\td{\vartheta},\td{u}\rangle-\frac{q_{t}}{2}\|\td{u}\|_{2}^{2}\big)\d\td{\nu}(\td{u})}{\int_{\bbR^{d^{*}}}\exp\big(\langle\td{\vartheta},\td{u}^{\prime}\rangle-\frac{q_{t}}{2}\|\td{u}^{\prime}\|_{2}^{2}\big)\d\td{\nu}(\td{u}^{\prime})},
\end{equation}
which is the $d^{*}$-dimensional counterpart of~\eqref{eq:tilted:measure}. Then the following holds.
\begin{enumerate}[(a)]
\item For each $(t,x)\in(0,1)\times\bbR^{d}$, it follows that
\begin{equation}\label{eq:manifold:core}
D^{*}(t,x)=P\td{D}^{*}(t,P^{\top}x)+c_{t}(I_{d}-PP^{\top})x=c_{t}x+a_{t}P\td{f}^{*}(t,p_{t}P^{\top}x).
\end{equation}
\item Lemma~\ref{lemma:conditional:exp:time} applies verbatim to the $d^{*}$-dimensional interpolant $\widetilde{X}_{t}$ in Definition~\ref{def:manifold b g}, with $\td{\nu}$ in place of $\nu$. In particular, $\|\td{f}^{*}(t,\td{\vartheta})\|_{\infty}\leq1$ for each $(t,\td{\vartheta})\in[0,1]\times\bbR^{d^{*}}$, the map $\td{f}^{*}$ is differentiable on $(0,1)\times\bbR^{d^{*}}$, and the bounds in~\eqref{eq:lemma:conditional:exp:time:bound} hold for $\td{f}^{*}$ with $d$ replaced by $d^{*}$ and $F_{\rm deno}$ replaced by $\td{F}_{\rm deno}\coloneqq(A_{1}+B_{1})(d^{*})^{3/2}/\delta_{0}^{2}$.
\end{enumerate}
\end{lemma}

\begin{proof}[Proof of Lemma~\ref{lemma:manifold:core}]
As observed above, the $d^{*}$-dimensional interpolant $\widetilde{X}_{t}$ satisfies Assumptions~\ref{assumption:init:dist:gaussian} and~\ref{assumption:target:dist} in dimension $d^{*}$, with $\td{\nu}$ in place of $\nu$ and $\supp(\td{\nu})\subseteq[0,1]^{d^{*}}$ by Assumption~\ref{assumption:target:dist:manifold}. Hence Lemma~\ref{lemma:conditional:exp:time} and the representation~\eqref{eq:denoiser:representation} apply verbatim to $\widetilde{X}_{t}$, which proves part~(b) and yields
\begin{equation}\label{eq:proof:lemma:manifold:core:1}
\td{D}^{*}(t,\td{x})=c_{t}\td{x}+a_{t}\td{f}^{*}(t,p_{t}\td{x}), \quad (t,\td{x})\in(0,1)\times\bbR^{d^{*}},
\end{equation}
where the coefficients $(a_{t},c_{t},p_{t})$ in~\eqref{eq:denoiser:precondition} remain unchanged, as they only depend on $\sigma$ and the coefficients $(\alpha_{t},\beta_{t})$.
\par We now turn to part~(a). Let $\td{\rho}_{t}$ denote the density of $\widetilde{X}_{t}$, as in the proof of Lemma~\ref{lemma:b g manifold decomp}. Applying Proposition~\ref{proposition:velocity:score} to the interpolants $X_{t}$ and $\widetilde{X}_{t}$, together with~\eqref{eq:manifold rho_t}, implies for each $(t,x)\in(0,1)\times\bbR^{d}$,
\begin{align*}
D^{*}(t,x)
&=\frac{1}{\beta_{t}}\Big(x+\alpha_{t}^{2}\nabla_{x}\log\rho_{t}(x)\Big) \\
&=\frac{1}{\beta_{t}}\Big(PP^{\top}x+\alpha_{t}^{2}P\nabla_{\td{x}}\log\td{\rho}_{t}(P^{\top}x)\Big)+\frac{1}{\beta_{t}}\Big(1-\frac{\alpha_{t}^{2}}{\Delta_{t}}\Big)(I_{d}-PP^{\top})x \\
&=P\td{D}^{*}(t,P^{\top}x)+c_{t}(I_{d}-PP^{\top})x,
\end{align*}
where the second equality holds from~\eqref{eq:manifold rho_t} and $x=PP^{\top}x+(I_{d}-PP^{\top})x$, and the last equality holds from Proposition~\ref{proposition:velocity:score} applied to $\widetilde{X}_{t}$ and $(1-\alpha_{t}^{2}/\Delta_{t})/\beta_{t}=\sigma^{2}\beta_{t}/\Delta_{t}=c_{t}$ by~\eqref{eq:denoiser:precondition}. This proves the first equality in~\eqref{eq:manifold:core}. Plugging~\eqref{eq:proof:lemma:manifold:core:1} into the right-hand side above yields
\begin{equation*}
D^{*}(t,x)=c_{t}PP^{\top}x+a_{t}P\td{f}^{*}(t,p_{t}P^{\top}x)+c_{t}(I_{d}-PP^{\top})x=c_{t}x+a_{t}P\td{f}^{*}(t,p_{t}P^{\top}x),
\end{equation*}
which proves the second equality in~\eqref{eq:manifold:core}. This completes the proof.
\end{proof}
}

\subsection{Convergence rate under low-dimensional linear sub-space assumption}
We first establish the approximation result for the manifold-structured denoiser $D^{*}$.

{
\begin{lemma}[Approximation error under the low-dimensional linear-subspace hypothesis]\label{lemma:approx:velocity:manifold}
Suppose that Assumptions~\ref{assumption:init:dist:gaussian} and~\ref{assumption:target:dist:manifold} hold. Let $T\in(1/2,1)$ and $R\in(1,+\infty)$. Set the hypothesis class $\scrD$ as a preconditioned deep neural network class, which is defined as
\begin{equation}\label{eq:scrD:manifold}
\scrD=\left\{
\begin{aligned}
&D(t,x)=c_{t}x+a_{t}f(t,p_{t}x):~f\in N(L,S), \\
&\|f(t,x)\|_{2}\leq\sqrt{d^{*}},~\|\nabla f(t,x)\|_{\op}\leq3d^{*}, \\
&\|\partial_{t}f(t,x)\|_{2}\leq3\td{F}_{\rm deno},~(t,x)\in[0,T]\times\bbR^{d}
\end{aligned}
\right\},
\end{equation}
where $\td{F}_{\rm deno}$ is the constant in Lemma~\ref{lemma:manifold:core}, and the depth and the number of parameters of the neural network are given, respectively, as $L=C$ and $S=CN^{d^{*}+1}$. Then the following inequality holds for each $N\in\bbN_{+}$,
\begin{equation}\label{eq:manifold-approx-rate}
\inf_{D\in\scrD}\calE_{T,R}(D)\leq CR^{2}N^{-2},
\end{equation}
where $C$ is a constant only depending on $d$, $d^{*}$, $\sigma$ and the coefficients $(\alpha_{t},\beta_{t})$.
\end{lemma}

\begin{proof}[Proof of Lemma~\ref{lemma:approx:velocity:manifold}]
\par By Lemma~\ref{lemma:manifold:core}, the ground-truth denoiser takes the preconditioned form $$D^{*}(t,x)=c_{t}x+a_{t}P\td{f}^{*}(t,p_{t}P^{\top}x)$$ for each $(t,x)\in(0,T]\times\bbR^{d}$, where the core $\td{f}^{*}$ is the $d^{*}$-dimensional posterior mean defined in~\eqref{eq:tilted:measure:manifold}. Hence it suffices to approximate $\td{f}^{*}$ by a $d^{*}$-dimensional neural network and to lift it to $\bbR^{d}$ through the matrix $P$, for which part~(b) of Lemma~\ref{lemma:manifold:core} provides the uniform bounds: for each $(t,\td{\vartheta})\in[0,T]\times\bbR^{d^{*}}$ and $1\leq k,\ell\leq d^{*}$,
\begin{equation}\label{eq:proof:lemma:approx:velocity:manifold:1}
\|\td{f}^{*}(t,\td{\vartheta})\|_{\infty}\leq1, \quad
|\partial_{\td{\vartheta}_{\ell}}\td{f}_{k}^{*}(t,\td{\vartheta})|\leq1, \quad
|\partial_{t}\td{f}_{k}^{*}(t,\td{\vartheta})|\leq\frac{\td{F}_{\rm deno}}{\sqrt{d^{*}}}.
\end{equation}
\par We now construct the neural network. Since $\|P^{\top}x\|_{\infty}\leq\|P^{\top}x\|_{2}\leq\|x\|_{2}\leq\sqrt{d}R$ for each $x\in\bbB_{R}^{\infty}$, and $p_{t}\leq1/\delta_{0}$ for each $t\in[0,1]$ by~\eqref{eq:pq},~\eqref{eq:delta:0} and $0\leq\beta_{t}\leq1$, write $\bar{R}\coloneqq\sqrt{d}R/\delta_{0}$ and $\td{\calX}\coloneqq[0,T]\times[-\bar{R},\bar{R}]^{d^{*}}$, so that $(t,p_{t}P^{\top}x)\in\td{\calX}$ whenever $(t,x)\in[0,T]\times\bbB_{R}^{\infty}$. According to Corollary~\ref{corollary:approx}, for each $1\leq k\leq d^{*}$, there exists a real-valued deep neural network $\td{f}_{k}$ with depth $L_{k}=\lceil\log_{2}(d^{*}+1)\rceil+3$ and number of parameters $S_{k}=(22(d^{*}+1)+6)(N+1)^{d^{*}+1}$, such that
\begin{equation}\label{eq:proof:lemma:approx:velocity:manifold:2}
\|\td{f}_{k}-\td{f}_{k}^{*}\|_{L^{\infty}(\td{\calX})}\leq2^{d^{*}+1}\Big(\frac{T}{N}\|\partial_{t}\td{f}_{k}^{*}\|_{L^{\infty}(\td{\calX})}+\sum_{\ell=1}^{d^{*}}\frac{2\bar{R}}{N}\|\partial_{\td{\vartheta}_{\ell}}\td{f}_{k}^{*}\|_{L^{\infty}(\td{\calX})}\Big)\leq\frac{CR}{N},
\end{equation}
where the last inequality follows from~\eqref{eq:proof:lemma:approx:velocity:manifold:1}, and $C$ is a constant only depending on $d$, $d^{*}$, $\sigma$ and the coefficients $(\alpha_{t},\beta_{t})$. Moreover, Corollary~\ref{corollary:approx} guarantees that
\begin{equation}\label{eq:proof:lemma:approx:velocity:manifold:3}
\|\td{f}_{k}\|_{L^{\infty}(\td{\calX})}=\|\td{f}_{k}^{*}\|_{L^{\infty}(\td{\calX})}\leq1, \quad
\|\partial_{t}\td{f}_{k}\|_{L^{\infty}(\td{\calX})}\leq3\|\partial_{t}\td{f}_{k}^{*}\|_{L^{\infty}(\td{\calX})}, \quad
\|\partial_{\td{\vartheta}_{\ell}}\td{f}_{k}\|_{L^{\infty}(\td{\calX})}\leq3\|\partial_{\td{\vartheta}_{\ell}}\td{f}_{k}^{*}\|_{L^{\infty}(\td{\calX})}.
\end{equation}
The bounds in~\eqref{eq:proof:lemma:approx:velocity:manifold:3} hold on $\td{\calX}$ only. As in the proof of Lemma~\ref{lemma:approx:velocity}, define the coordinatewise clipping map $\pi_{\bar{R}}(\td{\vartheta})\coloneqq(\min\{\max\{\td{\vartheta}_{\ell},-\bar{R}\},\bar{R}\})_{\ell=1}^{d^{*}}$, which is realized exactly by two additional ReLU layers with $\calO(d^{*})$ non-zero weights, and set
\begin{equation*}
f(t,\vartheta)\coloneqq P\td{f}\big(t,\pi_{\bar{R}}(P^{\top}\vartheta)\big), \quad \td{f}\coloneqq(\td{f}_{k})_{k=1}^{d^{*}},
\end{equation*}
where the linear maps $\vartheta\mapsto P^{\top}\vartheta$ and $\td{y}\mapsto P\td{y}$ are realized exactly by the input and output affine layers with $\calO(dd^{*})$ non-zero weights. Since $P$ has orthonormal columns and $\pi_{\bar{R}}$ is $1$-Lipschitz in each coordinate, it follows that for each $(t,\vartheta)\in[0,T]\times\bbR^{d}$,
\begin{equation*}
\|f(t,\vartheta)\|_{2}=\|\td{f}\|_{2}\leq\sqrt{d^{*}}, \quad
\|\nabla f(t,\vartheta)\|_{\op}\leq\|\nabla\td{f}\|_{\op}\leq\Big(\sum_{k=1}^{d^{*}}\sum_{\ell=1}^{d^{*}}|\partial_{\td{\vartheta}_{\ell}}\td{f}_{k}|^{2}\Big)^{1/2}\leq3d^{*}, \quad
\|\partial_{t}f(t,\vartheta)\|_{2}=\|\partial_{t}\td{f}\|_{2}\leq3\td{F}_{\rm deno},
\end{equation*}
where the second inequality in the second bound is led by the Frobenius norm, and the bounds combine~\eqref{eq:proof:lemma:approx:velocity:manifold:3} with~\eqref{eq:proof:lemma:approx:velocity:manifold:1}. The vector-valued network $f$ has depth $L=\max_{1\leq k\leq d^{*}}L_{k}+C$ and number of parameters $S=\sum_{k=1}^{d^{*}}S_{k}+C(dd^{*}+d^{*})\leq CN^{d^{*}+1}$. Hence $D(t,x)\coloneqq c_{t}x+a_{t}f(t,p_{t}x)$ belongs to the hypothesis class $\scrD$ in~\eqref{eq:scrD:manifold}.
\par It remains to bound the truncated risk of $D$. For each $(t,x)\in(0,T]\times\bbB_{R}^{\infty}$, we have $(t,p_{t}P^{\top}x)\in\td{\calX}$, on which $\pi_{\bar{R}}$ is the identity, and thus Lemma~\ref{lemma:manifold:core} together with $a_{t}\leq1$ and the orthonormality of the columns of $P$ implies
\begin{equation*}
\|D(t,x)-D^{*}(t,x)\|_{2}
=a_{t}\big\|P\big(\td{f}(t,p_{t}P^{\top}x)-\td{f}^{*}(t,p_{t}P^{\top}x)\big)\big\|_{2}
\leq\Big(\sum_{k=1}^{d^{*}}\|\td{f}_{k}-\td{f}_{k}^{*}\|_{L^{\infty}(\td{\calX})}^{2}\Big)^{1/2}
\leq\frac{C\sqrt{d^{*}}R}{N},
\end{equation*}
where the last inequality follows from~\eqref{eq:proof:lemma:approx:velocity:manifold:2}. Consequently,
\begin{equation*}
\calE_{T,R}(D)=\frac{1}{T}\int_{0}^{T}\bbE_{X_{t}\sim\mu_{t}}\Big[\|D(t,X_{t})-D^{*}(t,X_{t})\|_{2}^{2}\bbone\{\|X_{t}\|_{\infty}\leq R\}\Big]\dt\leq C^{\prime}R^{2}N^{-2},
\end{equation*}
where $C^{\prime}$ is a constant only depending on $d$, $d^{*}$, $\sigma$ and the coefficients $(\alpha_{t},\beta_{t})$. This completes the proof.
\end{proof}
}

\par With Lemma~\ref{lemma:approx:velocity:manifold} established, we are now ready to derive the convergence rate for {denoiser} matching under the low-dimensional linear-subspace hypothesis. The following theorem is the
manifold counterpart of Theorem~\ref{theorem:velocity:rate}.

{
\begin{theorem}[Convergence rate for denoiser matching under low-dimensional linear-subspace hypothesis]\label{theorem:velocity:rate:manifold}
Suppose Assumptions~\ref{assumption:init:dist:gaussian} and~\ref{assumption:target:dist:manifold} hold. Let $T\in(1/2,1)$. Set the hypothesis class $\scrD$ as the preconditioned deep neural network class~\eqref{eq:scrD:manifold} in Lemma~\ref{lemma:approx:velocity:manifold}, where the depth and the number of non-zero weights of the neural network are given, respectively, as $L=C$ and $S=Cn^{\frac{d^{*}+1}{d^{*}+3}}$. Then the following inequality holds
\begin{align*}
\bbE_{\euS}\big[\calE_{T}(\what{D})\big]\leq Cn^{-\frac{2}{d^{*}+3}}\log^{2}n,
\end{align*}
where $C$ is a constant only depending on $d$, $d^{*}$, $\sigma$ and the coefficients $(\alpha_{t},\beta_{t})$.
\end{theorem}

\begin{proof}[Proof of Theorem~\ref{theorem:velocity:rate:manifold}]
The proof follows the same strategy as that of Theorem~\ref{theorem:velocity:rate} (Appendix~\ref{section:proof:theorem:velocity}). First, for each $D\in\scrD$, since $\|f(t,x)\|_{2}\leq\sqrt{d^{*}}$, $a_{t}\leq1$ and $c_{t}\leq\sigma^{2}/\delta_{0}$ by~\eqref{eq:delta:0}, the growth condition~\eqref{eq:lemma:oracle:velocity:0} holds with $B_{\scrD}=\sqrt{d^{*}}+\sigma^{2}/\delta_{0}$, and hence the oracle inequality in Lemma~\ref{lemma:oracle:velocity} applies. Moreover, since $a_{t}>0$ for each $t\in[0,T]$, the sign-equivalence argument in the proof of Theorem~\ref{theorem:velocity:rate} together with Lemma~\ref{lemma:vcdim} yields $\vcdim(\Pi_{k}\scrD)\leq CN^{d^{*}+1}\log N$. Plugging this bound and the approximation error bound~\eqref{eq:manifold-approx-rate} in Lemma~\ref{lemma:approx:velocity:manifold} into the oracle inequality, and setting $N=Cn^{\frac{1}{d^{*}+3}}$ and $R^{2}=\theta^{-1}\log n$, yields the desired result.
\end{proof}
}

\par Following the same line of reasoning, the manifold counterparts of Lemma~\ref{lemma:discretization}, Theorem~\ref{theorem:error:euler}, and
Corollary~\ref{corollary:error:euler} can be readily obtained by replacing $d$ with $d^{*}$ in the corresponding convergence rates.
We omit these results here for brevity and proceed directly to the analysis of the
characteristic generator under the low-dimensional linear-subspace hypothesis, for which we require an
approximation result for the manifold-structured flow map $g^{*}$. The following lemma provides this approximation result.

\begin{lemma}[Approximation under manifold-structured flow]\label{lemma:approx:flow:manifold}
Let $T\in(1/2,1)$ and $R\in(1,+\infty)$. Suppose $g^{*}$ admits the decomposition in \eqref{eq:g* manifold decomp 2}:
\begin{equation}\label{eq:manifold-structure-flow}
g^{*}(t,s,x)=P\,\widetilde g^{*}(t,s,P^{\top}x)+\iota(t,s)(I_{d}-PP^{\top})x,
\end{equation}
where $P\in\mathbb{R}^{d\times d^{*}}$ has orthonormal columns, $\widetilde g^{*}$ satisfies Lemma~\ref{lemma:g_tilde properties}, and $\iota(t,s):=\sqrt{\frac{\alpha_{s}^{2}+\sigma^{2}\beta_{s}^{2}}{\alpha_{t}^{2}+\sigma^{2}\beta_{t}^{2}}}$. Set the hypothesis class $\scrG$ as a deep neural network class, which is defined as
\begin{equation*}
\scrG=\left\{g\in N(L,S):
\begin{aligned}
&\|g(t,s,x)\|_{\infty}\leq B_{\rm man}^{\flow}R,~\max\{\|\partial_{t}g\|_{2},\|\partial_{s}g\|_{2}\}\leq D_{\rm man}^{\flow}R, \\
&\|\nabla g(t,s,x)\|_{\op}\leq G_{\rm man}^{\flow},~0\le t \le s \le T,~ x\in\bbR^{d}
\end{aligned}
\right\},
\end{equation*}
where the depth and the number of nonzero weights of the neural network are given, respectively, as $L=C$ and $S=CN^{d^{*}+2}$. Then the following inequality holds for each $N\in\bbN_{+}$,
\begin{equation}\label{eq:manifold-approx-rate-flow}
\inf_{g\in\scrG}\,\mathcal{R}_{T,R}(g)\ \le\ CR^{2}N^{-2}.
\end{equation}
Using Lemma \ref{lemma:g_tilde properties}, we have admissible choices for $B_{\rm man}^{\flow}$, $D_{\rm man}^{\flow}$ and $G_{\rm man}^{\flow}$ as follows:
\[
B_{\rm man}^{\flow}=\sqrt d(\widetilde B_{\flow} \sqrt{d^*}+\iota_{\rm max}),\quad
D_{\rm man}^{\flow}=3d(2\widetilde B^{\prime}_{\flow}+\iota_{\rm max}),\quad
G_{\rm man}^{\flow}=3d^{*}\exp(\widetilde G) + \iota_{\rm max}
\]
with $\iota_{\rm max}:= \max\big\{\|\iota\|_{L^{\infty}}, \|\partial_{t}\iota\|_{L^{\infty}}, \|\partial_{s}\iota\|_{L^{\infty}}\big\}$ finite by Condition~\ref{condition:interpolant}. And $C$ denotes a constant may only depend on $d^*$, $d$ and $\sigma$.
\end{lemma}

{
\begin{proof}[Proof of Lemma~\ref{lemma:approx:flow:manifold}]
Write $g_{1}^{*}(t,s,x)\coloneqq P\td{g}^{*}(t,s,P^{\top}x)$ and $g_{2}^{*}(t,s,x)\coloneqq\iota(t,s)(I_{d}-PP^{\top})x$, so that $g^{*}=g_{1}^{*}+g_{2}^{*}$ by~\eqref{eq:manifold-structure-flow}. We complete the proof through the following three steps.
\vspace{2mm}

\noindent\emph{Step 1 (Approximate $g_{1}^{*}$).}
Set $R^{\prime}\coloneqq\sqrt{d}R$ and $\td{\calX}\coloneqq[0,T]\times[0,T]\times[-R^{\prime},R^{\prime}]^{d^{*}}$. Extend $\td{g}^{*}$ to $\td{\calX}$ by setting $\td{g}^{*}(t,s,\td{x})\coloneqq\td{g}^{*}(t,\max\{s,t\},\td{x})$ for $s<t$; the extension is Lipschitz on $\td{\calX}$, and by Lemma~\ref{lemma:g_tilde properties}, its first-order derivatives are bounded by $2\td{B}_{\flow}^{\prime}R^{\prime}$ in $t$ and $s$, and by $\exp(\td{G})$ in each coordinate of $\td{x}$. According to Corollary~\ref{corollary:approx}, for each $1\leq k\leq d^{*}$, there exists a real-valued deep neural network $\td{g}_{k}$ with depth $L_{k}=\lceil\log_{2}(d^{*}+2)\rceil+3$ and number of parameters $S_{k}=(22(d^{*}+2)+6)(N+1)^{d^{*}+2}$, such that
\begin{equation*}
\|\td{g}_{k}-\td{g}_{k}^{*}\|_{L^{\infty}(\td{\calX})}\leq2^{d^{*}+2}\Big(\frac{2T}{N}\cdot2\td{B}_{\flow}^{\prime}R^{\prime}+\sum_{\ell=1}^{d^{*}}\frac{2R^{\prime}}{N}\exp(\td{G})\Big)\leq\frac{C_{1}R}{N},
\end{equation*}
and moreover $\|\td{g}_{k}\|_{L^{\infty}(\td{\calX})}=\|\td{g}_{k}^{*}\|_{L^{\infty}(\td{\calX})}\leq\td{B}_{\flow}R^{\prime}$, while the first-order derivatives of $\td{g}_{k}$ are bounded by three times those of $\td{g}_{k}^{*}$. Set $g_{1}(t,s,x)\coloneqq P\td{g}(t,s,\pi_{R^{\prime}}(P^{\top}x))$ with $\td{g}\coloneqq(\td{g}_{k})_{k=1}^{d^{*}}$, where $\pi_{R^{\prime}}$ is the coordinatewise clipping map at level $R^{\prime}$ in $\bbR^{d^{*}}$, realized exactly by two additional ReLU layers with $\calO(d^{*})$ non-zero weights as in the proof of Lemma~\ref{lemma:approx:velocity}, and the linear maps $x\mapsto P^{\top}x$ and $\td{y}\mapsto P\td{y}$ are realized exactly by affine layers with $\calO(dd^{*})$ non-zero weights. Since $\|P^{\top}x\|_{\infty}\leq\|x\|_{2}\leq R^{\prime}$ for each $x\in\bbB_{R}^{\infty}$, on which $\pi_{R^{\prime}}$ is the identity, and $\|P\td{y}\|_{2}=\|\td{y}\|_{2}$, it follows that for each $0\leq t\leq s\leq T$ and $x\in\bbB_{R}^{\infty}$,
\begin{equation*}
\|g_{1}(t,s,x)-g_{1}^{*}(t,s,x)\|_{2}\leq\Big(\sum_{k=1}^{d^{*}}\|\td{g}_{k}-\td{g}_{k}^{*}\|_{L^{\infty}(\td{\calX})}^{2}\Big)^{1/2}\leq\frac{C_{1}\sqrt{d^{*}}R}{N},
\end{equation*}
while $\|g_{1}\|_{\infty}\leq\sqrt{dd^{*}}\td{B}_{\flow}R$, $\|\nabla g_{1}\|_{\op}\leq3d^{*}\exp(\td{G})$ and $\max\{\|\partial_{t}g_{1}\|_{2},\|\partial_{s}g_{1}\|_{2}\}\leq6\sqrt{dd^{*}}\td{B}_{\flow}^{\prime}R$ hold on $[0,T]\times[0,T]\times\bbR^{d}$.
\vspace{2mm}

\noindent\emph{Step 2 (Approximate $g_{2}^{*}$).}
By Corollary~\ref{corollary:approx}, there exists a two-dimensional neural network $\iota_{\theta}$ with depth $L=4$ and number of parameters $S=50(N+1)^{2}$, such that $\|\iota_{\theta}-\iota\|_{L^{\infty}([0,T]^{2})}\leq C_{2}N^{-1}$, $\|\iota_{\theta}\|_{L^{\infty}([0,T]^{2})}\leq\iota_{\rm max}$ and $$\max\{\|\partial_{t}\iota_{\theta}\|_{L^{\infty}([0,T]^{2})},\|\partial_{s}\iota_{\theta}\|_{L^{\infty}([0,T]^{2})}\}\leq3\iota_{\rm max}.$$ Set $g_{2}(t,s,x)\coloneqq\iota_{\theta}(t,s)\,\pi_{R^{\prime}}^{\perp}\big((I_{d}-PP^{\top})x\big)$, where $\pi_{R^{\prime}}^{\perp}$ is the coordinatewise clipping map at level $R^{\prime}$ in $\bbR^{d}$, realized by two additional ReLU layers with $\calO(d)$ non-zero weights; since $|((I_{d}-PP^{\top})x)_{j}|\leq\|(I_{d}-PP^{\top})x\|_{2}\leq\|x\|_{2}\leq R^{\prime}$ for each $x\in\bbB_{R}^{\infty}$ and $1\leq j\leq d$, the map $\pi_{R^{\prime}}^{\perp}$ is the identity there. The linear map $x\mapsto(I_{d}-PP^{\top})x$ is affine with $\calO(d^{2})$ non-zero weights, and the coordinate-wise products $\iota_{\theta}(t,s)\cdot(\pi_{R^{\prime}}^{\perp}((I_{d}-PP^{\top})x))_{j}$, $j=1,\ldots,d$, are realized via Lemma~\ref{lemma:product:nn}, adding one ReQU hidden layer and $12d$ parameters. Then for each $0\leq t\leq s\leq T$ and $x\in\bbB_{R}^{\infty}$,
\begin{equation*}
\|g_{2}(t,s,x)-g_{2}^{*}(t,s,x)\|_{2}\leq\|\iota_{\theta}-\iota\|_{L^{\infty}([0,T]^{2})}\big\|(I_{d}-PP^{\top})x\big\|_{2}\leq\frac{C_{2}\sqrt{d}R}{N},
\end{equation*}
while $\|g_{2}\|_{\infty}\leq\sqrt{d}\iota_{\rm max}R$, $\|\nabla g_{2}\|_{\op}\leq\iota_{\rm max}$ and $\max\{\|\partial_{t}g_{2}\|_{2},\|\partial_{s}g_{2}\|_{2}\}\leq3d\iota_{\rm max}R$ hold on $[0,T]\times[0,T]\times\bbR^{d}$.
\vspace{2mm}

\noindent\emph{Step 3 (Aggregation).}
Set $g\coloneqq g_{1}+g_{2}$ as the final network, which implements the add operation in the output affine layer with $\calO(d)$ non-zero weights. The total extra overhead is $\calO(dd^{*})+\calO(d^{2})+\calO(N^{2})$, which is absorbed into $L=C_{3}$ and $S=C_{4}N^{d^{*}+2}$. Combining Steps 1 and 2 gives, for each $0\leq t\leq s\leq T$ and $x\in\bbB_{R}^{\infty}$,
\begin{equation*}
\|g(t,s,x)-g^{*}(t,s,x)\|_{2}\leq\|g_{1}(t,s,x)-g_{1}^{*}(t,s,x)\|_{2}+\|g_{2}(t,s,x)-g_{2}^{*}(t,s,x)\|_{2}\leq\frac{C_{5}R}{N},
\end{equation*}
which yields $\calR_{T,R}(g)\leq C_{6}R^{2}N^{-2}$. Furthermore,
\begin{equation*}
\|g\|_{\infty}\leq\sqrt{d}\big(\td{B}_{\flow}\sqrt{d^{*}}+\iota_{\rm max}\big)R, \quad
\max\big\{\|\partial_{t}g\|_{2},\|\partial_{s}g\|_{2}\big\}\leq3d\big(2\td{B}_{\flow}^{\prime}+\iota_{\rm max}\big)R, \quad
\|\nabla g\|_{\op}\leq3d^{*}\exp(\td{G})+\iota_{\rm max},
\end{equation*}
so that $g$ satisfies the constraints of the hypothesis class $\scrG$ with the admissible constants $(B_{\rm man}^{\flow},D_{\rm man}^{\flow},G_{\rm man}^{\flow})$ displayed above. Here, $C_{1}$ to $C_{6}$ are constants which may only depend on $d$, $d^{*}$ and $\sigma$. This completes the proof.
\end{proof}
}

\par With Lemma~\ref{lemma:approx:flow:manifold}, we establish the convergence result for the characteristic generator under the low-dimensional linear-subspace hypothesis. This
theorem is the manifold counterpart of Theorem~\ref{theorem:error:characteristic},
demonstrating that the characteristic generator can achieve a convergence rate depending
on the intrinsic dimension $d^{*}$ rather than the ambient dimension $d$.

\begin{theorem}[Error analysis for characteristic generator under low-dimensional linear-subspace hypothesis]\label{theorem:error:characteristic:manifold}
{
Suppose Assumptions~\ref{assumption:init:dist:gaussian} and~\ref{assumption:target:dist:manifold} hold. Under the same conditions as Theorem~\ref{theorem:velocity:rate:manifold}. Further, set the hypothesis class $\scrG$ as a deep neural network class, which is defined as
\begin{equation*}
\scrG=\left\{g\in N(L,S):
\begin{aligned}
&\|g(t,s,x)\|_{\infty}\leq B_{\rm man}^{\flow}\log^{1/2}m, \\
&\|\partial_{s}g(t,s,x)\|_{2}\leq D_{\rm man}^{\flow}\log^{1/2}m, \\
&\|\partial_{t}g(t,s,x)\|_{2}\leq D_{\rm man}^{\flow}\log^{1/2}m, \\
&\|\nabla g(t,s,x)\|_{\op}\leq G_{\rm man}^{\flow}
\end{aligned}
\right\},
\end{equation*}
where the Lipschitz control parameters $B_{\rm man}^{\flow}$, $D_{\rm man}^{\flow}$, and $G_{\rm man}^{\flow}$ are given in Lemma~\ref{lemma:approx:flow:manifold} and depend only on $d$, $d^{*}$, and $\sigma$. The depth and the number of non-zero weights of the neural network are given, respectively, as $L=C$ and $S=Cm^{\frac{d^{*}+2}{d^{*}+4}}$. Then it follows that
\begin{align*}
\bbE_{\euS}\bbE_{\euZ}\big[\calD(\what{g})\big]\leq C\Phi_{T}\Big\{n^{-\frac{2}{d^{*}+3}}\log^{2}n+\frac{1}{K}\Big\}
+C\Big\{m^{-\frac{2}{d^{*}+4}}\log^{2}m+\frac{\log m}{K}\Big\},
\end{align*}
where $\Phi_{T}$ is the coefficient functional~\eqref{eq:Phi:T}, and the constant $C$ only depends on $d$, $d^{*}$, $\sigma$ and the coefficients $(\alpha_{t},\beta_{t})$. Furthermore, if the number of time steps $K$ for the exponential integrator and the number of samples $m$ for characteristic fitting satisfy
\begin{equation*}
K\geq Cn^{\frac{2}{d^{*}+3}}\log m, \quad
m\geq Cn^{\frac{d^{*}+4}{d^{*}+3}},
\end{equation*}
respectively, then the following inequality holds
\begin{equation*}
\bbE_{\euS}\bbE_{\euZ}\big[\calD(\what{g})\big]\leq C\Phi_{T}n^{-\frac{2}{d^{*}+3}}\log^{2}n.
\end{equation*}}
\end{theorem}

\begin{proof}[Proof of Theorem~\ref{theorem:error:characteristic:manifold}]
{Under the low-dimensional linear-subspace hypothesis, Lemma~\ref{lemma:approx:flow:manifold} establishes that there exists a neural network class $\scrG$ with $S=CN^{d^{*}+2}$ nonzero weights such that $\inf_{g\in\scrG}\mathcal{R}_{T,R}(g)\leq CR^{2}N^{-2}$. By Lemma~\ref{lemma:vcdim}, the VC-dimension satisfies $\vcdim(\Pi_{k}\scrG)\leq CN^{d^{*}+2}\log N$. Moreover, every denoiser in the hypothesis class~\eqref{eq:scrD:manifold} of Theorem~\ref{theorem:velocity:rate:manifold} is of the preconditioned form with $\|\nabla f(t,x)\|_{\op}\leq3d^{*}$, so the structured Lipschitz condition~\eqref{eq:discretization:lip} holds with $G=3d^{*}$, and inequality~\eqref{eq:proof:theorem:error:euler:1} together with Theorem~\ref{theorem:velocity:rate:manifold} gives $\bbE_{\euS}[\bar{e}_{K}^{2}]\leq C\Phi_{T}\big(n^{-\frac{2}{d^{*}+3}}\log^{2}n+1/K\big)$. Following the same proof strategy as Theorem~\ref{theorem:error:characteristic} (Appendix \ref{section:proof:analysis:characteristic}) and setting $N=Cm^{\frac{1}{d^{*}+4}}$, $R^{2}=\theta^{-1}\log m$ yields the desired result.}
\end{proof}

\par Theorems~\ref{theorem:velocity:rate:manifold} and~\ref{theorem:error:characteristic:manifold} demonstrate that when the target distribution
has a low-dimensional manifold structure, both the denoiser matching and the
characteristic generator can achieve convergence rates that depend on the intrinsic
dimension $d^{*}$ instead of the ambient dimension $d$, thereby effectively mitigating
the curse of dimensionality. Applying these results to the analysis of linear interpolant and F\"{o}llmer flow yields Corollaries \ref{corollary:rate:linear:manifold} and \ref{corollary:rate:follmer:manifold} in the main text.

{
\begin{proof}[Proof of Corollaries~\ref{corollary:rate:linear:manifold} and~\ref{corollary:rate:follmer:manifold}]
The proofs of Corollaries~\ref{corollary:rate:linear} and~\ref{corollary:rate:follmer} in Appendix~\ref{section:proof:analysis:applications} carry over verbatim, with Theorems~\ref{theorem:velocity:rate} and~\ref{theorem:error:characteristic} replaced by their manifold counterparts, Theorems~\ref{theorem:velocity:rate:manifold} and~\ref{theorem:error:characteristic:manifold}, and with the ambient dimension $d$ in all exponents replaced by the intrinsic dimension $d^{*}$. In particular, the closed forms~\eqref{eq:Phi:linear} and~\eqref{eq:Phi:follmer} of $\Phi_{T}$ and the moment bound~\eqref{eq:W2:moment} are unaffected by the manifold structure, and the exponential integrator bound of Corollary~\ref{corollary:error:euler} holds with the rate $n^{-\frac{2}{d^{*}+3}}\log^{2}n$, since the argument in the proof of Theorem~\ref{theorem:error:characteristic:manifold} verifies the structured Lipschitz condition~\eqref{eq:discretization:lip} with $G=3d^{*}$ and Theorem~\ref{theorem:velocity:rate:manifold} supplies the denoiser matching rate.
\end{proof}
}

{
\subsection{Relation between Assumption~\ref{assumption:target:dist:manifold} and PCA}\label{appendix:pca}

\subsubsection{PCA serves as a linear auto-encoder}
\par Without loss of generality, we assume $\mathbb{E}_{Z\sim\tilde{\nu}}[Z]=0$ and $\mathrm{Cov}_{Z\sim\tilde{\nu}}(Z)\succ 0$. From Assumption~\ref{assumption:target:dist:manifold}, $X_{1}\stackrel{\mathrm{d}}{=}PZ+\sigma W$ and thus $\mathbb{E}_{X_{1}\sim\mu_{1}}[X_{1}]=0$.

\par Let $X_{1}\in\mathbb{R}^{d}$ be the data variable, and let $F\in\mathbb{R}^{d^{*}\times d}$ be a linear encoder mapping from the data space to the latent space. Let $G\in\mathbb{R}^{d\times d^{*}}$ be a coloum-orthogonal matrix, i.e., $G^{\top}G=I_{d^{*}}$, serving as a decoder. The low-dimensional representation of $X_{1}$ is $F(X_{1})$, and the reconstruction of $X_{1}$ from $F(X_{1})$ is defined by $G\circ F(X_{1})$. The optimal linear encoder-decoder pair is determized by minimizing the expected $\ell^{2}$-reconstruction error
\begin{equation}\label{eq:linear:encoder}
\min_{F,G}\mathbb{E}_{X_{1}\sim\mu_{1}}\bigl[\|G\circ F(X_{1})-X_{1}\|_{2}^{2}\bigr], \quad \text{s.t.} ~ G^{\top}G=I_{d^{*}}.
\end{equation}

\par We first minimize the objective for a fixed decoder $G$ and a fixed data sample $X_{1}=x_{1}$, i.e.,
\begin{equation*}
\min_{F}\|G\circ F(X_{1})-x_{1}\|_{2}^{2}.
\end{equation*}
The normal equation of this least-squares problem reads $G^{\top}G\circ F^{*}(x_{1})=G^{\top}x_{1}$, which implies from $G^{\top}G=I_{d^{*}}$ that $F^{*}(x_{1})=G^{\top}x_{1}$. This means the optimal encoder $F^{*}$ given a column-orthogonal decoder $G$ satisfies $F^{*}=G^{\top}$. Consequently, the original optimization~\eqref{eq:linear:encoder} problem reduces to
\begin{equation}\label{eq:linear:encoder:1}
\min_{G}\mathbb{E}_{X_{1}\sim\mu_{1}}\bigl[\|G\circ G^{\top}(X_{1})-X_{1}\|_{2}^{2}\bigr], \quad \text{s.t.} ~ G^{\top}G=I_{d^{*}}.
\end{equation}
Using the fact that $G^{\top}G=I_{d^{*}}$, we have
\begin{align*}
\|G\circ G^{\top}(X_{1})-X_{1}\|_{2}^{2} = (G\circ G^{\top}(X_{1})-X_{1})^{\top}(G\circ G^{\top}(X_{1})-X_{1})=X_{1}^{\top}X_{1}-X_{1}^{\top}GG^{\top}X_{1}.
\end{align*}
Taking expectation with respect to $X_{1}\sim\mu_{1}$ yields
\begin{align*}
\mathbb{E}_{X_{1}\sim\mu_{1}}\bigl[\|G\circ G^{\top}(X_{1})-X_{1}\|_{2}^{2}\bigr]
&=\mathbb{E}_{X_{1}\sim\mu_{1}}\bigl[X_{1}^{\top}X_{1}-X_{1}^{\top}GG^{\top}X_{1}\bigr] \\
&=-\mathbb{E}_{X_{1}\sim\mu_{1}}\bigl[X_{1}^{\top}GG^{\top}X_{1}\bigr]+\mathbb{E}_{X_{1}\sim\mu_{1}}\bigl[X_{1}^{\top}X_{1}\bigr] \\
&=-\mathbb{E}_{X_{1}\sim\mu_{1}}\bigl[\mathrm{tr}(X_{1}^{\top}GG^{\top}X_{1})\bigr]+\mathbb{E}_{X_{1}\sim\mu_{1}}\bigl[X_{1}^{\top}X_{1}\bigr] \\
&=-\mathbb{E}_{X_{1}\sim\mu_{1}}\bigl[\mathrm{tr}(G^{\top}X_{1}X_{1}^{\top}G)\bigr]+\mathbb{E}_{X_{1}\sim\mu_{1}}\bigl[X_{1}^{\top}X_{1}\bigr] \\
&= -\mathrm{tr}(G^{\top}\Sigma G)+\mathbb{E}_{X_{1}\sim\mu_{1}}\bigl[X_{1}^{\top}X_{1}\bigr],
\end{align*}
where $\Sigma\in\mathbb{R}^{d^{*}\times d^{*}}$ is the covariance matrix of $\mu_{1}$. Since $\mathbb{E}_{X_{1}\sim\mu_{1}}[X_{1}^{\top}X_{1}]$ is independent of $G$,~\eqref{eq:linear:encoder:1} is equivalent to
\begin{equation}\label{eq:linear:encoder:2}
\max_{G}\mathrm{tr}(G^{\top}\Sigma G), \quad \text{s.t.} ~ G^{\top}G=I_{d^{*}}.
\end{equation}

\par Since $X_{1}\stackrel{\mathrm{d}}{=}PZ+\sigma W$, we have
\begin{equation*}
\Sigma = P\mathrm{Cov}_{Z\sim\tilde{\nu}}(Z)P^{\top}+\sigma^{2}I_{d}.
\end{equation*}
Since $\mathrm{Cov}_{Z\sim\tilde{\nu}}(Z)\succ 0$, denote by $\lambda_{1}\geq\cdots\geq\lambda_{d^{*}}>0$ the eigenvalues of $\mathrm{Cov}_{Z\sim\tilde{\nu}}(Z)\in\mathbb{R}^{d^{*}\times d^{*}}$. Let $\Sigma=V\Lambda V^{\top}$ be the singular value decomposition (SVD) of $\Sigma$, where
\begin{equation*}
\Lambda\coloneqq\mathrm{diag}(\lambda_{1}+\sigma^{2},\ldots,\lambda_{d^{*}}+\sigma^{2},\underbrace{\sigma^{2},\ldots,\sigma^{2}}_{d-d^{*}}), \quad V\coloneqq [v_{1} \mid v_{2} \mid \ldots \mid v_{d}]\in\mathbb{R}^{d\times d}.
\end{equation*}
Define $H\coloneqq V^{\top}G\in\mathbb{R}^{d\times d^{*}}$, and denote by $H_{ik}$ the $(i,k)$-entry of $H$. It is straightforward that
\begin{equation*}
H^{\top}H=G^{\top}VV^{\top}G=G^{\top}G=I_{d^{*}},
\end{equation*}
and, as a consequence, for all $k\in\{1,\ldots,d^{*}\}$,
\begin{equation*}
\sum_{i=1}^{d}H_{ik}^{2} = (H^{\top}H)_{kk} = 1.
\end{equation*}
Since $H^{\top}H=I_{d^{*}}$, both $HH^{\top}$ and $I_{d}-HH^{\top}$ are projection operators, and then $HH^{\top}\succ 0$ and $I_{d}-HH^{\top}\succ 0$. Let $e_{i}\in\mathbb{R}^{d}$ be the standard basis. Then
\begin{equation*}
w_{i}\coloneqq\sum_{k=1}^{d^{*}}H_{ik}^{2} = (HH^{\top})_{ii} = e_{i}^{\top}HH^{\top}e_{i} \geq 0, \quad\text{and}\quad e_{i}^{\top}(I_{d}-HH^{\top})e_{i}=1-w_{i}\geq 0.
\end{equation*}
As a result, $w_{i}\in[0,1]$. Then the objective in~\eqref{eq:linear:encoder:2} reads
\begin{equation}\label{eq:pca:1}
\mathrm{tr}(G^{\top}\Sigma G) = \mathrm{tr}(H^{\top}\Lambda H) = \sum_{i=1}^{d^{*}}(\lambda_{i}+\sigma^{2})\sum_{k=1}^{d^{*}}H_{ik}^{2}+\sum_{i=d^{*}+1}^{d}\sigma^{2}\sum_{k=1}^{d^{*}}H_{ik}^{2} = \sum_{i=1}^{d^{*}}(\lambda_{i}+\sigma^{2})w_{i}+\sum_{i=d^{*}+1}^{d}\sigma^{2}w_{i}.
\end{equation}
On the other hand, we have
\begin{equation}\label{eq:pca:2}
\sum_{i=1}^{d}w_{i}=\sum_{i=1}^{d}\sum_{k=1}^{d^{*}}H_{ik}^{2}=\sum_{k=1}^{d^{*}}\sum_{i=1}^{d}H_{ik}^{2} = \sum_{k=1}^{d^{*}}1=d^{*}.
\end{equation}
Therefore, maximizing~\eqref{eq:pca:1} under the constraints~\eqref{eq:pca:2} and $w_{i}\in[0,1]$ yields the maximizer $w_{i}^{*}=1$ for $i=\{1,\ldots,d^{*}\}$ and $w_{i}^{*}=0$ otherwise. Then
\begin{equation}\label{eq:pca:3}
H^{*} =
\begin{pmatrix}
I_{d^{*}} \\ 0
\end{pmatrix},
\quad\text{and}\quad G^{*}=VH^{*}=[v_{1}\mid \cdots\mid v_{d^{*}}]\in\mathbb{R}^{d\times d^{*}}.
\end{equation}

\begin{remark}
By replacing the covariance matrix by its empirical approximation in~\eqref{eq:linear:encoder:2} yields with PCA.
\end{remark}

\subsubsection{Proof of Proposition~\ref{proposition:pca}}

\begin{proof}[Proof of Proposition~\ref{proposition:pca}]
For any $u\in\mathrm{span}(P)$, there exists $\zeta\in\mathbb{R}^{d^{*}}$ such that $u=P\zeta$. Then
\begin{equation}\label{eq:pca:4}
\Sigma u = \Sigma(P\zeta) = \underbrace{(P\mathrm{Cov}_{Z\sim\tilde{\nu}}(Z)P^{\top}+\sigma^{2}I_{d})}_{\Sigma}P\zeta = P(\mathrm{Cov}_{Z\sim\tilde{\nu}}(Z)+\sigma^{2}I_{d})\zeta \in \mathrm{span}(P),
\end{equation}
which means $\Sigma$ map $\mathrm{span}(P)$ back into $\mathrm{span}(P)$. For any $w\in\mathrm{span}(P)^{\perp}$, we have $P^{\top}w=0$. Then
\begin{equation}\label{eq:pca:5}
\Sigma w = \underbrace{(P\mathrm{Cov}_{Z\sim\tilde{\nu}}(Z)P^{\top}+\sigma^{2}I_{d})}_{\Sigma}w = \sigma^{2}w \in \mathrm{span}(P)^{\perp}.
\end{equation}
For every $j\in\{1,\ldots,d^{*}\}$, suppose the $j$-th eigenvector $v_{j}$ in~\eqref{eq:pca:3} admits a decomposition $v_{j}=u+w$ with $u\in\mathrm{span}(P)$ and $w\in\mathrm{span}(P)^{\perp}$. Then it follows from~\eqref{eq:pca:4} and~\eqref{eq:pca:5} that
\begin{equation*}
\Sigma v_{j}=\Sigma(u+w) = \Sigma u+\Sigma w=\Sigma u+\sigma^{2}w, \quad \text{where} ~ \Sigma u\in\mathrm{span}(P), ~ \sigma^{2}w\in\mathrm{span}(P)^{\perp}.
\end{equation*}
On the other hand,
\begin{equation*}
\Sigma v_{j}=(\lambda_{j}+\sigma^{2})v_{j} = (\lambda_{j}+\sigma^{2})u+(\lambda_{j}+\sigma^{2})w, \quad \text{where} ~ (\lambda_{j}+\sigma^{2})u\in\mathrm{span}(P), ~ (\lambda_{j}+\sigma^{2})w\in\mathrm{span}(P)^{\perp}.
\end{equation*}
Combining these two sides yields $\sigma^{2}w=(\lambda_{j}+\sigma^{2})w$. Since $\mathrm{Cov}_{Z\sim\tilde{\nu}}(Z)\succ 0$, we have $\lambda_{j}>0$, and then $w=0$, i.e., $v_{j}\in\mathrm{span}(P)$. As a consequence, $\mathrm{span}(G^{*})\subset\mathrm{span}(P)$. Since both subspaces have the same dimension $d^{*}$, we have $\mathrm{span}(G^{*})=\mathrm{span}(P)$. This completes the proof.
\end{proof}
}

\section{Generalization Error Analysis for Least-Squares Regression}
\label{section:oracle}
\par In this section, we provide a generalization error analysis for nonparametric regression, which is used in establishing oracle inequalities for velocity matching (Lemma~\ref{lemma:oracle:velocity}) and characteristic fitting (Lemma~\ref{lemma:oracle:solution}).

\par Let $\calX\subset\bbR^{d}$ be a bounded domain, and let $X\in\calX$ be a random variable obeying the probability distribution $\mu$. Let $f^{*}:\calX\rightarrow\bbR$ be a measurable function. Define the population $L^{2}(\mu)$-risk for a function $f:\calX\rightarrow\bbR$ as
\begin{equation*}
R(f)=\|f-f^{*}\|_{L^{2}(\mu)}^{2}=\bbE_{X\sim\mu}\big[(f(X)-f^{*}(X))^{2}\big].
\end{equation*}
Let $\euD=\{X^{(i)}\}_{i=1}^{n}$ be a set of i.i.d. copies of $X\sim\mu$. Then define the empirical risk of $f$ by
\begin{equation*}
\what{R}_{n}(f)=\frac{1}{n}\sum_{i=1}^{n}(f(X^{(i)})-f^{*}(X^{(i)}))^{2}.
\end{equation*}

\par The following lemma relates the population risk to its empirical counterpart.

\begin{lemma}\label{lemma:generalization:1}
Suppose that $|f^{*}(x)|\leq B$ for each $x\in\calX$. Let $\scrF$ be a set of functions mapping from $\calX$ to $[-B,B]$. Then it follows that for each $n\geq\vcdim(\scrF)$,
\begin{equation*}
\bbE_{\euD}\Big[\sup_{f\in\scrF}R(f)-2\what{R}_{n}(f)\Big]\leq CB^{2}\frac{\vcdim(\scrF)}{n\log^{-1}n},
\end{equation*}
where $C$ is an absolute constant.
\end{lemma}

\par This lemma provides a generalization error bound with fast rate via the technique of the offset Rademacher complexity, which was first proposed by~\cite{Liang2015Learning}. In recent years, this technique has been applied to the convergence rate analysis for deep nonparametric regression, such as~\cite[Lemma 14]{Duan2023Convergence} and~\cite[Lemma 4.1]{ding2024semisupervised}.

\begin{proof}[Proof of Lemma~\ref{lemma:generalization:1}]
\par We define an auxiliary function class
\begin{equation*}
\scrH=\Big\{x\mapsto h(x)=(f(x)-f^{*}(x))^{2}:f\in\scrF\Big\}.
\end{equation*}
It is apparent that $0\leq h(x)\leq4B^{2}$ for each $x\in\calX$ and $h\in\scrH$. Then it is easy to show that
\begin{align*}
&\bbE_{\euD}\Big[\sup_{f\in\scrF}R(\what{f})-2\what{R}_{n}(\what{f})\Big]\leq\bbE_{\euD}\Big[\sup_{h\in\scrH}\bbE[h(X)]-\frac{2}{n}\sum_{i=1}^{n}h(X^{(i)})\Big] \\
&\leq\bbE_{\euD}\Big[\sup_{h\in\scrH}\bbE\Big[\frac{3}{2}h(X)-\frac{1}{8B^{2}}h^{2}(X)\Big]-\frac{1}{n}\sum_{i=1}^{n}\Big(\frac{3}{2}h(X^{(i)})+\frac{1}{8B^{2}}h^{2}(X^{(i)})\Big)\Big],
\end{align*}
where we used the fact that $h^{2}(x)\leq4B^{2}h(x)$ for each $x\in\calX$ and $h\in\scrH$.

\par Let us introduce a ghost sample $\euD^{\prime}=\{X^{(i),\prime}\}_{i=1}^{n}$, which is a set of $n$ i.i.d. random copies of $X\sim\mu$. Here the ghost sample $\euD^{\prime}$ is independent of $\euD=\{X^{(i)}\}_{i=1}^{n}$. Let $\xi=\{\xi^{(i)}\}_{i=1}^{n}$ be a set of i.i.d. Rademacher variables. Then replacing the expectation by the empirical mean based on the ghost sample $\euD^{\prime}$ yields
\begin{align*}
&\bbE_{\euD}\Big[\sup_{h\in\scrH}\bbE_{X}\Big[\frac{3}{2}h(X)-\frac{1}{8B^{2}}h^{2}(X)\Big]-\frac{1}{n}\sum_{i=1}^{n}\Big(\frac{3}{2}h(X^{(i)})+\frac{1}{8B^{2}}h^{2}(X^{(i)})\Big)\Big] \\
&=\bbE_{\euD}\Big[\sup_{h\in\scrH}\bbE_{\euD^{\prime}}\Big[\frac{1}{n}\sum_{i=1}^{n}\frac{3}{2}h(X^{(i),\prime})-\frac{1}{8B^{2}}h^{2}(X^{(i),\prime})\Big]-\frac{1}{n}\sum_{i=1}^{n}\Big(\frac{3}{2}h(X^{(i)})+\frac{1}{8B^{2}}h^{2}(X^{(i)})\Big)\Big] \\
&\leq\bbE_{\euD}\bbE_{\euD^{\prime}}\Big[\sup_{h\in\scrH}\frac{3}{2n}\sum_{i=1}^{n}(h(X^{(i),\prime})-h(X^{(i)}))-\frac{1}{8B^{2}n}\sum_{i=1}^{n}(h^{2}(X^{(i),\prime})+h^{2}(X^{(i)}))\Big] \\
&=\bbE_{\euD}\bbE_{\euD^{\prime}}\bbE_{\xi}\Big[\sup_{h\in\scrH}\frac{3}{2n}\sum_{i=1}^{n}\xi^{(i)}(h(X^{(i),\prime})-h(X^{(i)}))-\frac{1}{8B^{2}n}\sum_{i=1}^{n}(h^{2}(X^{(i),\prime})+h^{2}(X^{(i)}))\Big] \\
&=\bbE_{\euD}\bbE_{\xi}\Big[\sup_{h\in\scrH}\frac{3}{n}\sum_{i=1}^{n}\xi^{(i)}h(X^{(i)})-\frac{1}{4B^{2}n}\sum_{i=1}^{n}h^{2}(X^{(i)})\Big],
\end{align*}
where the inequality holds from Jensen's inequality. Combining the above results, we have
\begin{equation}\label{eq:lemma:generalization:1:1}
\bbE_{\euD}\Big[\sup_{f\in\scrF}R(\what{f})-2\what{R}_{n}(\what{f})\Big]\leq\bbE_{\euD}\bbE_{\xi}\Big[\sup_{h\in\scrH}\frac{3}{n}\sum_{i=1}^{n}\xi^{(i)}h(X^{(i)})-\frac{1}{4B^{2}n}\sum_{i=1}^{n}h^{2}(X^{(i)})\Big].
\end{equation}

\par We next estimate the expectation in the right-hand side of~\eqref{eq:lemma:generalization:1:1}. Let $\delta\in(0,4B^{2})$ and $\scrH_{\delta}$ be a $L^{\infty}(\euD)$ $\delta$-cover of $\scrH$ satisfying $|\scrH_{\delta}|=N(\delta,\scrH,L^{\infty}(\euD))$. Then for each $h\in\scrH$, there exists $h_{\delta}\in\scrH_{\delta}$ such that
\begin{equation*}
\max_{1\leq i\leq n}|h(X^{(i)})-h_{\delta}(X^{(i)})|\leq\delta.
\end{equation*}
Without loss of generality, we assume $|h_{\delta}(x)|\leq 4B^{2}$ for each $h_{\delta}\in\scrH_{\delta}$. Consequently, it follows from H{\"o}lder's inequality that
\begin{equation*}
\frac{1}{n}\sum_{i=1}^{n}\xi^{(i)}h(X^{(i)})-\frac{1}{n}\sum_{i=1}^{n}\xi^{(i)}h_{\delta}(X^{(i)})\leq\frac{1}{n}\sum_{i=1}^{n}|\xi^{(i)}||h(X^{(i)})-h_{\delta}(X^{(i)})|\leq\delta.
\end{equation*}
By the same argument, it holds that
\begin{equation*}
-\frac{1}{n}\sum_{i=1}^{n}h^{2}(X^{(i)})+\frac{1}{n}\sum_{i=1}^{n}h_{\delta}^{2}(X^{(i)})\leq8B^{2}\delta.
\end{equation*}
With the help of the above two inequalities, we have
\begin{align}
&\bbE_{\xi}\Big[\sup_{h\in\scrH}\frac{3}{n}\sum_{i=1}^{n}\xi^{(i)}h(X^{(i)})-\frac{1}{4B^{2}n}\sum_{i=1}^{n}h^{2}(X^{(i)})\Big] \nonumber \\
&\leq\bbE_{\xi}\Big[\max_{h_{\delta}\in\scrH_{\delta}}\frac{3}{n}\sum_{i=1}^{n}\xi^{(i)}h_{\delta}(X^{(i)})-\frac{1}{4B^{2}n}\sum_{i=1}^{n}h_{\delta}^{2}(X^{(i)})\Big]+5\delta. \label{eq:lemma:generalization:1:2}
\end{align}

\par Observe that $\{\xi^{(i)}h_{\delta}(X^{(i)})\}_{i=1}^{n}$ is a set of $n$ i.i.d. random variables with
\begin{equation*}
-h_{\delta}(X^{(i)})\leq\xi^{(i)}h_{\delta}(X^{(i)})\leq h_{\delta}(X^{(i)}), \quad 1\leq i\leq n.
\end{equation*}
Then it follows Hoeffding's inequality~\cite[Theorem D.2]{mohri2018foundations} that
\begin{align}
&\pr_{\xi}\Big\{\frac{3}{n}\sum_{i=1}^{n}\xi^{(i)}h_{\delta}(X^{(i)})>t+\frac{1}{4B^{2}n}\sum_{i=1}^{n}h_{\delta}^{2}(X^{(i)})\Big\} \nonumber \\
&=\pr_{\xi}\Big\{\sum_{i=1}^{n}\xi^{(i)}h_{\delta}(X^{(i)})>\frac{nt}{3}+\frac{1}{12B^{2}}\sum_{i=1}^{n}h_{\delta}^{2}(t^{(i)},X_{t}^{(i)})\Big\} \nonumber \\
&\leq\exp\Big(-\frac{(\frac{nt}{3}+\frac{1}{12B^{2}}\sum_{i=1}^{n}h_{\delta}^{2}(t^{(i)},X_{t}^{(i)}))^{2}}{2\sum_{i=1}^{n}h_{\delta}^{2}(t^{(i)},X_{t}^{(i)})}\Big)\leq\exp\Big(-\frac{nt}{18B^{2}}\Big), \label{eq:lemma:generalization:1:3}
\end{align}
where the first inequality follows from Hoeffding's inequality~\cite[Theorem D.2]{mohri2018foundations}, and the second inequality is due to $(a+b)^{2}/b\leq4a$ for each $a>0$ and $b\in\bbR$. As a consequence, for each $A>0$,
\begin{align*}
&\bbE_{\xi}\Big[\max_{h_{\delta}\in\scrH_{\delta}}\frac{3}{n}\sum_{i=1}^{n}\xi^{(i)}h_{\delta}(X^{(i)})-\frac{1}{4B^{2}n}\sum_{i=1}^{n}h_{\delta}^{2}(X^{(i)})\Big] \\
&=\int_{0}^{\infty}\pr_{\xi}\Big\{\max_{h_{\delta}\in\scrH_{\delta}}\frac{3}{n}\sum_{i=1}^{n}\xi^{(i)}h_{\delta}(X^{(i)})-\frac{1}{4B^{2}n}\sum_{i=1}^{n}h_{\delta}^{2}(X^{(i)})>t\Big\}\dt \\
&\leq A+|\scrH_{\delta}|\int_{T}^{\infty}\exp\Big(-\frac{nt}{18B^{2}}\Big)\dt=A+\frac{18B^{2}}{n}|\scrH_{\delta}|\exp\Big(-\frac{nA}{18B^{2}}\Big),
\end{align*}
where the inequality is owing to~\eqref{eq:lemma:generalization:1:3}. Letting $A=18B^{2}\log|\scrH_{\delta}|n^{-1}$ gives that
\begin{equation}\label{eq:lemma:generalization:1:4}
\bbE_{\xi}\Big[\max_{h_{\delta}\in\scrH_{\delta}}\frac{3}{n}\sum_{i=1}^{n}\xi^{(i)}h_{\delta}(X^{(i)})-\frac{1}{4B^{2}n}\sum_{i=1}^{n}h_{\delta}^{2}(X^{(i)})\Big]\leq18B^{2}\frac{\log|\scrH_{\delta}|+1}{n}.
\end{equation}
It remains to estimate the covering number $|\scrH_{\delta}|=N(\delta,\scrH,L^{\infty}(\euD))$. Noticing that
\begin{equation*}
|h(x)-h^{\prime}(x)|=|(f(x)-f^{*}(x))^{2}-(f^{\prime}(x)-f^{*}(x))^{2}|\leq 4B|f(x)-f^{\prime}(x)|,
\end{equation*}
we obtain that for $n\geq\vcdim(\scrF)$,
\begin{equation}\label{eq:lemma:generalization:1:5}
\log N(\delta,\scrH,L^{\infty}(\euD))\leq\log N\Big(\frac{\delta}{4B},\scrF,L^{\infty}(\euD)\Big)\leq \vcdim(\scrF)\log\Big(\frac{4eB^{2}n}{\delta}\Big),
\end{equation}
where the first and last inequalities follows from Lemmas~\ref{lemma:covering:lip} and~\ref{lemma:covering:vc}, respectively. Combining~\eqref{eq:lemma:generalization:1:1},~\eqref{eq:lemma:generalization:1:2},~\eqref{eq:lemma:generalization:1:4} and~\eqref{eq:lemma:generalization:1:5}, we have
\begin{align*}
\bbE_{\euD}\Big[\sup_{f\in\scrF}R(\what{f})-2\what{R}_{n}(\what{f})\Big]
&\leq\inf_{\delta>0}\Big\{36B^{2}\frac{\vcdim(\scrF)}{n}\log\Big(\frac{4eB^{2}n}{\delta}\Big)+5\delta\Big\}.
\end{align*}
Substituting $\delta=4B^{2}/n$ into the above inequality completes the proof.
\end{proof}

\begin{lemma}\label{lemma:generalization:2}
Suppose that $|f^{*}(x)|\leq B$ for each $x\in\calX$. Let $\scrF$ be a set of functions mapping from $\calX$ to $[-B,B]$. Let $\{\varepsilon^{(i)}\}_{i=1}^{n}$ be a set of independent $\sigma^{2}$-sub-Gaussian random variables. Then it follows that for each $n\geq\vcdim(\scrF)$,
\begin{equation*}
\bbE_{(\euD,\varepsilon)}\Big[\frac{1}{n}\sum_{i=1}^{n}\varepsilon^{(i)}\what{f}(X^{(i)})\Big]\leq\frac{1}{4}\bbE_{(\euD,\varepsilon)}\big[\what{R}(\what{f})\big]+C(B^{2}+\sigma^{2})\frac{\vcdim(\scrF)}{n\log^{-1}n},
\end{equation*}
where $C$ is an absolute constant.
\end{lemma}

\par This proof uses a technique similar to the proof of~\cite[Lemma 4]{schmidt2020nonparametric}.

\begin{proof}[Proof of Lemma~\ref{lemma:generalization:2}]
Let $\delta\in(0,B)$ and let $\scrF_{\delta}$ be a $L^{\infty}(\euD)$ $\delta$-cover of $\scrF$ with $|\scrF_{\delta}|=N(\delta,\scrF,L^{\infty}(\euD))$. Then there exists $\what{f}_{\delta}\in\scrF_{\delta}$, such that
\begin{equation*}
\max_{1\leq i\leq n}|\what{f}(X^{(i)})-\what{f}_{\delta}(X^{(i)})|\leq\delta.
\end{equation*}
Then it follows from H{\"o}lder's inequality that
\begin{equation}\label{eq:lemma:generalization:2:1}
\bbE_{(\euD,\varepsilon)}\Big[\frac{1}{n}\sum_{i=1}^{n}\varepsilon^{(i)}(\what{f}(X^{(i)})-\what{f}_{\delta}(X^{(i)}))\Big]\leq\delta\bbE_{(\euD,\varepsilon)}\Big[\frac{1}{n}\sum_{i=1}^{n}|\varepsilon^{(i)}|\Big]\leq\delta\sigma,
\end{equation}
where we used the fact that $\{\varepsilon^{(i)}\}_{i=1}^{n}$ are a set of $\sigma^{2}$-sub-Gaussian random variables. Additionally, according to the triangular inequality, we have
\begin{equation}\label{eq:lemma:generalization:2:2}
\what{R}_{n}^{1/2}(\what{f}_{\delta})-\what{R}_{n}^{1/2}(\what{f})\leq\Big(\frac{1}{n}\sum_{i=1}^{n}(\what{f}_{\delta}(X^{(i)})-\what{f}(X^{(i)}))^{2}\Big)^{1/2}\leq\delta.
\end{equation}
Consequently, we have
\begin{align}
\bbE_{(\euD,\varepsilon)}\Big[\frac{1}{n}\sum_{i=1}^{n}\varepsilon^{(i)}\what{f}(X^{(i)})\Big]
&=\bbE_{(\euD,\varepsilon)}\Big[\frac{1}{n}\sum_{i=1}^{n}\varepsilon^{(i)}(\what{f}(X^{(i)})-\what{f}_{\delta}(X^{(i)})+\what{f}_{\delta}(X^{(i)})-f^{*}(X^{(i)}))\Big] \nonumber \\
&\leq\bbE_{(\euD,\varepsilon)}\Big[\frac{1}{n}\sum_{i=1}^{n}\varepsilon^{(i)}(\what{f}_{\delta}(X^{(i)})-f^{*}(X^{(i)}))\Big]+\delta\sigma \nonumber \\
&\leq\frac{1}{\sqrt{n}}\bbE_{(\euD,\varepsilon)}\Big[\Big(\what{R}_{n}^{1/2}(\what{f})+\delta\Big)\sum_{i=1}^{n}\frac{\varepsilon^{(i)}(\what{f}_{\delta}(X^{(i)})-f^{*}(X^{(i)}))}{\sqrt{n}\what{R}_{n}^{1/2}(\what{f}_{\delta})}\Big]+\delta\sigma \nonumber \\
&\leq\frac{1}{\sqrt{n}}\Big(\bbE_{(\euD,\varepsilon)}^{1/2}\Big[\what{R}_{n}(\what{f})\Big]+\delta\Big)\bbE_{(\euD,\varepsilon)}^{1/2}\Big[\psi^{2}(\what{f}_{\delta})\Big]+\delta\sigma \nonumber \\
&\leq\frac{1}{4}\bbE_{(\euD,\varepsilon)}\Big[\what{R}_{n}(\what{f})\Big]+\frac{2}{n}\bbE_{(\euD,\varepsilon)}\big[\psi^{2}(\what{f}_{\delta})\big]+\frac{1}{4}\delta^{2}+\delta\sigma, \label{eq:lemma:generalization:2:3}
\end{align}
where $\psi(\what{f}_{\delta})$ is defined as
\begin{equation*}
\psi(\what{f}_{\delta})=\sum_{i=1}^{n}\frac{\what{f}_{\delta}(X^{(i)})-f^{*}(X^{(i)})}{\sqrt{n}\what{R}_{n}^{1/2}(\what{f}_{\delta})}\varepsilon^{(i)}.
\end{equation*}
Here the first inequality follows from~\eqref{eq:lemma:generalization:2:1}, the second inequality holds from~\eqref{eq:lemma:generalization:2:2}, the third inequality is due to Cauchy-Schwarz inequality, and the last inequality is owing to the weighted AM-GM inequality $ab\leq a/4+b$ for each $a,b\in\bbR$.

\par Observe that for each fixed function $f:\bbR^{d}\rightarrow\bbR$, the random variable $\psi(f)$ is sub-Gaussian with variance proxy $\sigma^{2}$ conditioning on $\euD=\{X^{(i)}\}_{i=1}^{n}$. Then it follows that
\begin{equation}\label{eq:lemma:generalization:2:4}
\bbE_{\varepsilon}\big[\psi^{2}(\what{f}_{\delta})\big]\leq\bbE_{\varepsilon}\Big[\max_{f_{\delta}\in\scrF_{\delta}}\psi^{2}(f_{\delta})\Big]\leq4\sigma^{2}(\log|\scrF_{\delta}|+1).
\end{equation}
We now estimate the covering number $|\scrF_{\delta}|=N(\delta,\scrF,L^{\infty}(\euD))$. It follows from Lemma~\ref{lemma:covering:vc} that for $n\geq\vcdim(\scrF)$,
\begin{equation}\label{eq:lemma:generalization:2:5}
\log N(\delta,\scrF,L^{\infty}(\euD))\leq \vcdim(\scrF)\log\Big(\frac{eBn}{\delta}\Big).
\end{equation}
Combining~\eqref{eq:lemma:generalization:2:3},~\eqref{eq:lemma:generalization:2:4} and~\eqref{eq:lemma:generalization:2:5}, and setting $\delta=B/n$ complete the proof.
\end{proof}

\section{Approximation by Deep Neural Networks with Lipschitz Constraint}\label{section:appendix:app:lip}

\par The approximation error analysis for deep neural networks has been investigated by~\cite{Yarotsky2017Error,yarotsky2018optimal,yarotsky2020phase,shen2019nonlinear,shen2020deep,lu2021deep,petersen2018optimal,Jiao2023deep,Duan2022Convergence}. However, limited work has been done for deep neural networks with Lipschitz constraint~\cite{Huang2022Error,chen2022distribution,Jiao2023Approximation,ding2024semisupervised}.

\par This proof is based on the proof of~\cite[Theorem 1]{Yarotsky2017Error}. The ReLU activation function is defined as $\relu(x)=\max\{0,x\}$, and the ReQU activation function is defined as the squared ReLU function $\requ(x)=(\max\{0,x\})^{2}$.

\begin{lemma}\label{lemma:maxmin:nn}
The maximum or minimum of two inputs can be implemented by a ReLU neural network with $1$ hidden layer and $7$ non-zero parameters.
\end{lemma}

\begin{proof}[Proof of Lemma~\ref{lemma:maxmin:nn}]
According to the equality $a=\relu(a)-\relu(-a)$, the identity mapping can be implemented by a ReLU neural network with $1$ hidden layer and $4$ non-zero parameters. We also notice that
\begin{equation*}
\max\{a,b\}=a+\relu(b-a)=\relu(a)-\relu(-a)+\relu(b-a),
\end{equation*}
which means that the maximum of two inputs can be implemented by $7$ non-zero parameters. By a same argument, with the aid of equality $\min\{a,b\}=a-\relu(a-b)$, we can obtain a same result for the minimum of two inputs. This completes the proof.
\end{proof}

\begin{lemma}\label{lemma:product:nn}
The product of two inputs can be implement by a ReQU neural network with $1$ hidden layer and $12$ non-zero parameters.
\end{lemma}

\begin{proof}[Proof of Lemma~\ref{lemma:product:nn}]
According to~\cite[Lemma 2.1]{Li2019Better}, the following identities hold:
\begin{equation*}
x_{1}x_{2}=\frac{1}{4}w_{3}^{T}\requ(w_{1}x_{1}+w_{2}x_{2}),
\end{equation*}
where $w_{1}=(1,-1,1,-1)^{T}$, $w_{2}=(1,-1,-1,1)^{T}$ and $w_{3}=(1,1,-1,-1)^{T}$, which completes the proof.
\end{proof}

\begin{lemma}\label{lemma:product:p:nn}
The product of $p$ inputs can be implement by a ReQU neural network with $\lceil\log_{2}p\rceil$ hidden layers and $6(p+1)$ non-zero parameters.
\end{lemma}

\begin{proof}[Proof of Lemma~\ref{lemma:product:p:nn}]
Define the augmented input vector $(x_{1},\ldots,x_{p},x_{p+1},\ldots,x_{n})$ where $n=2^{\lceil\log_{2}p\rceil}$ and $x_{i}=1$ for $p+1\leq i\leq n$. Observe that $\prod_{i=1}^{p}x_{i}=\prod_{i=1}^{n}x_{i}$. According to Lemma~\ref{lemma:product:nn}, the mapping $(x_{1},\ldots,x_{n})\rightarrow(x_{1}x_{2},\ldots,x_{n-1}x_{n})\in\bbR^{n/2}$ can be implemented by a ReQU neural network with $1$ hidden layer and $6n$ non-zero parameters. By a same argument, we can construct a ReQU neural network with $1$ hidden layer and $3n$ non-zero parameters, which maps $(x_{1}x_{2},\ldots,x_{n-1}x_{n})$ to $(x_{1}x_{2}x_{3}x_{4},\ldots,x_{n-3}x_{n-2}x_{n-1}x_{n})\in\bbR^{n/4}$. By induction on $n$, the number of layers is given as $\lceil\log_{2}p\rceil$, and the total number of non-zero parameters is given by
\begin{equation*}
12\times(1+2+2^{2}+\ldots+2^{\lceil\log_{2}p\rceil-1})\leq6(p+1).
\end{equation*}
This completes the proof.
\end{proof}

\begin{lemma}\label{lemma:trapezoid:nn}
The univariate trapezoid function
\begin{equation}\label{eq:proof:approx:3}
\psi(z)=\left\{
\begin{aligned}
&1, && |z|<1, \\
&2-|z|, && 1\leq|z|\leq2 , \\
&0, && 2<|z|,
\end{aligned}
\right.
\end{equation}
can be implement by a ReLU neural network with $3$ hidden layers and $14$ non-zero parameters.
\end{lemma}

\begin{proof}[Proof of Lemma~\ref{lemma:trapezoid:nn}]
We first implement the following hat-function by ReLU neural network
\begin{equation*}
\widetilde{\psi}(z)=\left\{
\begin{aligned}
&2-|z|, && |z|\leq2 , \\
&0, && 2<|z|.
\end{aligned}
\right.
\end{equation*}
Noticing that $\widetilde{\psi}(z)=\min\{\relu(z+2),\relu(-z+2)\}$, by applying Lemma~\ref{lemma:maxmin:nn}, we find that $\widetilde{\psi}$ can be implemented by a ReLU neural network with $2$ hidden layers and $11$ non-zero parameters. Further, according to the equality $\psi(z)=\min\{1,\widetilde{\psi}(z)\}$, the univariate trapezoid function $\psi$ can be implemented by a ReLU neural network with $3$ hidden layers and $14$ non-zero parameters. This completes the proof.
\end{proof}

\begin{lemma}[Approximation error]\label{lemma:approx}
Let $p\in\bbN_{+}$, and let $\{N_{k}\}_{k=1}^{p}$ be a set of positive integer. Set the deep neural network class $N(L,S)$ as $L=\lceil\log_{2}p\rceil+3$ and $S=(22p+6)\prod_{k=1}^{p}(N_{k}+1)$. Then for each $u^{*}\in W^{1,\infty}([0,1]^{p})$, there exists a deep neural network $u\in N(L,S)$ such that
\begin{equation*}
\|u-u^{*}\|_{L^{\infty}([0,1]^{p})}\leq2^{p}\sum_{k=1}^{p}\frac{1}{N_{k}}\|\partial_{k}u^{*}\|_{L^{\infty}(\calX)}.
\end{equation*}
Further, it holds that the following holds for each $1\leq k\leq p$:
\begin{equation*}
\|u\|_{L^{\infty}([0,1]^{p})}=\|u^{*}\|_{L^{\infty}([0,1]^{p})} \quad\text{and}\quad
\|\partial_{k}u\|_{L^{\infty}([0,1]^{p})}\leq 3\|\partial_{k}u^{*}\|_{L^{\infty}([0,1]^{p})}.
\end{equation*}
\end{lemma}

\begin{proof}[Proof of Lemma~\ref{lemma:approx}]

\par The proof is divided into three steps. In the first step, we approximate the target function based on a partition of unity and the degree-$0$ Taylor expansion. Then we implement this piece-wise linear function using deep neural network exactly in the second step. Finally, in the last step, we estimate the Lipschitz constant of the deep neural network.

\par\noindent\emph{Step 1. Approximate the target function by a piecewise linear function.}

\par Consider a partition of unity formed by a grid of $\prod_{k=1}^{p}(N_{k}+1)$ functions $\phi_{m}$ on the domain $[0,1]^{p}$:
\begin{equation}\label{eq:proof:approx:1}
\sum_{m}\phi_{m}(x)\equiv 1, \quad x\in[0,1]^{p},
\end{equation}
where the multi-index $m$ is defined as $m=(m_{1},\ldots,m_{p})^{T}$ with $m_{k}\in\{0,\ldots,N_{k}\}$, and the function $\phi_{m}$ is defined as the product
\begin{equation}\label{eq:proof:approx:2}
\phi_{m}(x)=\prod_{k=1}^{p}\psi\Big(3N_{k}\Big(x_{k}-\frac{m_{k}}{N_{k}}\Big)\Big).
\end{equation}
Here $\psi$ is the univariate trapezoid function defined as~\eqref{eq:proof:approx:3}. It is noticeable that for each $m$,
\begin{equation}\label{eq:proof:approx:5}
\sup_{z\in[0,1]}|\psi(z)|=1, \quad \sup_{x\in[0,1]^{p}}|\phi_{m}(x)|=1,
\end{equation}
and
\begin{equation}\label{eq:proof:approx:6}
\supp(\phi_{m})\subset\Big\{x\in[0,1]^{p}:\Big|x_{k}-\frac{m_{k}}{N_{k}}\Big|\leq\frac{2}{3N_{k}},~1\leq k\leq p\Big\}.
\end{equation}
\par Now define a piecewise linear approximation to $u^{*}$ by
\begin{equation}\label{eq:proof:approx:4}
u(x)=\sum_{m}\phi_{m}(x)u^{*}\Big(\frac{m_{1}}{N_{1}},\ldots,\frac{m_{p}}{N_{p}}\Big).
\end{equation}
Then it follows that
\begin{align*}
&|u^{*}(x)-u(x)|\leq\sum_{m}\Big|\phi_{m}(x)\Big(u^{*}(x)-u^{*}\Big(\frac{m_{1}}{N_{1}},\ldots,\frac{m_{p}}{N_{p}}\Big)\Big)\Big| \\
&\leq\sum_{m}\Big|u^{*}(x)-u^{*}\Big(\frac{m_{1}}{N_{1}},\ldots,\frac{m_{p}}{N_{p}}\Big)\Big|\bbone\Big\{m:\Big|x_{k}-\frac{m_{k}}{N_{k}}\Big|\leq\frac{2}{3N_{k}},~k\in[p]\Big\} \\
&\leq 2^{p}\max_{m}\Big|u^{*}(x)-u^{*}\Big(\frac{m_{1}}{N_{1}},\ldots,\frac{m_{p}}{N_{p}}\Big)\Big|\bbone\Big\{m:\Big|x_{k}-\frac{m_{k}}{N_{k}}\Big|\leq\frac{2}{3N_{k}},~k\in[p]\Big\} \\
&\leq 2^{p}\max_{m}\sum_{k=1}^{p}\esssup_{x\in[0,1]^{p}}|\partial_{k}u^{*}(x)|\Big|x_{k}-\frac{m_{k}}{N_{k}}\Big|\bbone\Big\{m:\Big|x_{k}-\frac{m_{k}}{N_{k}}\Big|\leq\frac{2}{3N_{k}},~k\in[p]\Big\} \\
&\leq 2^{p}\sum_{k=1}^{p}\esssup_{x\in[0,1]^{p}}|\partial_{k}u^{*}(x)|\frac{2}{3N_{k}},
\end{align*}
where the first inequality follows from the triangular inequality, the second inequality is due to~\eqref{eq:proof:approx:5} and~\eqref{eq:proof:approx:6}, the third inequality used the observation that any $x\in[0,1]^{p}$ belongs to the support of at most $2^{d}$ functions $\phi_{m}$, the forth inequality used Taylor's theorem of degree-$0$, and the last inequality holds from H{\"o}lder's inequality. Consequently, we obtain the following inequality
\begin{equation*}
\|u-u^{*}\|_{L^{\infty}([0,1]^{p})}\leq2^{p}\sum_{k=1}^{p}\frac{1}{N_{k}}\|\partial_{k}u^{*}\|_{L^{\infty}([0,1]^{p})}.
\end{equation*}

\par\noindent\emph{Step 2. Implement the piecewise linear function by a deep neural network.}

\par In this step, we implement the piececwise linear approximation~\eqref{eq:proof:approx:4} by a deep neural network. Using Lemmas~\ref{lemma:product:p:nn} and~\ref{lemma:trapezoid:nn}, for each $m$, the function $\phi_{m}$ defined as~\eqref{eq:proof:approx:2} can be implemented by a deep neural network with $\lceil\log_{2}p\rceil+3$ layers and $16p+6(p+1)=22p+6$ non-zero parameters. Since $u$ defined in~\eqref{eq:proof:approx:4} is a linear combination of $\prod_{k=1}^{p}(N_{k}+1)$ functions $\phi_{m}$, it can be implemented by a deep neural network with $\lceil\log_{2}p\rceil+3$ layers and $(22p+6)\prod_{k=1}^{p}(N_{k}+1)$ non-zero parameters.

\par\noindent\emph{Step 3. Compute the Lipschitz constant of the deep neural network.}

\par According to \emph{Step 2}, the piecewise linear approximation $u$ can be implemented by a deep neural network with no error. Therefore, it suffices to compute the Lipschitz constant of $u$ in~\eqref{eq:proof:approx:4}. Taking derivative on both sides of~\eqref{eq:proof:approx:4} with respect to $x_{k}$ yields
\begin{align}
\partial_{k}u(x)
&=\sum_{m}u^{*}\Big(\frac{m_{1}}{N_{1}},\ldots,\frac{m_{p}}{N_{p}}\Big)\partial_{k}\phi_{m}(x) \nonumber \\
&=\sum_{m}u^{*}\Big(\frac{m_{1}}{N_{1}},\ldots,\frac{m_{p}}{N_{p}}\Big)A_{k}(m)\partial_{k}\psi\Big(3N_{k}\Big(x_{k}-\frac{m_{k}}{N_{k}}\Big)\Big) \nonumber \\
&=\sum_{m}u^{*}\Big(\frac{m_{1}}{N_{1}},\ldots,\frac{m_{p}}{N_{p}}\Big)A_{k}(m)\partial_{k}\psi\Big(3N_{k}\Big(x_{k}-\frac{m_{k}}{N_{k}}\Big)\Big)\bbone\Big\{m:\Big|x_{k}-\frac{m_{k}}{N_{k}}\Big|\leq\frac{2}{3N_{k}}\Big\}, \label{eq:proof:approx:7}
\end{align}
where the constant is given as
\begin{equation*}
A_{k}(m)=\prod_{\ell\neq k}\psi\Big(3N_{\ell}\Big(x_{\ell}-\frac{m_{\ell}}{N_{\ell}}\Big)\Big)\bbone\Big\{m:\Big|x_{\ell}-\frac{m_{\ell}}{N_{\ell}}\Big|\leq\frac{2}{3N_{\ell}},~\ell\neq k\Big\}.
\end{equation*}
It is evident that $0\leq A_{k}(m)\leq1$ for each $1\leq k\leq p$ and $m$.

\par We next estimate~\eqref{eq:proof:approx:7} in the following cases.

\begin{enumerate}[(i)]
\item If there exists $m_{k}^{*}\in\{1,\ldots,N_{k}\}$ such that $|x_{k}-\frac{m_{k}^{*}}{N_{k}}|\leq\frac{1}{3N_{k}}$, then
\begin{equation*}
\partial_{k}u(x)=u^{*}\Big(\frac{m_{1}}{N_{1}},\ldots,\frac{m_{k}^{*}}{N_{k}},\ldots,\frac{m_{p}}{N_{p}}\Big)A_{k}(m^{*})\partial_{k}\psi\Big(3N_{k}\Big(x_{k}-\frac{m_{k}^{*}}{N_{k}}\Big)\Big)=0,
\end{equation*}
where $m^{*}=(m_{1},\ldots,m_{k}^{*},\ldots,m_{p})^{T}$.
\item If there exists $m_{k}^{*}\in\{1,\ldots,N_{k}\}$ such that $\frac{m_{k}^{*}}{N_{k}}+\frac{1}{3N_{k}}\leq x_{k}\leq\frac{m_{k}^{*}}{N_{k}}+\frac{2}{3N_{k}}$, then
\begin{align*}
\partial_{k}u(x)
&=u^{*}\Big(\frac{m_{1}}{N_{1}},\ldots,\frac{m_{k}^{*}}{N_{k}},\ldots,\frac{m_{p}}{N_{p}}\Big)A_{k}(m^{*})\partial_{k}\psi\Big(3N_{k}\Big(x_{k}-\frac{m_{k}^{*}}{N_{k}}\Big)\Big) \\
&\quad+u^{*}\Big(\frac{m_{1}}{N_{1}},\ldots,\frac{m_{k}^{*}+1}{N_{k}},\ldots,\frac{m_{p}}{N_{p}}\Big)A_{k}(m_{+}^{*})\partial_{k}\psi\Big(3N_{k}\Big(x_{k}-\frac{m_{k}^{*}+1}{N_{k}}\Big)\Big) \\
&=-3u^{*}\Big(\frac{m_{1}}{N_{1}},\ldots,\frac{m_{k}^{*}}{N_{k}},\ldots,\frac{m_{p}}{N_{p}}\Big)A_{k}(m^{*})N_{k} \\
&\quad+3u^{*}\Big(\frac{m_{1}}{N_{1}},\ldots,\frac{m_{k}^{*}+1}{N_{k}},\ldots,\frac{m_{p}}{N_{p}}\Big)A_{k}(m_{+}^{*})N_{k} \\
&\leq3N_{k}\Big|u^{*}\Big(\frac{m_{1}}{N_{1}},\ldots,\frac{m_{k}^{*}+1}{N_{k}},\ldots,\frac{m_{p}}{N_{p}}\Big)-u^{*}\Big(\frac{m_{1}}{N_{1}},\ldots,\frac{m_{k}^{*}}{N_{k}},\ldots,\frac{m_{p}}{N_{p}}\Big)\Big| \\
&\leq3\esssup_{x\in[0,1]^{p}}|\partial_{k}u^{*}(x)|,
\end{align*}
where $m_{+}^{*}=(m_{1},\ldots,m_{k}^{*}+1,\ldots,m_{p})^{T}$, the first inequality follows from the fact that $|A_{k}(m)|\leq1$, and the last inequality is due to Taylor's theorem.
\item If there exists $m_{k}^{*}\in\{1,\ldots,N_{k}\}$ such that $\frac{m_{k}^{*}}{N_{k}}-\frac{2}{3N_{k}}\leq x_{k}\leq\frac{m_{k}^{*}}{N_{k}}-\frac{1}{3N_{k}}$, then by a same argument, we have
\begin{align*}
\partial_{k}u(x)
&\leq3N_{k}\Big|u^{*}\Big(\frac{m_{1}}{N_{1}},\ldots,\frac{m_{k}^{*}}{N_{k}},\ldots,\frac{m_{p}}{N_{p}}\Big)-u^{*}\Big(\frac{m_{1}}{N_{1}},\ldots,\frac{m_{k}^{*}-1}{N_{k}},\ldots,\frac{m_{p}}{N_{p}}\Big)\Big| \\
&\leq3\esssup_{x\in[0,1]^{p}}|\partial_{k}u^{*}(x)|.
\end{align*}
\end{enumerate}
Combining the three cases above, we obtain the following inequality
\begin{equation*}
\|\partial_{k}u\|_{L^{\infty}([0,1]^{p})}\leq 3\|\partial_{k}u^{*}\|_{L^{\infty}([0,1]^{p})}, \quad 1\leq k\leq p.
\end{equation*}
This completes the proof.
\end{proof}

\par We have investigated the approximation error of a target function on the hypercube $[0,1]^{p}$ in Lemma~\ref{lemma:approx}. In the following corollary, we extend our analysis to target functions on general bounded domain $[0,T]\times[-R,R]^{d}$.

\begin{corollary}\label{corollary:approx}
Let $p\in\bbN_{+}$, $\calX=\prod_{k=1}^{p}[a_{i},b_{i}]$, and let $\{N_{k}\}_{k=1}^{p}$ be a set of positive integer. Set the deep neural network class $N(L,S)$ as $L=\lceil\log_{2}p\rceil+3$ and $S=(22p+6)\prod_{k=1}^{p}(N_{k}+1)$. Then for each $u^{*}\in W^{1,\infty}(\calX)$, there exists a deep neural network $u\in N(L,S)$ such that
\begin{equation*}
\|u-u^{*}\|_{L^{\infty}(\calX)}\leq2^{p}\sum_{k=1}^{p}\frac{b_{k}-a_{k}}{N_{k}}\|\partial_{k}u^{*}\|_{L^{\infty}(\calX)}.
\end{equation*}
Further, it holds that for each $1\leq k\leq p$:
\begin{align*}
\|u\|_{L^{\infty}(\calX)}=\|u^{*}\|_{L^{\infty}(\calX)} \quad\text{and}\quad
\|\partial_{k}u\|_{L^{\infty}(\calX)}\leq 3\|\partial_{k}u^{*}\|_{L^{\infty}(\calX)}.
\end{align*}
\end{corollary}

\begin{proof}[Proof of Corollary~\ref{corollary:approx}]
We first define a variable transformation on $u^{*}\in W^{1,\infty}(\calX)$ as
\begin{align*}
\phi:W^{1,\infty}(\calX)&\rightarrow W^{1,\infty}([0,1]^{p}) \\
u^{*}(x)&\mapsto (\phi\circ u^{*})(x^{\prime})=u^{*}(a+(b-a)x^{\prime}),
\end{align*}
where $(b-a)x^{\prime}=((b_{k}-a_{k})x_{k}^{\prime})_{k=1}^{p}\in\bbR^{p}$. Then it is noticeable that
\begin{equation*}
\esssup_{x^{\prime}\in[0,1]^{p}}\Big|\frac{\partial(\phi\circ u^{*})}{\partial x_{k}^{\prime}}(x^{\prime})\Big|=(b_{k}-a_{k})\esssup_{x\in\calX}\Big|\frac{\partial u^{*}}{\partial x_{k}}(x)\Big|.
\end{equation*}
Set the deep neural network class $N(L,S)$ as $L=\lceil\log_{2}p\rceil+3$ and $S=(22p+6)\prod_{k=1}^{p}(N_{k}+1)$. According to Lemma~\ref{lemma:approx}, there exists a neural network $u\in N(L,S)$ such that
\begin{align}
|u(x^{\prime})-(\phi\circ u^{*})(x^{\prime})|
&\leq2^{p}\sum_{k=1}^{p}\esssup_{x^{\prime}\in[0,1]^{p}}\Big|\frac{\partial(\phi\circ u^{*})}{\partial x_{k}^{\prime}}(x^{\prime})\Big|\frac{1}{N_{k}} \nonumber \\
&\leq2^{p}\sum_{k=1}^{p}\frac{b_{k}-a_{k}}{N_{k}}\esssup_{x\in\calX}\Big|\frac{\partial u^{*}}{\partial x_{k}}(x)\Big|, \label{eq:proof:approx:8}
\end{align}
and for each $1\leq k\leq p$, the following inequality holds:
\begin{equation}\label{eq:proof:approx:9}
\esssup_{x^{\prime}\in[0,1]^{p}}\Big|\frac{\partial u}{\partial x_{k}^{\prime}}(x^{\prime})\Big|\leq3\esssup_{x^{\prime}\in[0,1]^{p}}\Big|\frac{\partial(\phi\circ u^{*})}{\partial x_{k}^{\prime}}(x^{\prime})\Big|=3(b_{k}-a_{k})\esssup_{x\in\calX}\Big|\frac{\partial u^{*}}{\partial x_{k}}(x)\Big|.
\end{equation}
We next define the inverse transform on $u\in W^{1,\infty}([0,1]^{p})$ as
\begin{align*}
\psi:W^{1,\infty}([0,1]^{p})&\rightarrow W^{1,\infty}(\calX) \\
u(x^{\prime})&\mapsto(\psi\circ u)(x)=u\Big(\frac{x^{\prime}-a}{b-a}\Big),
\end{align*}
where $(x^{\prime}-a)/(b-a)=((x_{k}^{\prime}-a_{k})/(b_{k}-a_{k}))_{k=1}^{p}\in\bbR^{p}$. It follows from~\eqref{eq:proof:approx:9} that
\begin{equation*}
\esssup_{x\in\calX}\Big|\frac{\partial(\psi\circ u)}{\partial x_{k}}(x)\Big|=\frac{1}{b_{k}-a_{k}}\esssup_{x^{\prime}\in[0,1]^{p}}\Big|\frac{\partial u}{\partial x_{k}^{\prime}}(x^{\prime})\Big|\leq3\esssup_{x\in\calX}\Big|\frac{\partial u^{*}}{\partial x_{k}}(x)\Big|,
\end{equation*}
for each $1\leq k\leq p$. Then composing $\psi$ on both sides of~\eqref{eq:proof:approx:8} yields the desired inequality.
\end{proof}

\section{Denoiser parameterization} \label{app:denoiser_param}
In practice, we parameterize the network $D_\theta(t, x)$ following~\cite{karras2022elucidating}:
\begin{equation}
\label{eq:denoiser-param}
D_\theta(t, x) = \cskip(t) x + \cout(t) F_\theta\left(\cnoise(t), \cin(t) x\right),
\end{equation}
where $F_\theta$ is the neural network to be trained, $\cskip(t)$ scale the skip connection, $\cin(t)$ and $\cout(t)$ scale the input and output of $F_\theta$, and $\cnoise(t)$ scales time $t$.

Now~{\eqref{eq:velocity:popu:risk}} becomes
\begin{align}
\calL(F) = & \int_0^1 \bbE_{X_0} \bbE_{X_1} \left[ \omega(t) \|\cskip(t) X_t + \cout(t) F\left(\cnoise(t), \cin(t) X_t\right) - x_1 \|^2 \right] \dt \label{eq:dsm-formal-loss} \\
= & \int_0^1 \bbE_{X_0} \bbE_{X_1} \left[ \omega(t) \cout^2(t) \|F\left(\cnoise(t), \cin(t)X_t\right) - \frac{X_1 - \cskip(t)X_t}{\cout(t)}\|^2 \right] \dt \notag \\
= & \int_0^1 \bbE_{X_0} \bbE_{z} \left[ \lambda(t) \|F_{\text{pred}} - F_{\text{target}}\|^2 \right] \dt, \notag
\end{align}
where $\lambda(t) = \omega(t) \cout^2(t), F_{\text{pred}} = F\left(\cnoise(t), \cin(t)X_t\right)$ and $F_{\text{target}} = \frac{X_1 - \cskip(t)X_t}{\cout(t)}$.

Let $\sigma$ denote the standard deviation of $\mu_1$. We now design $\cin$ so that the spatial inputs of $F$ has unit variance.
\begin{align*}
& \var[\cin(t) x_t] = 1, \\
\Leftrightarrow & \cin(t) = \sqrt{\frac{1}{\var[X_t]}} = \sqrt{\frac{1}{\alpha_t^2 \sigma^2 + \beta_t^2}}.
\end{align*}

We then design $\cout$ so that the $F_{\text{target}}$ has unit variance.
\begin{align}
& \var[\frac{X_1 - \cskip(t) X_t}{\cout(t)}] = 1, \notag \\
\Leftrightarrow & \cout(t) = \sqrt{\var[(1-\alpha_t\cskip(t)) X_1 - \cskip(t)\beta_t z]}, \notag \\
\Leftrightarrow & \cout(t) = \sqrt{(1-\alpha_t\cskip(t))^2 \sigma^2 + \cskip(t)^2 \beta_t^2}. \label{eq:cout-raw}
\end{align}

We then design $\cskip$ that minimizes $\cout$ so that the errors of $F$ are amplified as little as possible.
Let $\frac{\partial \cout^2(t)}{\partial \cskip(t)} = 0$, we obtain
\begin{align}
& -\alpha_t (1-\alpha_t\cskip(t)) \sigma^2 + \beta_t^2 \cskip(t) = 0, \notag \\
\Leftrightarrow & \cskip(t) = \frac{\alpha_t \sigma^2}{\alpha_t^2 \sigma^2 + \beta_t^2}. \label{eq:cskip}
\end{align}
We can check that~\eqref{eq:cskip} is indeed the minima of $\cout$.
Meanwhile,~\eqref{eq:cout-raw} yields
\[
\cout(t) = \frac{\beta_t \sigma}{\sqrt{\alpha_t^2 \sigma^2 + \beta_t^2}}.
\]
We can then design $\omega(t)$ so that $\lambda(t) = 1$ uniformly on $[0, 1]$.
\[
\omega(t) = \frac{1}{\cout^2(t)} = \frac{\alpha_t^2 \sigma^2 + \beta_t^2}{\beta_t^2 \sigma^2}.
\]

We conclude the form of coefficients in Table~\ref{tab:denoiser_param}.

\begin{table}[ht]
\centering
\caption{Denoiser parameterization.}
\label{tab:denoiser_param}
\begin{tabular}{ccc}
\toprule
function & requirements & form \\
\midrule
$\cnoise(t)$ & - & free choice \\
$\cin(t)$ & $\var[\cin(t) X_t] = 1$ & $\sqrt{\frac{1}{\alpha_t^2 \sigma^2 + \beta_t^2}}$ \\
$\cout(t)$ & $\var[\frac{X_1 - \cskip(t) X_t}{\cout(t)}] = 1$ & $\frac{\beta_t \sigma}{\sqrt{\alpha_t^2 \sigma^2 + \beta_t^2}}$ \\
$\cskip(t)$ & $\frac{\partial \cout^2(t)}{\partial \cskip(t)} = 0$ & $\frac{\alpha_t \sigma^2}{\alpha_t^2 \sigma^2 + \beta_t^2}$ \\
$\omega(t)$ & $\lambda(t) = 1$ & $\frac{\alpha_t^2 \sigma^2 + \beta_t^2}{\beta_t^2 \sigma^2}$ \\
\bottomrule
\end{tabular}
\end{table}

Now~\eqref{eq:dsm-formal-loss} can be used as the working denoiser matching loss.

\section{ Extra experiment details}
\label{sec:extra experiment details}

This section reports the detailed settings of the experiments conducted on image datasets.
The details are shown in Table \ref{tab:hyperparam}.

\begin{table}
\centering
\small
\caption{Hyperparameters.}
\label{tab:hyperparam}
\begin{tabular}{cccccc}
\toprule
& MNIST & CIFAR-10 & CelebAHQ-256 & CelebAHQ-512 & { ImageNet-256} \\
\midrule
encoder/decoder & \ding{55} & \ding{55} & \ding{51} & \ding{51} & \ding{51} \\
number of GPUs & 4 $\times$ A800 & 4 $\times$ A800 & 12 $\times$ A100 & 8 $\times$ A100 & { 4 $\times$ A100}  \\
architecture & DDPM++ & DDPM++ & DiT-B/2 & DiT-XL/2 & { DiT-B/2} \\
channel multiplier & 32 & 128 & - & - \\
channels per resolution & 1, 2, 2 & 1, 2, 2, 2 & - & - & - \\
model size (GB) & 0.02 & 0.5 & 0.25 & 1.3 & { 0.25} \\
optimizer & RAdam & RAdam & AdamW & AdamW & { AdamW} \\
learning rate & 1e-3 & 1e-4 & 1e-4 & 1e-4 & { 1e-4} \\
LR ramp-up (Mimg) & 0.5 & 0.5 & 10 & 10 & { 0}\\
EMA rate & 0.999 & 0.9999 & - & - & - \\
EMA half-life (Mimg) & - & - & 0.5 & 0.5 & { 0.5} \\
%weight decay & 0 & 0 & 0.1 & 0.1 \\
teacher duration (Mimg) &  10 & 100 & - & - & - \\
student duration (Mimg) &  2 & 2 & 100 & 15 & { 100} \\
batch size & 1024 & 1024 & 192 & 128 & { 256} \\
training steps & \multicolumn{4}{c}{duration / batch\_size} \\
stochastic interploants & F\"ollmer & F\"ollmer & Linear & Linear & { Linear}\\
global loss metric & LPIPS~\cite{zhang2018perceptual} & LPIPS & $L^2$ & $L^2$ \\
adaptive weightning~\cite{Esser2021Taming,kim2024consistency} & \ding{51} & \ding{51} & \ding{55} & \ding{55} & \ding{55} \\
NFE of teacher (max) & 19 & 19 & 2 & 2 & { 2} \\
dropout & 0.1 & 0.1 & 0.13 & 0.13 & { 0.1} \\
\bottomrule
\end{tabular}
\end{table}

In practice, when there is an available teacher network $D_{\calT}$, we initialize $D_{\calS}$ from the pre-trained $D_{\calT}$, and then embed another temporal input into it to construct the characteristic generator as~\cite{kim2024consistency} did.
For the choice of $u$ in~\eqref{eq:char:pop:risk:2}, we apply a weighted random strategy which is more likely to choose a long range interval $[t, u]$ to ensure the precision of the teacher solver, as~\cite{kim2024consistency} did.
Inspired by~\cite{Esser2021Taming,kim2024consistency}, we adaptively balance the global characteristic matching loss and the local denoiser matching loss on MNIST and CIFAR-10.
We calculate FID in original pixel space for MNIST, and in the feature space (dimension=2048) extracted by a pre-trained Inceptionv3 network for other image datasets.

{
\section{Ablation Studies}
\label{app:ablation}

This section supply several small ablation studies on the CIFAR10 dataset.

\subsection{Ablation Study on Euler and Exponential Integrator}
In principle, more accurate reference solution given by the pre-trained teacher model yields better student model. We show the superiority of the exponential integrator over Euler by comparing the generation quality of the teacher model using both discretization method under the same NFE (NFE=32). The FID of the samples generated with exponential integrator and Euler is 5.55 and 12.27 respectively, empirically validating the theoretical finding. As a result, the reference solution generated by the exponential integrator has better quality than that by the Euler method, offering a more stable distillation target.

\subsection{Ablation Study on Self-Distillation}\label{section:ablation:student:teacher}
We introduced two training variations of characteristic generator: a student-teacher distillation type where a pre-trained teacher model is explicitly required during training to provide reference solution, and a self-distillation type where the generator itself provides the reference solution, eliminating the need of a pre-trained teacher model. We compares the two distillation types on the CIFAR10 dataset. Concerning 1-NFE FID, the student-teacher distillation characteristic generator outperforms than the self-distillation one by 2.40 versus 3.35. The results is reasonable as the teacher model provides a solid generation target, stabilizing the training process, while the self-distillation might suffer from mode collapse in the early training stage, as the generator might not be able to generate meaningful reference samples at initialization.
}

\section{Remarks on Time Consumption}
\label{sec:remark time}
For numerical experiments on the MNIST and CIFAR-10 dataset, a pre-trained denoiser network (teacher model) is explicitly involved, to generate the reference trajectories, from which the characteristic generator is distilled.

Take the CIFAR-10 dataset for example. The total time consumption is summarized as follows:
we first spend around 40 NVIDIA A800 hours during the pre-train stage of the denoiser network.
During the training of the characteristic generator, the maximum number of steps to generate each reference trajectory is set to 19.
On average, the reference trajectory is generated on-the-fly with speed 0.007 sec per image.
In total, the training time of the characteristic generator is around 3 NVIDIA A800 hours, 38\% of which is spent on generating the reference trajectory from the teacher model.

For the CelebA 256 $\times$ 256 dataset, we first spend around 78 NVIDIA A800 hours during the pre-train stage of the denoiser network.
Then we use the pre-trained result from the first stage as the source of parameter initialization and teacher model. The training time of characteristic generator is around 480 NVIDIA A800 hours, around 45\% of which are spent on generating a reference trajectory, by evaluating the characteristic generator itself twice.

For the CelebA 512 $\times$ 512 dataset, there is no pre-training stage. The training time of characteristic generator is around 480 NVIDIA A800 hours.

{ For the ImageNet 256 $\times$ 256 dataset, there is no pre-training stage. The training time of characteristic generator is around 320 NVIDIA A100 hours.}

As for the generation process, since the extra parameters of the characteristic generator is relatively small compared to those of the original velocity/denoiser network, the time consumption of each evaluation of the characteristic generator is similar to that of each Euler step.

It is important to note that these training times are one-time offline costs.
Once trained, the characteristic generator can significantly reduce the number of function evaluations (NFE) during inference.
Traditional sampling methods for diffusion models require hundreds or thousands of NFE: DDPM requires 1000 NFE, DDIM requires 100 NFE, and Score SDE requires 2000 NFE to achieve their reported FID scores.
In contrast, our characteristic generator achieves competitive performance with substantially fewer NFE.
For instance, CG achieves FID of 4.59 with only NFE=1, and FID of 2.83 with NFE=4, which is already comparable to Score SDE (FID=2.20) that requires 2000 NFE.
This demonstrates that our approach enables significantly faster inference by reducing the number of function evaluations by orders of magnitude.
Thus, the offline training cost is an one-time investment that enables efficient inference once and for all.

\section{Comparison with Consistency Trajectory Models}
\label{section:comparison:ctm}

In this section, we discuss the \textbf{implementation details} compared with the Consistency Trajectory Models (CTM).
\subsection*{Similarities}
\begin{itemize}
\item Our work and the CTM share similar neural network initialization strategy (from a pre-trained teacher model, if any).
\item On MNIST and CIFAR-10, where the teacher and student model are two different networks, our work and CTM shares similar strategies for generating reference solutions from the teacher model: running some ODE solver on the given time interval, with a threshold on the maximum number of steps allowed.
\item Also on MNIST and CIFAR-10, we and CTM both use the Learned Perceptual Image Patch Similarity (LPIPS) metric instead of $L^2$-norm to measure distance in the feature space instead of in the original pixel space~\cite{zhang2018perceptual}.
We also share similar loss balance techniques to balance between the global loss and the local loss.
\end{itemize}

\subsection*{Differences}
\begin{itemize}
\item We extend the distillation framework to the latent space on CelebAHQ-256 and CelebA-512, significantly reducing the training and generating cost, while CTM only experiments on the original pixel space.
We also manage to shift from the U-Net architecture to the DiT architecture.
\item On CelebAHQ, We extend the distillation framework to self-distillation, and offer a candidate design for the self-guided teacher model, which takes only 2 NFE to generate a reference solution, while CTM still needs dozens of NFEs.
Under this setting, the training process of our proposed method would be significantly faster than CTM.
\item The incorporation of GAN contributes a lot to the low FID of CTM, but we find it introduces extra training instability and requires some manual tuning.
Our work is more like a clean and minimal template for the distillation-based diffusion models, and the experiments results indicates that our implementation works well enough and could be easily adapted to downstream tasks.
We believe that less is more when it comes to high-level framework.
\end{itemize}

\section{Comparisons Between One-Step Exponential Integrator and Characteristic Generator}
\label{section:comparison}
{
\begin{itemize}
\item Error bound for one-step exponential integrator (EI) generation. Recall the error bound for the EI sampling in Theorem~\ref{theorem:error:euler}:
\begin{equation*}
\text{Error of EI}\leq C\Phi_{T}\Big\{n^{-\frac{2}{d+3}}\log^{2}n+\frac{1}{K_{\mathrm{EI}}}\Big\},
\end{equation*}
where $K_{\mathrm{EI}}$ is the number of steps of the EI sampling. One-step EI generation means $K_{\mathrm{EI}}=1$, thus
\begin{equation*}
\text{Error of one-step EI}\leq C\Phi_{T}\Big\{n^{-\frac{2}{d+3}}\log^{2}n+1\Big\}.
\end{equation*}
The error is dominated by the constant term, and does not converge even for a sufficiently large number of samples $n$. This large inherent error from using a single, coarse step explains the poor image quality seen in Figure~\ref{fig:comparison}.
\item Error bound for one-step characteristic generation.
The characteristic generator is fundamentally different. It is a one-step model that is trained to approximate the result of a high-quality, multi-step numerical sampler. In the characteristic learning, the generator is learned from the trajectories generated using an exponential integrator sampler with $K_{\mathrm{CG}}$ steps. Recall the error bound for the characteristic generator in Theorem~\ref{theorem:error:characteristic}:
\begin{equation*}
\text{Error of CG}\leq C\Phi_{T}\Big\{n^{-\frac{2}{d+3}}\log^{2}n+\frac{1}{K_{\mathrm{CG}}}\Big\}+C\Big\{m^{-\frac{2}{d+4}}\log^{2}m+\frac{\log m}{K_{\mathrm{CG}}}\Big\}.
\end{equation*}
In practice, we can choose a sufficiently large $K_{\mathrm{CG}}$. As a result, the final error of the one-step CG is dominated by the sample complexity terms:
\begin{equation*}
\text{Error of CG}\leq C\Phi_{T}n^{-\frac{2}{d+3}}\log^{2}n+Cm^{-\frac{2}{d+4}}\log^{2}m,
\end{equation*}
which converge to zero as the dataset sizes ($n$ and $m$) increase.
\end{itemize}}
In essence, the CG distills the knowledge of a precise, multi-step solver into a fast, single-step network. Our theory correctly predicts that its error should be much lower than that of a naive one-step EI sampler, which aligns perfectly with our empirical findings in Figure~\ref{fig:comparison}.